\pdfoutput=1
\PassOptionsToPackage{sort}{natbib}  
\documentclass{article}

\usepackage[preprint,nonatbib]{neurips_2022}

\usepackage{graphicx}
\usepackage{subcaption}

\usepackage{tikz}
\usepackage{comment}
\usepackage{amsmath,amssymb} 
\usepackage{color}
\usepackage{bbding}
\usepackage{pifont}

\usepackage[accsupp]{axessibility}  

\usepackage{tikz}
\usepackage{comment} 
\usepackage{amsmath,amssymb} 
\usepackage{color}

\usepackage{booktabs}
\usepackage{multirow}
\usepackage{tabularx}
\usepackage{microtype}
\usepackage[normalem]{ulem}
\usepackage{ dsfont }
\usepackage{xcolor}
\usepackage{ulem}
\usepackage{arydshln}
\usepackage{tabularx}
\usepackage{amsmath}
\usepackage{parskip}

\usepackage[shortlabels]{enumitem}
\usepackage{algorithmic}
\usepackage{algorithm}
\long\def\ignorethis#1{}

\newcolumntype{Y}{>{\centering\arraybackslash}X}
\newcolumntype{L}{>{\arraybackslash}X}\newbox\jsavebox

\usepackage[percent]{overpic}

\newcommand{\cmark}{\text{\ding{51}}}
\newcommand{\xmark}{\text{\ding{55}}}

\usepackage[pagebackref=true,breaklinks=true,letterpaper=true,colorlinks,bookmarks=false,citecolor=blue,linkcolor=blue,urlcolor=blue]{hyperref}

\begin{document}

\newcommand*{\XT}[1]{\textcolor{orange}{[XT: #1]}}
\newcommand*{\AJ}[1]{\textcolor{blue}{[AJ: #1]}}
\newcommand*{\SF}[1]{\textcolor{red}{[SF: #1]}}

\title{Autoregressive 3D Shape Generation via \\ Canonical Mapping} 

\author{
An-Chieh Cheng$^{1*}$, 
 \textbf{Xueting Li$^{2*}$}, \
\textbf{Sifei Liu$^{2}$\thanks{Equal contribution}}, \
 \textbf{Min Sun$^{1}$}, \
 \textbf{Ming-Hsuan Yang$^{3}$},  \
\\ $^{1}$National Tsing-Hua University, \
 $^{2}$NVIDIA, \
 $^{3}$University of California, Merced}

\maketitle

\begin{abstract}

With the capacity of modeling long-range dependencies in sequential data, transformers have shown remarkable performances in a variety of generative tasks such as image, audio, and text generation.
Yet, taming them in generating less structured and voluminous data formats such as high-resolution point clouds have seldom been explored due to ambiguous sequentialization processes and infeasible computation burden.
In this paper, we aim to further exploit the power of transformers and employ them for the task of 3D point cloud generation.
The key idea is to decompose point clouds of one category into semantically aligned sequences of shape compositions, via a learned canonical space.
These shape compositions can then be quantized and used to learn a context-rich composition codebook for point cloud generation.
Experimental results on point cloud reconstruction and unconditional generation show that our model performs favorably
against state-of-the-art approaches. 
Furthermore, our model can be easily extended to multi-modal shape completion as an application for conditional shape generation.

\end{abstract}

\section{Introduction}

In the past few years, transformers not only dominate the natural language processing area~\cite{vaswani2017attention,devlin2018bert,brown2020language}, but also consistently show remarkable performance in a variety of vision tasks such as image classification~\cite{dosovitskiy2020image}, semantic and instance segmentation~\cite{liu2021swin,strudel2021segmenter} and image generation~\cite{esser2021taming}.
Compared to convolutional neural networks, transformers learn dependencies between visual elements from scratch without making any prior assumptions about data structure.
As a result, they are more flexible and capable of capturing long-range dependencies in sequential data.
Such property is especially desirable in the autoregressive generation of globally coherent long-range sequential data, such as high-resolution images.
Indeed, promising performance of autoregressive generation via transformers has been demonstrated in \cite{esser2021taming} for image generation.

\begin{figure}[h]
\centering
\includegraphics[width=1.0\linewidth]{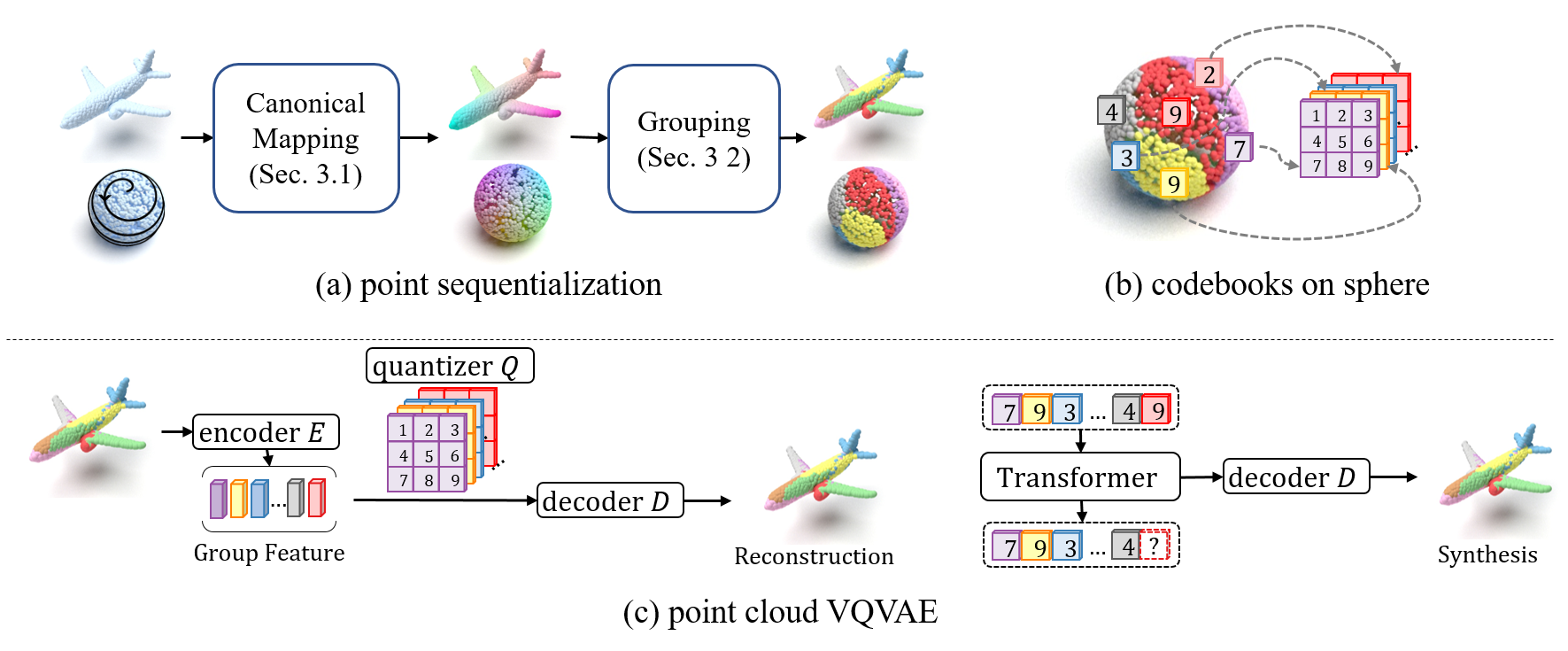}
\caption{Given a point cloud, we first decompose it into a sequence of perceptually meaningful shape compositions via a canonical auto-encoder in (a). A group of codebooks is then learned on the sequentialized shape compositions in (b). Finally, we introduce an autoregressive model for point cloud generation in (c).
}
\label{fig:teaser}
\vspace{-1.5em}
\end{figure}

However, employing transformers for autoregression generation on less structured data, such as raw point clouds, has seldom been explored hitherto. The main challenge is that the sequentialization of such data is non-trivial. Naively arranging
a point cloud as a sequence of points will break shape structural information and is computationally infeasible. To resolve the limitation, similar to the grid-like patches applied to 2D images \cite{esser2021taming,razavi2019generating}, one can uniformly divide a point cloud into several groups and lay them out as a sequence. However, learning the sequential shape representation can be difficult since such shape compositions are entirely random.

In this paper, we resolve these issues and take the first step to employ transformers in 3D point cloud generation.
The key idea is to decompose a point cloud into a sequence of semantically meaningful shape compositions, which are further encoded by an autoregressive model for point cloud generation.
Specifically, we first learn a mapping function that maps each point cloud onto a shared canonical sphere primitive.
Through a canonical auto-encoder with a few self-supervised objectives, the mapping function ensures that corresponding parts (e.g., tails of two airplanes) from different instances overlap when mapped onto the canonical sphere, i.e., dense correspondences of different instances are established and are explicitly represented via a canonical sphere (see Fig.~\ref{fig:teaser} (a) middle).
Grouping is carried out on the canonical sphere to obtain the shape compositions (see Fig.~\ref{fig:teaser} (a) right).
Thanks to the correspondence constraint, each group on the canonical sphere essentially corresponds to the same semantic part on all point cloud instances.
As a result, each point cloud can be sequentialized into a set of shape compositions that are semantically aligned across different instances.
%
Finally, we train a vector-quantized autoencoder (VQVAE) using these sequentialized point cloud sequences, followed by learning a transformer that resolves the point cloud generation task. 

Thanks to the sequentialization of point clouds, we are able to learn an independent codebook for each shape composition, as opposed to the conventional VQVAE that uses a single, large codebook.
Such design leads to high utilization of codes, reduced number of codes, and the dimension of each code.
The learned VQVAE and transformer can be readily applied to point cloud reconstruction and unconditional generation, performing favorably against state-of-the-art approaches. 
Furthermore, our model can be easily extended to multi-modal shape completion as an application for conditional shape generation.
The main contributions of this work include:
\begin{itemize}
\item We propose a novel transformer-based autoregressive model for point cloud generation.
\item We introduce a canonical autoencoder and a self-supervised point grouping network to sequentialize point clouds into semantically aligned sequences of shape compositions.
%
\item We train a VQVAE with group-specific codebooks, followed by learning a transformer model using the sequentialized point clouds to resolve the task of point cloud generation.
\item Qualitative and quantitative comparisons demonstrate that our model can achieve state-of-the-art performance for point cloud auto-encoding and generation. We also extend our model to multi-modal shape completion as an application for conditional shape generation.
\end{itemize}

\section{Related work}

\subsection{3D Shape Generation}
3D shape generation targets at learning generative models on 3D shapes including but not limited to point clouds~\cite{yang2019pointflow}, voxels~\cite{wu2016learning}, implicit surfaces~\cite{Chan2021}, etc.
Some early works generate point clouds with a fixed-dimensional matrix~\cite{achlioptas2018learning,gadelha2018multiresolution}. Although these models can easily be plugged into existing generative models (e.g., \cite{gadelha2018multiresolution} a variational auto-encoder or ~\cite{achlioptas2018learning} a generative adversarial network), they are restricted to generating a fixed number of points and are not permutation invariant. Several works~\cite{yang2018foldingnet,groueix2018papier,li2021sp} mitigate this issue by mapping a primitive to a point cloud. Specifically, they attach a global latent code to each sampled point on the primitive and then apply the transformation to each concatenation. In this work, we also generate point clouds from a shared primitive. Different from existing works, we decompose the primitive into different compositions and represent each group as a local latent code. Our model can generate point clouds with more fine-grained details thanks to the local latent code representation.

Recently works consider point clouds as samples from a distribution and propose different probabilistic models to capture the distribution. For example, PointFlow (PF)~\cite{yang2019pointflow} applies normalizing flow to 3D point clouds. ShapeGF~\cite{cai2020learning} learns the gradient of the log-density field of shapes and generates point clouds using Langevin dynamics. DFM~\cite{luo2021diffusion} and PVD~\cite{zhou20213d} are both diffusion models that learn a probabilistic model over a denoising process on inputs. 

Most related to our work, PointGrow~\cite{sun2020pointgrow} and AutoSDF~\cite{mittal2022autosdf} also use autoregressive models to generate 3D shapes.
Specifically, the PointGrow discretizes point coordinates of a point cloud to fixed values and generate a shape in a point-wise manner following the spatial order. However, due to the large amount of points in each point cloud, the size of generated point clouds is limited. Instead, our model decomposes a point cloud into compact shape compositions that are both semantically meaningful and more efficient to process.
%
The AutoSDF learns an autoregressive model on volumetric Truncated-Signed Distance Field (T-SDF), where a 3D shape is represented as a randomly permuted sequence of latent variables, while the proposed method takes raw point clouds as inputs and decomposes them into ordered sequences of shape compositions.


\subsection{Transformers for Point Clouds}

Recently, more works start to apply transformers in model point clouds due to their impressive representation capacity.
For example, Zhao et al.~\cite{zhao2021point} introduce a point transformer layer using vector self-attention operations and show improvement in point cloud classification and segmentation. Nico et al.~\cite{engel2021point} propose to extract local and global features and relate both representations using a transformer-based model. Guo et al.~\cite{Guo_2021} use a transformer to capture local context within the point cloud and thereby enhance the input embedding. Xiang et al.~\cite{xiang2021snowflakenet} leverage a transformer architecture to extract shape context features with a focus on enhancing shape completion. Kim et al.~\cite{kim2021setvae} propose to encode a hierarchy of latent variables for flexible subset structures using a transformer. However, these works only employ transformer architectures on the encoding side to utilize its representation learning ability. In contrast, we use transformer architectures as the decoder and focus on the autoregressive generation process.

\section{Method}

We propose a framework that employs transformers in the task of point cloud generation.
The overview of our framework is illustrated in Fig.~\ref{fig:teaser}.
Given a point cloud, our method first maps it onto a canonical sphere in Section~\ref{subsec:mapping}.
By adopting few self-supervised training objectives, we ensure that the semantically corresponding points from different instances overlap on the canonical sphere.
Thus, by grouping points on the canonical sphere and serializing them as a sequence, we equivalently decompose each point cloud into an ordered sequence of shape compositions.
This process is described in Section~\ref{subsec:pcl_seq}.
We then learn a vector-quantized variational auto-encoder (VQVAE) using the sequentialized point clouds in Section~\ref{subsec:vqvae}, with codebooks as a library of the shape compositions in the point clouds.
Finally, a transformer is trained for point cloud generation in Section~\ref{subsec:transformer}.

\begin{figure}[t]
\centering
\includegraphics[width=1\linewidth]{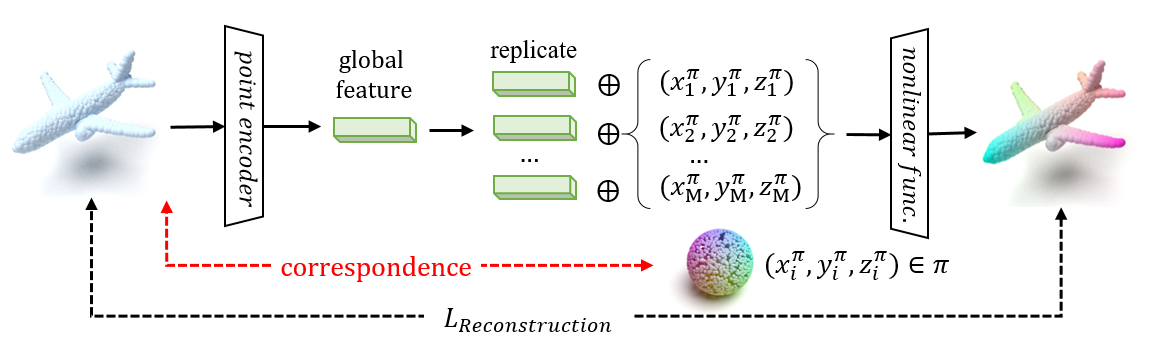}

\caption{Our Canonical Auto-encoder contains two parts: a point encoder that produces the shape feature of an input point cloud; a nonlinear function that decodes the canonical sphere, along with the shape feature, back to the input shape.}

\label{fig:cmf}
\vspace{-1.5em}
\end{figure}

\subsection{Point Cloud Sequentialization}
\label{subsec:mapping}
Different from convolutional neural networks, transformers model requires sequential data as inputs.
Taking the transformer in~\cite{esser2021taming} as an example,
an image is first sequentialized by starting from the top left patch and sequentially moving to the bottom right patch in a zigzag order.
The underline key is that all images are sequentialized by the same ``zigzag'' order, allowing the transformer to learn to predict patches based on their surrounding context.
However, when it comes to an orderless data structure such as point clouds, it remains challenging to sequentialize the unordered points in a similar way as images.

To resolve this issue, two key questions need to be answered: a) what forms an unit in a sequentialized point cloud? b) how to sort these units in a consistent order for different point clouds?
In this section, for the ease of understanding, we consider a single point as an unit in each point cloud and demonstrate how to sort these points in a consistent order for different point clouds. We discuss how to learn more semantically meaningful and memory friendly units (i.e., shape compositions) in the next section.

To sequentialize all point clouds in a consistent order, we first map them onto a shared canonical sphere. Each point on the sphere is from semantically corresponding points on different point clouds. Thus, by finding an order for points on the sphere, all the point clouds can be sequentialized accordingly.
%
%
For instance, if a point on the canonical sphere is labeled as the $k$th point in a sequence, then all its corresponding points in different point clouds are also labeled as the $k$th point in the sequentialized point cloud.
%
In the following, we first discuss how to map point clouds to a canonical sphere and then describe the order we choose to sort all points.

\subsubsection{Mapping point clouds to a canonical sphere}

We learn a nonlinear function to associate all point clouds with a canonical sphere $\pi$ and thus obtain their correspondences, as inspired by~\cite{cheng2021learning}. 
As shown in Fig.~\ref{fig:cmf}, given an input point cloud $x\in\mathcal{R}^{M\times3}$ including $M$ points, we first encode it as a 256-dimensional global latent code by an encoder, e.g., DGCNN~\cite{wang2019dynamic}. We then replicate the global code and concatenate it with points sampled from a canonical unit sphere as the input to the nonlinear function. At the output end of the function, we reconstruct the input point cloud via a Chamfer loss. The function thus performs a nonlinear transformation between the sphere and an individual instance, conditioned on its latent shape feature. We name the combination of the shape encoder and the nonlinear function as \textit{Canonical Auto-encoder}. For any point $x_i$ from an input point cloud $x$, to locate its corresponding point on the sphere, we can (1) search its nearest neighbor $\hat{x}_i$ on the reconstructed point cloud $\hat{x}$, then (2) trace the point $\pi_i$ on the sphere where $\hat{x}_i$ is mapped from.

As being proved by Cheng et al.~\cite{cheng2021learning}, points of all the reconstructed shapes are ``re-ordered'' according to the point indices of the canonical sphere (see Fig.~\ref{fig:cmf}). We note that as the key difference, we remove the ``point cloud-to-sphere mapping network'' designed in \cite{cheng2021learning} to avoid processing the points that are inaccurately mapped to locations far away from the sphere surface. Our design also simplifies the following process in Sec. \ref{subsec:pcl_seq}, i.e., the grouping and sequentialization can be conducted on a complete sphere instead of a subset of it \cite{cheng2021learning}. For brevity, in the following, we denote the canonical mapping process as $\Phi$, the corresponding point of $x_i\in x$ on the sphere $\pi$ as $\Phi_{x\rightarrow \pi}(x_i)$, and the corresponding point of $\pi_{i}\in \pi$ on $x$ as $\Phi^{-1}_{\pi\rightarrow x}(\pi_{i})$.

\subsubsection{Canonical sphere serialization}
\label{subsubsec:pcl_fib}
Since all point clouds are aligned with the canonical sphere by the canonical mapping function, any order defined on the canonical sphere can be easily transferred to any point cloud .
%
%
In this paper, we traverse the canonical sphere from the pole with a Fibonacci spiral (see Fig.~\ref{fig:teaser} (a)) and serialize the points in the spiral order along the way.
As a result, the index of a point in a point cloud can be easily determined as the index of its corresponding point on the canonical sphere.

\begin{figure}[t]
\centering
\includegraphics[width=1.0\linewidth]{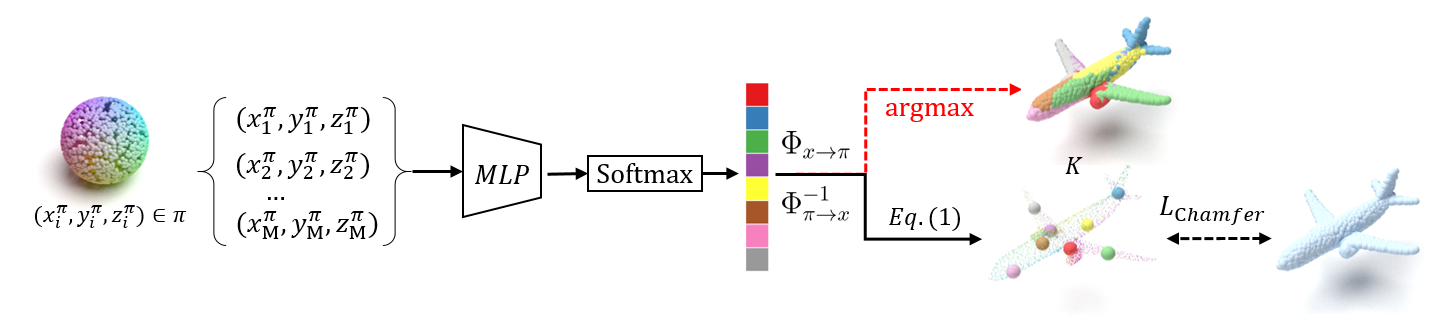}

\caption{We learn a self-supervised network to decompose the canonical sphere into non-overlapping groups. With canonical mapping, point clouds are simultaneously decomposed into semantically aligned shape compositions. See Sec.~\ref{subsec:pcl_seq}.}

\label{fig:cgf}
\vspace{-1.5em}
\end{figure}

\subsection{Shape Composition Learning}
\label{subsec:pcl_seq}

Though the re-ordered point clouds in Sec.~\ref{subsec:mapping} can be readily represented as sequences of points, such sequences usually include thousands of points and are intractable to be modeled by autoregressive models.
In this section, we introduce a more semantically meaningful and memory efficient unit for point cloud sequentialization.

Specifically, we decompose the points of each point cloud instance into $G$ groups ($G=128$ throughout all experiments). We call each group a shape composition, which is analogous to an image patch in the 2D domain.
As discussed above, since each point cloud is aligned to the canonical sphere, decomposing the point clouds is thus equivalent to decomposing the canonical sphere.
A straightforward way is to uniformly divide the sphere by randomly sampling $G$ points as center points and assigning each point to the nearest center point. 
However, this approach does not take the semantic prior into consideration and often produces discontinuous shape compositions. 

Instead, we introduce a self-supervised grouping network as shown in Fig.~\ref{fig:cgf}.
For each point on the sphere, we predict its group assignment by a multi-layer perceptron (MLP), followed by a SoftMax activation function.
Both are shared by all the points on the sphere.
This results in an assignment probability map $P$ for all points $q\in \pi$, where $P_i^j$ indicates the probability to assign point $\pi_i$ to the $j$th group.

To train this network and produce reasonable shape compositions, at the output end, we transfer the grouping assignment probability from each point $\pi_{i}\in \pi$ on the canonical sphere to its corresponding point $\Phi_{\pi\rightarrow x}^{-1}(\pi_{i})$ on the input point cloud.
%
%
As a result, we obtain an assignment probability map for each point cloud instance.
%
To ensure the learned grouping captures the structure of a point cloud and formulates a decent abstraction of it, we compute $G$ structure points $K\in\mathcal{R}^{G\times3}$~\cite{chen2020unsupervised}, where each $K_j$ is computed as:
%
\begin{equation}\label{eq:integration}
K_j = \sum_{i=1}^{m} \Phi^{-1}_{\pi\rightarrow x}(\pi_i) P_i^j  \quad \textrm{with} \quad \sum_{i=1}^{m} P_i^j=1 \quad \textrm{for}  \quad j=1,2,...,G
\end{equation}
Finally, a Chamfer distance is applied between the predicted structure points $K$ and the input point cloud $x$, as $\mathcal{L}_{CD}(K, x)$.
%

After training, we assign each point on $\pi$ to the group with the highest probability.
To assign each point $x_i\in x$ to a group, we simply let it take the group label of its corresponding point $\Phi_{x\rightarrow \pi}(x_i)$ on the sphere.
In different point clouds, points on corresponding semantic parts share the same grouping assignment through the canonical sphere.
As a result, any point cloud instance is decomposed into a set of shape compositions, each of which includes the points assigned to this group.

These shape compositions form the basic shape units and are further sorted into a sequence following the Fibonacci sipral order described in Sec.~\ref{subsubsec:pcl_fib}.
In the following sections, we still denote each point cloud as $x$ for brevity, but we assume that all point clouds have been processed into sequences of shape compositions using the method described above.

\subsection{Point Cloud Reconstruction through VQVAE}
\label{subsec:vqvae}
Now we introduce how to utilize the sequentialized point clouds to learn a VQVAE.
Our VQVAE includes three components, an encoder $E$, a decoder $D$, and a vector quantizer $Q$, as shown in Fig.~\ref{fig:teaser}(c).
We discuss each component in detail in the following.

\subsubsection{Point cloud encoding} Given a sequentialized instance $x$, we first compute the point-wise feature by the encoder $E$.
To compute the feature of each shape composition, we apply max-pooling to aggregate features of all points belonging to this shape composition.
We denote the feature of the $j$th group as $z^j$.

\subsubsection{Point cloud sequence quantization} Next, we quantize the group feature vectors $z$ by a group of jointly learned codebooks.
In conventional VQVAEs, a single codebook is learned and shared by all the compositions (e.g., image patches in 2D VQVAEs~\cite{van2017neural}).
%
However, we found that this strategy often leads to low code utilization.
%
The model struggles to capture the diverse feature of all groups via only a few codes while leaving all the others unused.
Such design leads to the usage of an unnecessarily large codebook and inferior reconstruction results. 


To resolve this issue, we learn an independent codebook for each group where at least one code from each codebook will be utilized. Since each codebook is only responsible for representing one particular shape composition, we can safely reduce the number of codes and the dimension of each code without degrading the performance. Specifically, given $z^j$ for group $j$, we first reduce its dimension from 256 to 4 to obtain a low dimensional feature $\hat{z}^j$ by learning a linear projection. We then quantize $\hat{z}^j$ into $z_q^{low}$ by finding its nearest neighbor token from the corresponding group codebook $Z^j$. Note that each group codebook $Z^j$ contains 50 4-dimensional latent codes. Finally, we project the matched codebook token back to the high-dimension embedding space and denote the quantized group feature as $z_q$. We note that the recent work~\cite{yu2021vector} also shows that this dimension reduction process improves the reconstruction quality. We show in Sec.~\ref{sec:exp:abl} that our design choices for codebook significantly increases codebook usage. 


\subsubsection{Point cloud sequence decoding}
To recover the input point cloud from $z_q$, we concatenate each point in the canonical sphere $\pi$ with the corresponding quantized group feature and feed the concatenation to the decoder $D$.

\subsubsection{VQVAE training} We use the same network architecture as in Sec.~\ref{subsec:mapping} for both $E$ and $D$.
We train them together with the codebooks by applying the Chamfer and Earth Mover Distance between the reconstructed point cloud $\hat{x}$ and the input point cloud $x$:
\begin{equation}
\begin{aligned}
\mathcal{L}_{Quantization}=\mathcal{L}_{CD}(x,\hat{x}) + \mathcal{L}_{EMD}(x,\hat{x}) & + \left \| sg[z_{q}^{low}]-\hat{z} \right \|_{2}^{2}
\end{aligned}
\end{equation}
where $sg[\cdot]$ is the stop-gradient operation. We use exponential moving average (EMA) \cite{cai21eman} to maintain the embeddings in each of the group codebooks $Z^j$.

\subsection{Point Cloud Generation through Transformers}
\label{subsec:transformer}
Given the learned codebooks, we can represent a point cloud sequence as a sequence of codebook token indices in order to learn an auto-regressive model. 
%
Specifically, we represent the codebook token indices as ${s_{1},s_{2},...,s_{G}}$, where $G$ is the total group number.
Given indices $s_{<i}$, we train a transformer model to predict the distribution of possible next indices $s_{i}$ based on its preceeding codebook tokens as:
\begin{equation}
\prod _{i=1}^{G}p(s_{i} | s_{1},s_{2},...,s_{i-1})
\end{equation}
The training objective is to minimize the negative log-likelihood by
\begin{equation}
\mathcal{L}_{Transformer}=\mathbb{E}_{x\mathtt{\sim}p(x))}\left [ -\log p(s) \right ]
\end{equation}
The architecture of our transformer model is similar as~\cite{esser2021taming}, where the indices are projected into the embedding space at each position together with an additive positional embedding. 
However, since each of our groups owns its own codebook, we do not use a shared embedding space for all codebook token indices. 
Instead, each index $s_i$ is mapped to the embedding space using a separate linear layer.

\subsubsection{Unconditional generation}
\label{subsubsec:transformer:uncond}
With the learned transformer model, unconditional shape generation is carried out by sampling token-by-token from the output distribution. 
The sampled tokens are then fed into the decoder $D$ in the VQVAE to decode output shapes.

\subsubsection{Conditional generation}
\label{subsubsec:transformer:cond}
Going beyond unconditional generation, we further incorporate our transformer with a conditional input.
Specifically, given a condition $c$ (e.g., a depth image), we use our transformer to generate a shape that matches the semantic meaning of $c$. 
For instance, if $c$ is a depth image, then the transformer is expected to generate a 3D shape that renders the depth image from the given viewpoint.
To this end, we first encode the condition $c$ into a feature vector in the same dimension of token embedding, then prepend the feature vector before the first token embedding.

\begin{table*}[t]
\centering
\caption{Shape auto-encoding on the ShapeNet dataset. The best results are highlighted in bold. 
CD is multiplied by $10^4$ and EMD is multiplied by $10^2$.}
\label{tab:rec}
\footnotesize
\begin{tabularx}{\textwidth}{l*{9}{Y}}
\toprule

                          &           & \multicolumn{2}{c}{AtlasNet} & \multirow{2}{*}{PF} & \multirow{2}{*}{ShapeGF} & \multirow{2}{*}{DPM} & \multirow{2}{*}{Ours} & \multirow{2}{*}{Oracle} \\ 
\cmidrule(lr){3-4} 
Dataset                   & Metric    & Sphere       & Patches       &                            &                       &      &    &                   \\ \midrule
\multirow{2}{*}{Airplane} & CD      & 1.002     & 0.969     & 1.208     & 0.966  & 0.997     & \textbf{0.889} & 0.837 \\
                          & EMD     & 2.672     & 2.612     & 2.757     & 2.562  & 2.227     & \textbf{2.122} & 2.062 \\
                          \cmidrule{1-9}
\multirow{2}{*}{Chair}    & CD      & 6.564     & 6.693     & 10.120    & \textbf{5.599}  & 7.305     & 6.177  & 3.201\\
                          & EMD     & 5.790     & 5.509     & 6.434     & 4.917  & 4.509     & \textbf{4.218} & 3.297 \\
                          \cmidrule{1-9}
\multirow{2}{*}{Car}     & CD       & 5.392     & 5.441     & 6.531     & 5.328  & 5.749     & \textbf{5.050} & 3.904 \\
                         & EMD      & 4.587     & 4.570     & 5.138     & 4.409  & 4.141     & \textbf{3.614} & 3.251 \\
\bottomrule

\end{tabularx}
\vspace{-1.5em}
\end{table*}
\begin{figure}[H]
    \begin{center}
    \newcommand{\sizea}{0.19\linewidth}
    \newcommand{\sizeb}{0.14\linewidth}
    \newcommand{\tare}{4cm}
    \newcommand{\tale}{4.5cm}
    \newcommand{\tal}{4cm}
    \newcommand{\tab}{3.5cm}
    \newcommand{\tar}{4cm}
    \newcommand{\tat}{5.5cm}
    \newcommand{\tcl}{3.0cm}
    \newcommand{\tcb}{3cm}
    \newcommand{\tcr}{4cm}
    \newcommand{\tct}{4.2cm}
    \newcommand{\thl}{3.0cm}
    \newcommand{\thb}{0.0cm}
    \newcommand{\thr}{3cm}
    \newcommand{\tht}{2cm}
    \setlength{\tabcolsep}{-1pt}
    \renewcommand{\arraystretch}{0}
    \begin{tabular}{@{}cccc:cccc@{}}
        \includegraphics[width=\sizea, trim={\tale} {\tab} {\tare} {\tat},clip]{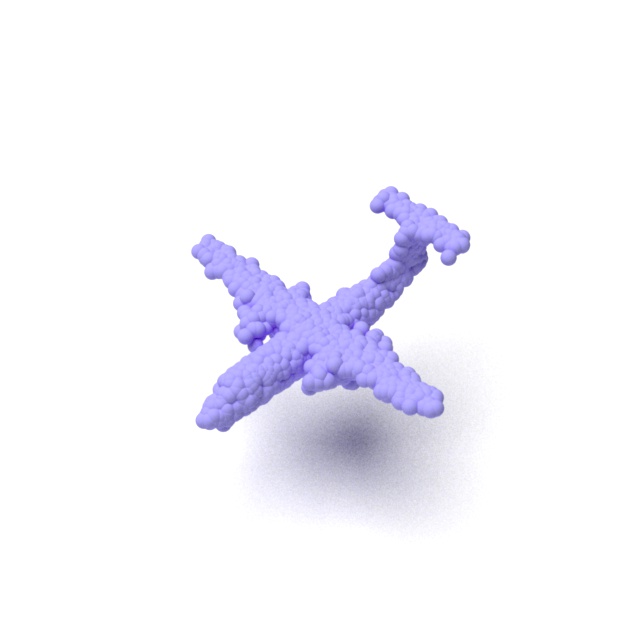}&
        \includegraphics[width=\sizea, trim={\tale} {\tab} {\tare} {\tat},clip]{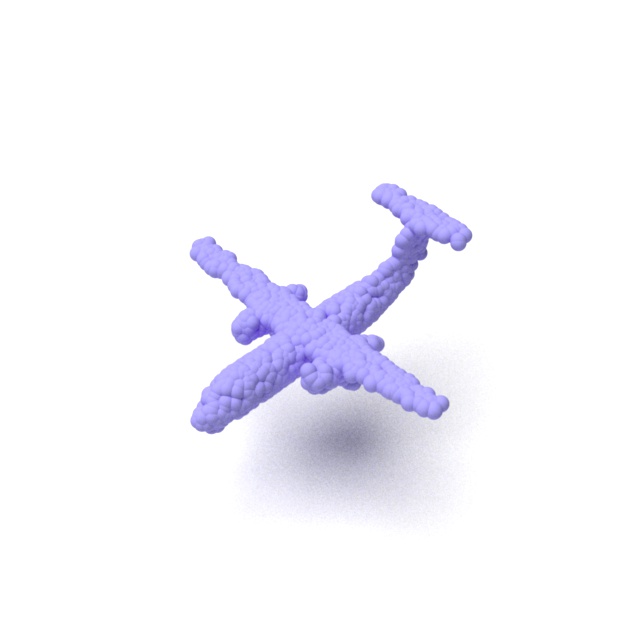}&
        \includegraphics[width=\sizea, trim={\tale} {\tab} {\tare} {\tat},clip]{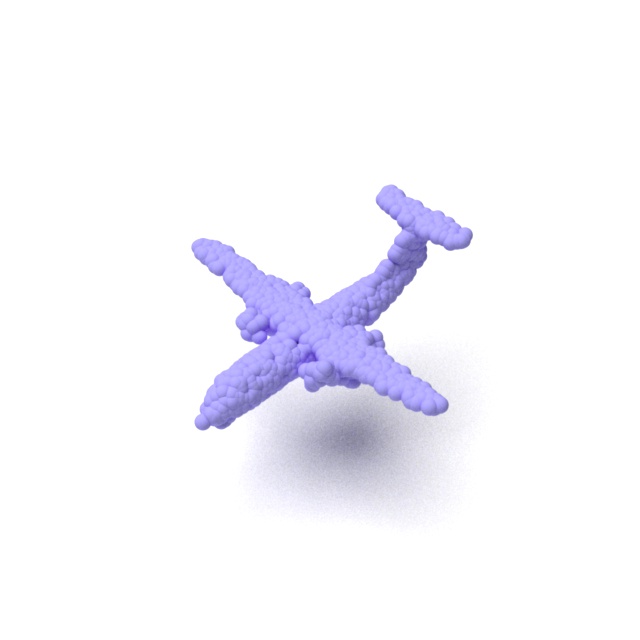}&
        \includegraphics[width=\sizea, trim={\tale} {\tab} {\tare} {\tat},clip]{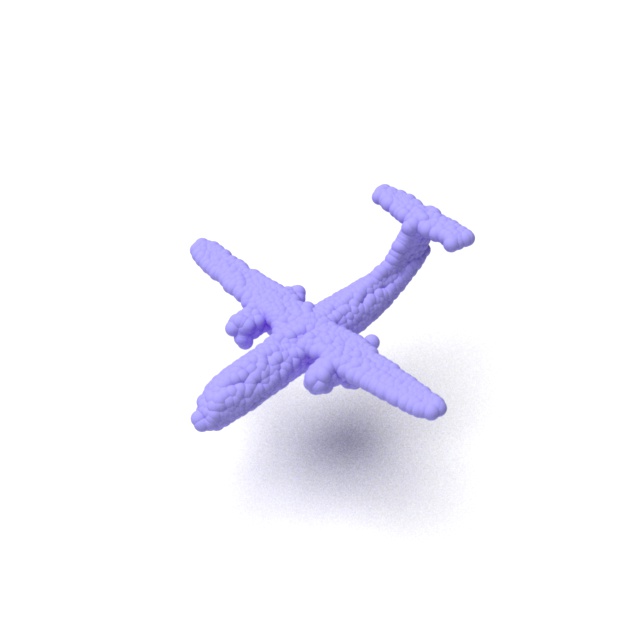}&
        \includegraphics[width=\sizea, trim={\tale} {\tab} {\tare} {\tat},clip]{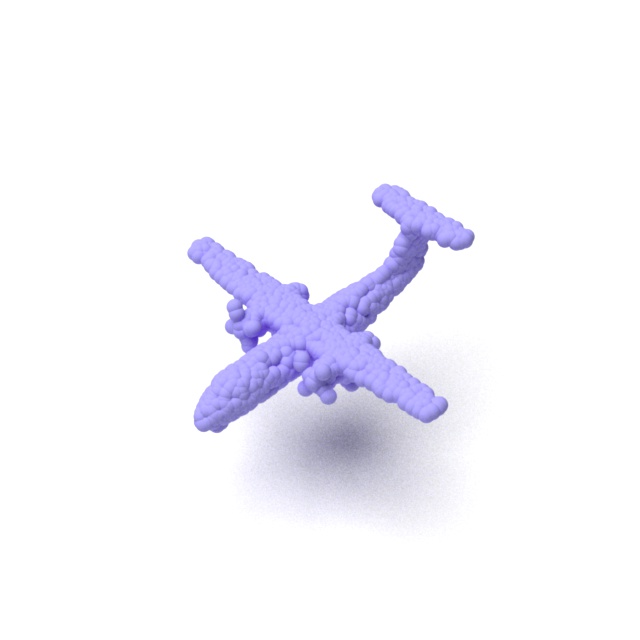}
        \\
        \includegraphics[width=\sizea, trim={\tale} {\tab} {\tare} {\tat},clip]{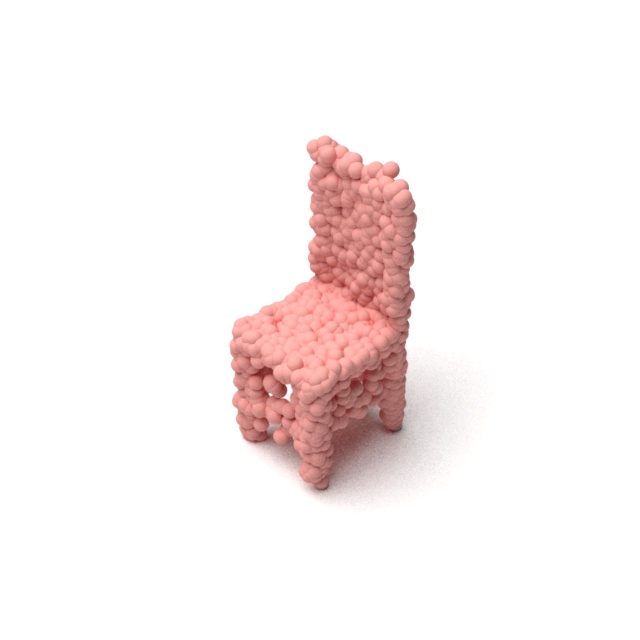}&
        \includegraphics[width=\sizea, trim={\tale} {\tab} {\tare} {\tat},clip]{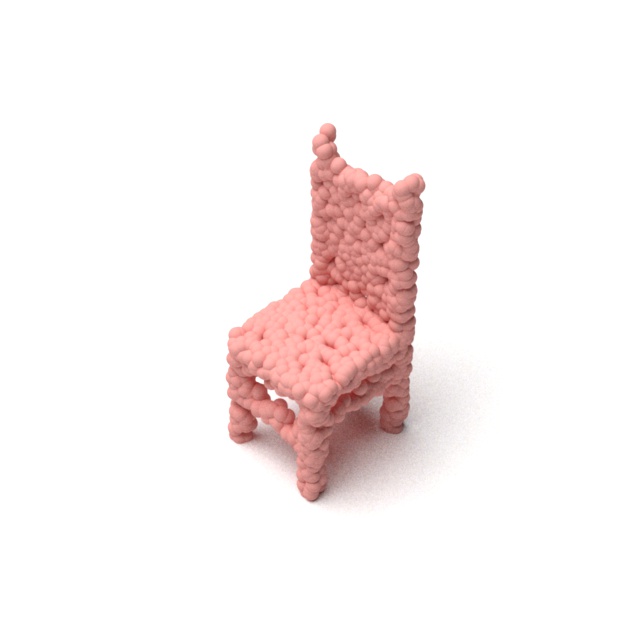}&
        \includegraphics[width=\sizea, trim={\tale} {\tab} {\tare} {\tat},clip]{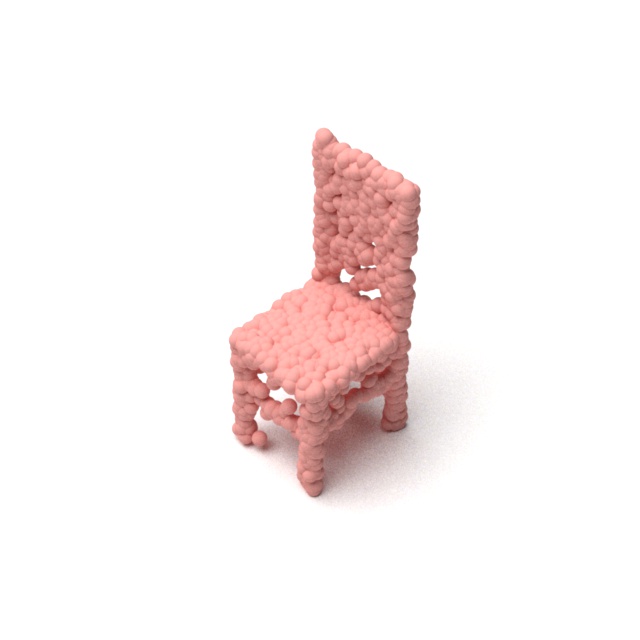}&
        \includegraphics[width=\sizea, trim={\tale} {\tab} {\tare} {\tat},clip]{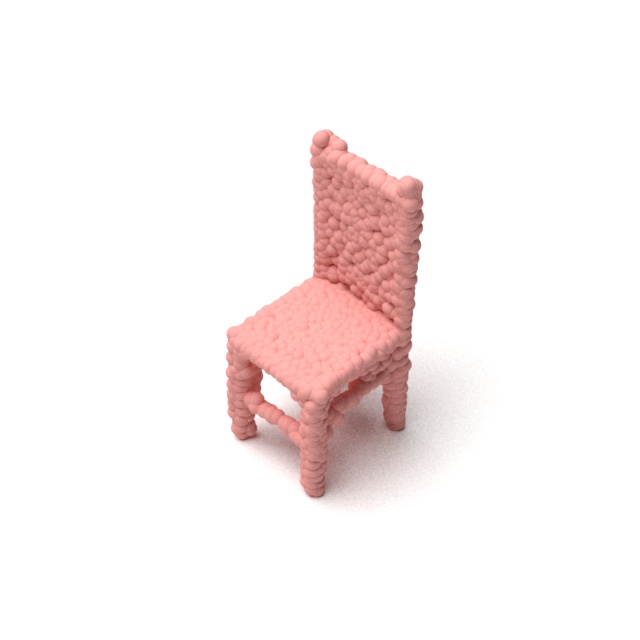}&
        \includegraphics[width=\sizea, trim={\tale} {\tab} {\tare} {\tat},clip]{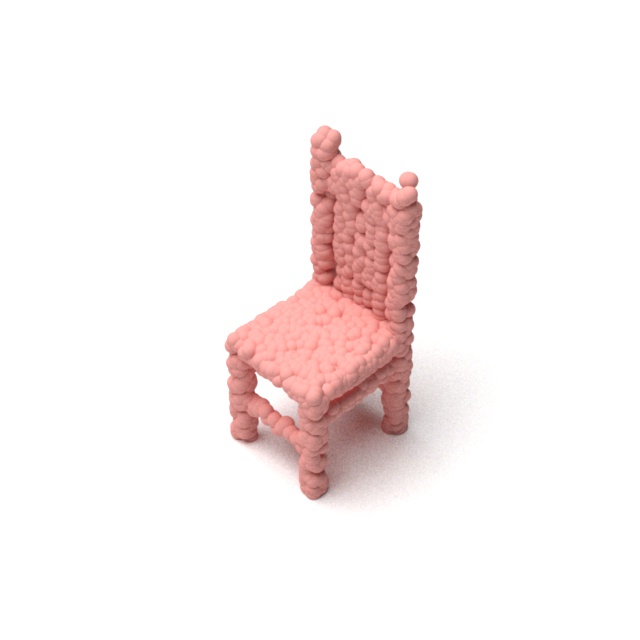}
        \\
        \includegraphics[width=\sizea, trim={\tale} {\tab} {\tare} {\tat},clip]{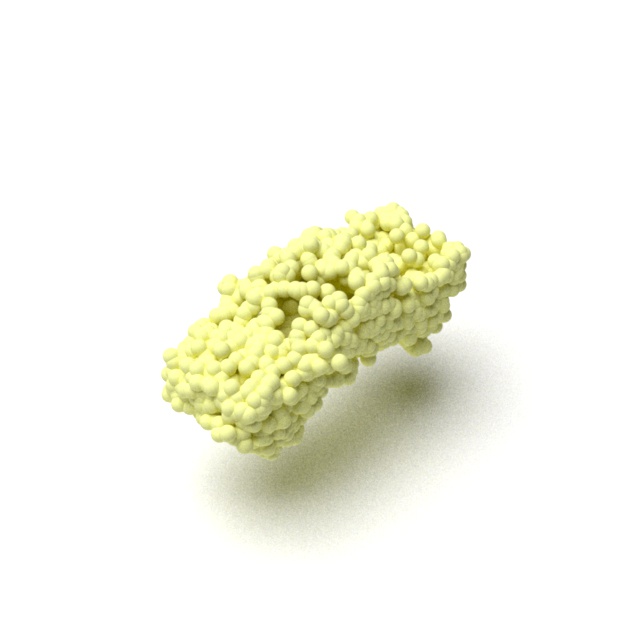}&
        \includegraphics[width=\sizea, trim={\tale} {\tab} {\tare} {\tat},clip]{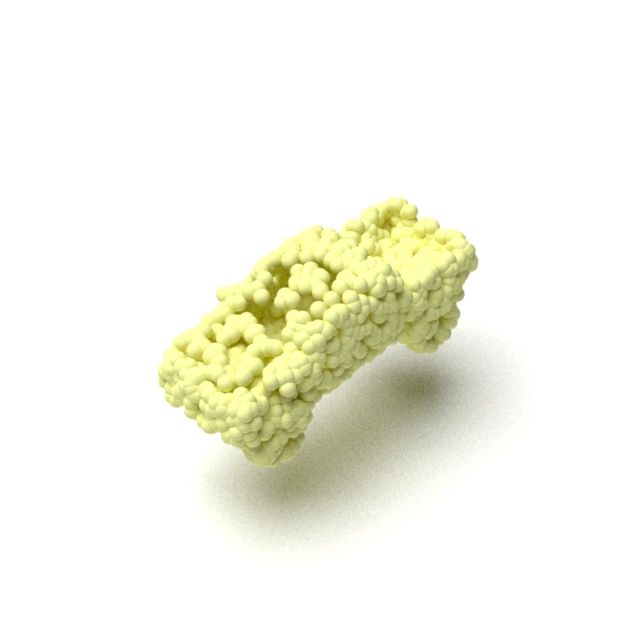}&
        \includegraphics[width=\sizea, trim={\tale} {\tab} {\tare} {\tat},clip]{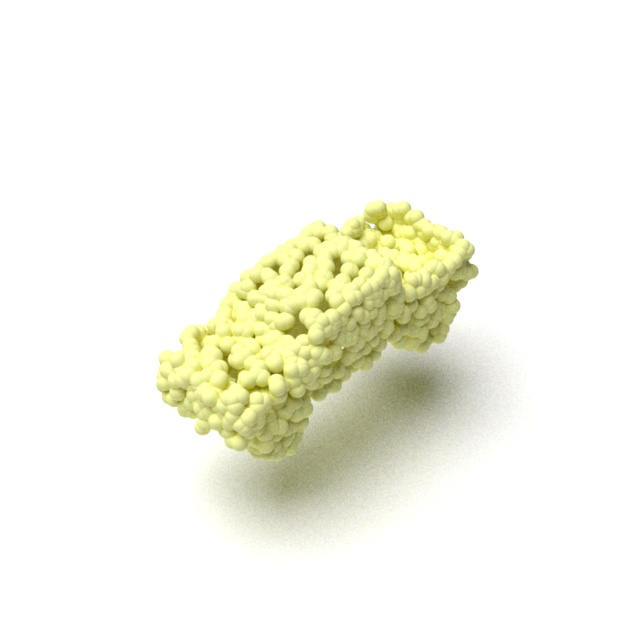}&
        \includegraphics[width=\sizea, trim={\tale} {\tab} {\tare} {\tat},clip]{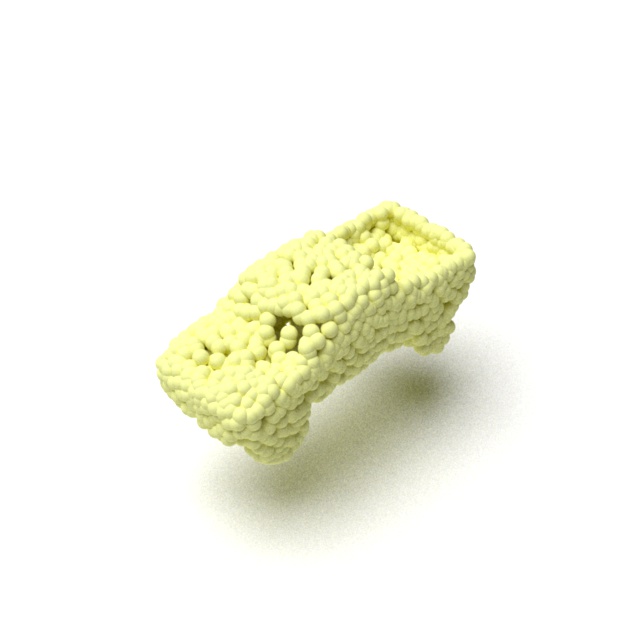}&
        \includegraphics[width=\sizea, trim={\tale} {\tab} {\tare} {\tat},clip]{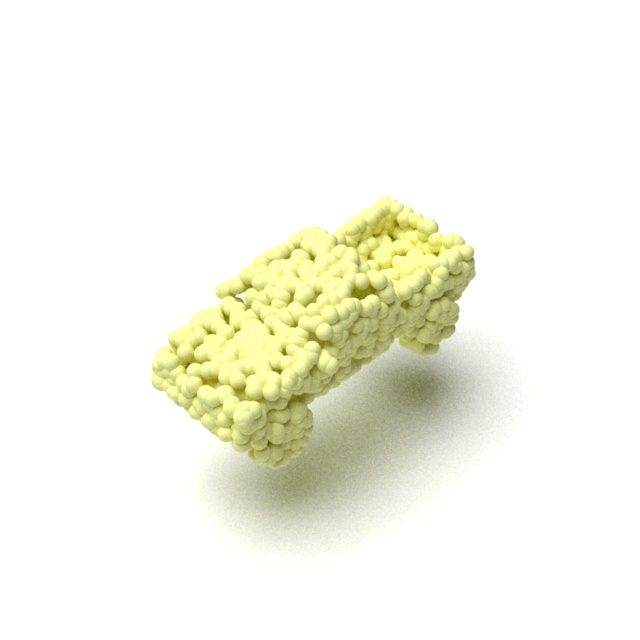}
        \\
        PF & ShapeGF & DPM & Ours & Input
    \end{tabular}

    \end{center}
    \caption{Auto-encoding (reconstruction) results. We also shown results from PF (PointFlow)~\cite{yang2019pointflow}, ShapeGF~\cite{cai2020learning}, and DPM~\cite{luo2021diffusion} on the left for comparison.
    }
    \label{fig:recon}
    \vspace{-1.5em}
\end{figure}

\section{Experiments}
In this section, we first evaluate our model’s performance by point cloud auto-encoding (Sec.~\ref{sec:exp:shape-quantization}), unconditional generation (Sec.~\ref{sec:exp:shape-generaton}) and conditional generation (Sec.~\ref{sec:exp:shape-completion}). Then, we present ablation studies in Sec.~\ref{sec:exp:abl}.

\paragraph{Datasets.}
Following previous works~\cite{yang2019pointflow,cai2020learning,kim2021setvae}, we conduct our auto-encoding and generation experiments on the airplane, chair, and car category from the ShapeNet~\cite{chang2015shapenet} dataset. We use data provided by~\cite{yang2019pointflow}, which includes 15,000 points sampled uniformly from the mesh surface for each object. We sub-sample 2048 points for training and testing, respectively. To train our model, we normalize all point clouds to a unit sphere as in~\cite{yang2018foldingnet}. For the multi-modal shape completion task, we follow~\cite{zhou20213d} which uses the ShapeNet rendering data from Genre~\cite{genre}. Genre contains 20 random views with ground-truth depth maps for each object in the ShapeNet. For baselines that take additional partial point clouds as inputs, we use the data provided by~\cite{zhou20213d}.

\paragraph{Evaluation metrics.}
For fair comparison, we follow prior works~\cite{yang2018foldingnet,yang2019pointflow,cai2020learning} and use the symmetric Chamfer Distance (CD) as well as the Earth Mover’s Distance (EMD) to evaluate the quality of the reconstructed point clouds. 
To evaluate the quality of the unconditionally generated point clouds, we use the Minimum Matching Distance (MMD)~\cite{achlioptas2018learning}, the Coverage Score (COV)~\cite{achlioptas2018learning}, and the 1-NN classifier accuracy (1-NNA)~\cite{yang2019pointflow}. 
To evaluate the multi-modal shape completion performance for the conditional generation task, we follow Wu et al.~\cite{wu2020multimodal} that uses a) the Total Mutual Difference (TMD) to measure the generation diversity and b) the Minimal Matching Distance (MMD) to measure the completion quality with Chamfer Distance (CD) as distance measure. We provide details about each of the metrics in the Appendix. For all generation experiments, we normalize each point cloud to a unit sphere before measuring the metrics. This is to ensure the metric focuses more on the geometry of the shape rather than on the scale.



\subsection{Shape Auto-encoding}\label{sec:exp:shape-quantization}
We first evaluate how well our model can approximate a shape with quantized features. We quantitatively compare our results  against the following state-of-the-art point cloud auto-encoders: AtlasNet~\cite{groueix2018papier} variants that deform from patches and from sphere, respectively, PointFlow (PF)~\cite{yang2019pointflow}, ShapeGF~\cite{cai2020learning}, and DPM~\cite{luo2021diffusion}. We also report the lower bound of the reconstruction errors in the “Oracle” column. This bound is obtained by measuring the reconstruction error between two different point clouds with the same number of points sampled from the identical underlying meshes. As shown in Table~\ref{tab:rec}, our method consistently outperforms other methods when measured by EMD. We note that EMD is usually considered a better metric for measuring a shape’s visual quality~\cite{zhou20213d} because it forces model outputs to have the same density as the ground-truth shapes ~\cite{liu2020morphing}. This suggests that our reconstructed point clouds have points that are more uniformly distributed on the surface. We also provide qualitative results compared to baselines in Figure~\ref{fig:recon} to validate the effectiveness of our model.

\subsection{Unconditional Generation}\label{sec:exp:shape-generaton}
\begin{table*}[t]
\centering
\caption{Shape generation results. $\uparrow$ means the higher the better, $\downarrow$ means the lower the better. 
MMD-CD is multiplied by $10^3$ and MMD-EMD is multiplied by $10^2$.}
\label{tab:generation}
\footnotesize
\begin{tabularx}{\textwidth}{LL*{2}{Y}l*{2}{Y}l*{2}{Y}}
\toprule
                          &           & \multicolumn{2}{c}{MMD ($\downarrow$)} &  & \multicolumn{2}{c}{COV (\%, $\uparrow$)} &  & \multicolumn{2}{c}{1-NNA (\%, $\downarrow$)} \\ \cmidrule(lr){3-4} \cmidrule(lr){6-7} \cmidrule(l){9-10} 
Category                  & Model     & CD         & EMD        &  & CD         & EMD       &  & CD     & EMD         \\ \midrule
\multirow{4}{*}{Airplane} & PointGrow
                          & 3.07            & 11.64         &  & 10.62          & 10.62     &  & 99.38  & 99.38 \\
                          & ShapeGF             
                          & \textcolor{black}{\textbf{1.02}}            & 6.53           &  & \textcolor{black}{\textbf{41.48}}         & 32.84     &  & 80.62  & 88.02         \\
                          & SP-GAN 
                          & 1.49            & 8.03           &  & 30.12         & 23.21     &  & 96.79  & 98.40 \\
                          & PF 
                          & 1.15            & 6.31          &  & 36.30          & 38.02     &  & 85.80  & 83.09 \\
                          & SetVAE 
                          & 1.04            & \textcolor{black}{\textbf{6.16}}          &  & 39.51          & 38.77     &  & 89.51  & 87.65 \\
                          & DPM
                          & 1.10            & 7.11          &  & 36.79          & 25.19     &  & 86.67  & 90.49 \\
                          & PVD
                          & 1.12            & 6.17    &  & 40.49    & \underline{\textbf{45.68}}  &  & \textcolor{black}{\textbf{80.25}}  & \textcolor{black}{\textbf{77.65}} \\

                          & Ours            & \underline{\textbf{0.83}}  & \underline{\textbf{5.50}}  &  & \underline{\textbf{45.67}}   & \textcolor{black}{\textbf{44.19}} &  & \underline{\textbf{63.45}}  & \underline{\textbf{71.60}} \\
                          \cmidrule(l{-1pt}r{0pt}){2-10} 
                          & Train           & 0.97           & 5.80         &  & 45.68          & 46.67          &  & 71.36          & 69.63          \\\midrule
\multirow{4}{*}{Chair}    & PointGrow
                          & 16.23           & 18.83         &  & 12.08          & 13.75     &  & 98.05  & 99.10 \\
                          & ShapeGF             
                          & \textcolor{black}{\textbf{7.17}}            & \textcolor{black}{\textbf{11.85}}         &  & \textcolor{black}{\textbf{45.62}}         & 44.71     &  & 61.78  & 64.27         \\
                          & SP-GAN 
                          & 8.51            & 13.09         &  & 34.74         & 26.28     &  & 77.87  & 84.29 \\
                          & PF 
                          & 7.26            & 12.12         &  & 42.60          & \textcolor{black}{\textbf{45.47}}     &  & 65.56  & 65.79 \\
                          & SetVAE 
                          & 7.60            & 12.10          &  & 42.75          & 40.48     &  & 65.79  & 70.39 \\
                          & DPM
                          & \underline{\textbf{6.81}}            & 11.91          &  & 43.35          & 42.75     &  & 64.65  & 69.26 \\
                          & PVD
                          & 7.65            & 11.87          &  & \underline{\textbf{45.77}}         & 45.02     &  & \underline{\textbf{60.05}}  & \underline{\textbf{59.52}} \\

                          & Ours            & 7.37  & \underline{\textbf{11.75}}  &  & \underline{\textbf{45.77}}          & \underline{\textbf{46.07}} &  & \textcolor{black}{\textbf{60.12}}  & \textcolor{black}{\textbf{61.93}} \\
                          \cmidrule(l{-1pt}r{0pt}){2-10} 
                          & Train           & 7.636           & 11.88         &  & 48.34          & 49.09          &  & 55.89          & 57.48          \\\midrule
\multirow{4}{*}{Car}     & PointGrow
                          & 14.12            & 18.33         &  & 6.82          & 11.65     &  & 99.86  & 98.01 \\
                          & ShapeGF             
                          & \textcolor{black}{\textbf{3.63}}            & 9.11          &  & \underline{\textbf{48.30}}         & 44.03     &  & \textcolor{black}{\textbf{60.09}}  & 61.36         \\
                          & PF 
                          & 3.69            & \textcolor{black}{\textbf{9.03}}          &  & \textcolor{black}{\textbf{44.32}}          & \textcolor{black}{\textbf{45.17}}     &  & 63.78  & \underline{\textbf{57.67}} \\
                          & SetVAE 
                          & \textcolor{black}{\textbf{3.63}}            & 9.05           &  & 39.77          & 37.22     &  & 65.91  & 67.61 \\
                          & DPM
                          & 3.70            & 9.39          &  & 38.07          & 30.40     &  & 74.01  & 73.15 \\
                          & PVD
                          & 3.74            & 9.31          &  & 43.47          & 39.49     &  & 65.62  & 63.35 \\

                          & Ours            & \underline{\textbf{3.31}}  & \underline{\textbf{8.89}}  &  & 41.76          & \underline{\textbf{47.72}} &  & \underline{\textbf{55.68}}  & \textcolor{black}{\textbf{57.81}} \\
                          \cmidrule(l{-1pt}r{0pt}){2-10} 
                          & Train           & 3.74           & 9.38         &  & 53.12          & 47.16          &  & 52.70          & 54.26          \\\bottomrule

\end{tabularx}
\vspace{-1.5em}
\end{table*}
\begin{figure}[t]
    \begin{center}
    \newcommand{\sizea}{0.124\linewidth}
    \newcommand{\sizeb}{0.10\linewidth}
    \newcommand{\sizec}{0.08\linewidth}
    \newcommand{\tare}{5cm}
    \newcommand{\tale}{4cm}
    \newcommand{\tal}{3.5cm}
    \newcommand{\tab}{3cm}
    \newcommand{\tar}{3.5cm}
    \newcommand{\tat}{3.5cm}
    \newcommand{\tcl}{3.0cm}
    \newcommand{\tcb}{3cm}
    \newcommand{\tcr}{4cm}
    \newcommand{\tct}{4.2cm}
    \newcommand{\thl}{3.0cm}
    \newcommand{\thb}{0.0cm}
    \newcommand{\thr}{3cm}
    \newcommand{\tht}{2cm}
    \setlength{\tabcolsep}{0pt}
    \renewcommand{\arraystretch}{0}
    \begin{tabular}{@{}ccccc:ccc@{}}
        \includegraphics[width=\sizea, trim={\tale} {\tab} {4.5cm} {\tat},clip]{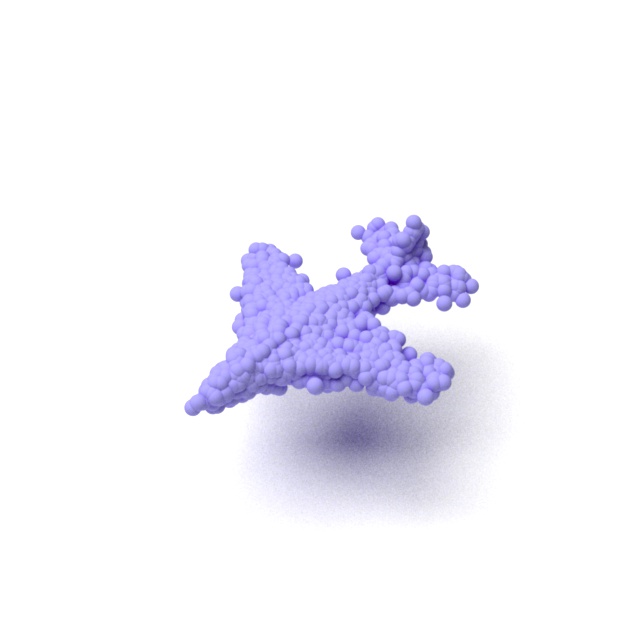}&
        \includegraphics[width=\sizea, trim={\tale} {\tab} {4.5cm} {\tat},clip]{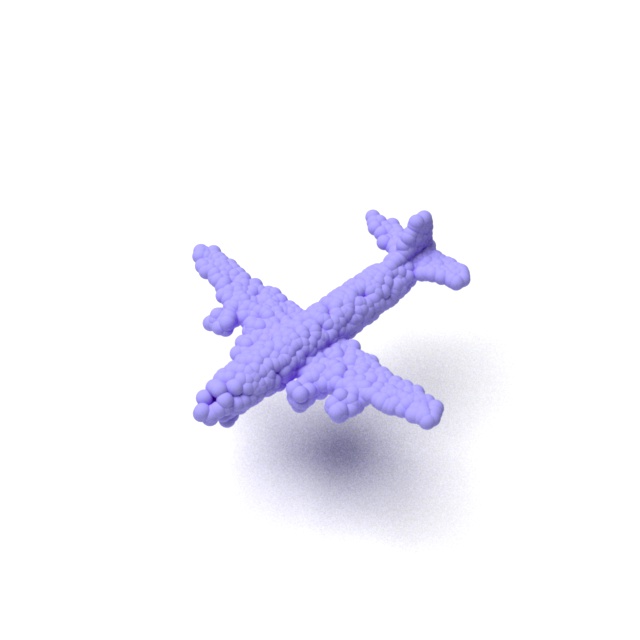}&
        \includegraphics[width=\sizea, trim={\tale} {\tab} {4.5cm} {\tat},clip]{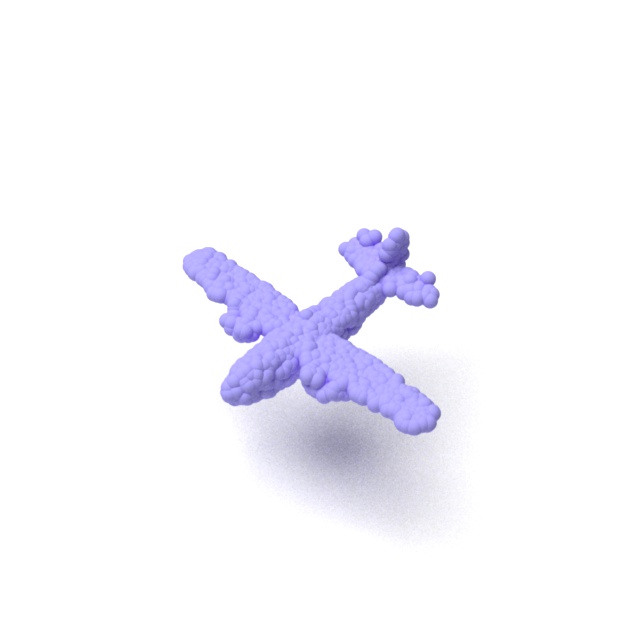}&
        \includegraphics[width=\sizea, trim={\tale} {\tab} {4.5cm} {\tat},clip]{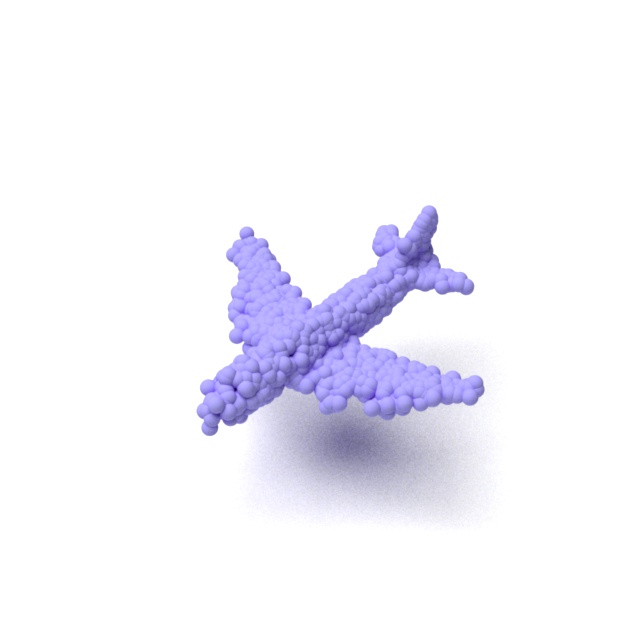}&
        \includegraphics[width=\sizea, trim={\tale} {\tab} {4.5cm} {\tat},clip]{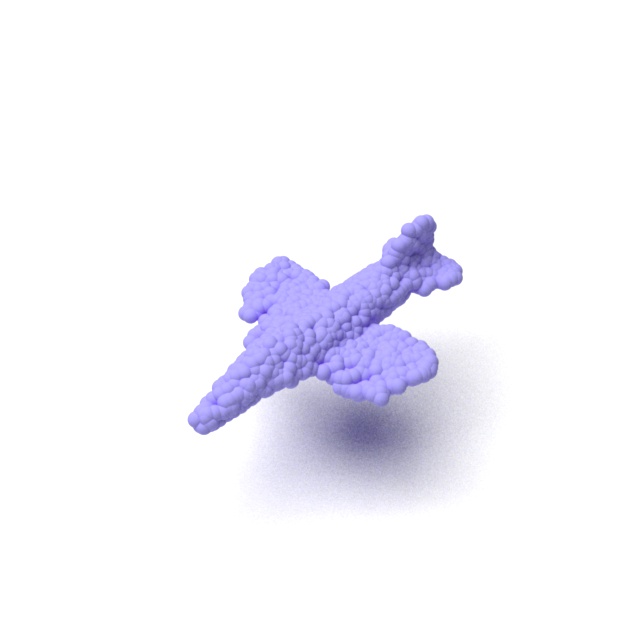}& 
        \includegraphics[width=\sizea, trim={\tal} {\tab} {4.5cm} {\tat},clip]{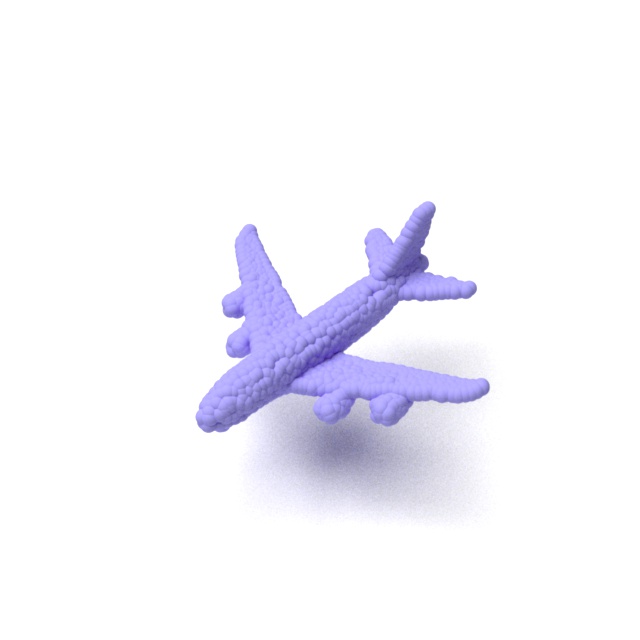}& 
        \includegraphics[width=\sizea, trim={\tal} {\tab} {\tar} {\tat},clip]{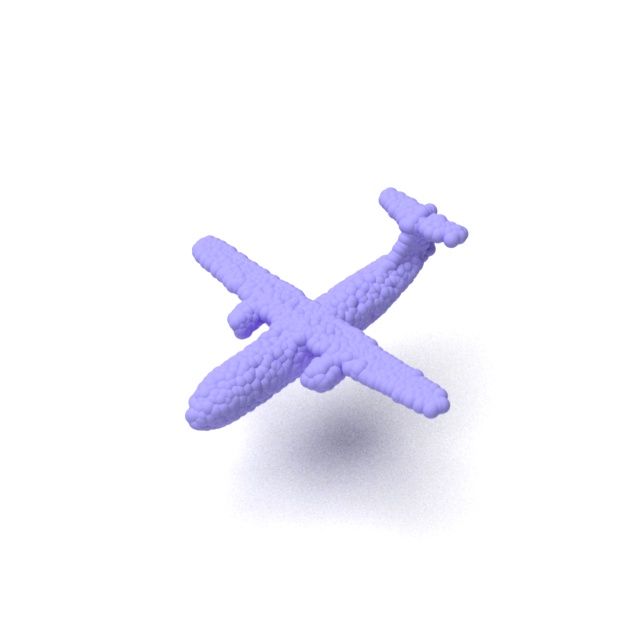}&
        \includegraphics[width=\sizea, trim={\tal} {\tab} {\tar} {\tat},clip]{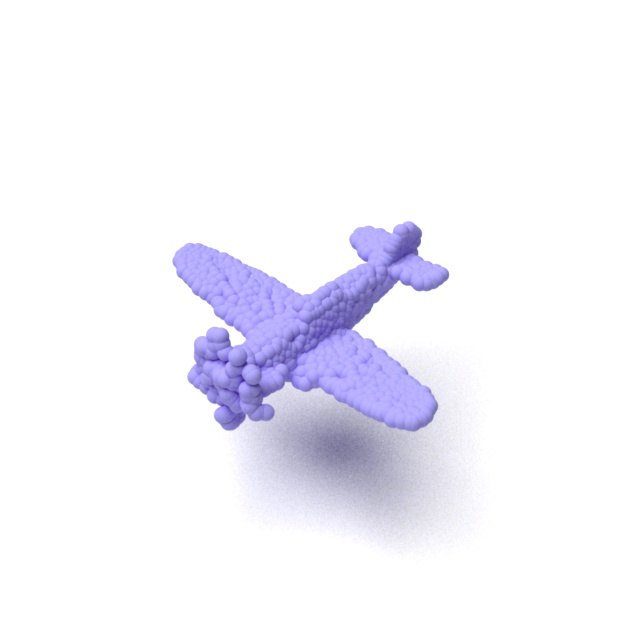}
        \\
        \includegraphics[width=\sizeb, trim={\tale} {\tab} {4.5cm} {\tat},clip]{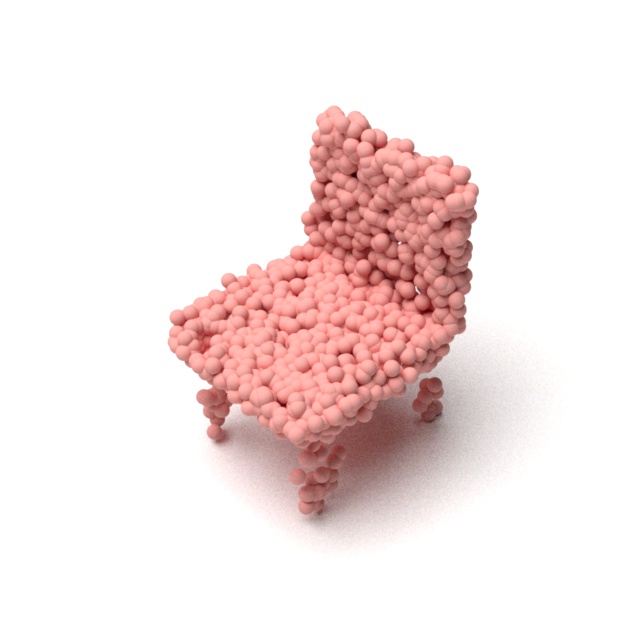}&
        \includegraphics[width=\sizeb, trim={\tale} {\tab} {4.5cm} {\tat},clip]{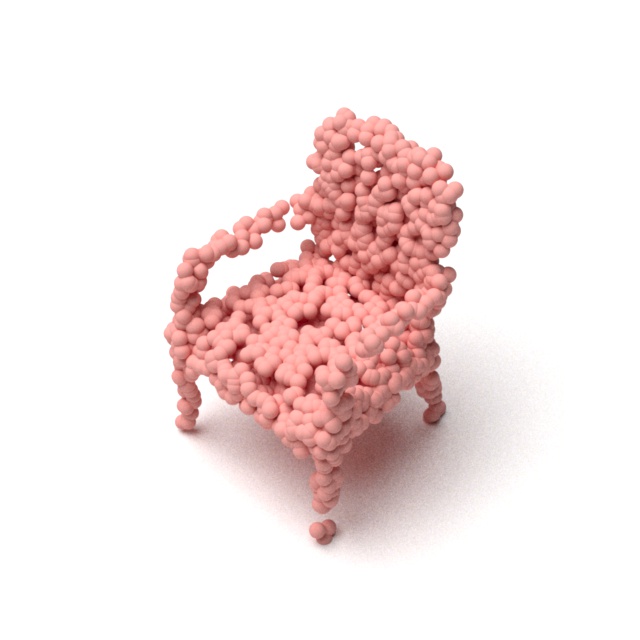}&
        \includegraphics[width=\sizeb, trim={\tale} {\tab} {4.5cm} {\tat},clip]{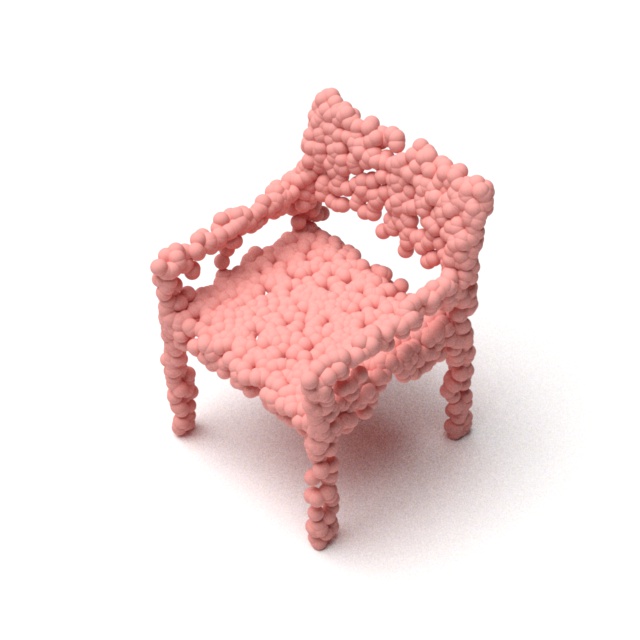}&
        \includegraphics[width=\sizeb, trim={\tale} {\tab} {4.5cm} {\tat},clip]{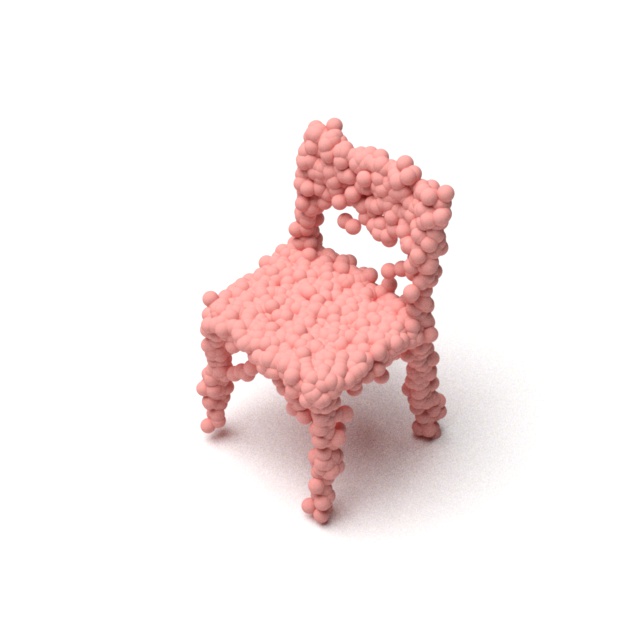}&
        \includegraphics[width=\sizeb, trim={\tale} {\tab} {4.5cm} {\tat},clip]{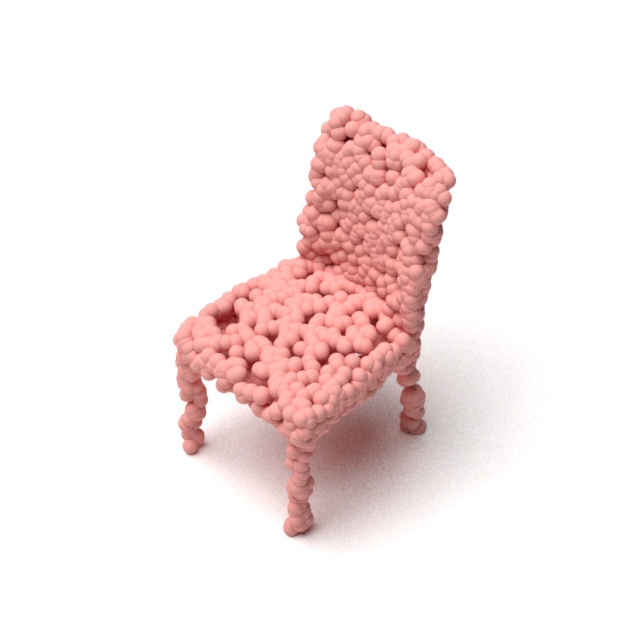}& 
        \includegraphics[width=\sizeb, trim={\tal} {\tab} {\tar} {\tat},clip]{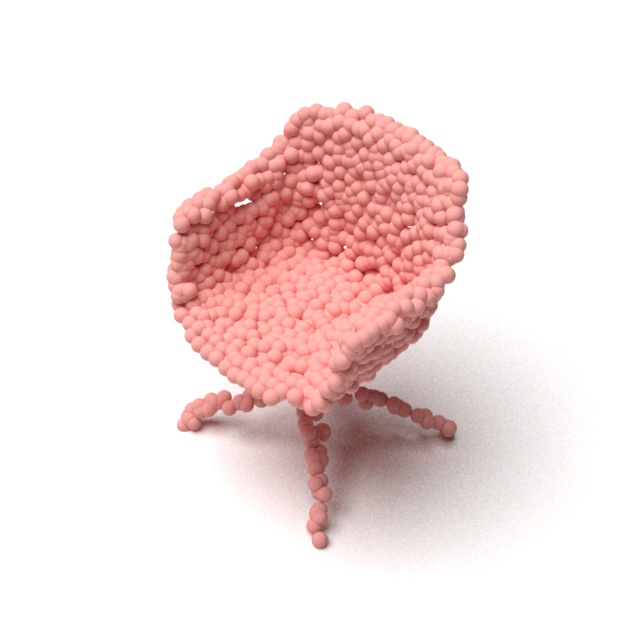}& 
        \includegraphics[width=\sizeb, trim={\tal} {\tab} {\tar} {\tat},clip]{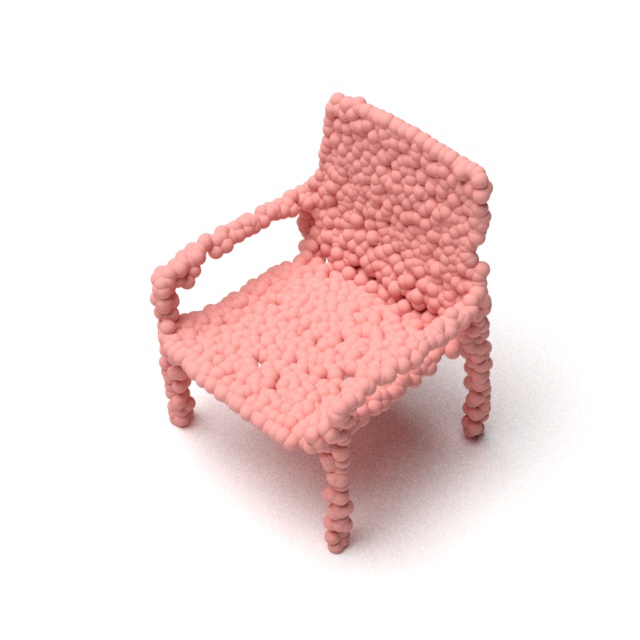}&
        \includegraphics[width=\sizeb, trim={\tal} {\tab} {\tar} {\tat},clip]{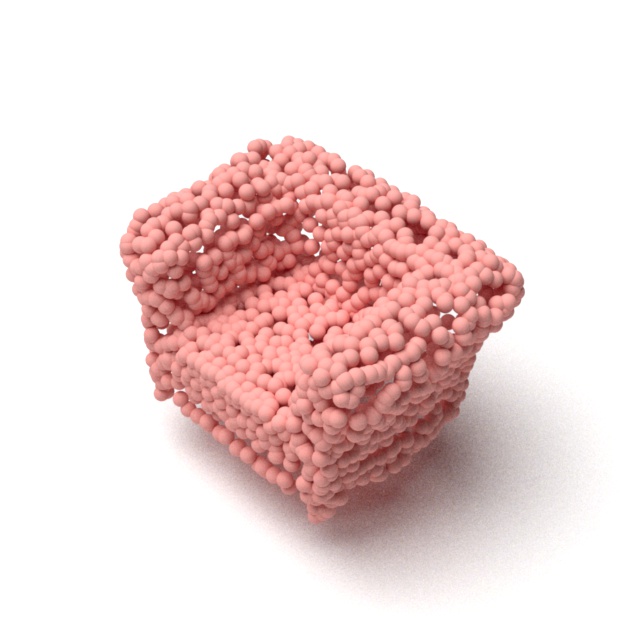}
        \\
        \includegraphics[width=\sizea, trim={\tale} {\tab} {4.5cm} {\tat},clip]{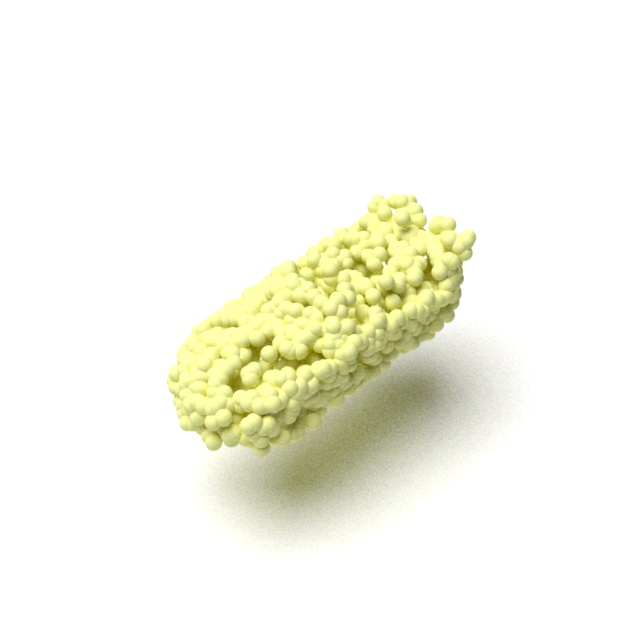}&
        \includegraphics[width=\sizea, trim={\tale} {\tab} {4.5cm} {\tat},clip]{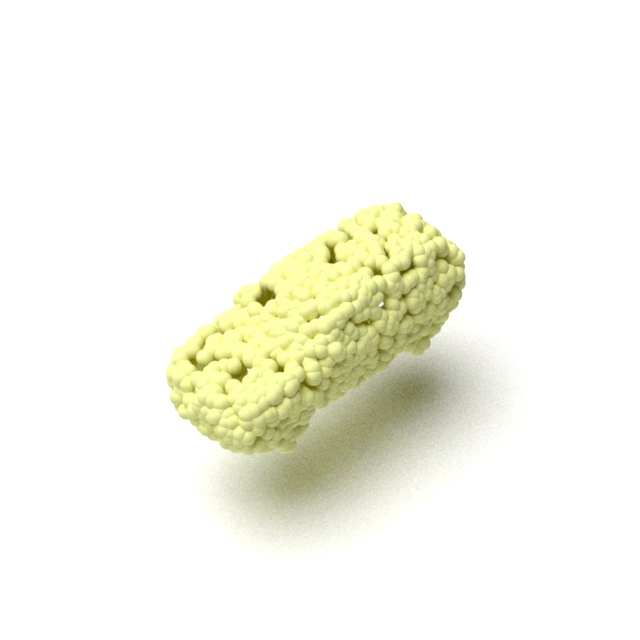}&
        \includegraphics[width=\sizea, trim={\tale} {\tab} {4.5cm} {\tat},clip]{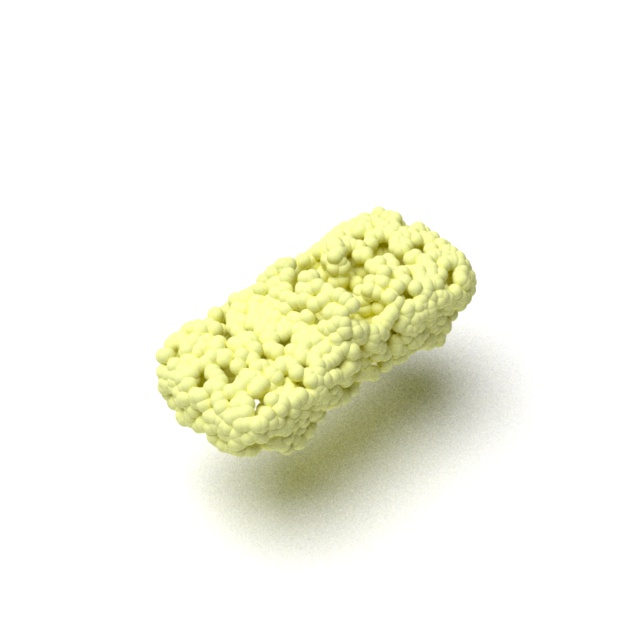}&
        \includegraphics[width=\sizea, trim={\tale} {\tab} {4.5cm} {\tat},clip]{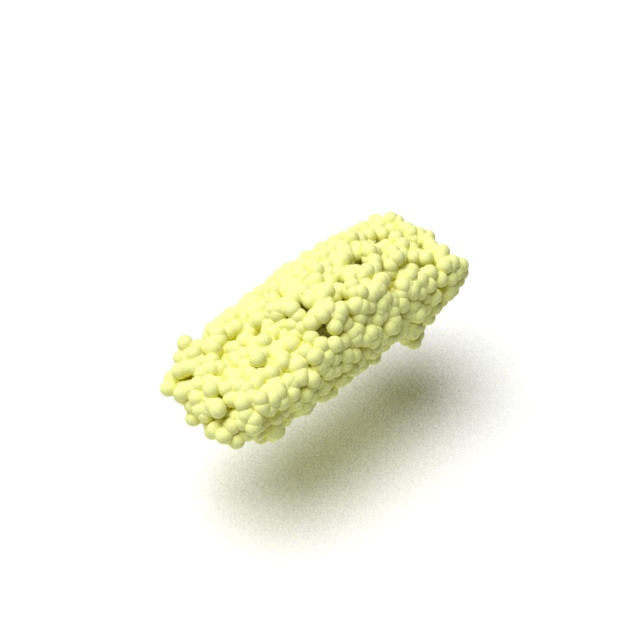}&
        \includegraphics[width=\sizea, trim={\tale} {\tab} {4.5cm} {\tat},clip]{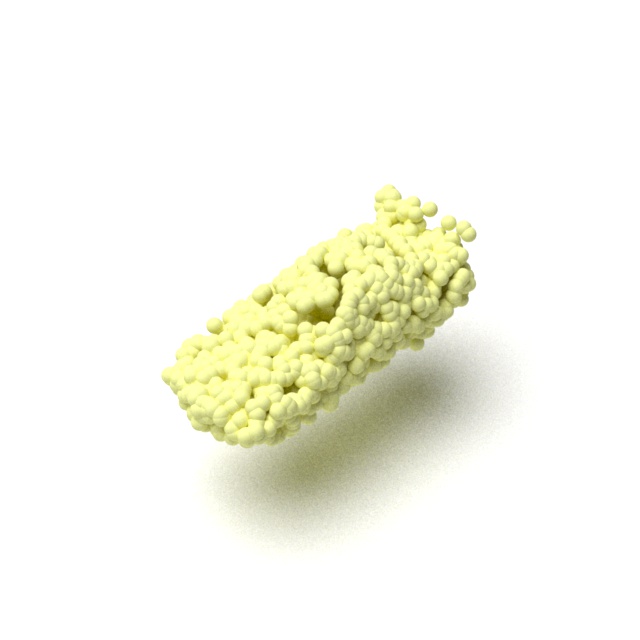}& 
        \includegraphics[width=\sizea, trim={\tal} {\tab} {\tar} {\tat},clip]{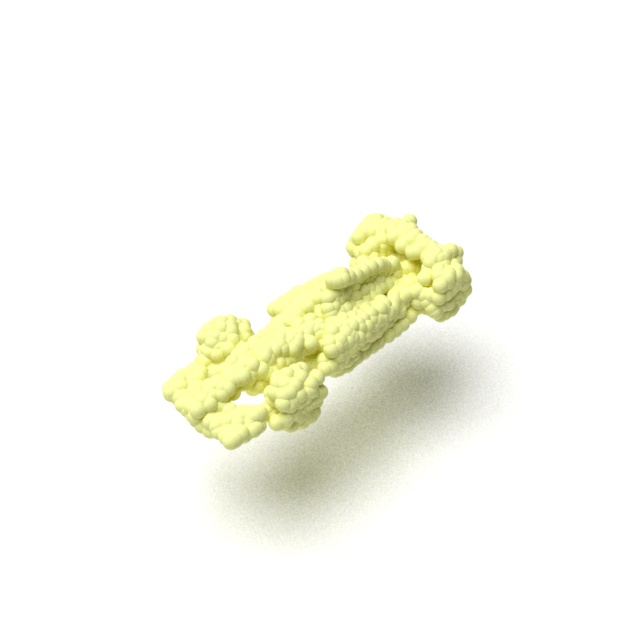}& 
        \includegraphics[width=\sizea, trim={\tal} {\tab} {\tar} {\tat},clip]{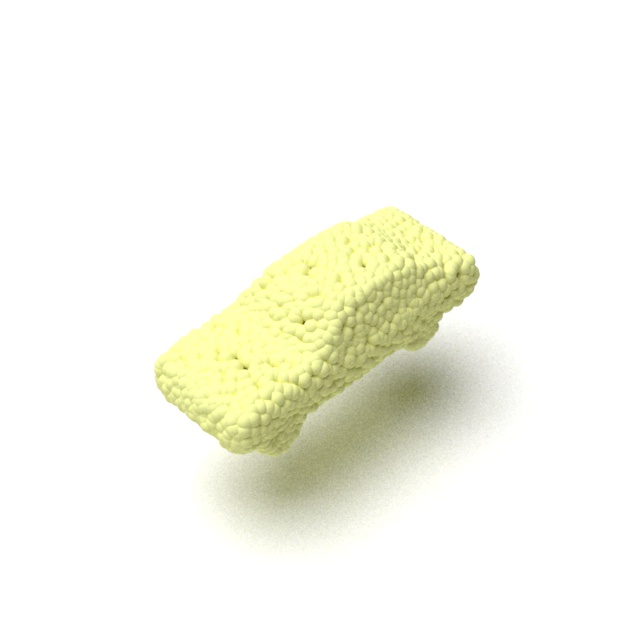}&
        \includegraphics[width=\sizea, trim={\tal} {\tab} {\tar} {\tat},clip]{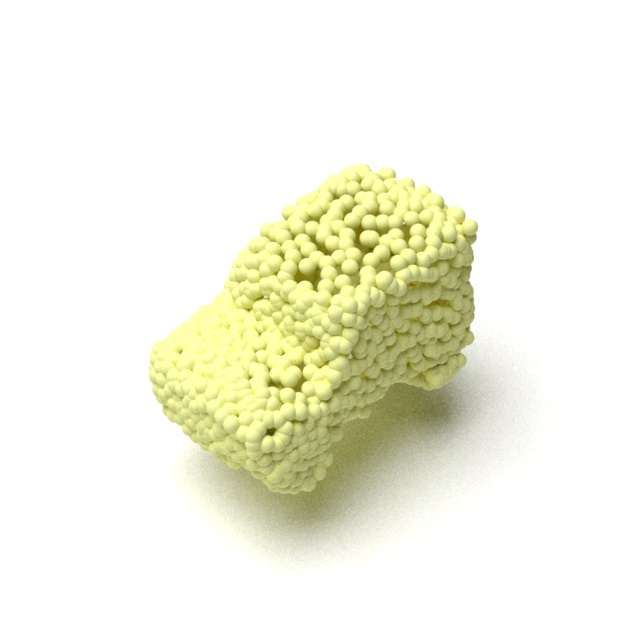}
        \\
        PF & ShapeGF & SetVAE & DPM & PVD & \multicolumn{3}{c}{Ours}
    \end{tabular}

    \end{center}
    \caption{Shape generation results. We shown results from PF (PointFlow)~\cite{yang2019pointflow}, ShapeGF~\cite{cai2020learning}, SetVAE~\cite{kim2021setvae}, DPM~\cite{luo2021diffusion}, and PVD~\cite{zhou20213d}.
    }
    \label{fig:gen}
    \vspace{-1.5em}

\end{figure}

\begin{figure}[t]
    \begin{center}
    \newcommand{\sizea}{0.16\linewidth}
    \newcommand{\sizeb}{0.10\linewidth}
    \newcommand{\sizec}{0.08\linewidth}
    \newcommand{\tare}{5cm}
    \newcommand{\tale}{4cm}
    \newcommand{\tal}{3.5cm}
    \newcommand{\tab}{3cm}
    \newcommand{\tar}{3.5cm}
    \newcommand{\tat}{3.5cm}
    \newcommand{\tcl}{3.0cm}
    \newcommand{\tcb}{3cm}
    \newcommand{\tcr}{4cm}
    \newcommand{\tct}{4.2cm}
    \newcommand{\thl}{3.0cm}
    \newcommand{\thb}{0.0cm}
    \newcommand{\thr}{3cm}
    \newcommand{\tht}{2cm}
    \setlength{\tabcolsep}{0pt}
    \renewcommand{\arraystretch}{0}
    \begin{tabular}{@{}cc:cccc@{}}
        \includegraphics[width=\sizea, trim={\tale} {\tab} {4.5cm} {\tat},clip]{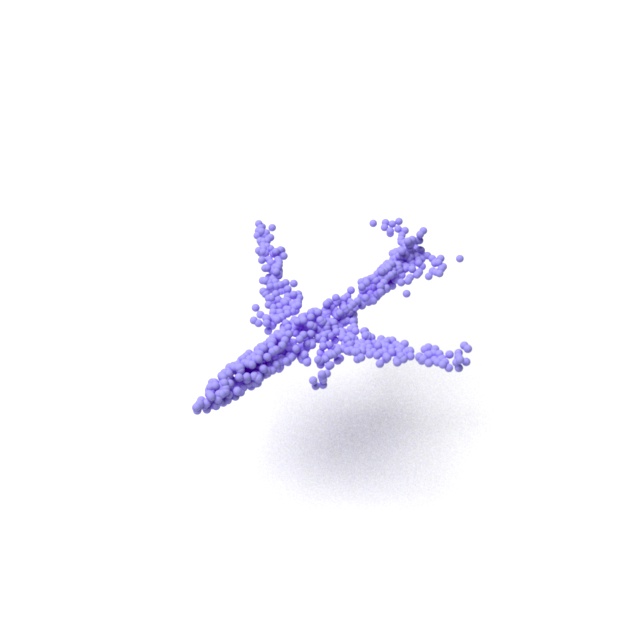}&
        \includegraphics[width=\sizea, trim={\tale} {\tab} {4.5cm} {\tat},clip]{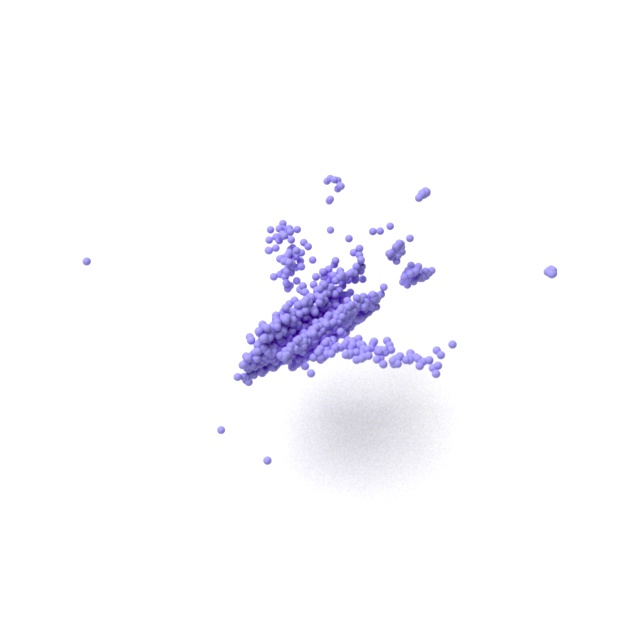}&
        \includegraphics[width=\sizea, trim={\tale} {\tab} {4.5cm} {\tat},clip]{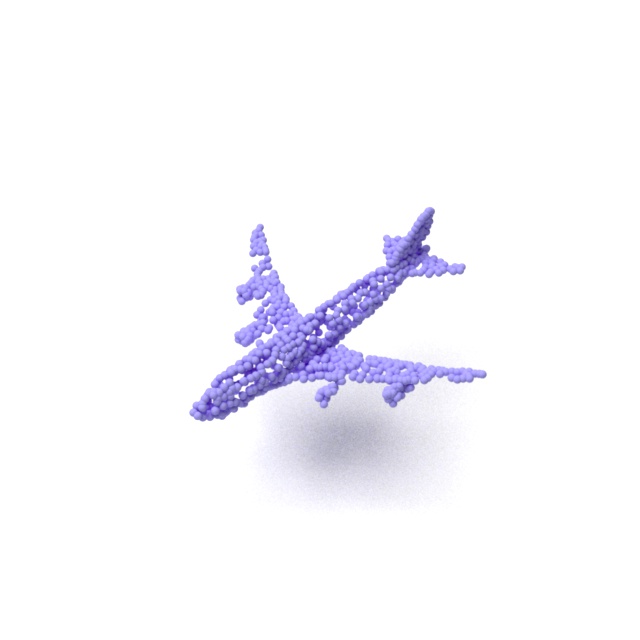}&
        \includegraphics[width=\sizea, trim={\tale} {\tab} {4.5cm} {\tat},clip]{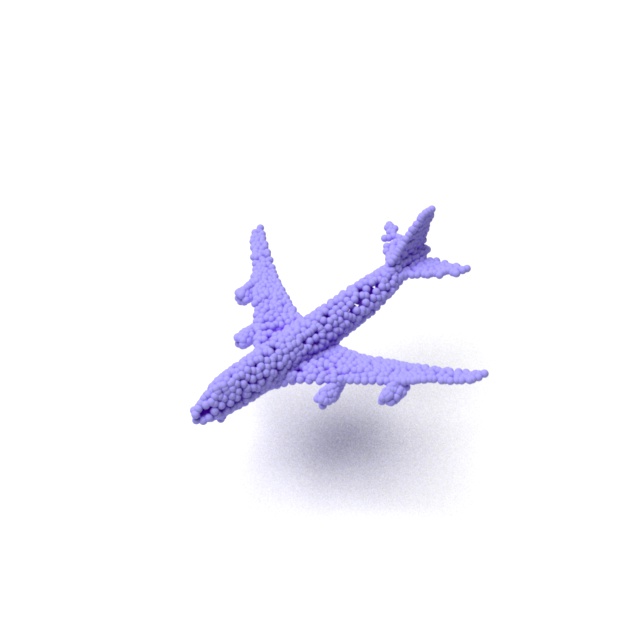}&
        \includegraphics[width=\sizea, trim={\tale} {\tab} {4.5cm} {\tat},clip]{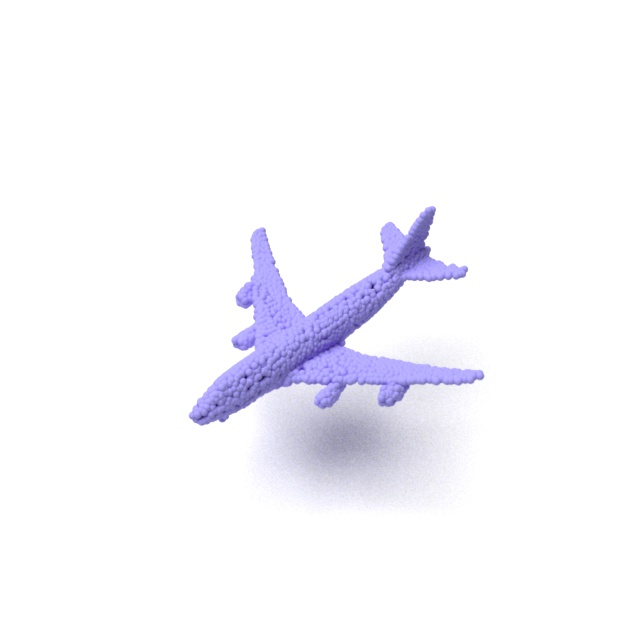}&
        \includegraphics[width=\sizea, trim={\tal} {\tab} {4.5cm} {\tat},clip]{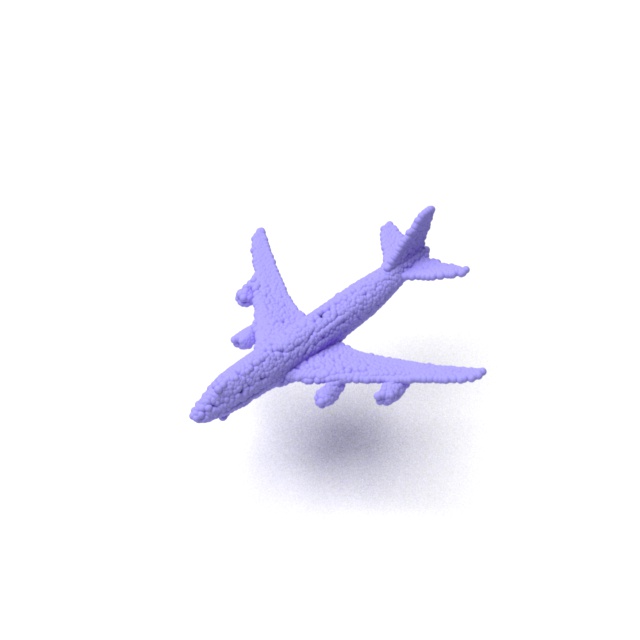} \\
        $n=1024$ & $n=2048$ & $n=1024$ & $n=2048$ & $n=4096$ & $n=8194$ \vspace{0.75em}
        \\
        \multicolumn{2}{c}{PointGrow} & \multicolumn{4}{c}{Ours} 
    \end{tabular}
    \end{center}
    \caption{High-resolution generation results comparing to PointGrow~\cite{sun2020pointgrow}. $n$ refers to the number of output points.
    }

    \label{fig:pointgrow}
    \vspace{-1.5em}
\end{figure}

We quantitatively compare our method with the following state-of-the-art generative models: PointGrow~\cite{sun2020pointgrow}, ShapeGF~\cite{cai2020learning}, SP-GAN~\cite{li2021sp}, PointFlow (PF)~\cite{yang2019pointflow}, SetVAEF~\cite{kim2021setvae}, DPM~\cite{luo2021diffusion}, and PVD\cite{zhou20213d}. We summarize the quantitative results in Table~\ref{tab:generation}. We also report the performance of point cloud sampled from the training set in the “Train” row. For most of the metrics, our model has comparable, if not better, performance than other baselines. This suggests that our model is capable of generating diverse and realistic samples. We provide qualitative results comparing to baselines in Figure~\ref{fig:gen}. 
\\
Among these baselines, PointGrow is most relevant to our work that generates point clouds in an autoregressive manner. Our model significantly outperforms PointGrow in all metrics because PointGrow scales poorly~\cite{sun2020pointgrow} when generating large point sets. In contrast, our method can generate shapes in an arbitrary resolution ranging from low to high with sharp details within a single model. We show point clouds generated with different resolutions comparing to PointGrow in Figure~\ref{fig:pointgrow}.

\subsection{Conditional Generation}\label{sec:exp:shape-completion}

\begin{table*}[t]
\centering
\caption{Multi-modal completion on the Chair dataset. $t$ denotes the temperature scaling factor. $\uparrow$ means the higher the better, $\downarrow$ means the lower the better.
MMD and TMD are both multiplied by $10^3$.}
\label{tab:completion}
\footnotesize
\begin{tabularx}{1\textwidth}{l*{6}{Y}}
\toprule
Category & Metric  &   Input  & MMD ($\downarrow$) & TMD ($\uparrow$)  \\ \midrule
\multirow{3}{*}{Airplane} & MSC     & Depth+Camera & 1.475  & 0.925             \\
& PVD     & Depth+Camera & 1.012 & 2.108 \\
\cmidrule(lr){2-6} 
& Ours (t=1)     & Depth & 0.663 & 1.449  \\
& Ours (t=2)     & Depth & 0.673 & 2.406 \\
& Ours (t=3)     & Depth & 0.684 & 2.352  \\
\midrule
\multirow{3}{*}{Chair} & MSC     & Depth+Camera & 6.372  & 5.924    \\
& PVD     & Depth+Camera & 5.042 & 7.524 \\
\cmidrule(lr){2-6} 
& Ours (t=1)     & Depth & 5.142 & 6.553  \\
& Ours (t=2)     & Depth & 5.261 & 8.174  \\
& Ours (t=3)     & Depth & 6.427 & 13.341 
\\ \bottomrule
\end{tabularx}
\end{table*}

\begin{figure}[t]
    \begin{center}
    \newcommand{\sizea}{0.124\linewidth}
    \newcommand{\sizeb}{0.124\linewidth}
    \newcommand{\sizec}{0.124\linewidth}
    \newcommand{\tare}{5cm}
    \newcommand{\tale}{3.5cm}
    \newcommand{\tal}{3.5cm}
    \newcommand{\tab}{2.5cm}
    \newcommand{\tar}{3.5cm}
    \newcommand{\tat}{3cm}
    \newcommand{\tcl}{3.0cm}
    \newcommand{\tcb}{3cm}
    \newcommand{\tcr}{4cm}
    \newcommand{\tct}{4.2cm}
    \newcommand{\thl}{3.0cm}
    \newcommand{\thb}{0.0cm}
    \newcommand{\thr}{3cm}
    \newcommand{\tht}{2cm}
    \setlength{\tabcolsep}{0pt}
    \renewcommand{\arraystretch}{0}
    \begin{tabular}{@{}cccc:cccc@{}}
        \includegraphics[width=\sizea, trim={\tale} {\tab} {2cm} {\tat},clip]{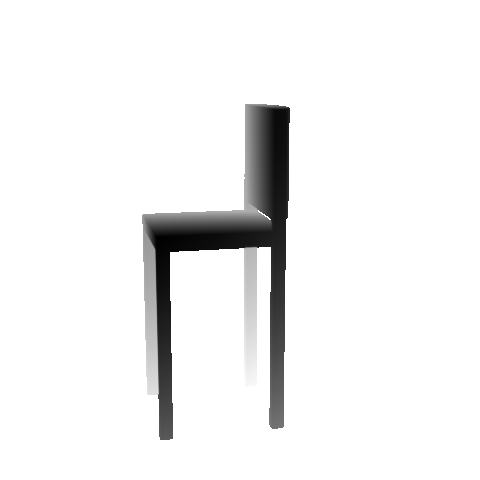} &
        \includegraphics[width=\sizec, trim={\tale} {\tab} {2cm} {\tat},clip]{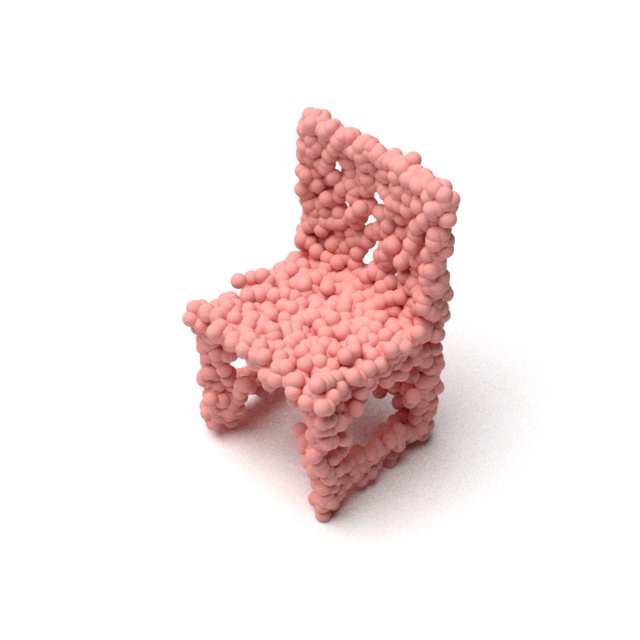} &
        \includegraphics[width=\sizeb, trim={\tale} {\tab} {2cm} {\tat},clip]{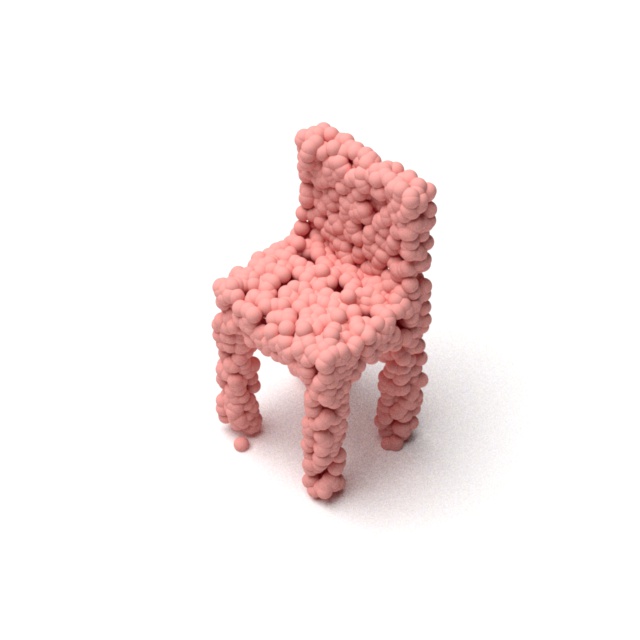} &
        \includegraphics[width=\sizea, trim={\tale} {\tab} {2cm} {\tat},clip]{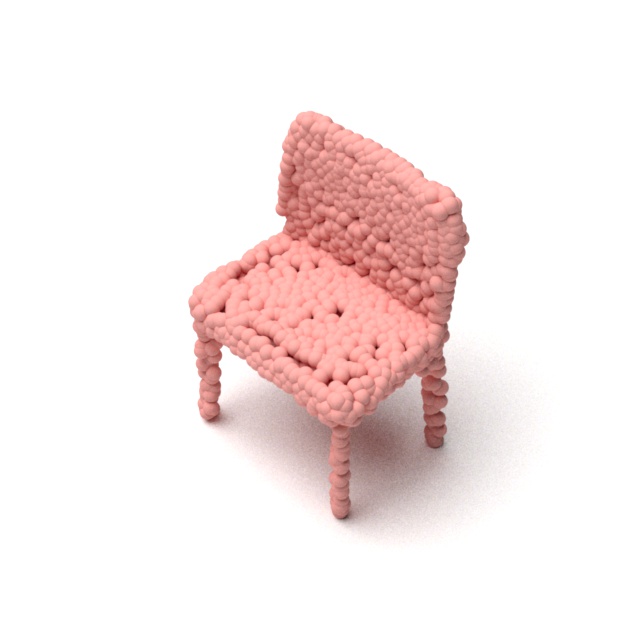} &
        \includegraphics[width=\sizea, trim={\tale} {\tab} {2cm} {\tat},clip]{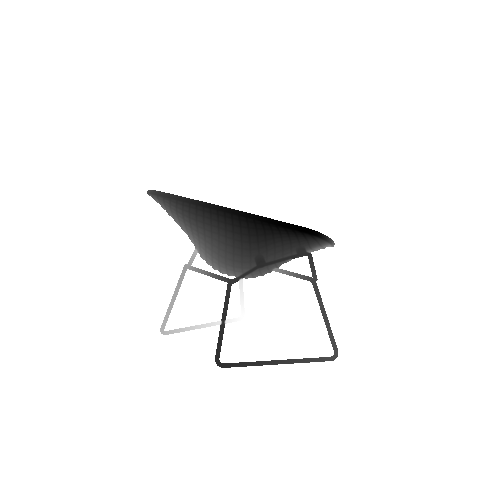} &
        \includegraphics[width=\sizec, trim={\tale} {\tab} {2cm} {\tat},clip]{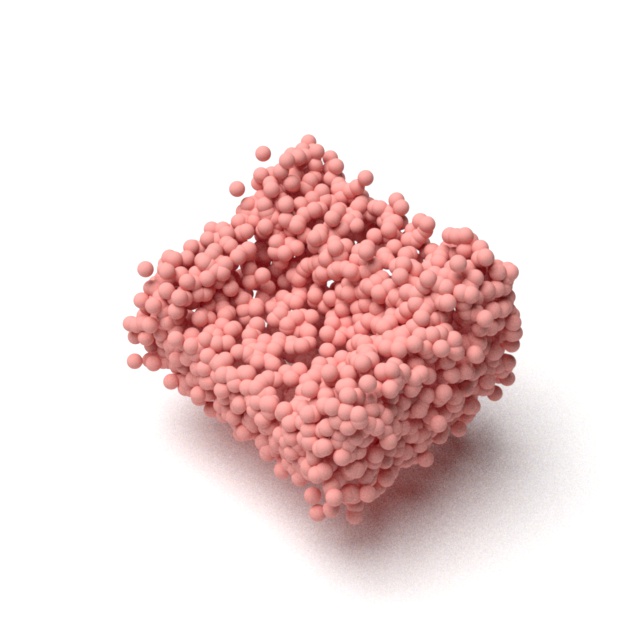} &
        \includegraphics[width=\sizeb, trim={\tale} {\tab} {2cm} {\tat},clip]{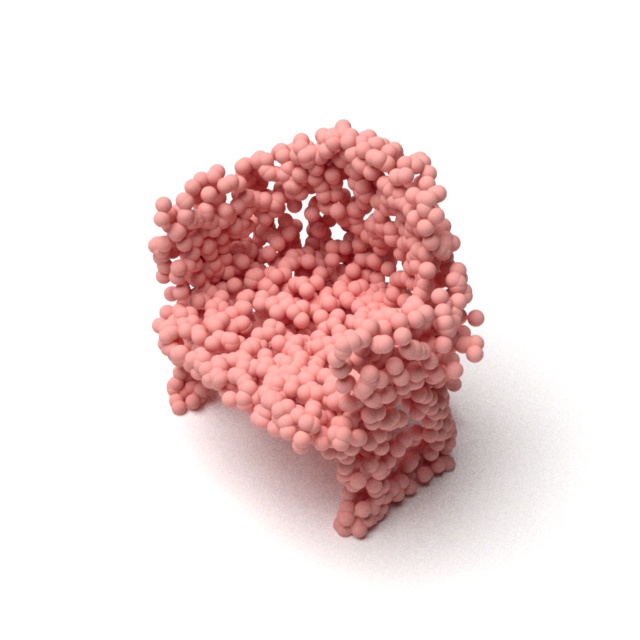} &
        \includegraphics[width=\sizeb, trim={\tale} {\tab} {2cm} {\tat},clip]{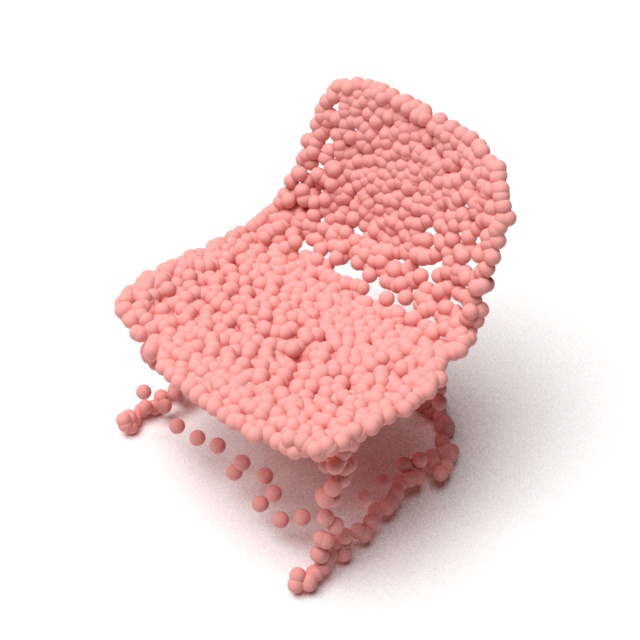}
        \\
        \includegraphics[width=\sizea, trim={\tale} {\tab} {2cm} {\tat},clip]{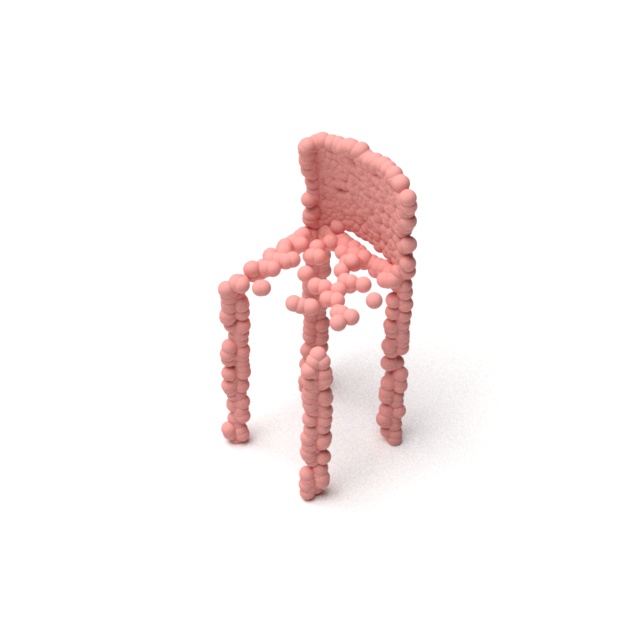} &
        \includegraphics[width=\sizec, trim={\tale} {\tab} {2cm} {\tat},clip]{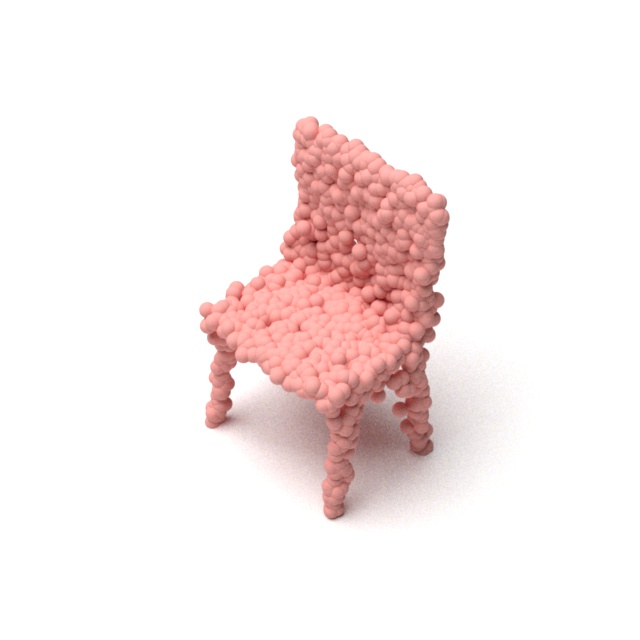} &
        \includegraphics[width=\sizeb, trim={\tale} {\tab} {2cm} {\tat},clip]{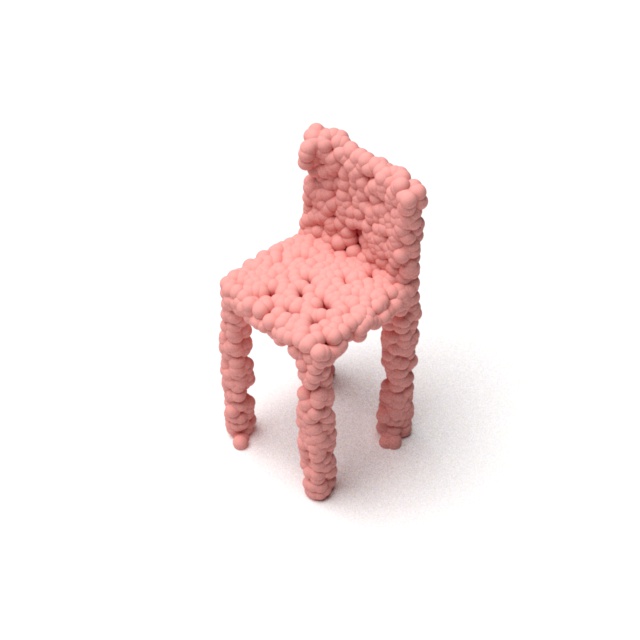} &
        \includegraphics[width=\sizea, trim={\tale} {\tab} {2cm} {\tat},clip]{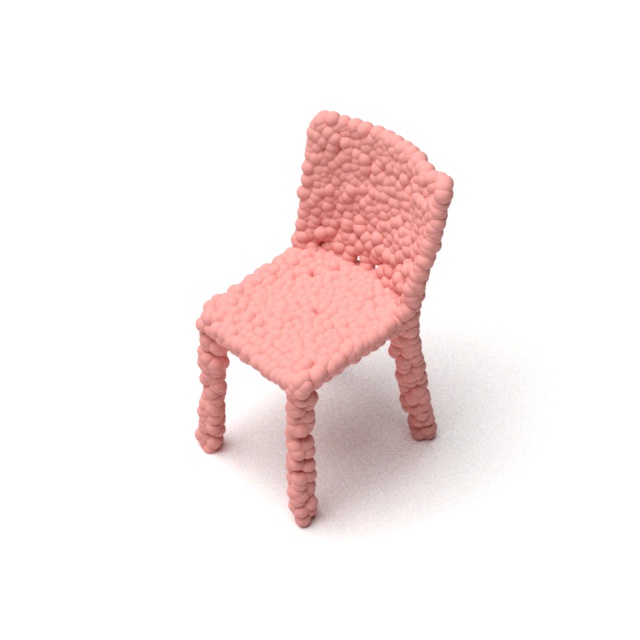} &
        \includegraphics[width=\sizea, trim={\tale} {\tab} {2cm} {\tat},clip]{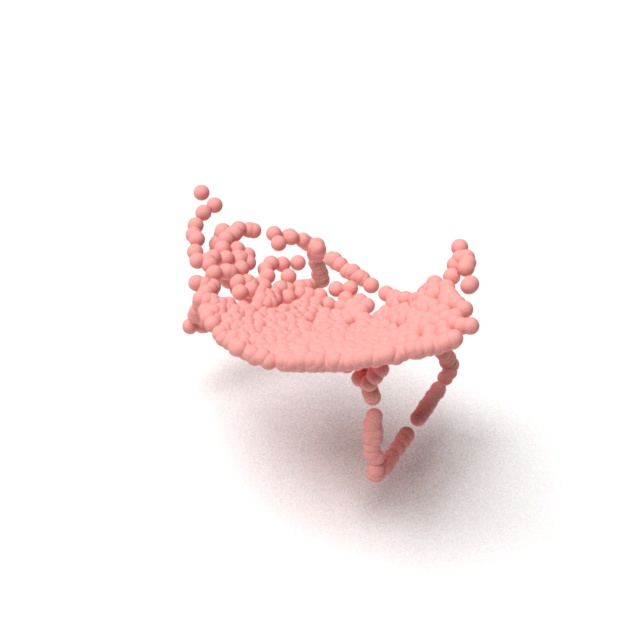} &
        \includegraphics[width=\sizec, trim={\tale} {\tab} {2cm} {\tat},clip]{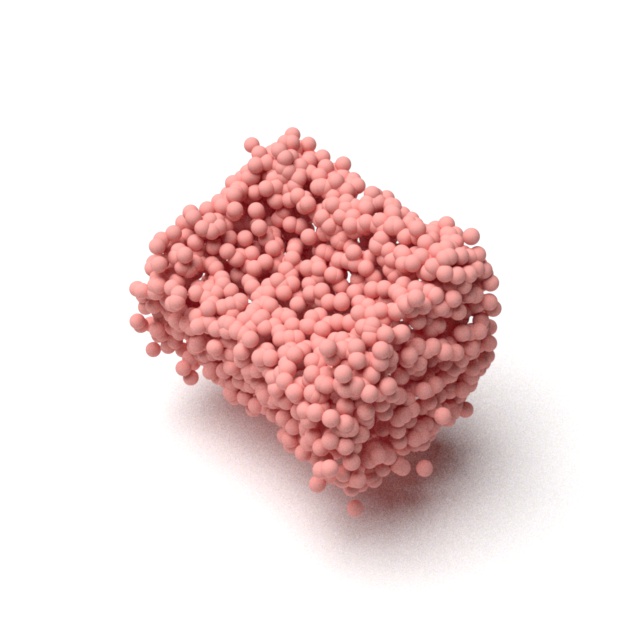} &
        \includegraphics[width=\sizeb, trim={\tale} {\tab} {2cm} {\tat},clip]{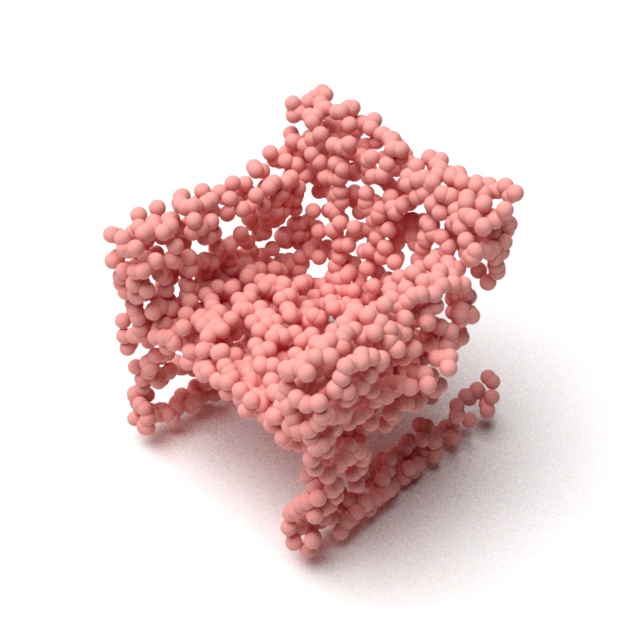} &
        \includegraphics[width=\sizeb, trim={\tale} {\tab} {2cm} {\tat},clip]{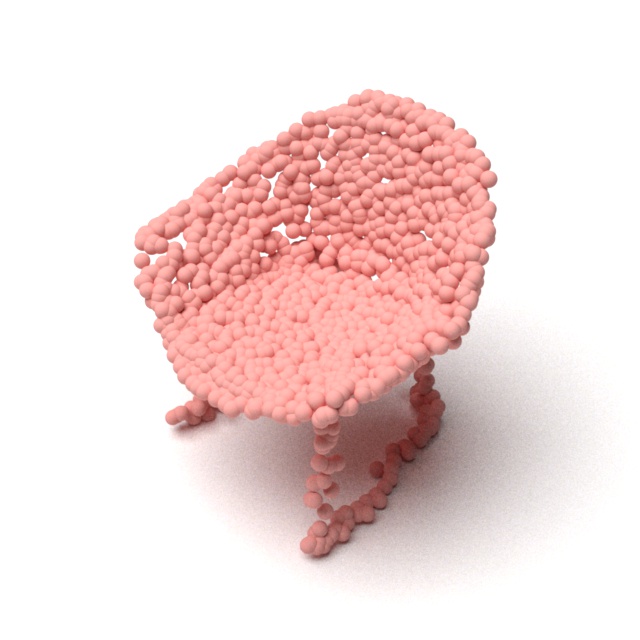}
        \\
        \includegraphics[width=\sizea, trim={\tale} {\tab} {2cm} {\tat},clip]{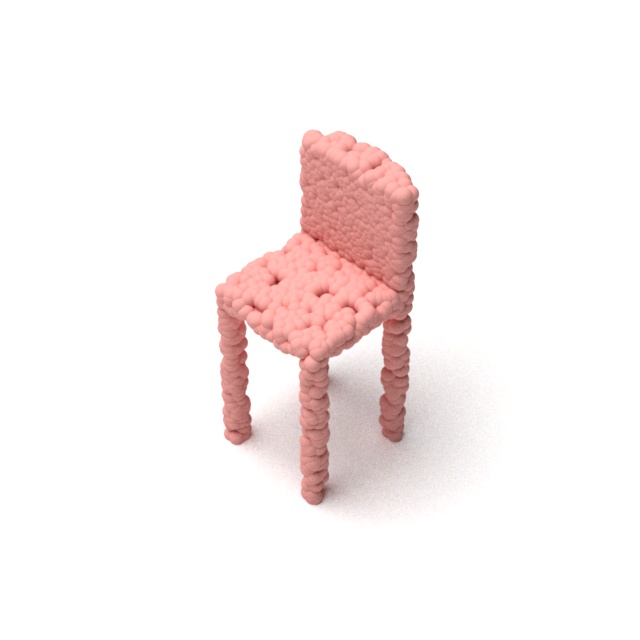} &
        \includegraphics[width=\sizec, trim={\tale} {\tab} {2cm} {\tat},clip]{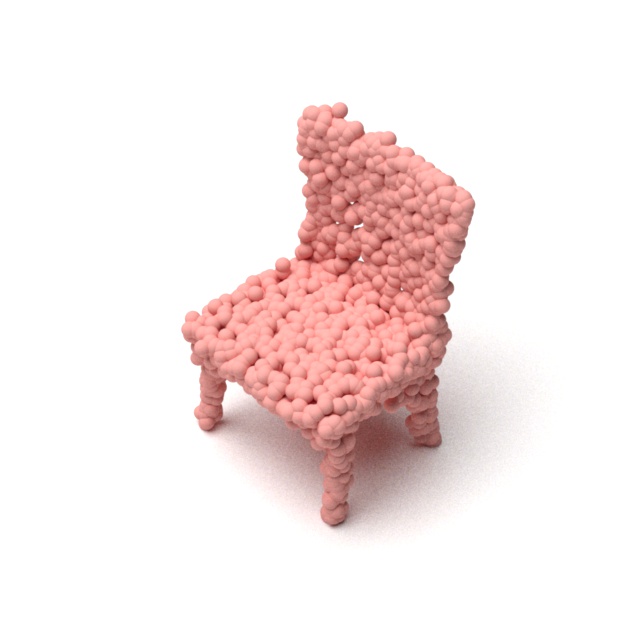} &
        \includegraphics[width=\sizeb, trim={\tale} {\tab} {2cm} {\tat},clip]{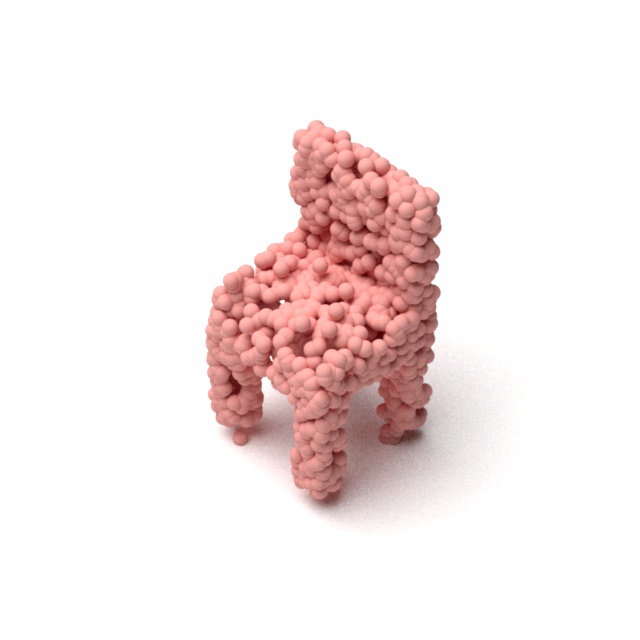} &
        \includegraphics[width=\sizea, trim={\tale} {\tab} {2cm} {\tat},clip]{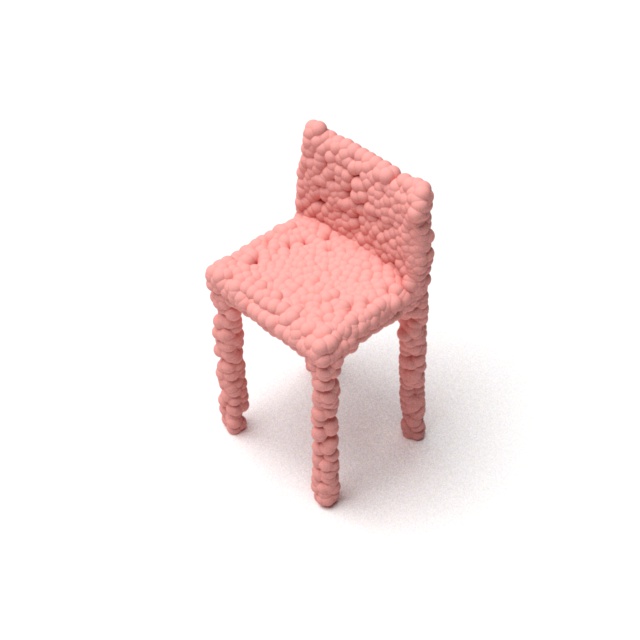} &
        \includegraphics[width=\sizea, trim={\tale} {\tab} {2cm} {\tat},clip]{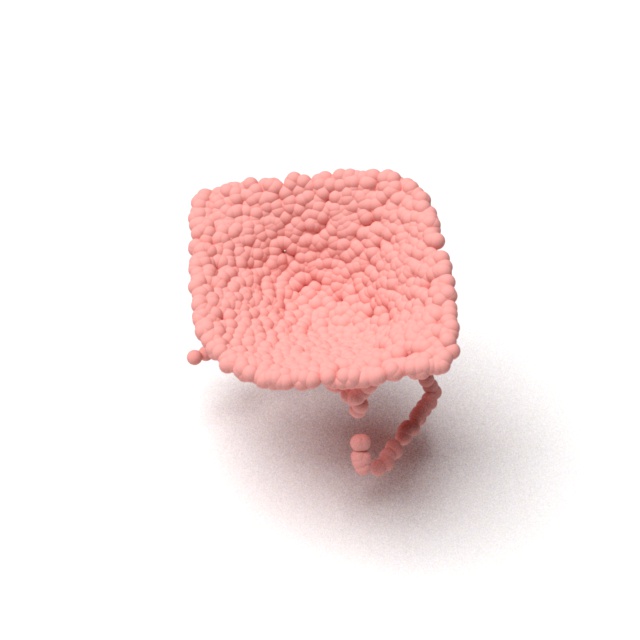} &
        \includegraphics[width=\sizec, trim={\tale} {\tab} {2cm} {\tat},clip]{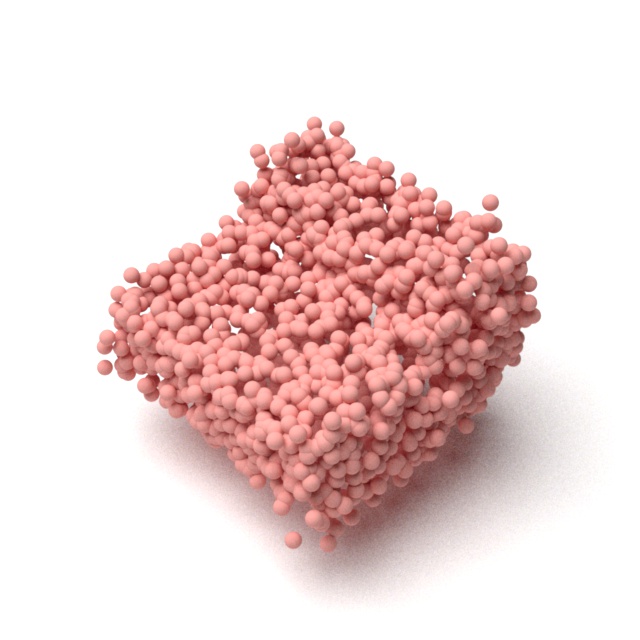} &
        \includegraphics[width=\sizeb, trim={\tale} {\tab} {2cm} {\tat},clip]{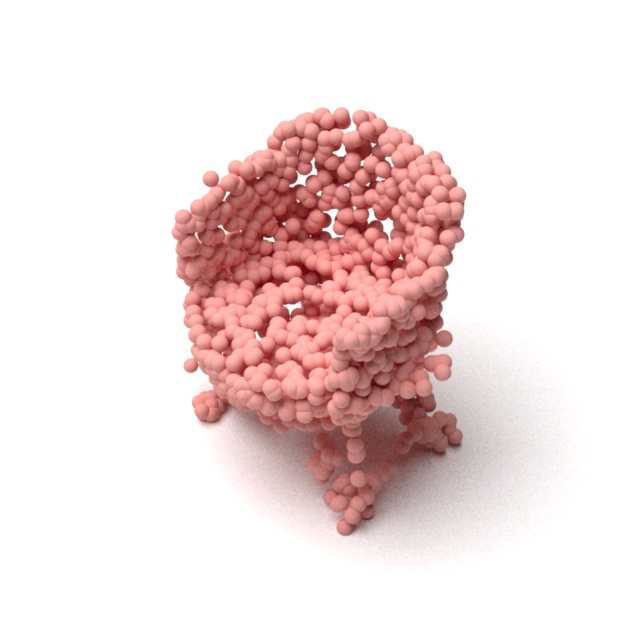} &
        \includegraphics[width=\sizeb, trim={\tale} {\tab} {2cm} {\tat},clip]{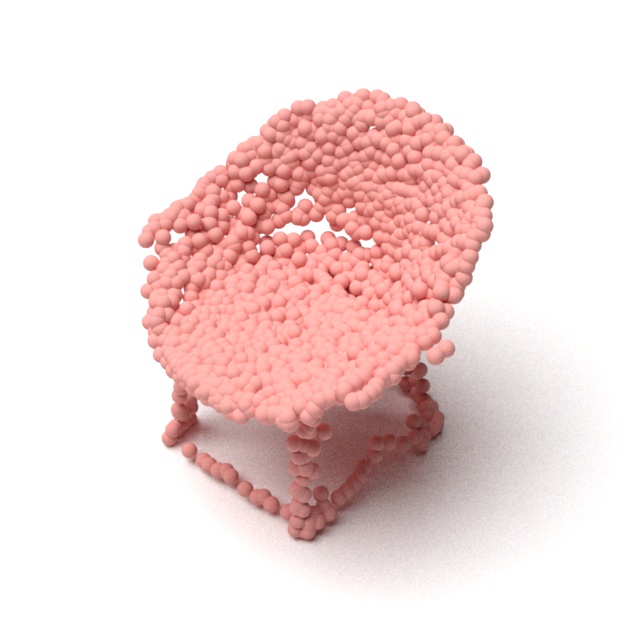}
        \\
        Input$/$GT & MSC & PVD & Ours & Input$/$GT & MSC & PVD & Ours
    \end{tabular}

    \end{center}
    \caption{Multi-modal shape completion results with comparison to MSC and PVD. In the first column, we show the input depth-map, partial point cloud, and reference ground-truth shape for each sample respectively from top to bottom. 
    }

    \label{fig:completion}
    \vspace{-1.5em}
\end{figure}
During inference, our transformer model generates a sequence by probabilistic sampling each token, which naturally allows multi-modal generation. On the other hand, the shape completion problem is multi-modal in nature since the incompleteness introduces significant ambiguity~\cite{wu2020multimodal}. Motivated by this property, we extend our approach to shape completion as an application for conditional shape generation. Specifically, we use a depth map as the input condition to the transformer model. We employ a ResNet50~\cite{he2016deep} encoder to extract global feature from the depth map and prepend the feature vector to the transformer as discussed in follow Sec.~\ref{subsubsec:transformer:cond}.

To evaluate the performance, we compare with two state-of-the-art approaches on multi-modal shape completion: MSC~\cite{wu2020multimodal} and PVD~\cite{zhou20213d}. Note that both MSC~\cite{wu2020multimodal} and PVD~\cite{zhou20213d} take aligned point clouds as inputs. Therefore, they require additional camera parameters to obtain partial scans from the depth map. We present quantitative comparison results in Table~\ref{tab:completion} and show that our approach can achieve comparable or even better performance without requiring additional camera parameters. We also report results using different temperatures, which suggests that our model can provide controllable diversity by scaling the temperature parameter.

To further investigate the reason that our MMD metric in the chair category is slightly inferior to PVD, we visualize qualitative results in Figure~\ref{fig:completion}. In general, our model produces shapes with better visual quality. However, due to the inherent ambiguity in single-view reconstruction, it is difficult to infer the real scale of objects without knowing the camera parameters, especially for depth maps rendered from the side views. For instance, given a depth map of a chair from side viewpoints, our model generates plausible but wider chairs than the ground truth 3D shapes, as shown in the first row of Fig.~\ref{fig:completion}.

\begin{figure}[ht]
\minipage{0.245\textwidth}
\includegraphics[width=\linewidth]{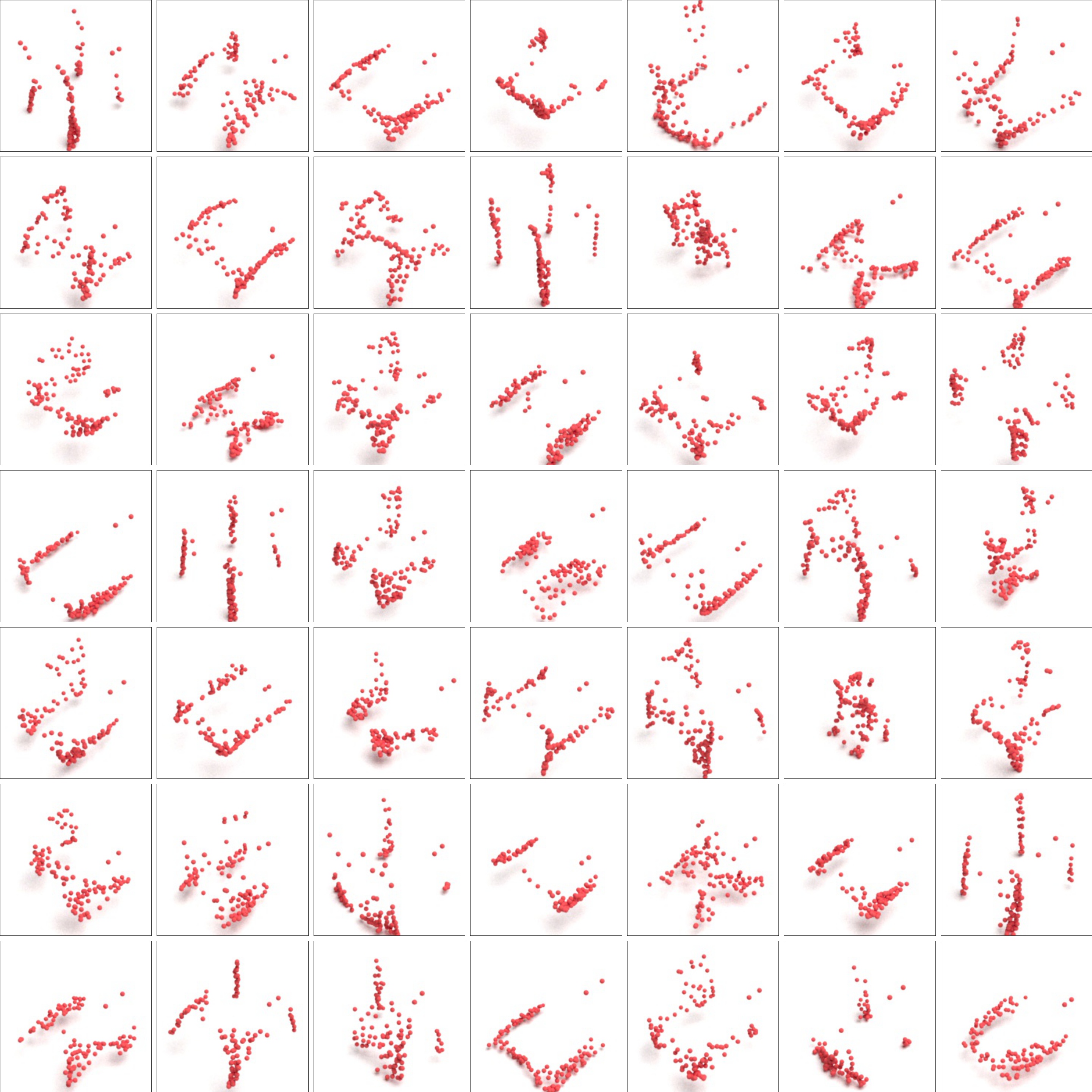}
\endminipage\hfill
\minipage{0.245\textwidth}
\includegraphics[width=\linewidth]{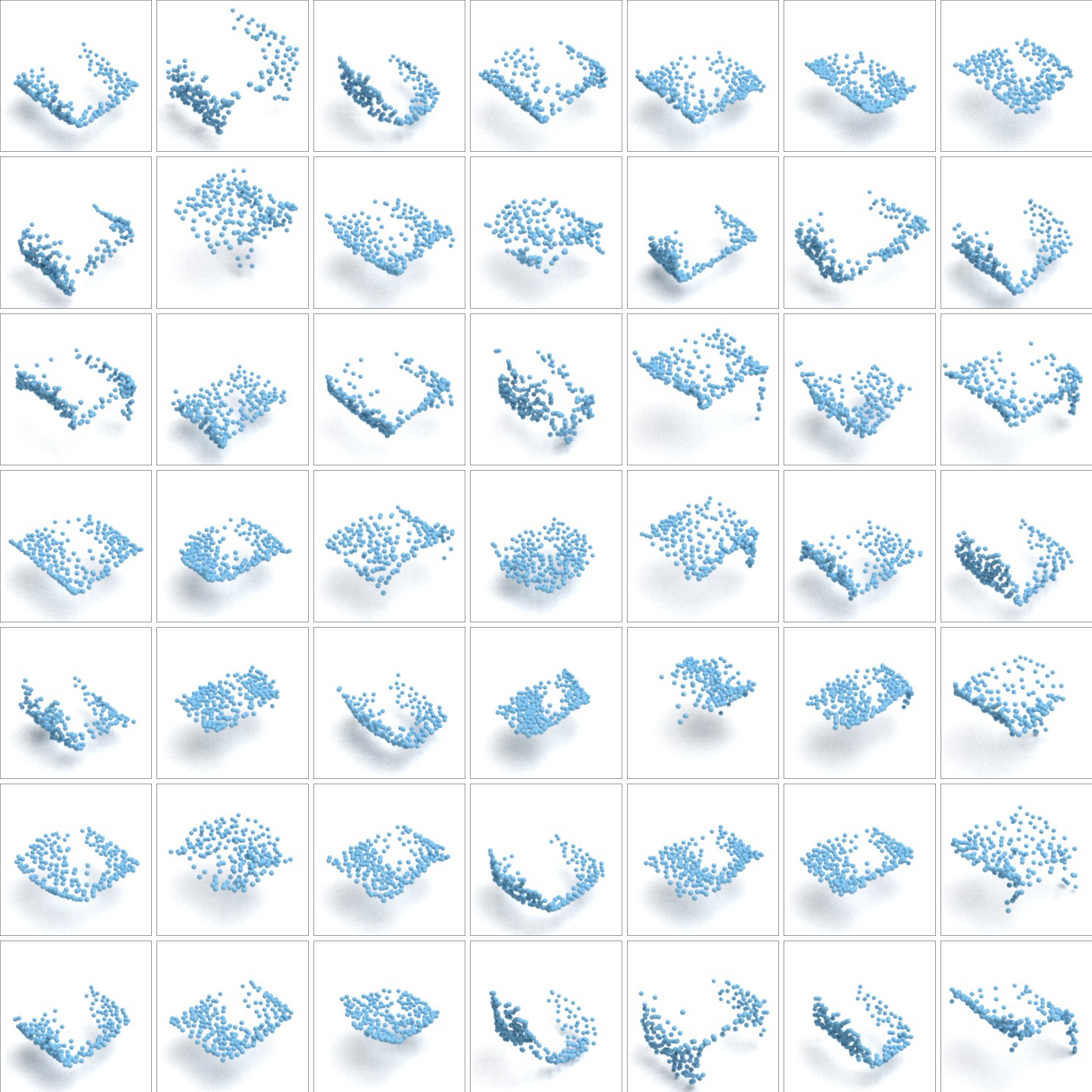}
\endminipage\hfill
\minipage{0.245\textwidth}%
\includegraphics[width=\linewidth]{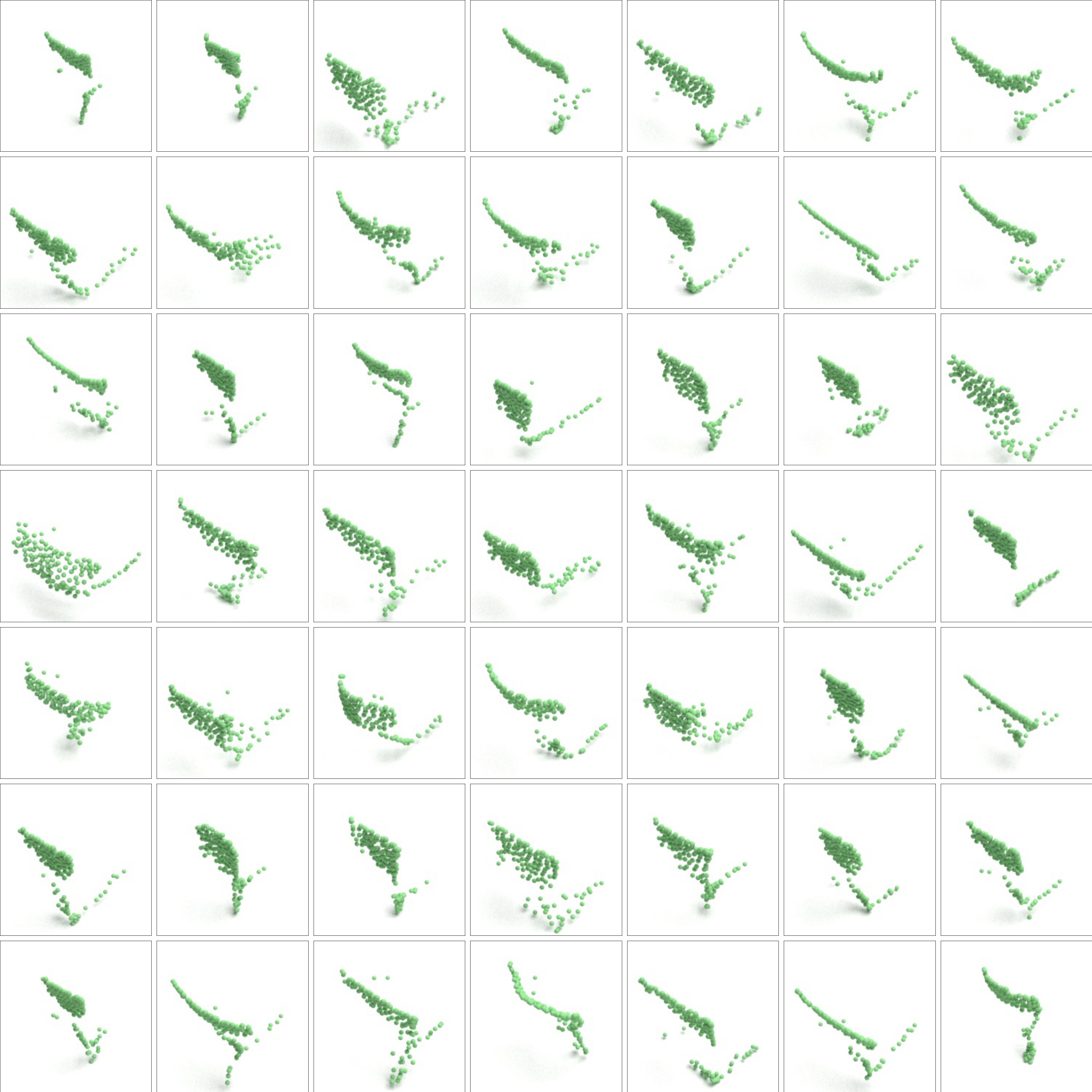}
\endminipage
\minipage{0.245\textwidth}%
\includegraphics[width=\linewidth]{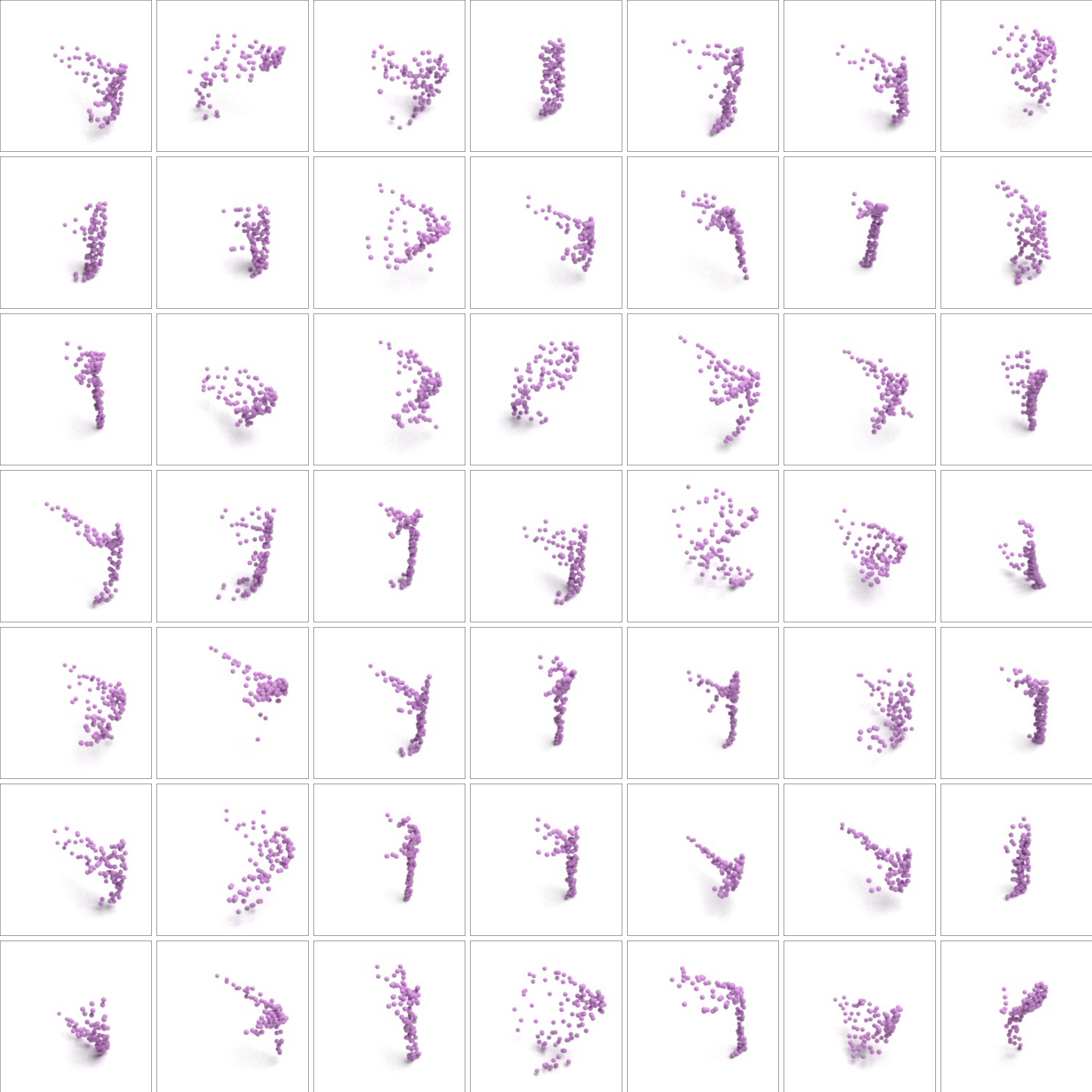}
\endminipage
\caption{Visulization of codebooks. Each of the $7\times7$ grid corresponds to a learned group codebook and each shape in the inner square represents a shape composition decoded from a codebook token.
}
\label{fig:codebook}
\end{figure}

\subsection{Ablation Study}\label{sec:exp:abl}

\begin{table*}[h]
\centering
\caption{Ablation study on the effectiveness of primitive grouping on the Chair dataset. CD is multiplied by $10^4$ and EMD is multiplied by $10^2$.}
\label{tab:ablation_pg}
\footnotesize
\begin{tabularx}{\textwidth}{l*{9}{Y}}
\toprule
& \multicolumn{3}{c}{Canonical Grouping} &  \multicolumn{3}{c}{Uniform Grouping} \\ 
\cmidrule(lr){2-4} \cmidrule(lr){5-7} 
$\# groups$    & 16 & 128 & 256 & 16 & 128 & 256 \\ \midrule
 CD ($\downarrow$)& 7.542 & 6.177 & 7.933 & 8.646  & 7.676 & 7.278 \\
 EMD ($\downarrow$)& 4.466 & 4.218 & 4.645 & 4.772  & 4.541 & 4.627 \\
\bottomrule
\end{tabularx}

\end{table*}


\begin{table}[H]
\centering
\caption{Ablation study on using different vector quantization on the Chair dataset. CD is multiplied by $10^4$ and EMD is multiplied by $10^2$.}
\label{tab:ablation_vq}
\footnotesize
\begin{tabularx}{\textwidth}{l*{9}{Y}}
\toprule

Dimension Reduction & $\xmark$ & 256$\rightarrow$64 & 256$\rightarrow$4 & 256$\rightarrow$64 & 256$\rightarrow$4 \\
Grouped Codebook & $\xmark$ & $\xmark$ & $\xmark$ & $\cmark$ & $\cmark$ \\ \midrule
 CD ($\downarrow$)& 7.139 & 6.561 & 6.298  & 6.442 & 6.177  \\
 EMD ($\downarrow$)& 4.376 & 4.272 & 4.283  & 4.228 & 4.218 \\
 Codebook Usage (\%, $\uparrow$) & 11.72 & 21.04  & 35.72 & 70.22 & 79.28 \\
\bottomrule
\end{tabularx}
\end{table}

\subsubsection{Effectiveness of the canonical mapping function}
Table~\ref{tab:ablation_pg} shows ablations on the effectiveness of our canonical mapping function, with empirical results on CD and EMD metrics using the ShapeNet Chair dataset. We report variants of our model using a different number of groups $G$ and an alternative way to segment the canonical sphere (in the “Uniform Grouping” column). Specifically, we uniformly sample $G$ points on the sphere as centers and use the nearest neighbor search to assign all points on the sphere to its nearest center point. Though straightforward, this approach does not provide any semantic correspondence across shape instances; thereby, our model consistently performs better than this baseline in different $G$ settings. For the “Uniform Grouping” setting, the auto-encoding results directly relate to the $G$ because a larger $G$ results in a finer segmentation. However, our model performs the best with a moderate $G = 128$. Since the size of each group is automatically determined in our model (i.e. rather than explicitly restricted to an equal size), therefore, some groups may include only a few points when $G$ is large. This tends to hurt the encoding performance and results in over-fitting.

\subsubsection{Effectiveness of group-wise codebooks}

Table~\ref{tab:ablation_vq} demonstrates the effectiveness of the latent reduction and group-wise codebook (see Sec.~\ref{subsec:vqvae}) in our vector quantizer $Q$. In addition to CD and EMD, we also report the codebook usage for each model. The usage is computed as the percentage of codes that have been utilized at least once over the entire test set. For a fair comparison with our full model that uses 128 group codebooks in size 50, we use a global codebook in size 5000 ($\approx$ 128$\times$50) for each variant that does not use group-wise codebooks. Our full model performs the best with dimension reduction from 256 to 4 together with a group-wise codebook. Reducing the lookup dimension in the codebook and using a group-wise codebook significantly boost the codebook usage and thereby achieve better auto-encoding quality. To further show that the proposed model can learn codebooks as a library of local shapes, we visualize the learned VQ codes in different groups in Figure~\ref{fig:codebook}, where each codebook clearly captures one meaningful part of the chair category.
\section{Conclusions}
In this work, we propose a transformer-based autoregressive model for point cloud generation. The key idea is to decompose a point cloud into a sequence of semantically aligned shape compositions in a learned canonical space. We show that these compositions can be further used to learn a group of context-rich codebooks for point cloud generation. Experimental results demonstrate that the proposed method can achieve state-of-the-art performance for point cloud auto-encoding and generation. Finally, we show that our model can be easily extended to multi-modal shape completion as an application for conditional shape generation.

\newpage
\appendix

\section*{\Large Appendix}

In this Appendix, we provide more details about implementation, model efficiency and additional results.
We start by introducing implementation details in Section~\ref{sec:details}, followed by model size and computational time discussion in Section~\ref{sec:efficiency}.
We then show ablation studies on different shape composition serialization methods in Section~\ref{sec:ablation}.
In Section~\ref{sec:rec}, Section~\ref{sec:uncon_gen} and Section~\ref{sec:con_gen}, we demonstrate more qualitative results on auto-encoding, unconditional generation and conditional generation.
Finally, we discuss limitations of our method in Section~\ref{sec:limitation}.

\section{Implementation Details}
\label{sec:details}
\begin{algorithm}[H]
\caption{\textbf{:} The training phase of our approach consists of learning four components: (1) Canonical Auto-encoder (2) Canonical Grouping Network (3) VQVAE (4) Transformer}
\label{algo:overview}
\vspace{0.2em}
\begin{algorithmic}
    \STATE\hspace{-0.8em} 
    \underline{\sc{\textbf{(A) Canonical Auto-encoder}}} \textit{\small \hfill $\vartriangleright$ \underline{8 hours on Airplane category}}
\end{algorithmic}
\vspace{0.1em}
\begin{algorithmic}[1]
    \STATE \hspace{0.5em} Sub-sample $M$ points from the input point cloud $x$ and canonical sphere $\pi$;
    \STATE \hspace{0.5em} Initialize weight of the encoder $E_{c}(\cdot)$ and decoder $D_{c}(\cdot)$;
    
    \STATE \hspace{0.5em} \textbf{while} not converged
    \textbf{do}
    \STATE \hspace{1.5em} \textbf{foreach} {iteration} \textbf{do}
    \STATE \hspace{2.5em} $\boldsymbol{z_x} \leftarrow \, E_{c}(x)$;
    \STATE \hspace{2.5em} $\boldsymbol{\hat{x}} \leftarrow \, D_{c}([\pi_{i},z_A])$, where $\pi_{i} \in \pi$;
    \STATE \hspace{2.5em} Obtain reconstruction loss $L_{CD}(\hat{x},x)$ and $L_{EMD}(\hat{x},x)$;
    \STATE \hspace{2.5em} Update weight;
\end{algorithmic}

\vspace{-0.7em}\hrulefill \\ [-1.2em]
\begin{algorithmic}
    \STATE \hspace{-0.8em} \underline{\sc{\textbf{(B) Canonical Grouping Network}}} \textit{\small \hfill $\vartriangleright$ \underline{2 hours on Airplane category}}
\end{algorithmic}
\vspace{0.1em}

\begin{algorithmic}[1]
    \STATE \hspace{0.5em} Generate randomly $M$ points from the canonical sphere $\pi$;
    \STATE \hspace{0.5em} Initialize weight of the canonical grouping network $MLP(\cdot)$;
    \STATE \hspace{0.5em} \textbf{while} not converged \textbf{do}
    \STATE \hspace{1.5em} \textbf{foreach} {iteration} \textbf{do}
    \STATE \hspace{2.5em} Obtain the corresponding point of $\pi_{i}\in \pi$ on $x$ as $\Phi^{-1}_{\pi\rightarrow x}(\pi_{i})$;
    \STATE \hspace{2.5em} $\boldsymbol{P_{i}} \leftarrow \ MLP([x_{i},\pi_{i}])$, where $x_{i} \in x, \pi_{i} \in \pi$;
    \STATE \hspace{2.5em} $\boldsymbol{K_j} \leftarrow \ \sum_{i=1}^{m} \Phi^{-1}_{\pi\rightarrow x}(\pi_i) P_i^j$   , where $j=1,2,...,G$;
    \STATE \hspace{2.5em} Obtain loss $L_{CD}(K,x)$;
    \STATE \hspace{2.5em} Update weight;
\end{algorithmic}

\vspace{-0.7em}\hrulefill \\ [-1.2em]
\begin{algorithmic}
    \STATE\hspace{-0.8em} \underline{\sc{\textbf{(C) VQVAE}}} \textit{\small \hfill $\vartriangleright$ \underline{24 hours on Airplane category}}
\end{algorithmic}
\vspace{0.1em}
\begin{algorithmic}[1]
    \STATE \hspace{0.5em} Sub-sample $M$ points from the input point cloud $x$ and canonical sphere $\pi$;
    \STATE \hspace{0.5em} Sequentialize $x$;
    \STATE \hspace{0.5em} Initialize weight of $E(\cdot)$, $D(\cdot)$, and $VQ(\cdot)$;
    \STATE \hspace{0.5em} \textbf{while} not converged \textbf{do}
    \STATE \hspace{1.5em} \textbf{foreach} {iteration} \textbf{do}
    \STATE \hspace{2.5em} Obtain group feature $\boldsymbol{z} \leftarrow \ E(x)$ 
    \STATE \hspace{2.5em} $\boldsymbol{z_{q}} \leftarrow \ VQ(z)$
    \STATE \hspace{2.5em} $\boldsymbol{\hat{x}} \leftarrow \, D([\pi_{i},z_{q}])$, where $\pi_{i} \in \pi$;
    \STATE \hspace{2.5em} Obtain loss $L_{Quantization}$;
    \STATE \hspace{2.5em} Update weight;
\end{algorithmic}
    
\vspace{-0.7em}\hrulefill \\ [-1.2em]
\begin{algorithmic}
    \STATE\hspace{-0.8em} \underline{\sc{\textbf{(D) Transformer}}} \textit{\small \hfill $\vartriangleright$ \underline{16 hours on Airplane category}}
\end{algorithmic}
\vspace{0.1em}
\begin{algorithmic}[1]
    \STATE \hspace{0.5em} Sub-sample $M$ points from the input point cloud $x$;
    \STATE \hspace{0.5em} Initialize weight of the transformer;
    \STATE \hspace{0.5em} Vector Quantize $x$ to a sequence $s$;
    \STATE \hspace{0.5em} \textbf{while} not converged \textbf{do}
    \STATE \hspace{1.5em} \textbf{foreach} {iteration} \textbf{do}
    \STATE \hspace{2.5em} Obtain the probability distribution for each $s_{i} \in s$ autoregressively.
    \STATE \hspace{2.5em} Obtain loss $L_{Transformer}$;
    \STATE \hspace{2.5em} Update weight;
\end{algorithmic}
\label{pseudo}
\end{algorithm}

\subsection{Model Architecture}
We use the same model architecture for all encoders and decoders used in our method. Specifically, we adopt the encoder structure from DGCNN~\cite{wang2019dynamic}, which contains 3 EdgeConv layers using neighborhood size 20. Our decoder follows the two branch structure as in SP-GAN~\cite{li2021sp}. Given a matrix that consists of sphere points and features, the decoder first feeds sphere points to a graph attention module to extract point-wise spatial features. On the other branch, we use a nonlinear feature embedding to extract style features from the latent. Then, we use adaptive instance normalization~\cite{dumoulin2016learned} to fuse the local styles with the spatial features. We repeat the process with another round of style embedding and fusion, then predict the final output from the fused feature. For our grouping network, we follow~\cite{chen2020unsupervised}, using a two-layer 128-neuron MLP with ReLU activations and BatchNorm layers. Our transformer model is modified from~\cite{esser2021taming}, where we reduce their number of layers and heads to 24 and 16, respectively.

\subsection{Training Details}
We use the airplane category in ShapeNet as an example to illustrate the training pipeline. We perform all the experiments on a workstation with Intel Xeon Gold 6154 CPU (3.00GHz) and 4 NVIDIA Tesla V100 (32GB) GPUs. We implement our framework with Pytorch 1.10. Please see Algorithm~\ref{pseudo} for more details. \textit{The source code will be released to public upon publication.}

\subsection{Inference Details}
Esser et al.~\cite{esser2021taming} introduce several test-time hyper-parameters (e.g., top-k and top-p heuristics, temperature scaling factor $t$) for transformer to obtain best results. Following~\cite{esser2021taming}, we use top-p sampling heuristic for the transformer model which we empirically set $p=0.92$ throughout all experiments. We do not use top-k sampling heuristic and the temperature scaling factor $t$ is set to 1 unless otherwise specified. We provide unconditional generation results on ShapeNet Chair using different $p$ in Table~\ref{tab:abl-topp}.

\begin{table*}[ht]
\centering
\caption{\footnotesize Shape generation results on ShapeNet Chair. $\uparrow$ means the higher the better, $\downarrow$ means the lower the better. 
MMD-CD is multiplied by $10^3$ and MMD-EMD is multiplied by $10^2$.}
\label{tab:abl-topp}
\footnotesize
\begin{tabularx}{\textwidth}{L*{2}{Y}l*{2}{Y}l*{2}{Y}}
\toprule
& \multicolumn{2}{c}{MMD ($\downarrow$)} &  & \multicolumn{2}{c}{COV (\%, $\uparrow$)} &  & \multicolumn{2}{c}{1-NNA (\%, $\downarrow$)} \\ \cmidrule(lr){2-3} \cmidrule(lr){5-6} \cmidrule(l){8-9}
 Model     & CD         & EMD        &  & CD         & EMD       &  & CD     & EMD         \\ \midrule
$p=0.85$
& 7.70  & 11.87   && 41.84 & 44.25  && 61.40  & 64.72\\
$p=0.92$
 & 7.37 & 11.75  &&  \textbf{45.77}  & \textbf{46.07} && \textbf{60.12} & \textbf{61.93} \\
$p=0.99$           
&  \textbf{7.22} & \textbf{11.73} && 44.86 &  45.46 && 60.19 & 62.38
\\\bottomrule

\end{tabularx}
\vspace{-4mm}
\end{table*}

\subsection{Evaluation Metrics}
\begin{itemize}
    \item \textbf{Minimum matching distance (MMD)~\cite{yang2019pointflow}} measures the fidelity of the generated point clouds. For each sample in the reference point clouds, we compute the distance to its nearest neighbor in the generated point cloud. The final MMD is the average of the distances. Note that the nearest neighbor can be calculated with different distance measurements such as Chamfer distance or Earth Mover distance. 
    \item \textbf{Coverage (COV)~\cite{yang2019pointflow}} detects mode-collapse by measuring the fraction of samples in the reference point clouds that are matched to at least one sample in the generated point clouds. Specifically, for each sample in the generated point clouds, we mark its nearest neighbor in the reference point clouds as a match. Similar to MMD, the nearest neighbor can be calculated with different distance measurements such as Chamfer distance or Earth Mover distance. 
    \item \textbf{1-nearest neighbor accuracy (1-NNA)~\cite{yang2019pointflow}} performs two-sample tests~\cite{lopez2016revisiting} on the generated point clouds and the reference point clouds. If the generated point clouds seem to be drawn from the reference point, then the classifier will perform like a random guess (i.e. results in near 50\% accuracy).
    \item \textbf{Total Mutual Difference (TMD)~\cite{wu2020multimodal}} measures the completion diversity given a conditional input (e.g., depth-map, partial point cloud). Specifically, for each shape $i$ in the $k$ generated shapes, we calculate its average Chamfer distance $d^{CD}_i$ to the other $k-1$ shapes. The total TMD is calculated as $\sum _{i=1}^{k}d^{CD}_i$.
\end{itemize}

\clearpage
\section{Computational Time and Model Size}\label{sec:efficiency}
\vspace{-1em}
\begin{table*}[ht]
\centering
\caption{The parameter size and inference time for different models.}
\label{tab:modelsize}
\begin{tabularx}{0.75\textwidth}{LL*{1}{Y}l*{1}{Y}l}
\toprule
Model  & $\#$ Parameters & Inference Time            \\ \midrule
PointGrow~\cite{sun2020pointgrow} & 0.31M & 5303.70 &  \\
ShapeGF~\cite{cai2020learning} & 0.13M & 0.2659 \\
SP-GAN~\cite{li2021sp} & 0.58M & 0.2407 \\
PointFlow~\cite{yang2019pointflow} & 1.61M & 0.3506 \\
SetVAE~\cite{kim2021setvae} & 0.55M & 0.0158 \\ 
DPM~\cite{luo2021diffusion} & 3.87M & 0.0943 \\
PVD~\cite{zhou20213d} & 27.6M & 38.35 \\ \midrule
Ours (Transformer) & 20.6M & 1.5391\\
Ours (VQ) & 0.91M & 0.0981\\ \midrule
Ours (Total) & 21.5M & 1.6372\\
\bottomrule
\end{tabularx}
\end{table*}
We report the inference time and model size for different models in Table~\ref{tab:modelsize}. To be precise, the inference time and model size for each model is measured as the time and the number of parameters needed for generating a shape instance. All results are measured with their official implementation on a workstation with Intel Xeon Gold 6154 CPU (3.00GHz) and a single NVIDIA Tesla V100 (32GB). Note that PointGrow requires forwarding the model the same time as the desired number of points (e.g., 2048), therefore, is slow to compute. PVD is a diffusion-based approach which involves multi-step refinement from a random noise, therefore, is computationally intensive, too.

\section{Ablation on Shape Composition Serialization}
\label{sec:ablation}
To analyze the effect of different shape composition serialization, we train our transformer model with (1) random order (2) Fibonacci spiral order (3) inverse Fibonacci spiral order (Spiral$\star$). As shown in Table~\ref{tab:abl-spiral}, using Fibonacci spiral order in either direction is generally better than using a random order.
\begin{table*}[ht]
\centering
\caption{\footnotesize Shape generation results on ShapeNet Chair. $\uparrow$ means the higher the better, $\downarrow$ means the lower the better. 
MMD-CD is multiplied by $10^3$ and MMD-EMD is multiplied by $10^2$.}
\label{tab:abl-spiral}
\footnotesize
\begin{tabularx}{\textwidth}{L*{2}{Y}l*{2}{Y}l*{2}{Y}}
\toprule
& \multicolumn{2}{c}{MMD ($\downarrow$)} &  & \multicolumn{2}{c}{COV (\%, $\uparrow$)} &  & \multicolumn{2}{c}{1-NNA (\%, $\downarrow$)} \\ \cmidrule(lr){2-3} \cmidrule(lr){5-6} \cmidrule(l){8-9}
 Model     & CD         & EMD        &  & CD         & EMD       &  & CD     & EMD         \\ \midrule
Random
& 7.44  & 11.85   && 43.20 & 42.14  && 61.02  & 65.18\\
Spiral
 & 7.37 & 11.75  &&  \textbf{45.77}  & \textbf{46.07} && 60.12 & \textbf{61.93} \\
Spiral$\star$           
&  \textbf{7.17} & \textbf{11.61} && 44.56 &  44.71 && \textbf{59.36} & 62.23
\\\bottomrule

\end{tabularx}
\vspace{-4mm}
\end{table*}

\clearpage
\section{Qualitative Results of Auto-encoding}
\label{sec:rec}
\begin{figure}[h]
    \begin{center}
    \newcommand{\sizea}{0.18\linewidth}
    \newcommand{\sizeb}{0.14\linewidth}
    \newcommand{\tare}{4cm}
    \newcommand{\tale}{4.5cm}
    \newcommand{\tal}{4cm}
    \newcommand{\tab}{3.5cm}
    \newcommand{\taba}{4.5cm}
    \newcommand{\tar}{4cm}
    \newcommand{\tat}{3.5cm}
    \newcommand{\tata}{7.5cm}
    \newcommand{\tcl}{3.0cm}
    \newcommand{\tcb}{3cm}
    \newcommand{\tcr}{4cm}
    \newcommand{\tct}{4.2cm}
    \newcommand{\thl}{3.0cm}
    \newcommand{\thb}{0.0cm}
    \newcommand{\thr}{3cm}
    \newcommand{\tht}{2cm}
    \setlength{\tabcolsep}{-1pt}
    \renewcommand{\arraystretch}{0}
    \begin{tabular}{@{}cccc:cccc@{}}
        \includegraphics[width=\sizea, trim={\tale} {\taba} {\tare} {\tata},clip]{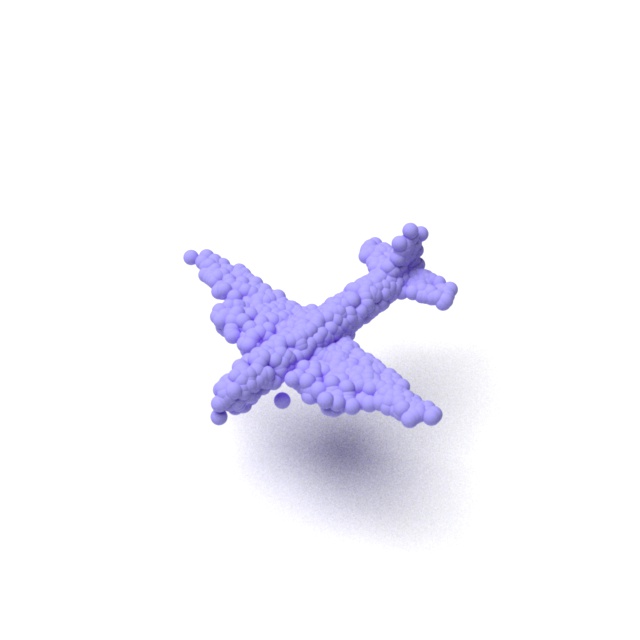}&
        \includegraphics[width=\sizea, trim={\tale} {\taba} {\tare} {\tata},clip]{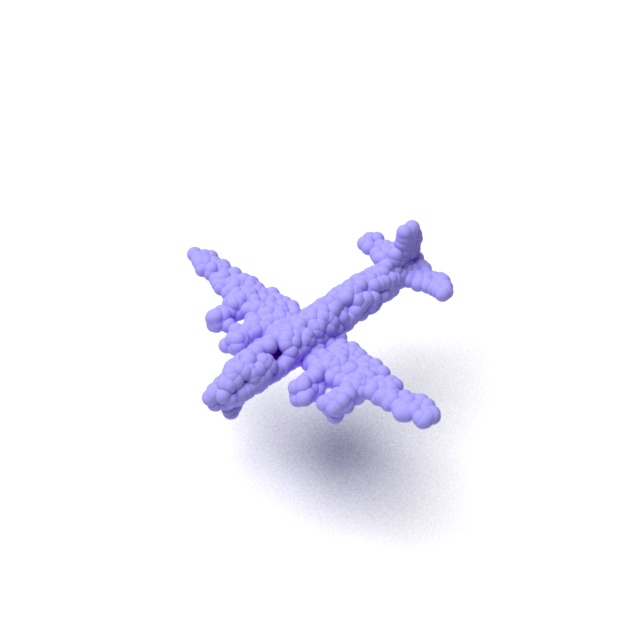}&
        \includegraphics[width=\sizea, trim={\tale} {\taba} {\tare} {\tata},clip]{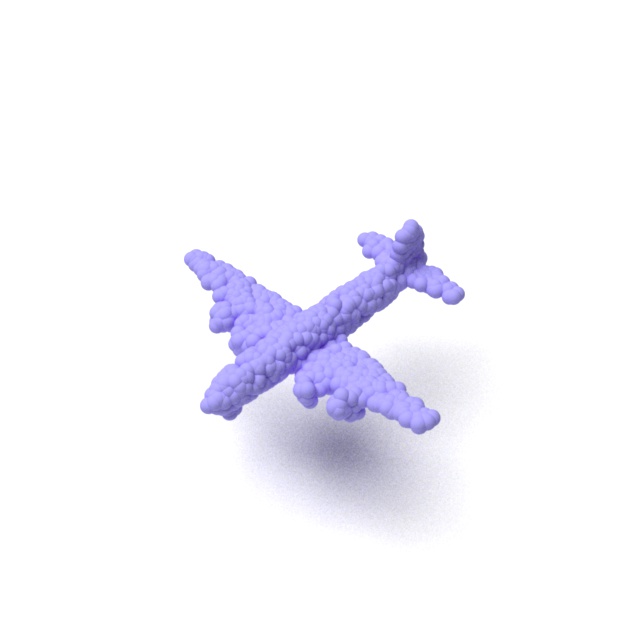}&
        \includegraphics[width=\sizea, trim={\tale} {\taba} {\tare} {\tata},clip]{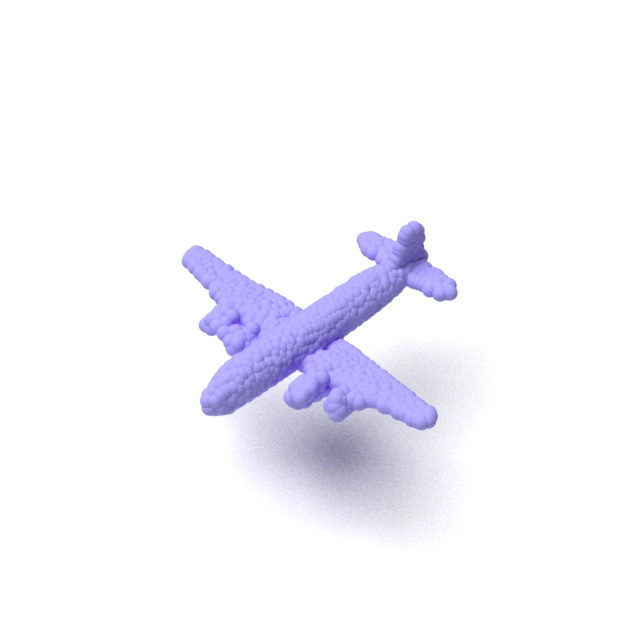}&
        \includegraphics[width=\sizea, trim={\tale} {\taba} {\tare} {\tata},clip]{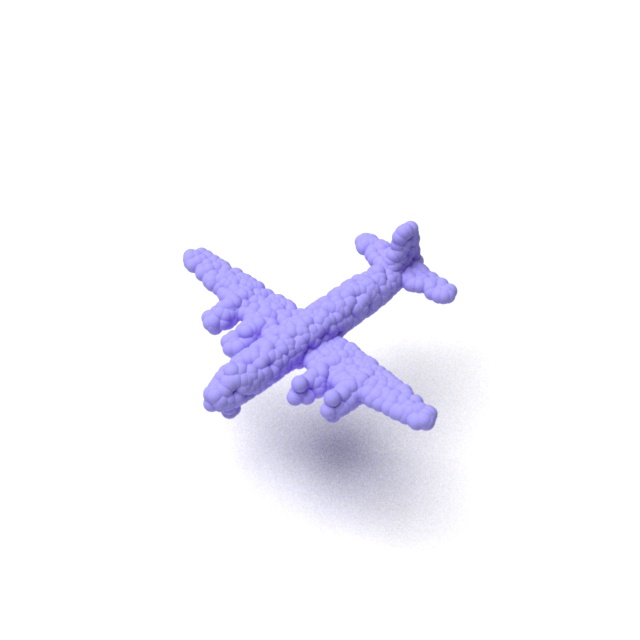}
        \\
        \includegraphics[width=\sizea, trim={\tale} {\taba} {\tare} {\tata},clip]{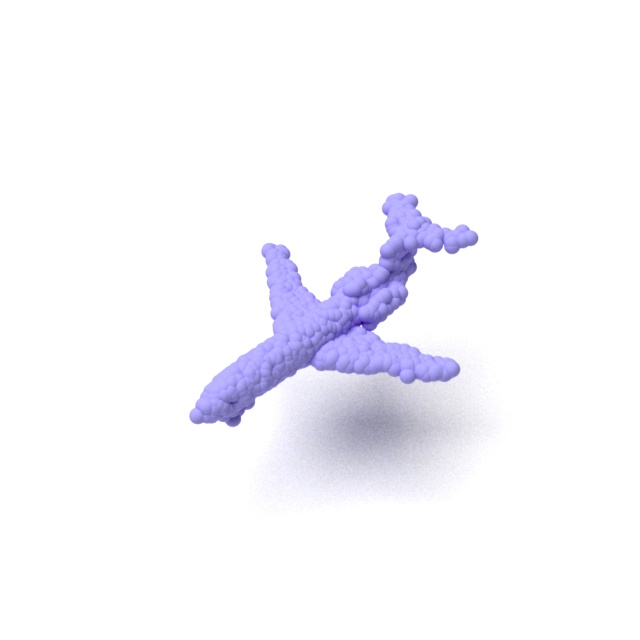}&
        \includegraphics[width=\sizea, trim={\tale} {\taba} {\tare} {\tata},clip]{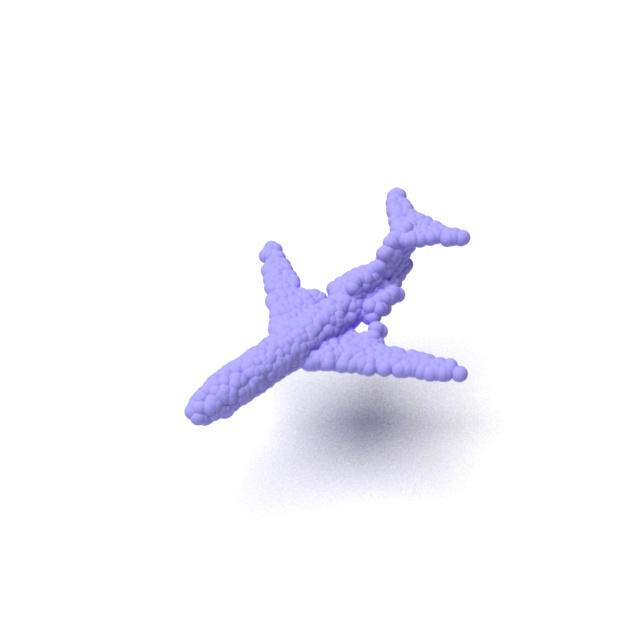}&
        \includegraphics[width=\sizea, trim={\tale} {\taba} {\tare} {\tata},clip]{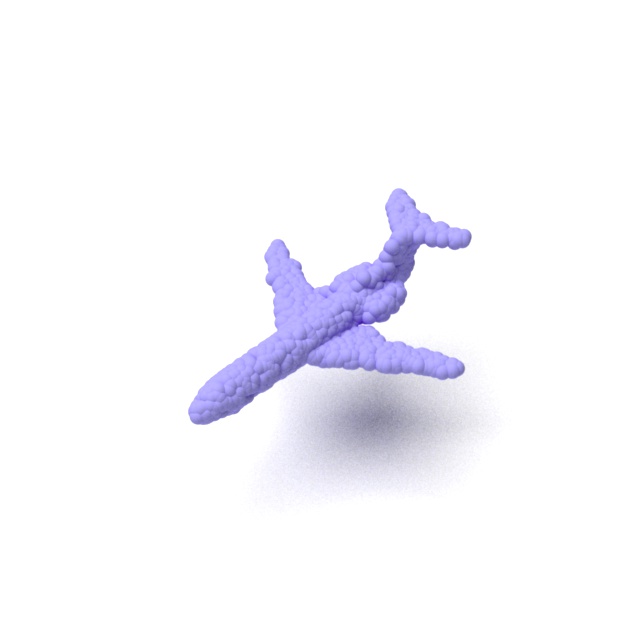}&
        \includegraphics[width=\sizea, trim={\tale} {\taba} {\tare} {\tata},clip]{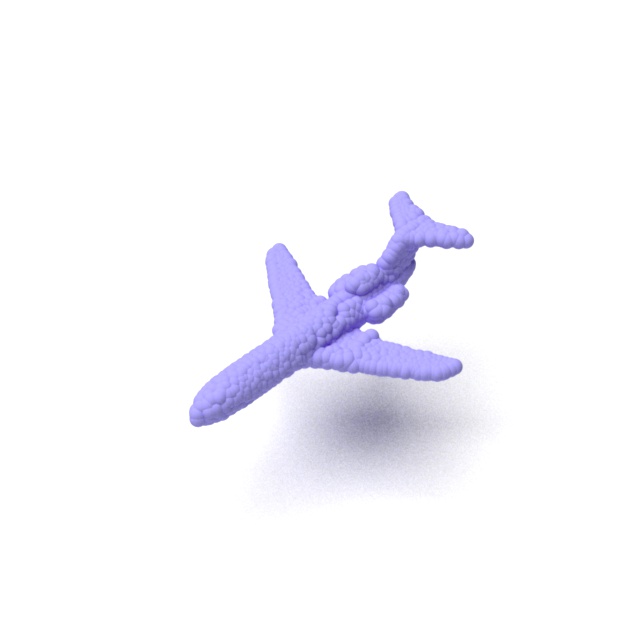}&
        \includegraphics[width=\sizea, trim={\tale} {\taba} {\tare} {\tata},clip]{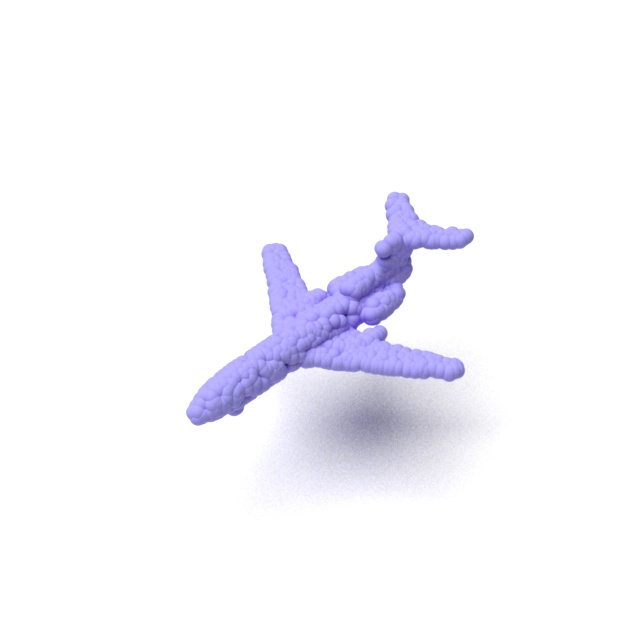}
        \\
        \includegraphics[width=\sizea, trim={\tale} {\taba} {\tare} {\tata},clip]{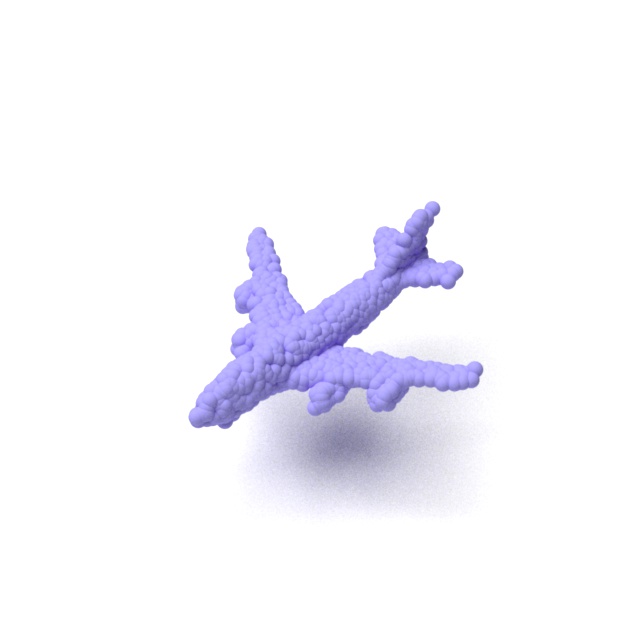}&
        \includegraphics[width=\sizea, trim={\tale} {\taba} {\tare} {\tata},clip]{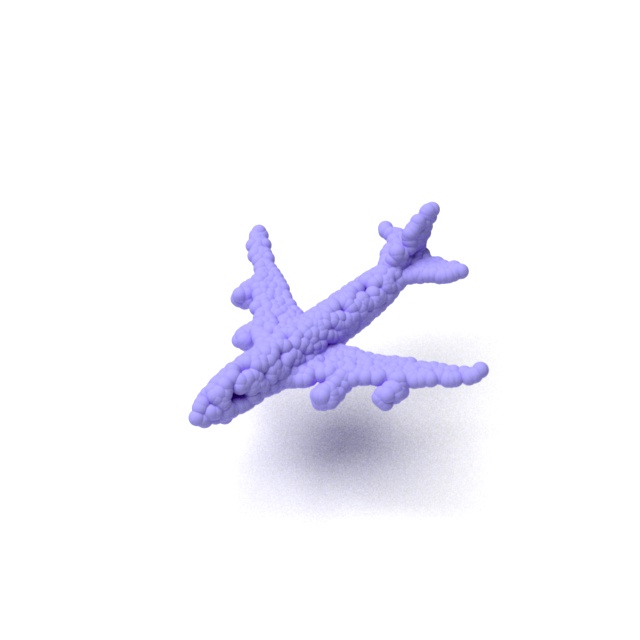}&
        \includegraphics[width=\sizea, trim={\tale} {\taba} {\tare} {\tata},clip]{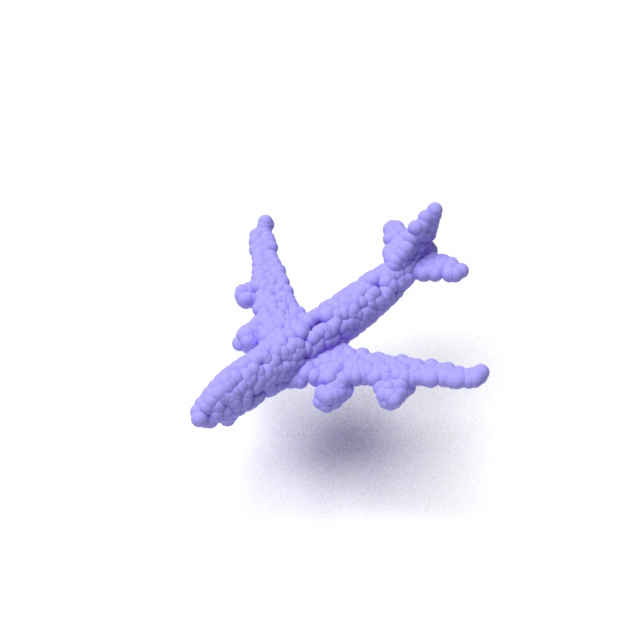}&
        \includegraphics[width=\sizea, trim={\tale} {\taba} {\tare} {\tata},clip]{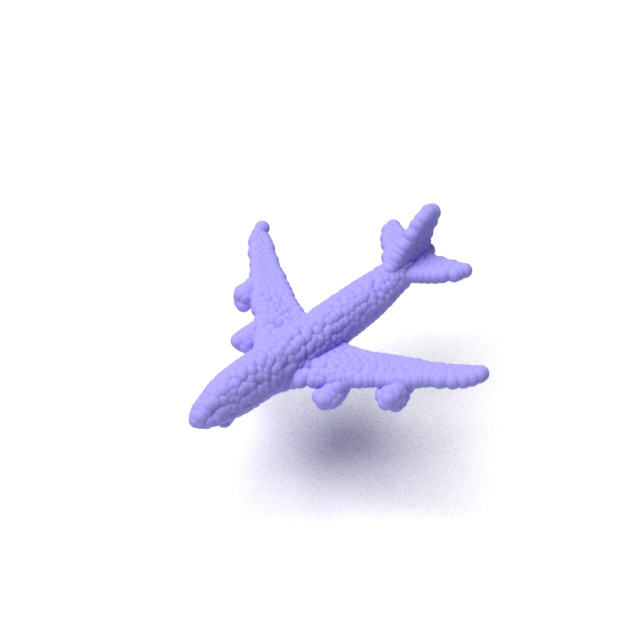}&
        \includegraphics[width=\sizea, trim={\tale} {\taba} {\tare} {\tata},clip]{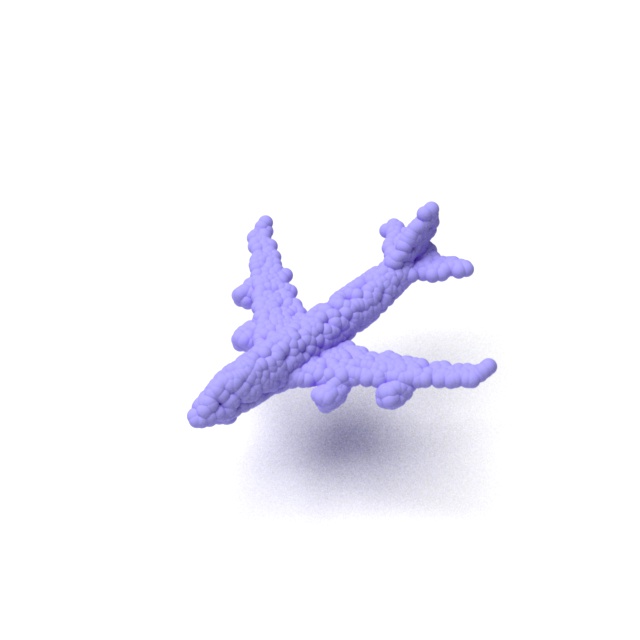}
        \\
        \includegraphics[width=\sizea, trim={\tale} {\tab} {\tare} {\tat},clip]{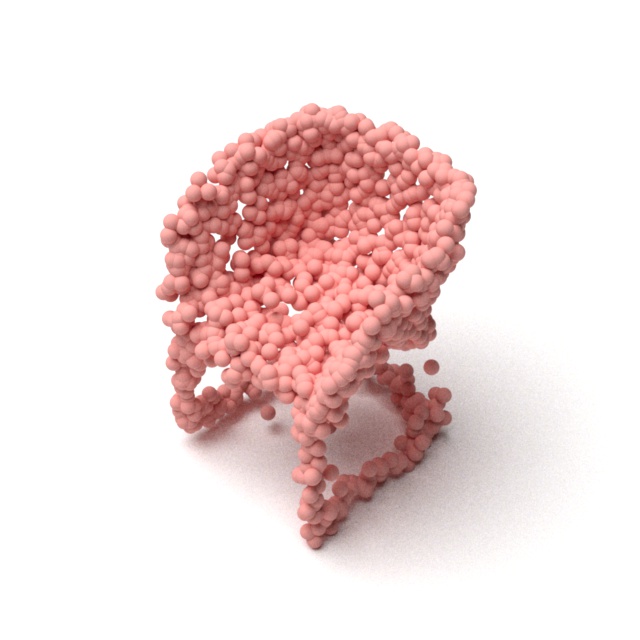}&
        \includegraphics[width=\sizea, trim={\tale} {\tab} {\tare} {\tat},clip]{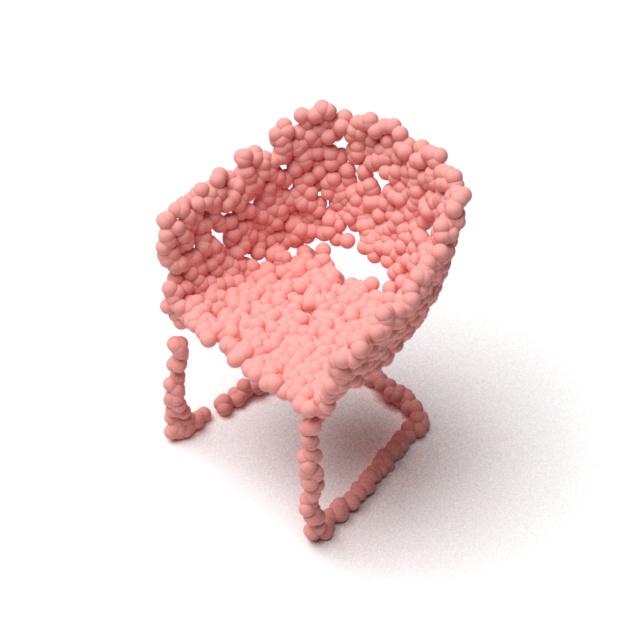}&
        \includegraphics[width=\sizea, trim={\tale} {\tab} {\tare} {\tat},clip]{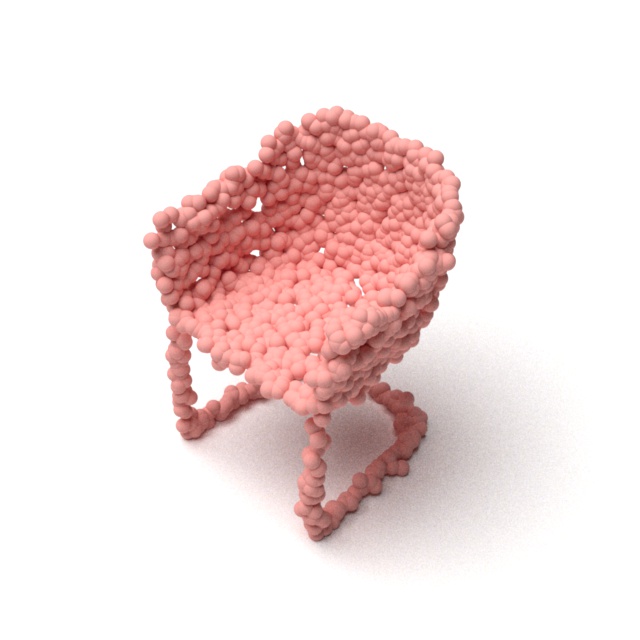}&
        \includegraphics[width=\sizea, trim={\tale} {\tab} {\tare} {\tat},clip]{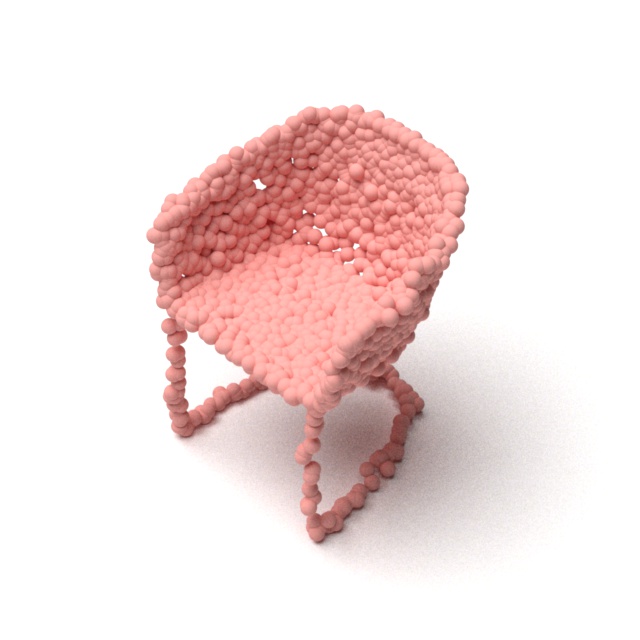}&
        \includegraphics[width=\sizea, trim={\tale} {\tab} {\tare} {\tat},clip]{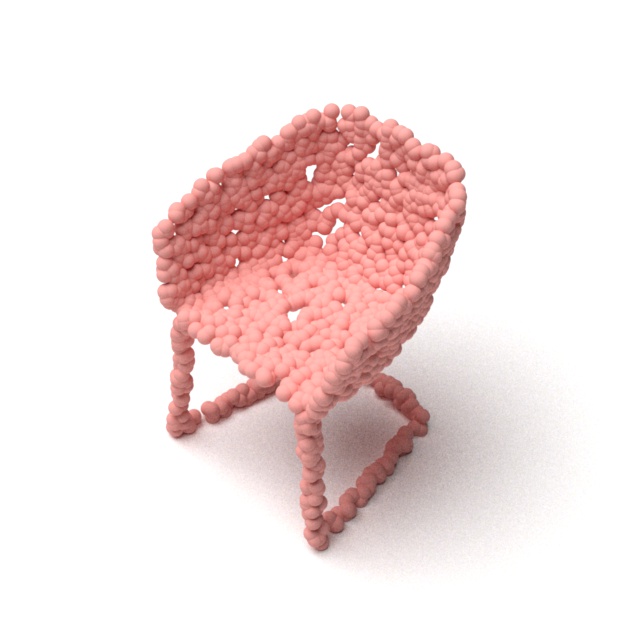}
        \\
        \includegraphics[width=\sizea, trim={\tale} {\tab} {\tare} {\tat},clip]{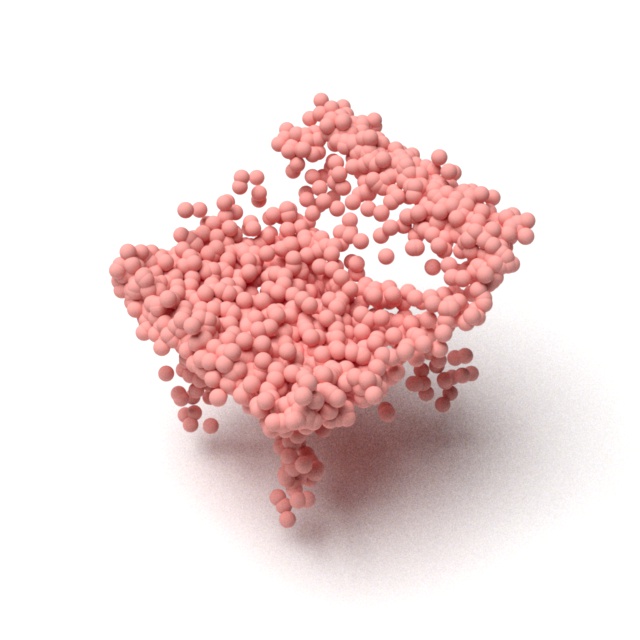}&
        \includegraphics[width=\sizea, trim={\tale} {\tab} {\tare} {\tat},clip]{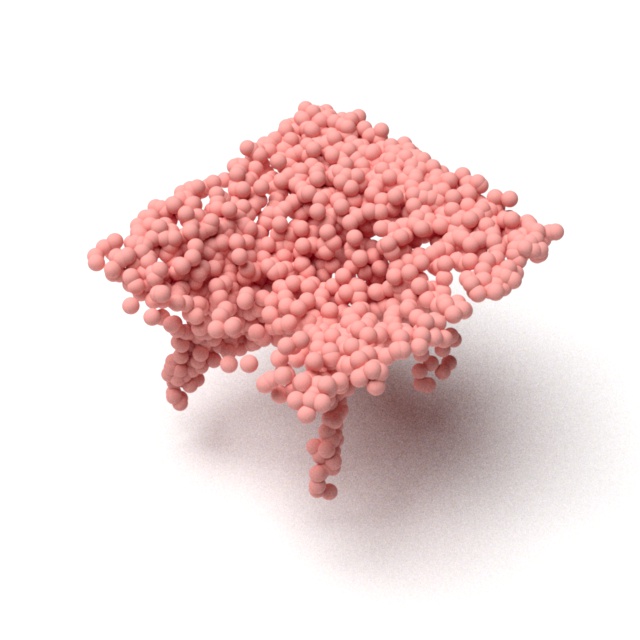}&
        \includegraphics[width=\sizea, trim={\tale} {\tab} {\tare} {\tat},clip]{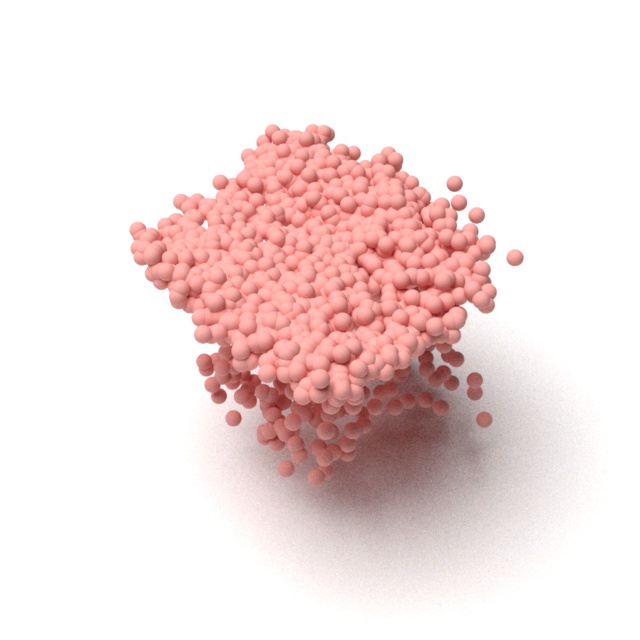}&
        \includegraphics[width=\sizea, trim={\tale} {\tab} {\tare} {\tat},clip]{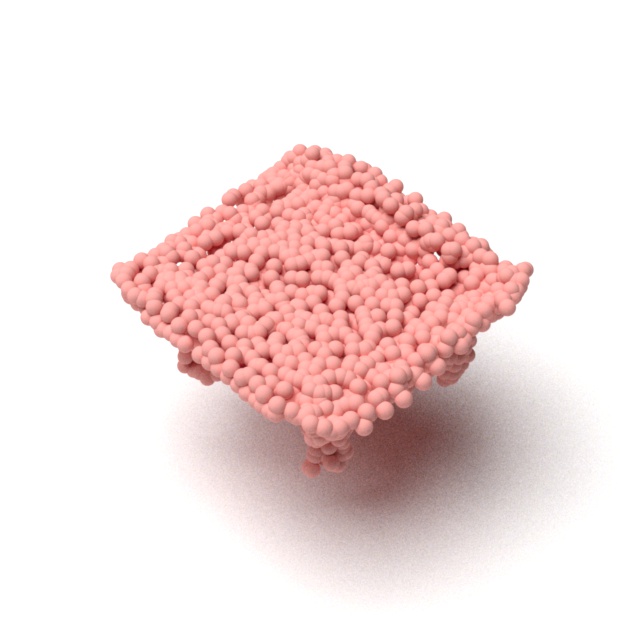}&
        \includegraphics[width=\sizea, trim={\tale} {\tab} {\tare} {\tat},clip]{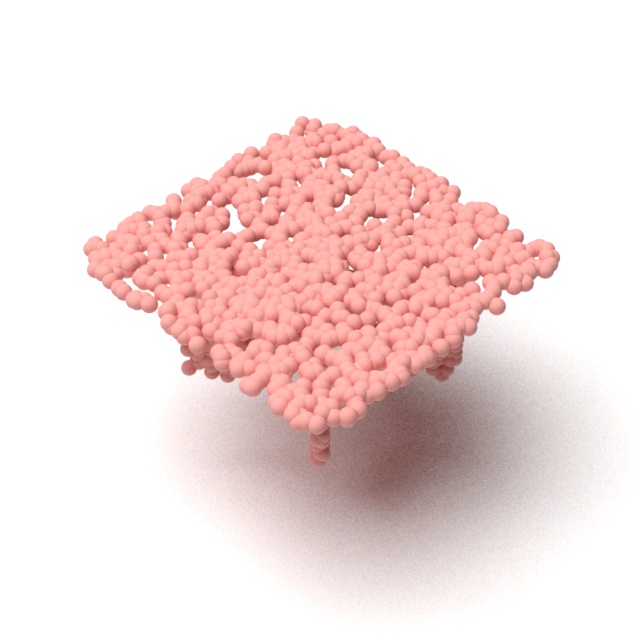}
        \\
        \includegraphics[width=\sizea, trim={\tale} {\tab} {\tare} {\tat},clip]{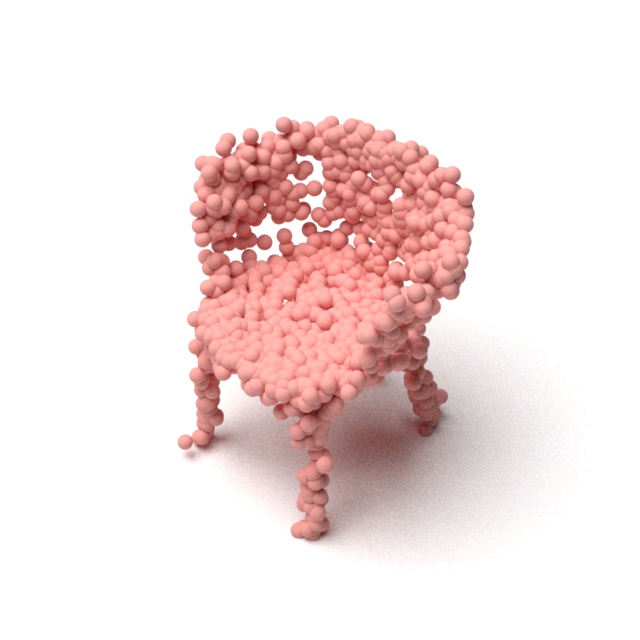}&
        \includegraphics[width=\sizea, trim={\tale} {\tab} {\tare} {\tat},clip]{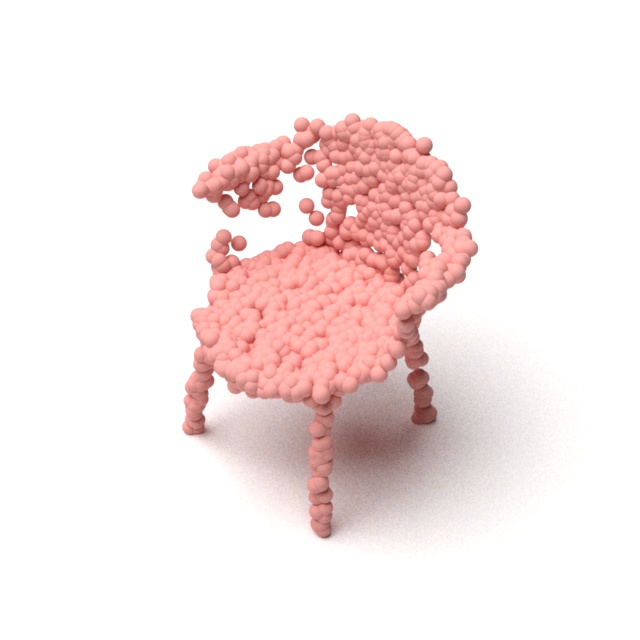}&
        \includegraphics[width=\sizea, trim={\tale} {\tab} {\tare} {\tat},clip]{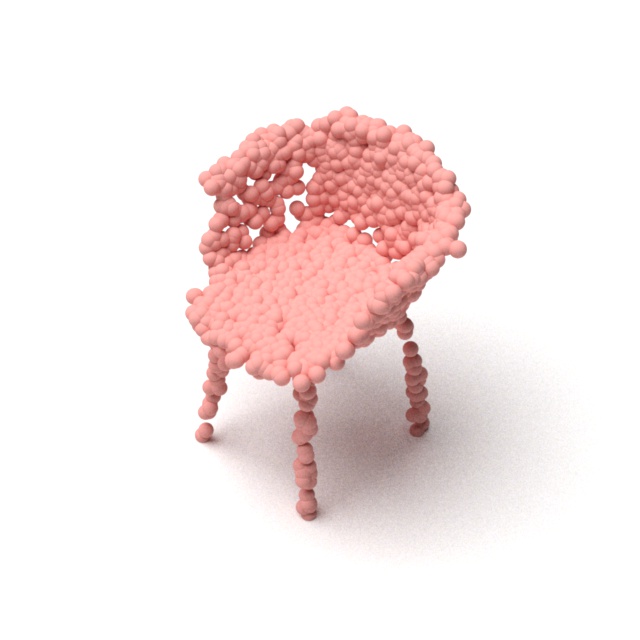}&
        \includegraphics[width=\sizea, trim={\tale} {\tab} {\tare} {\tat},clip]{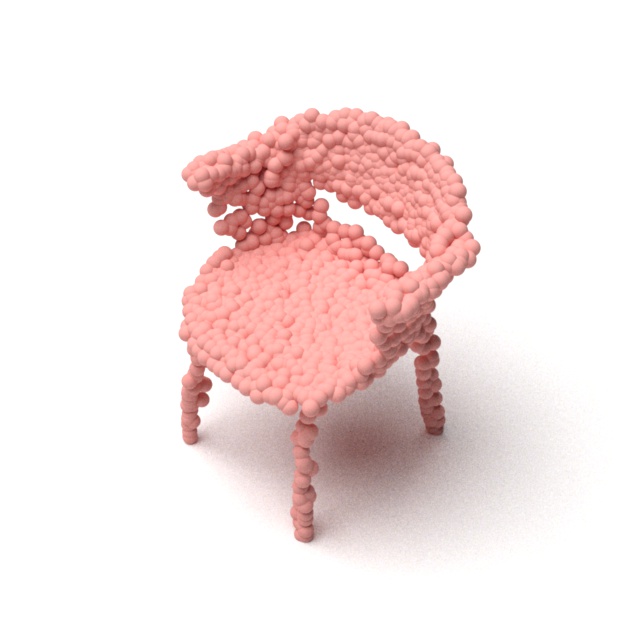}&
        \includegraphics[width=\sizea, trim={\tale} {\tab} {\tare} {\tat},clip]{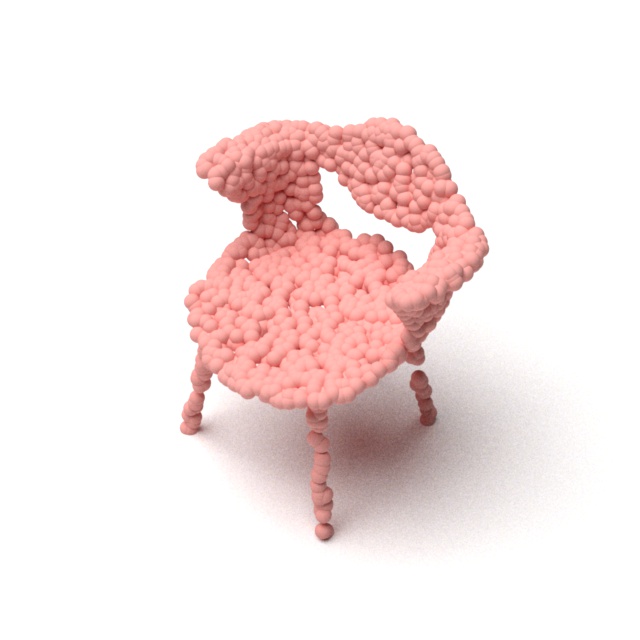}
        \\
        \includegraphics[width=\sizea, trim={\tale} {\taba} {\tare} {\tata},clip]{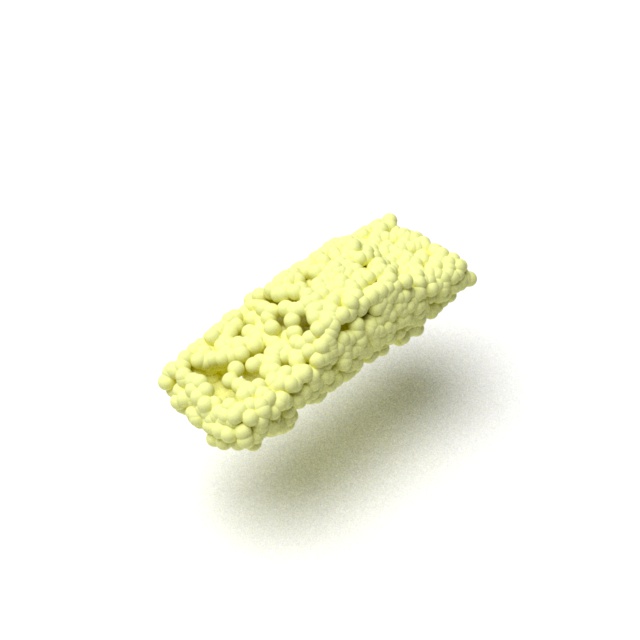}&
        \includegraphics[width=\sizea, trim={\tale} {\taba} {\tare} {\tata},clip]{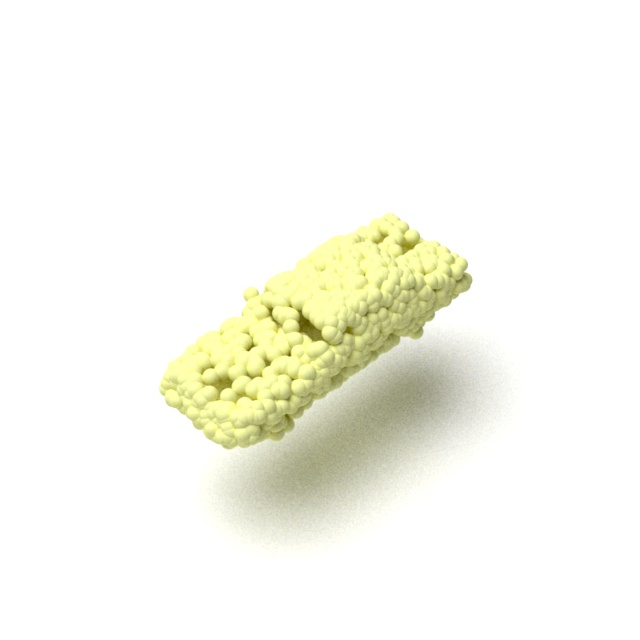}&
        \includegraphics[width=\sizea, trim={\tale} {\taba} {\tare} {\tata},clip]{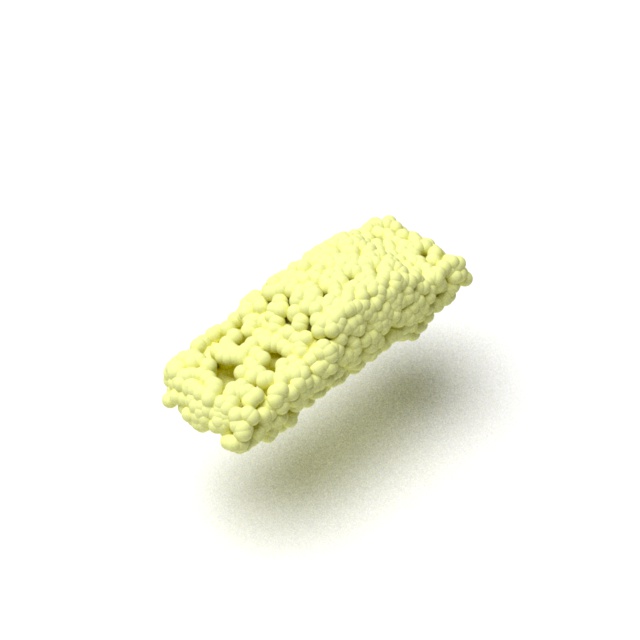}&
        \includegraphics[width=\sizea, trim={\tale} {\taba} {\tare} {\tata},clip]{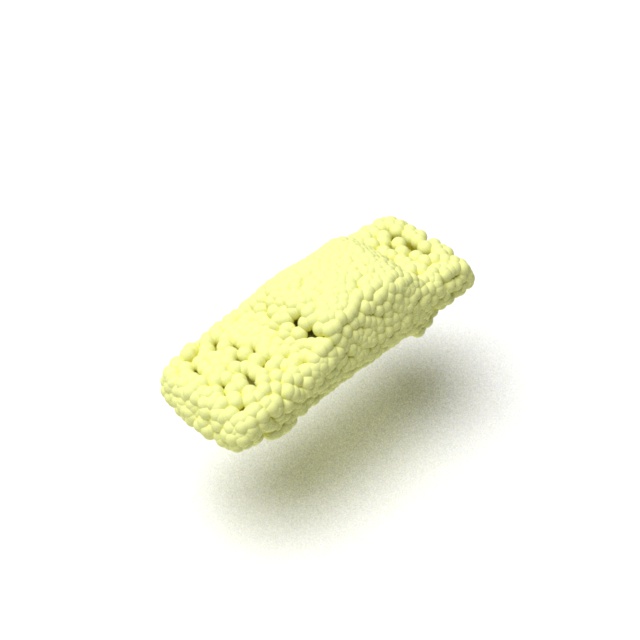}&
        \includegraphics[width=\sizea, trim={\tale} {\taba} {\tare} {\tata},clip]{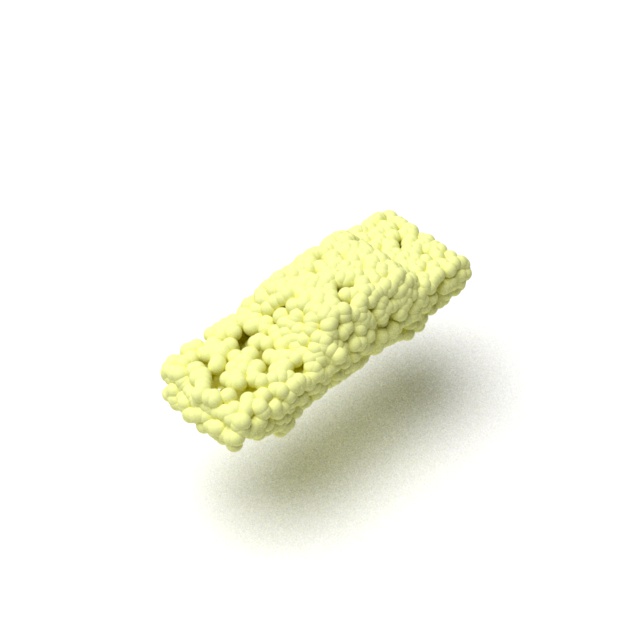}
        \\
        \includegraphics[width=\sizea, trim={\tale} {\taba} {\tare} {\tata},clip]{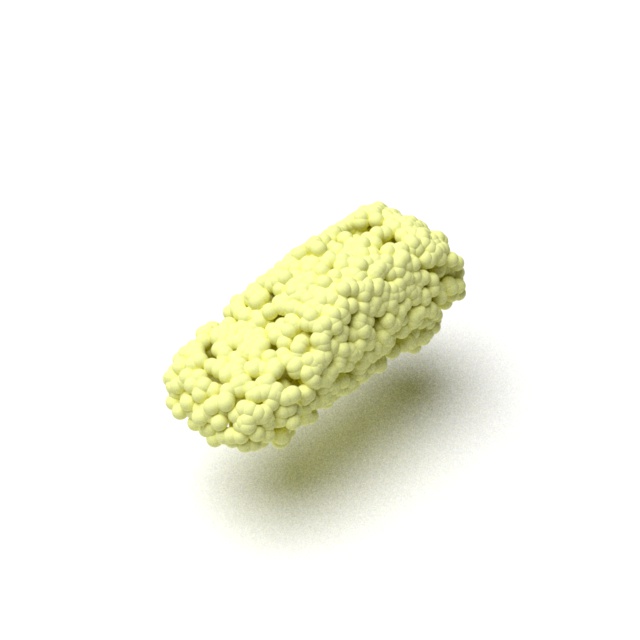}&
        \includegraphics[width=\sizea, trim={\tale} {\taba} {\tare} {\tata},clip]{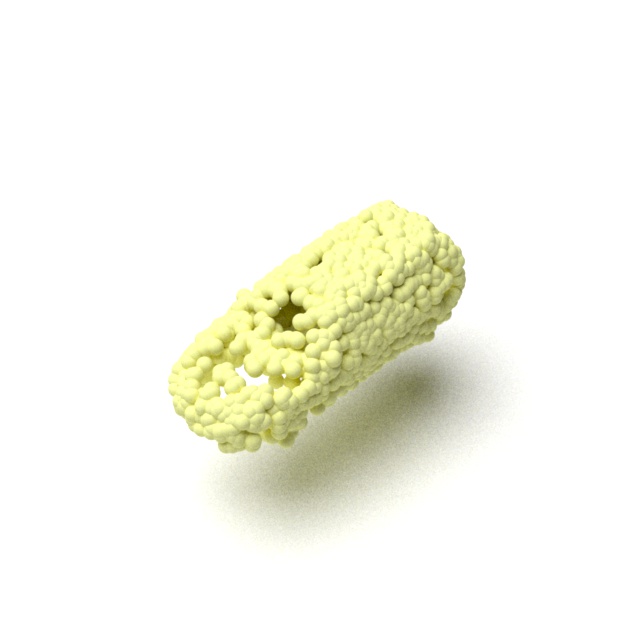}&
        \includegraphics[width=\sizea, trim={\tale} {\taba} {\tare} {\tata},clip]{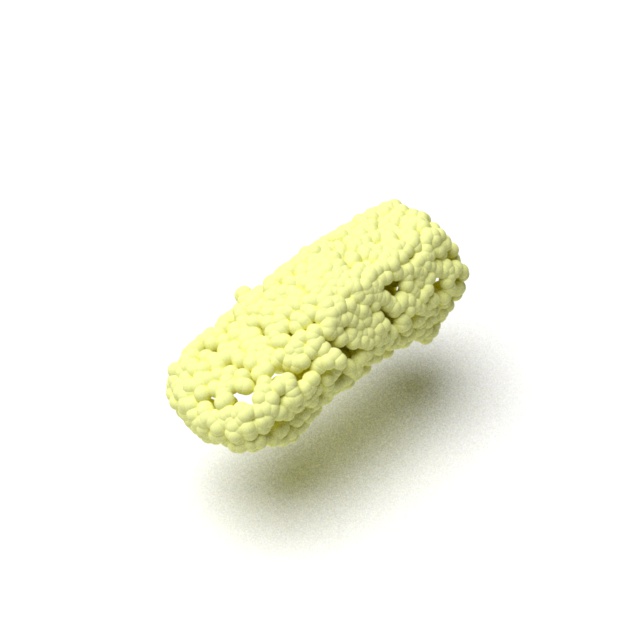}&
        \includegraphics[width=\sizea, trim={\tale} {\taba} {\tare} {\tata},clip]{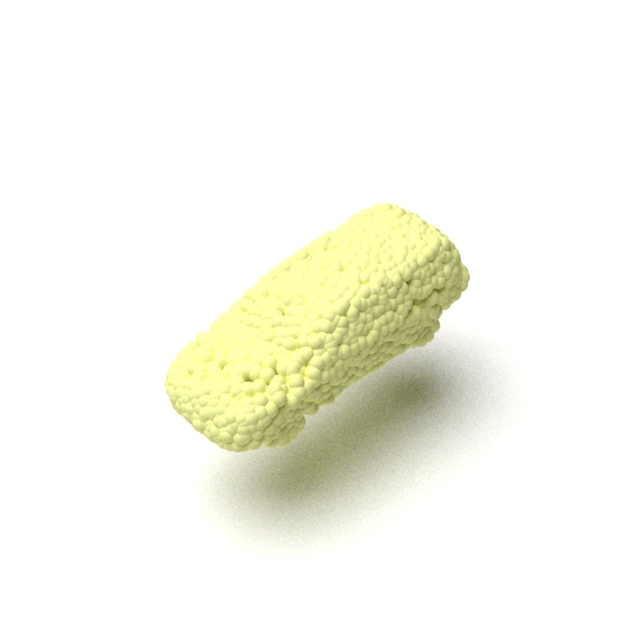}&
        \includegraphics[width=\sizea, trim={\tale} {\taba} {\tare} {\tata},clip]{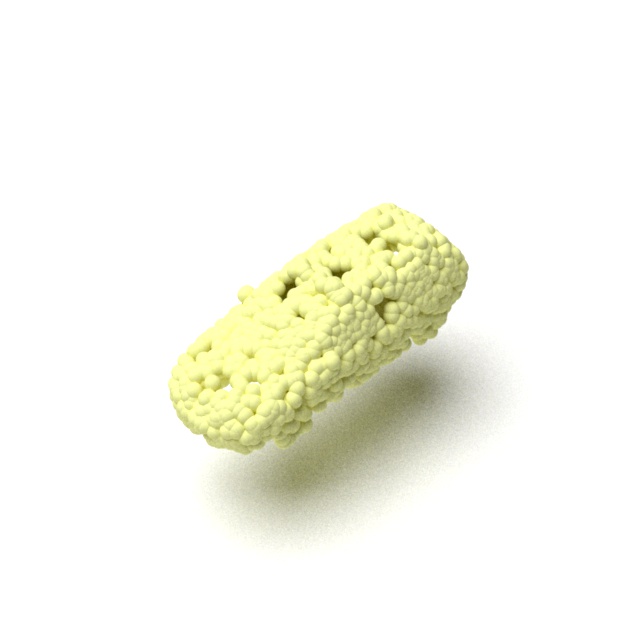}
        \\
        \includegraphics[width=\sizea, trim={\tale} {\taba} {\tare} {\tata},clip]{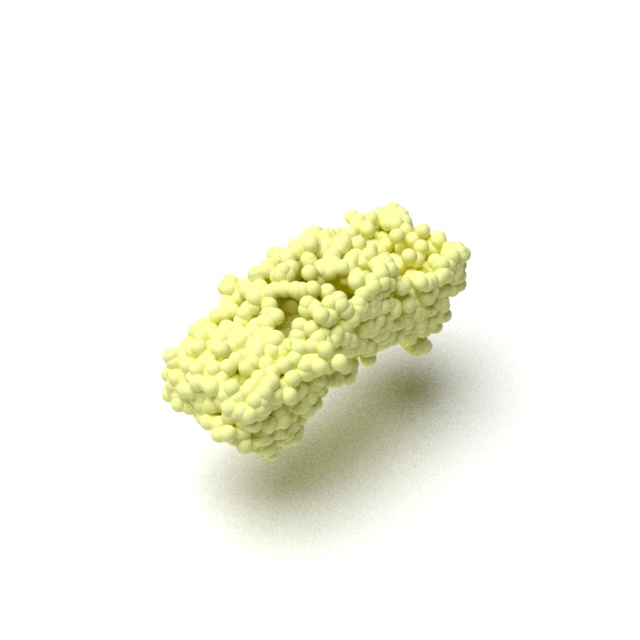}&
        \includegraphics[width=\sizea, trim={\tale} {\taba} {\tare} {\tata},clip]{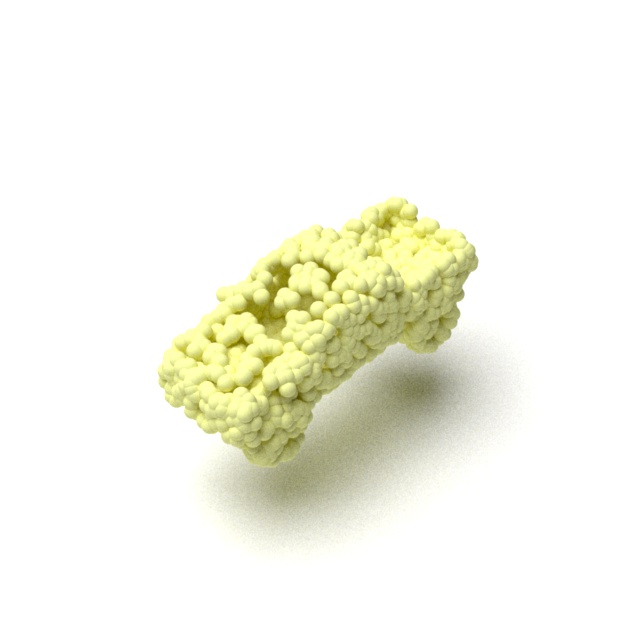}&
        \includegraphics[width=\sizea, trim={\tale} {\taba} {\tare} {\tata},clip]{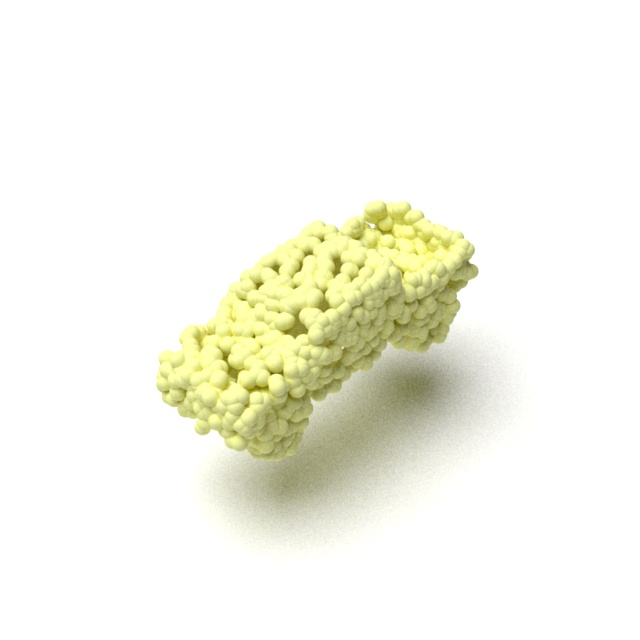}&
        \includegraphics[width=\sizea, trim={\tale} {\taba} {\tare} {\tata},clip]{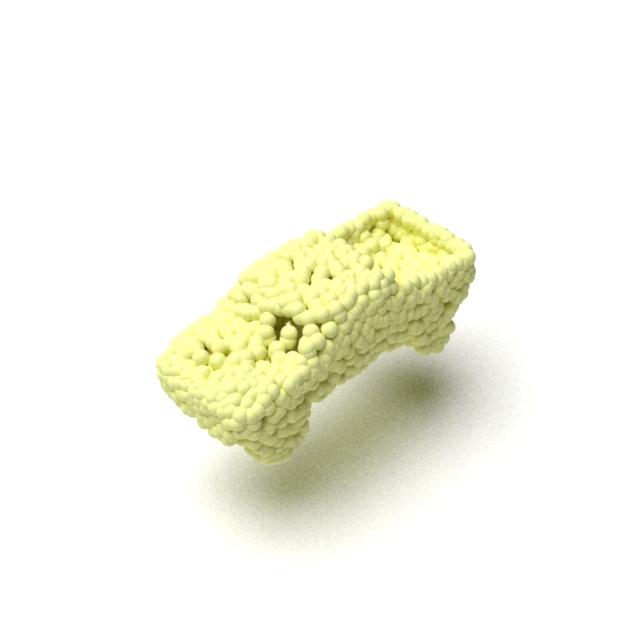}&
        \includegraphics[width=\sizea, trim={\tale} {\taba} {\tare} {\tata},clip]{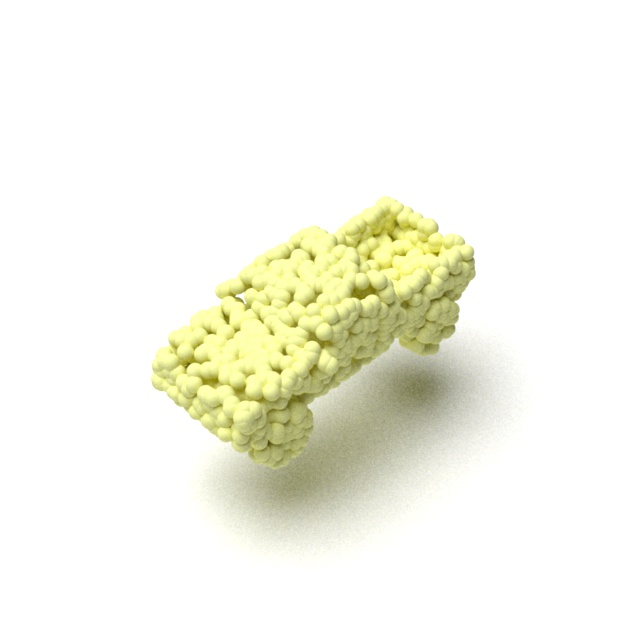}
        \\
        PF & ShapeGF & DPM & Ours & Input
    \end{tabular}
    \end{center}
    \caption{Auto-encoding (reconstruction) results. We also shown results from PF (PointFlow)~\cite{yang2019pointflow}, ShapeGF~\cite{cai2020learning}, and DPM~\cite{luo2021diffusion} on the left for comparison.
    }
    \vspace{-8.5em}
    \label{fig:sup:recon}
\end{figure}

In Figure~\ref{fig:sup:recon}, we show more auto-encoding results. Thanks to the context-rich codebook, our model is able to reconstruct shapes with better local details.
Moreover, the points are more uniformly distributed among the surface.

\clearpage
\section{Qualitative Results of Unconditional Generation}
\label{sec:uncon_gen}
\begin{figure}[h]
    \begin{center}
    \newcommand{\sizea}{0.165\linewidth}
    \newcommand{\sizeb}{0.165\linewidth}
    \newcommand{\sizec}{0.165\linewidth}
    \newcommand{\tare}{5cm}
    \newcommand{\tale}{4cm}
    \newcommand{\tal}{3.5cm}
    \newcommand{\tab}{3.5cm}
    \newcommand{\tar}{3.5cm}
    \newcommand{\tat}{7.5cm}
    \newcommand{\tcl}{3.0cm}
    \newcommand{\tcb}{3cm}
    \newcommand{\tcr}{4cm}
    \newcommand{\tct}{4.2cm}
    \newcommand{\thl}{3.0cm}
    \newcommand{\thb}{0.0cm}
    \newcommand{\thr}{3cm}
    \newcommand{\tht}{2cm}
    \setlength{\tabcolsep}{0pt}
    \renewcommand{\arraystretch}{0}
    \begin{tabular}{@{}ccccc:c@{}}
        \includegraphics[width=\sizea, trim={\tale} {\tab} {4.5cm} {\tat},clip]{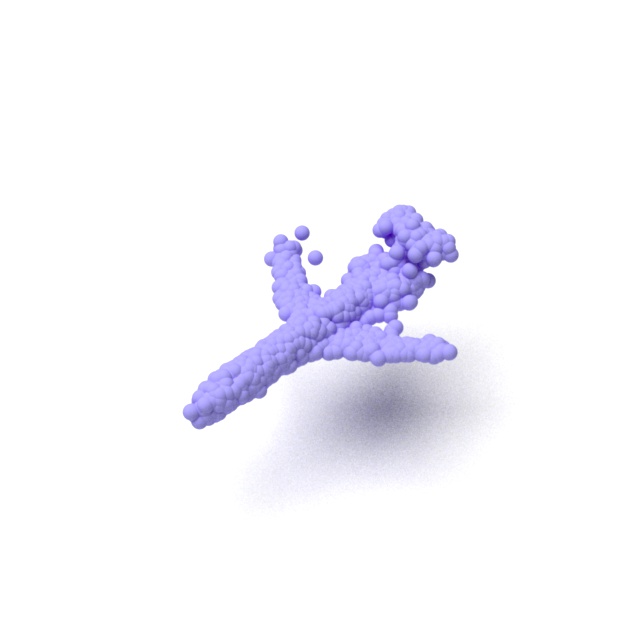} &
        \includegraphics[width=\sizea, trim={\tale} {\tab} {4.5cm} {\tat},clip]{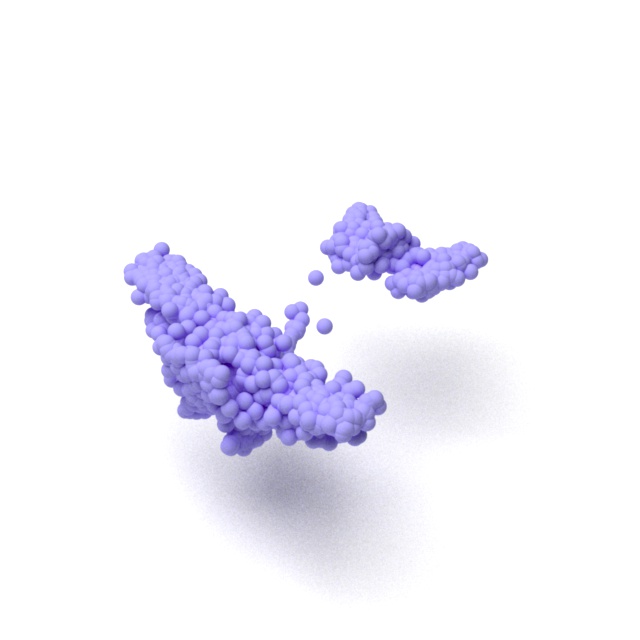} &
        \includegraphics[width=\sizea, trim={\tale} {\tab} {4.5cm} {\tat},clip]{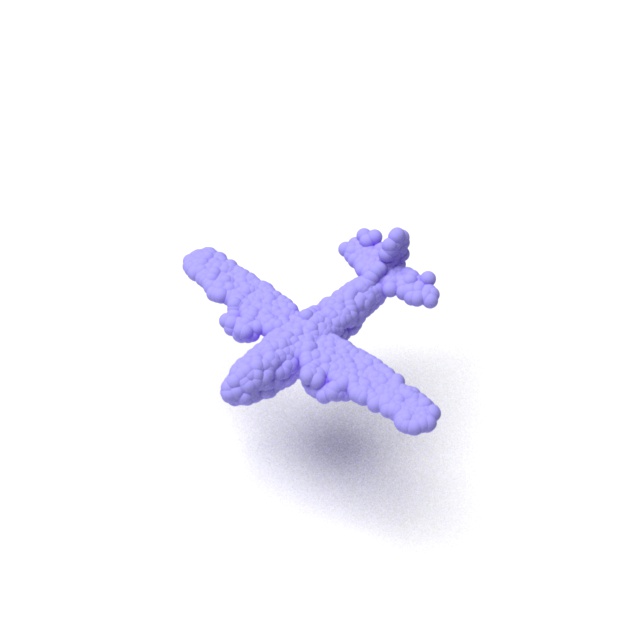} &
        \includegraphics[width=\sizea, trim={\tale} {\tab} {4.5cm} {\tat},clip]{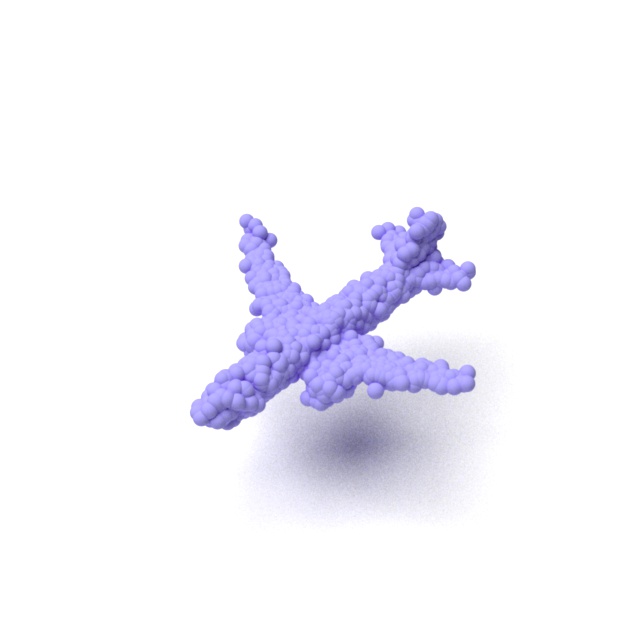} &
        \includegraphics[width=\sizea, trim={\tale} {\tab} {4.5cm} 
        {\tat},clip]{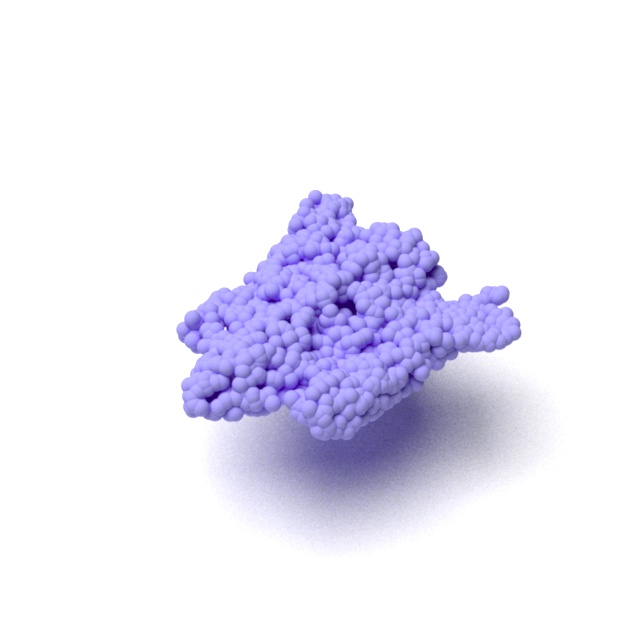} &
        \includegraphics[width=\sizea, trim={\tal} {\tab} {4.5cm} {\tat},clip]{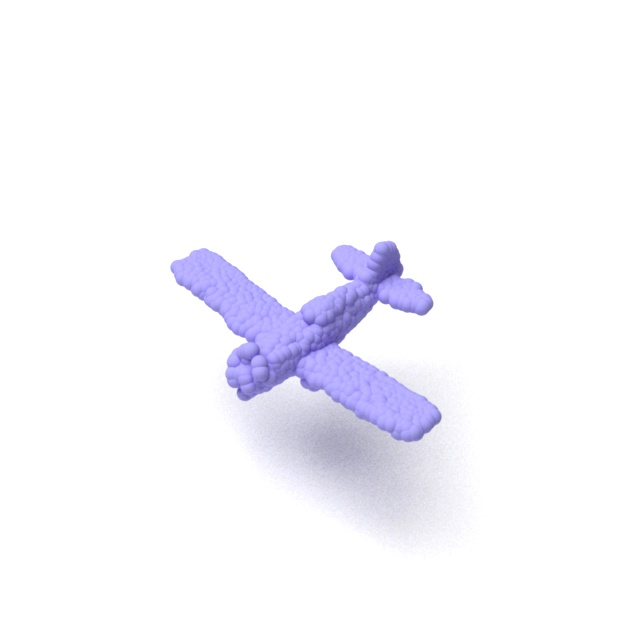} 
        \\
        \includegraphics[width=\sizea, trim={\tale} {\tab} {4.5cm} {\tat},clip]{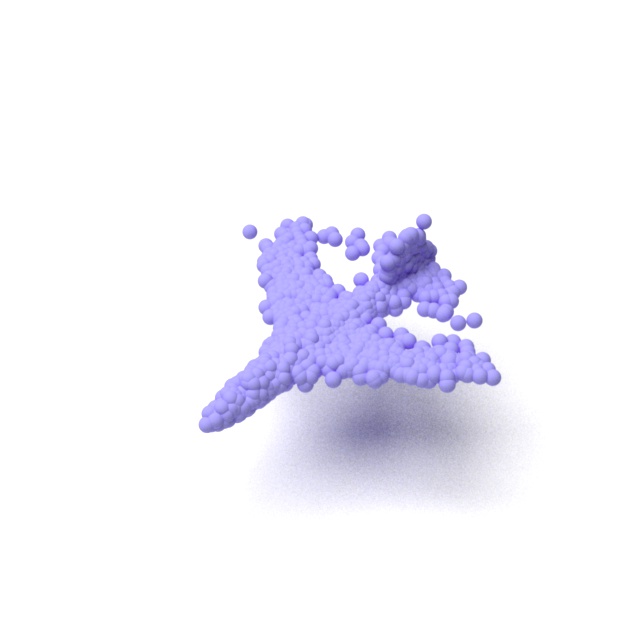} &
        \includegraphics[width=\sizea, trim={\tale} {\tab} {4.5cm} {\tat},clip]{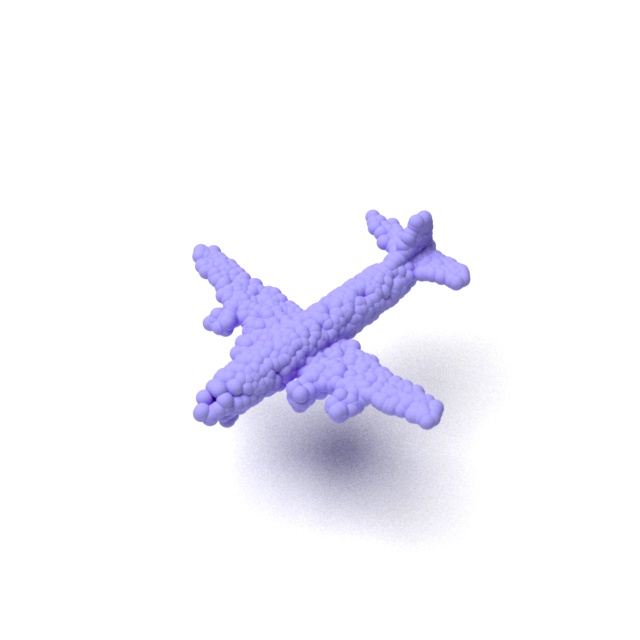} &
        \includegraphics[width=\sizea, trim={\tale} {\tab} {4.5cm} {\tat},clip]{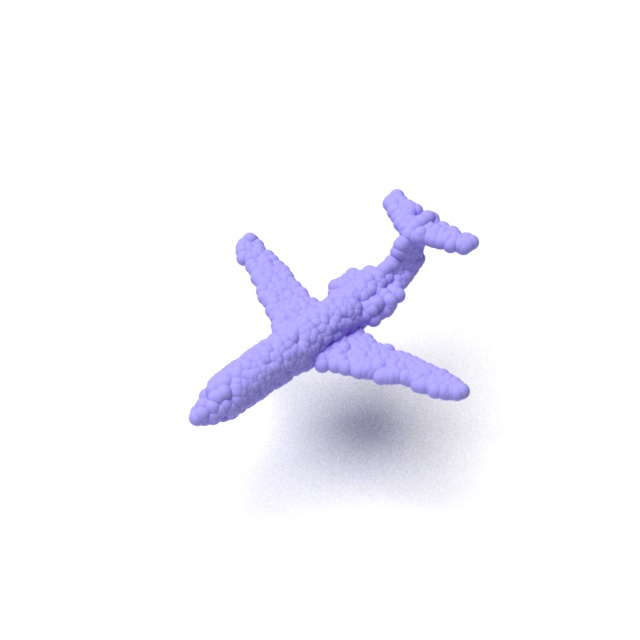} &
        \includegraphics[width=\sizea, trim={\tale} {\tab} {4.5cm} {\tat},clip]{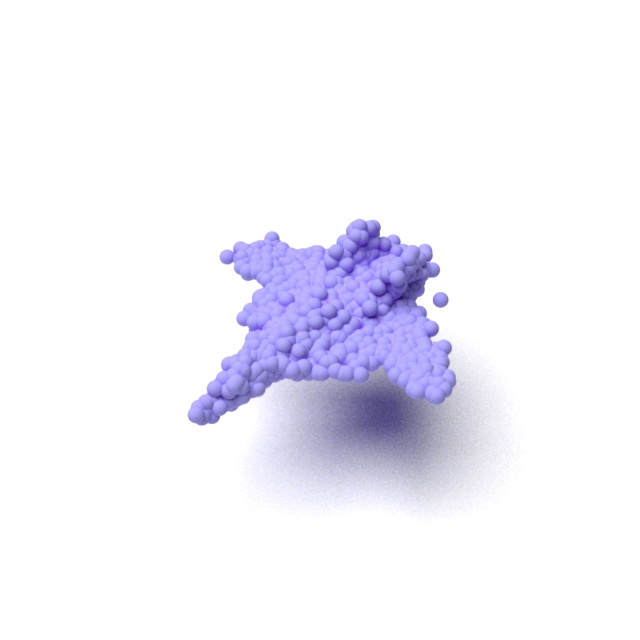} &
        \includegraphics[width=\sizea, trim={\tale} {\tab} {4.5cm} 
        {\tat},clip]{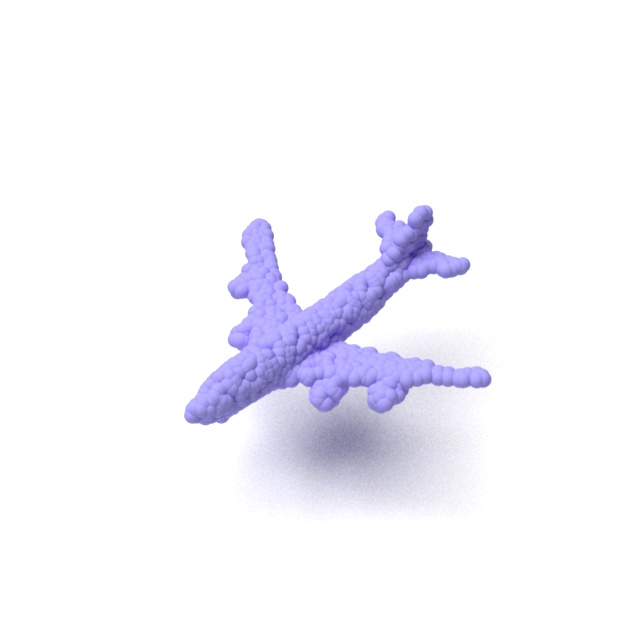} &
        \includegraphics[width=\sizea, trim={\tal} {\tab} {4.5cm} {\tat},clip]{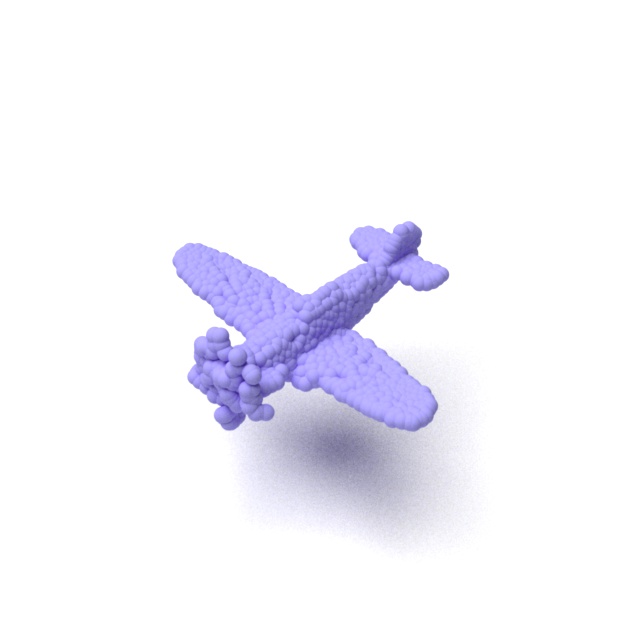} 
        \\
        \includegraphics[width=\sizea, trim={\tale} {\tab} {4.5cm} {\tat},clip]{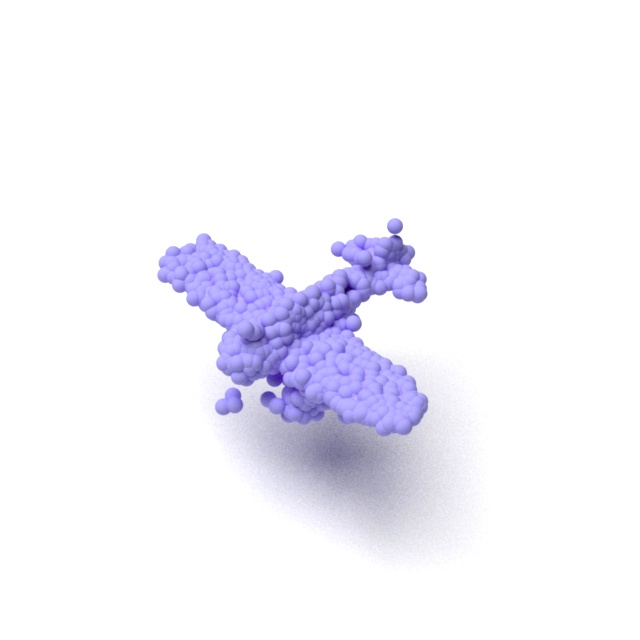} &
        \includegraphics[width=\sizea, trim={\tale} {\tab} {4.5cm} {\tat},clip]{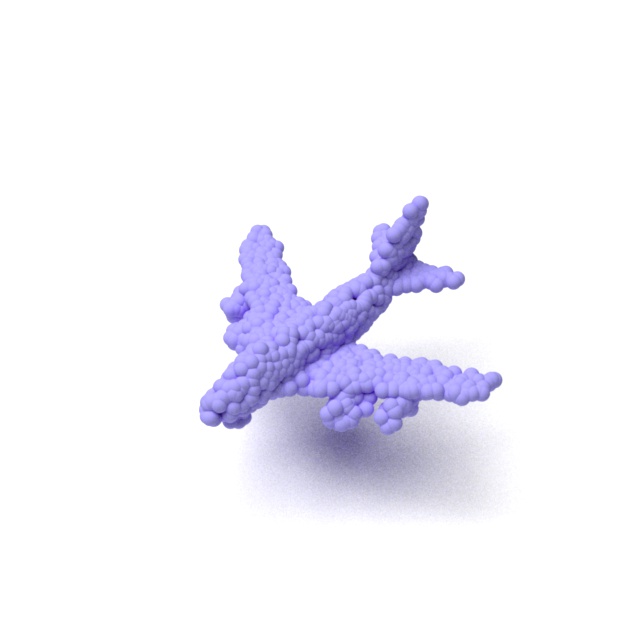} &
        \includegraphics[width=\sizea, trim={\tale} {\tab} {4.5cm} {\tat},clip]{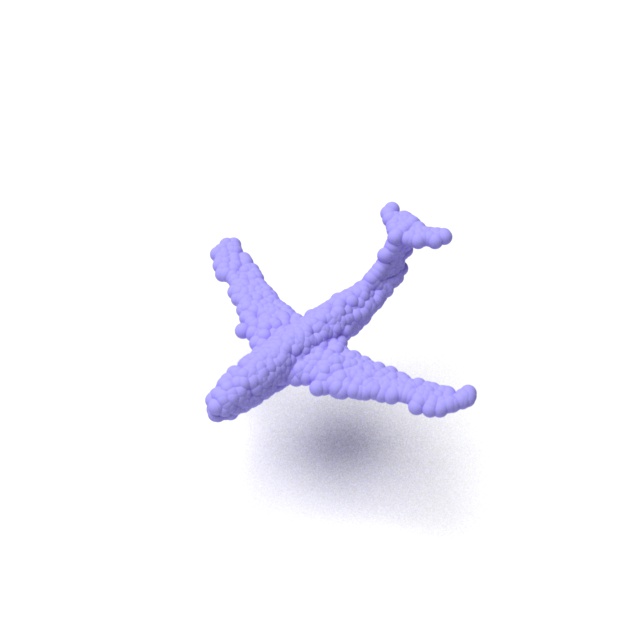} &
        \includegraphics[width=\sizea, trim={\tale} {\tab} {4.5cm} {\tat},clip]{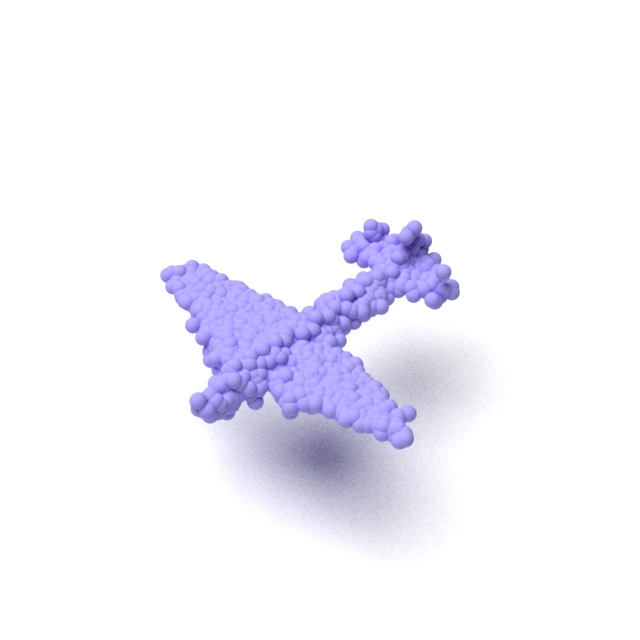} &
        \includegraphics[width=\sizea, trim={\tale} {\tab} {4.5cm} 
        {\tat},clip]{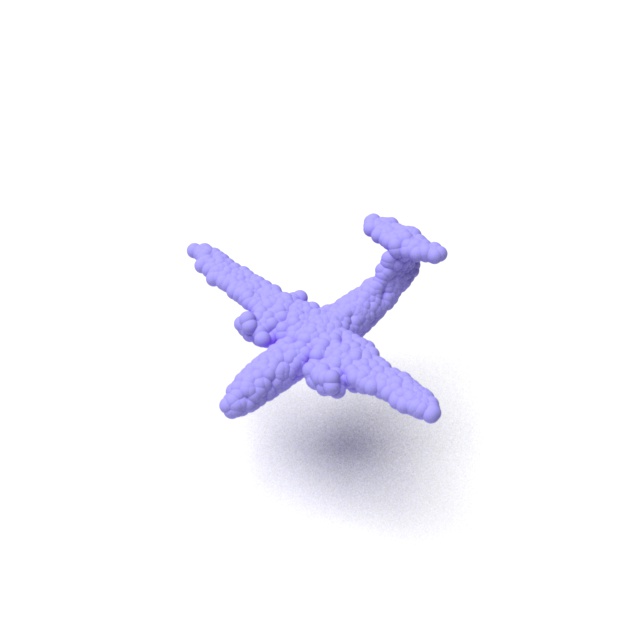} &
        \includegraphics[width=\sizea, trim={\tal} {\tab} {4.5cm} {\tat},clip]{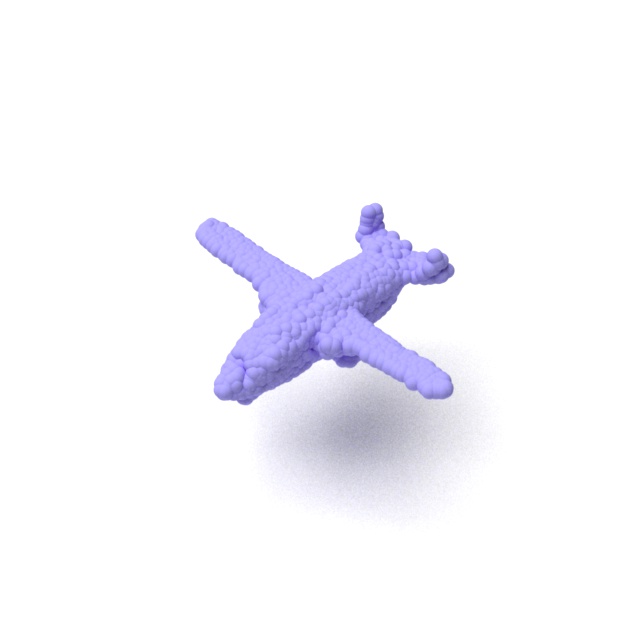} 
        \\
        \includegraphics[width=\sizea, trim={\tale} {\tab} {4.5cm} {\tat},clip]{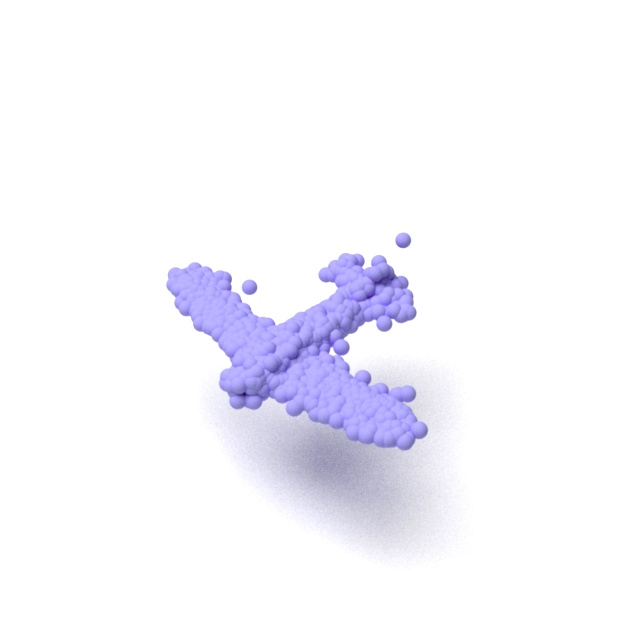} &
        \includegraphics[width=\sizea, trim={\tale} {\tab} {4.5cm} {\tat},clip]{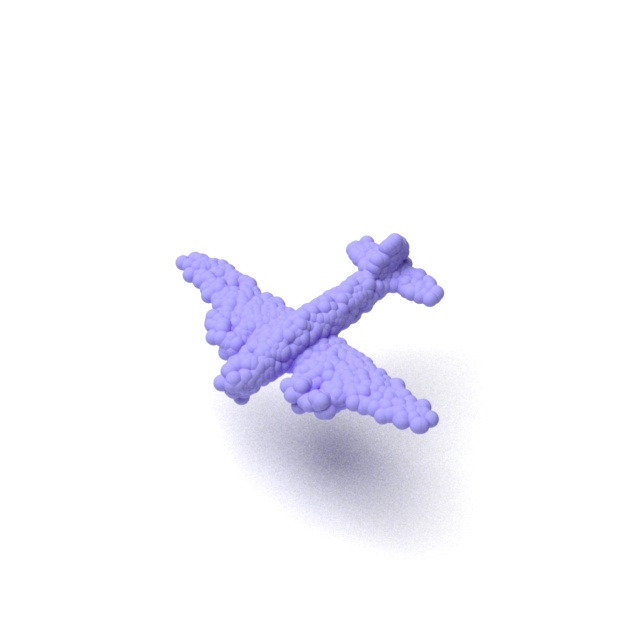} &
        \includegraphics[width=\sizea, trim={\tale} {\tab} {4.5cm} {\tat},clip]{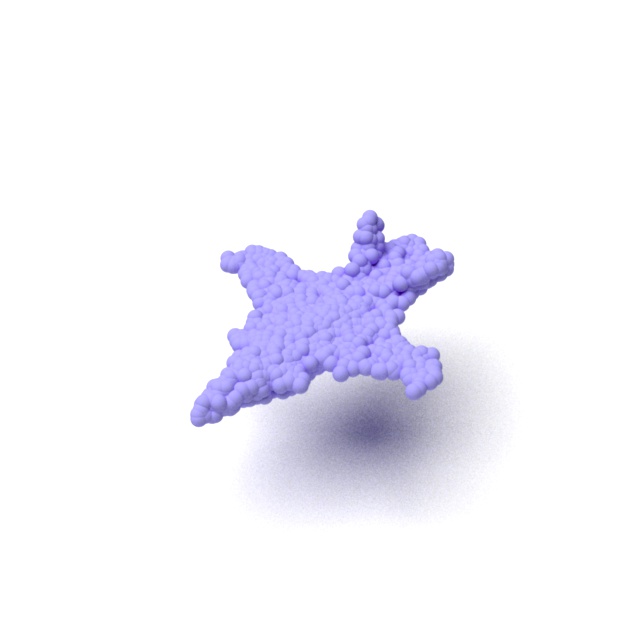} &
        \includegraphics[width=\sizea, trim={\tale} {\tab} {4.5cm} {\tat},clip]{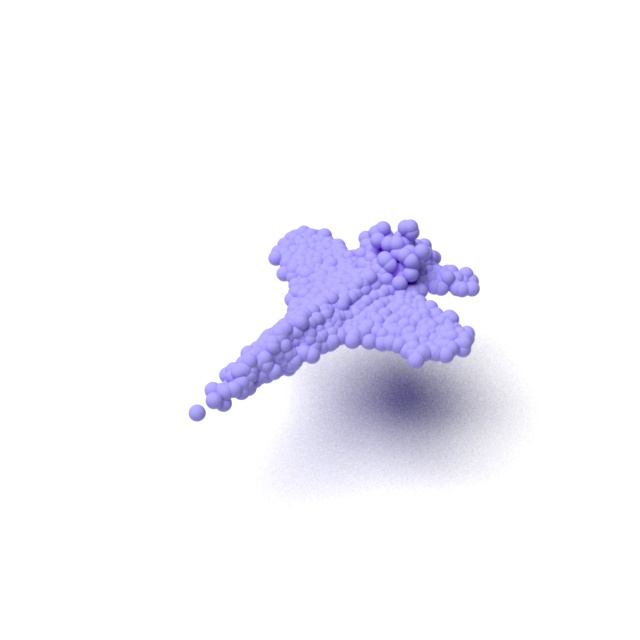} &
        \includegraphics[width=\sizea, trim={\tale} {\tab} {4.5cm} 
        {\tat},clip]{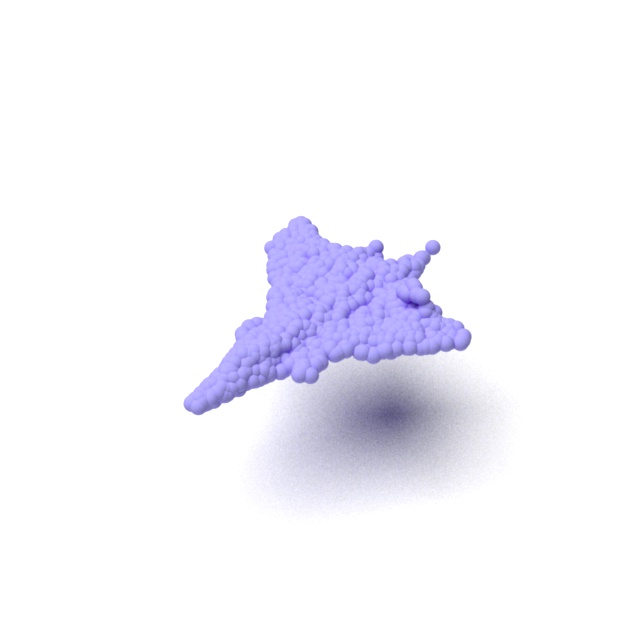} &
        \includegraphics[width=\sizea, trim={\tal} {\tab} {4.5cm} {\tat},clip]{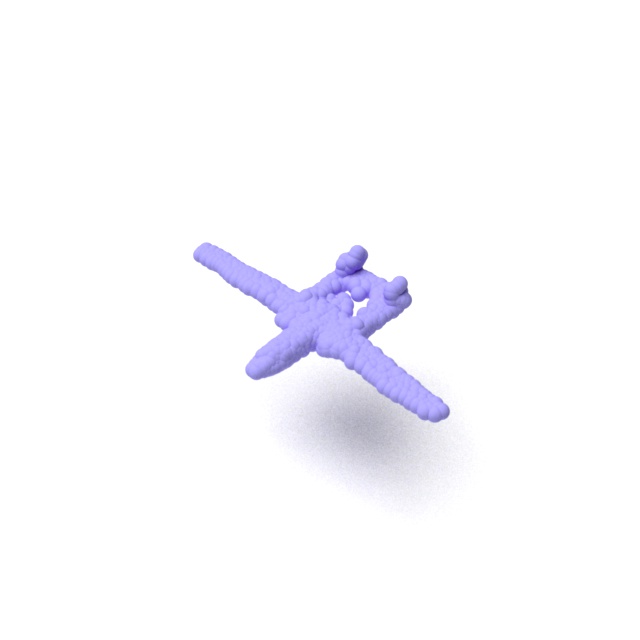} 
        \\
        \includegraphics[width=\sizea, trim={\tale} {\tab} {4.5cm} {\tat},clip]{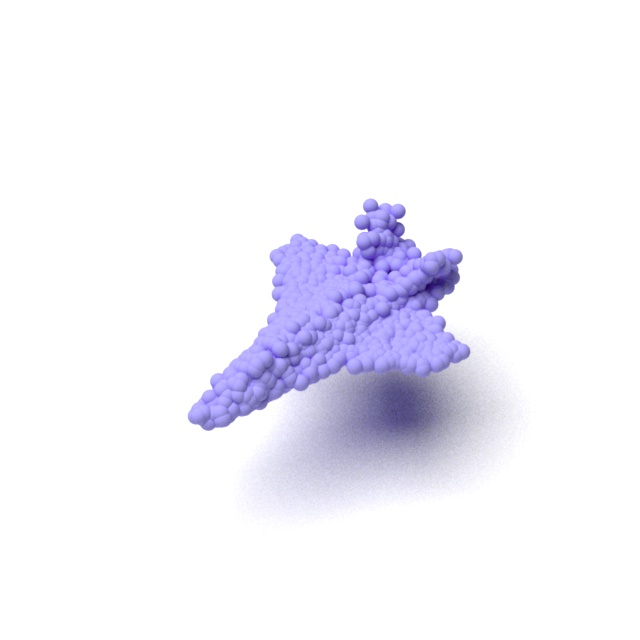} &
        \includegraphics[width=\sizea, trim={\tale} {\tab} {4.5cm} {\tat},clip]{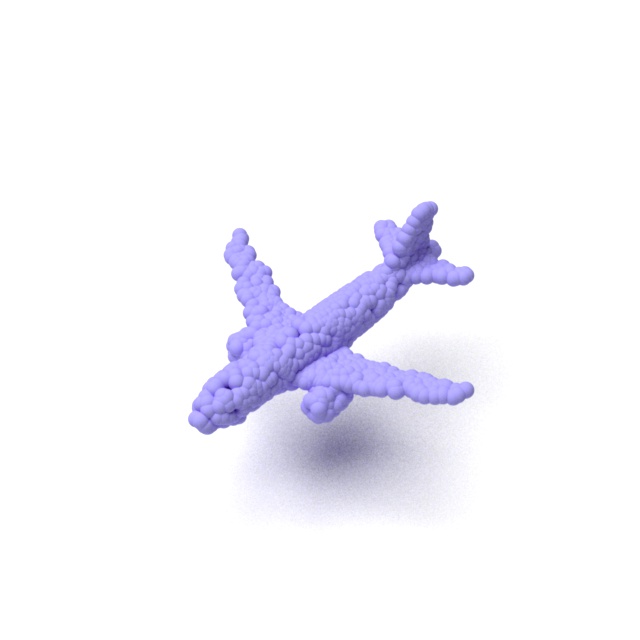} &
        \includegraphics[width=\sizea, trim={\tale} {\tab} {4.5cm} {\tat},clip]{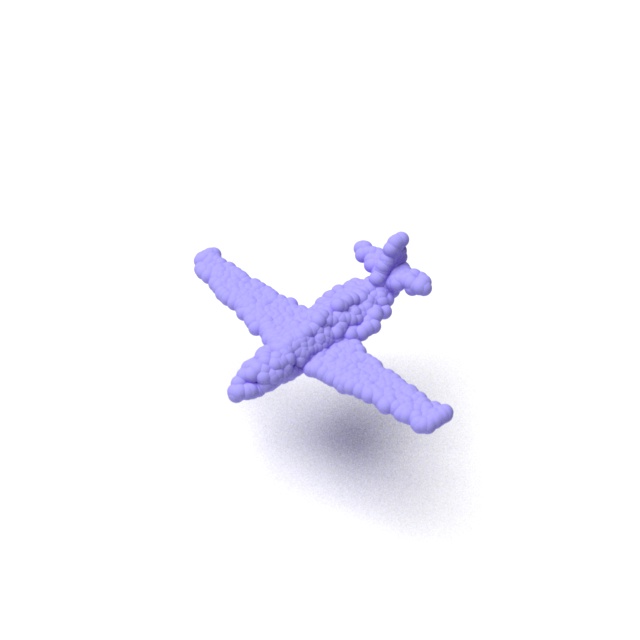} &
        \includegraphics[width=\sizea, trim={\tale} {\tab} {4.5cm} {\tat},clip]{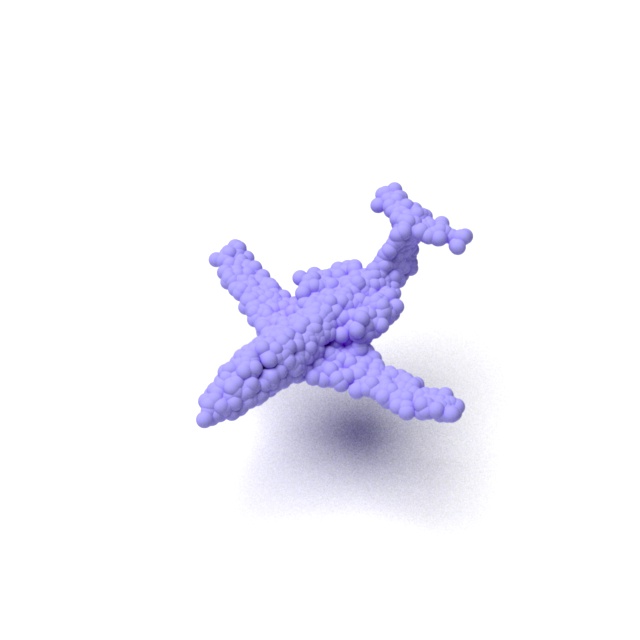} &
        \includegraphics[width=\sizea, trim={\tale} {\tab} {4.5cm} 
        {\tat},clip]{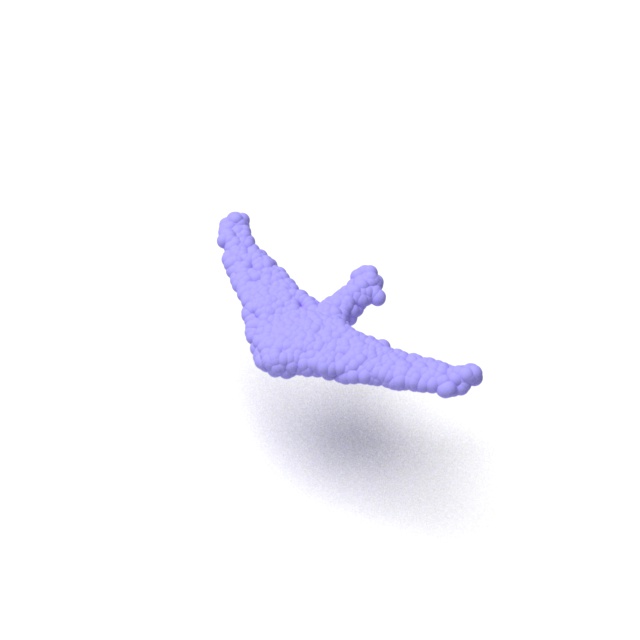} &
        \includegraphics[width=\sizea, trim={\tal} {\tab} {4.5cm} {\tat},clip]{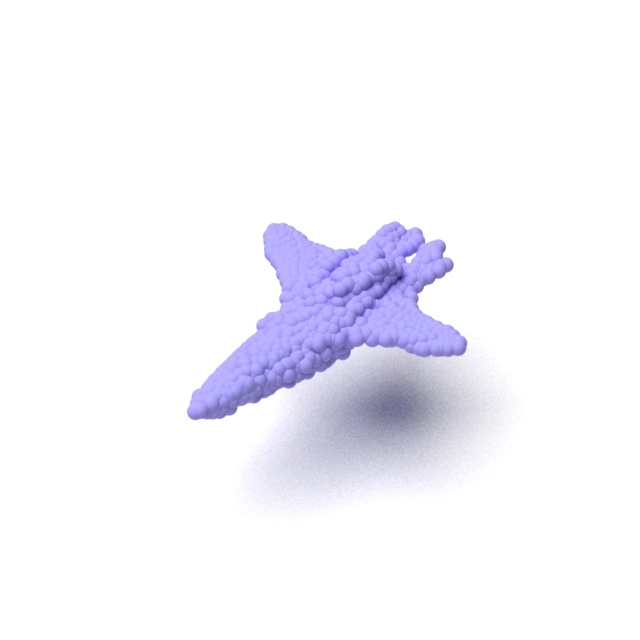} 
        \\
        \includegraphics[width=\sizea, trim={\tale} {\tab} {4.5cm} {\tat},clip]{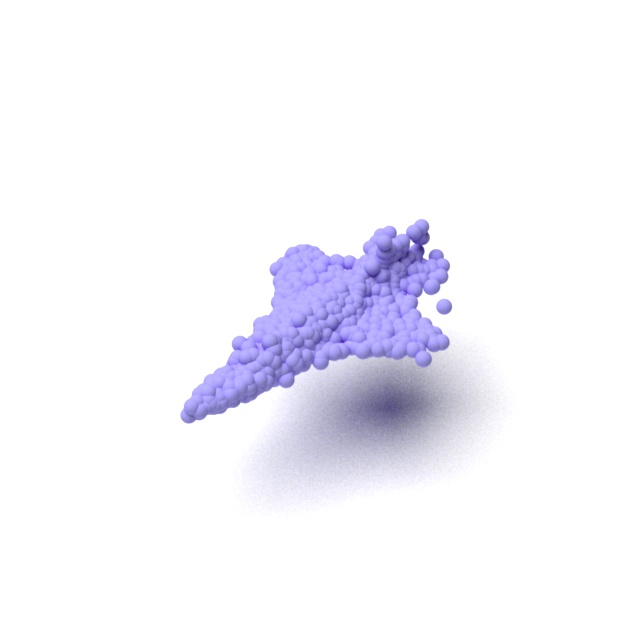} &
        \includegraphics[width=\sizea, trim={\tale} {\tab} {4.5cm} {\tat},clip]{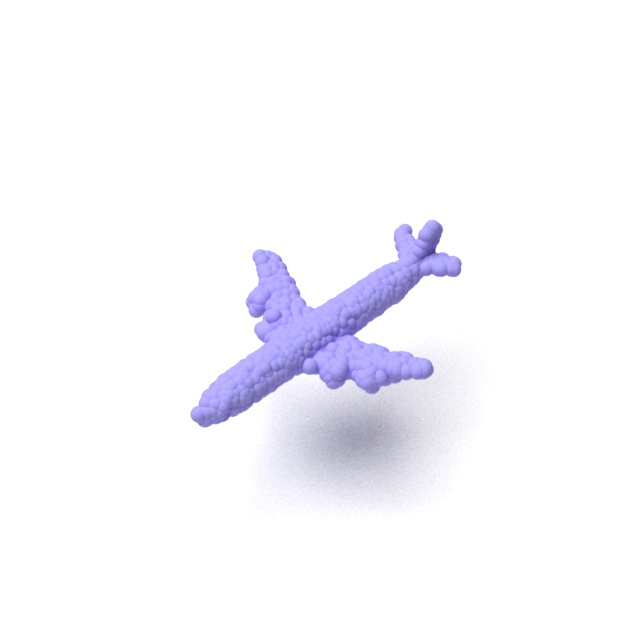} &
        \includegraphics[width=\sizea, trim={\tale} {\tab} {4.5cm} {\tat},clip]{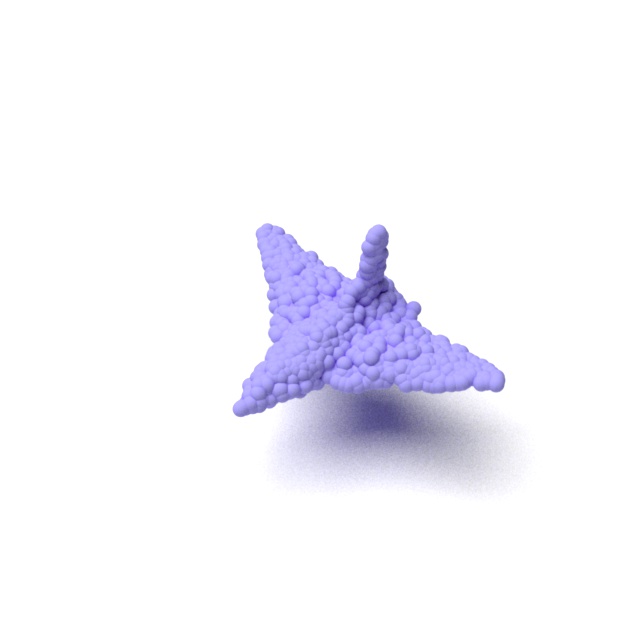} &
        \includegraphics[width=\sizea, trim={\tale} {\tab} {4.5cm} {\tat},clip]{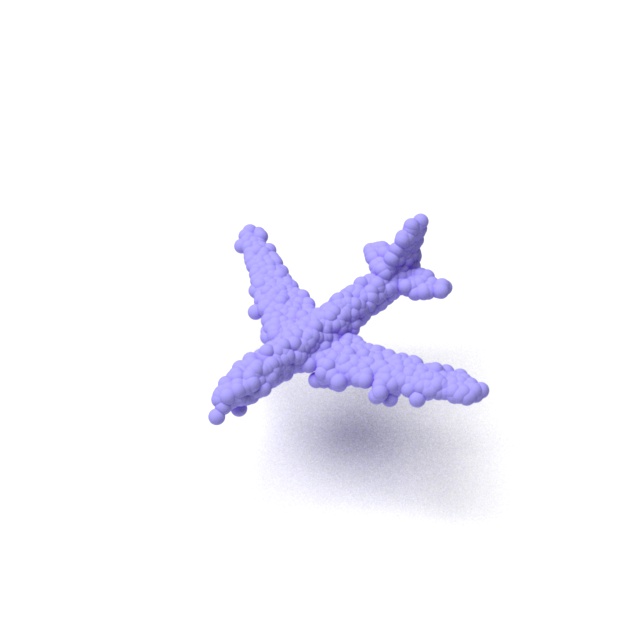} &
        \includegraphics[width=\sizea, trim={\tale} {\tab} {4.5cm} 
        {\tat},clip]{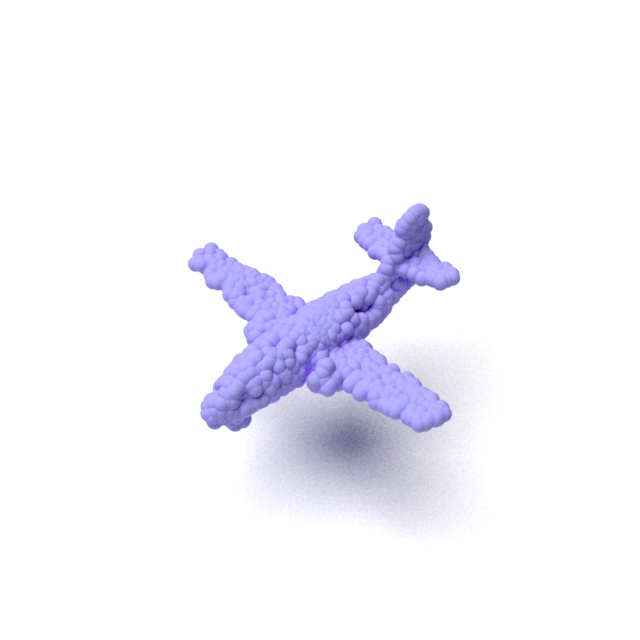} &
        \includegraphics[width=\sizea, trim={\tal} {\tab} {4.5cm} {\tat},clip]{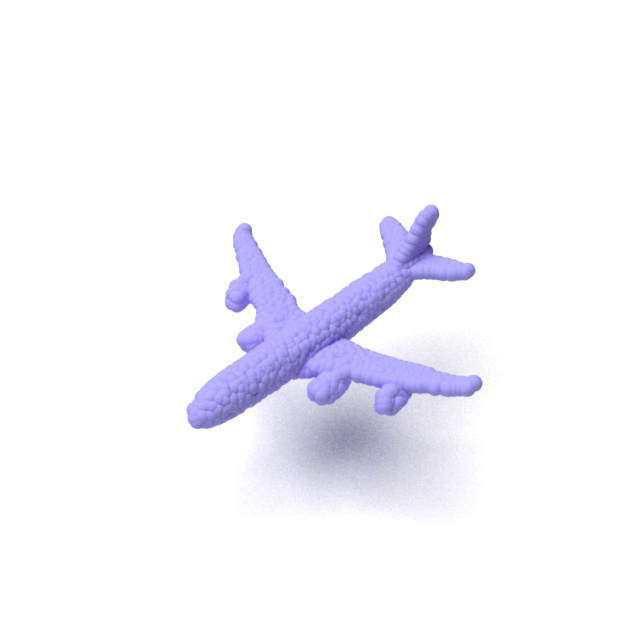} 
        \\
        \includegraphics[width=\sizea, trim={\tale} {\tab} {4.5cm} {\tat},clip]{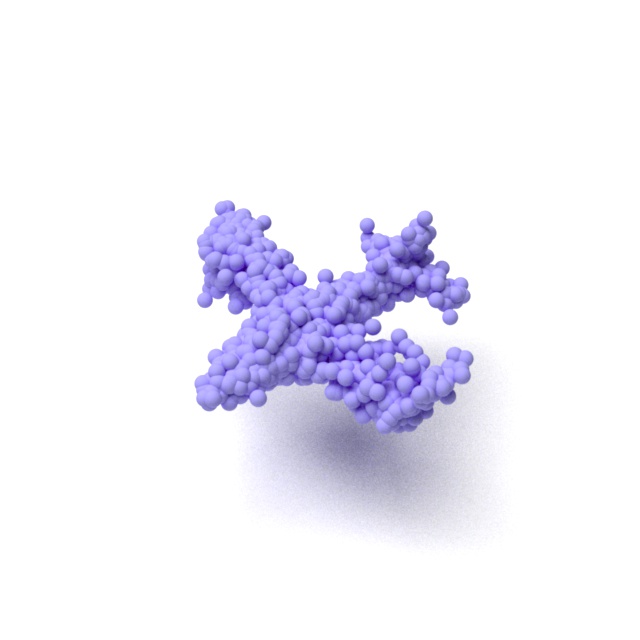} &
        \includegraphics[width=\sizea, trim={\tale} {\tab} {4.5cm} {\tat},clip]{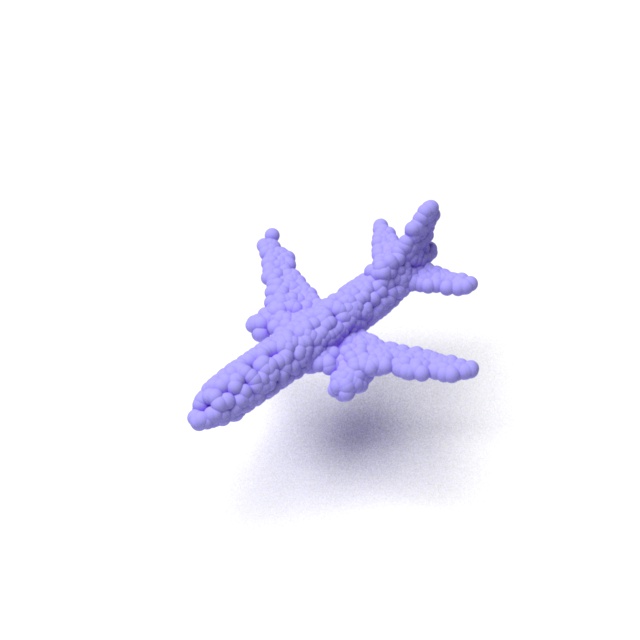} &
        \includegraphics[width=\sizea, trim={\tale} {\tab} {4.5cm} {\tat},clip]{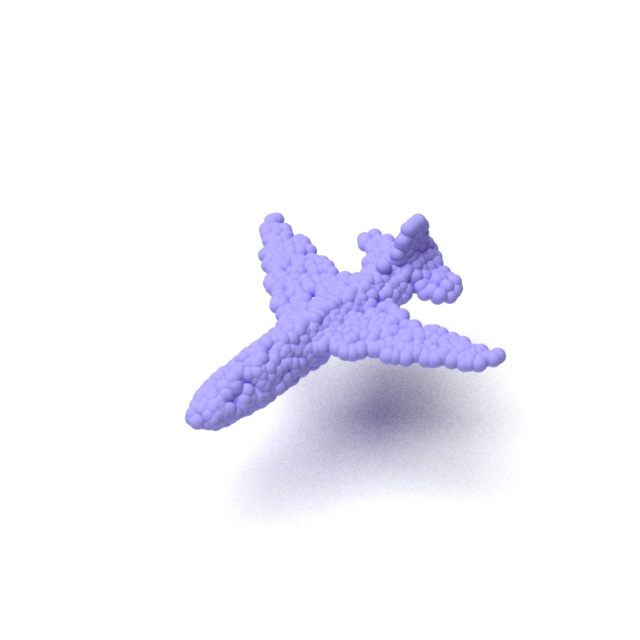} &
        \includegraphics[width=\sizea, trim={\tale} {\tab} {4.5cm} {\tat},clip]{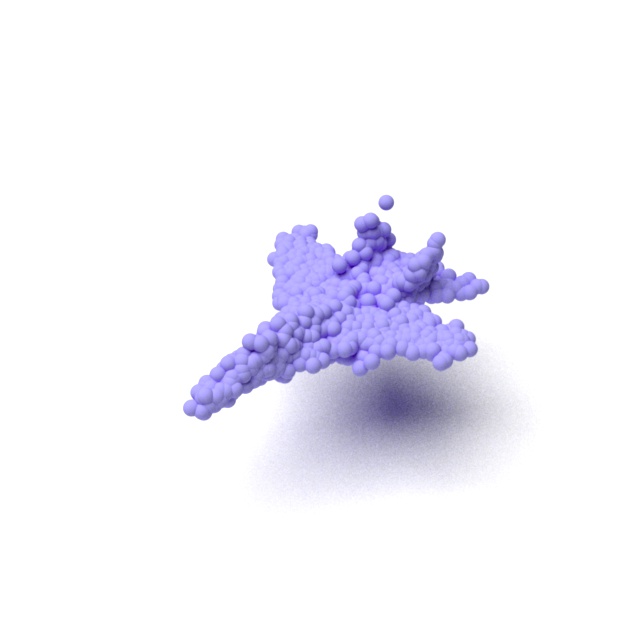} &
        \includegraphics[width=\sizea, trim={\tale} {\tab} {4.5cm} 
        {\tat},clip]{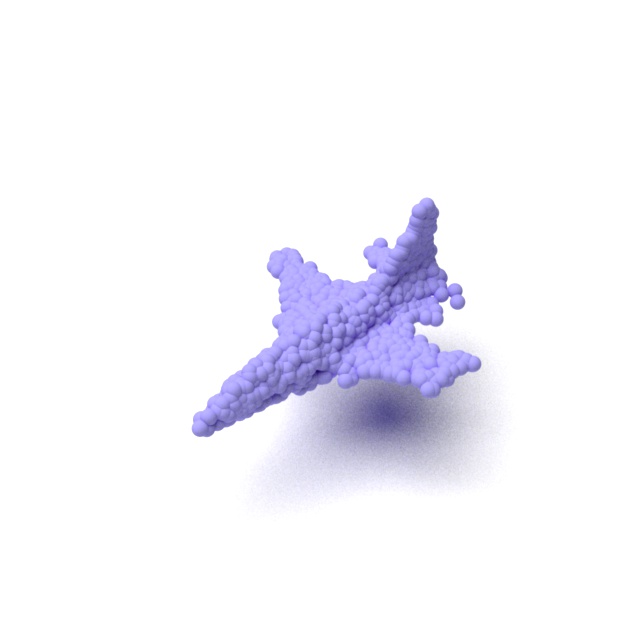} &
        \includegraphics[width=\sizea, trim={\tal} {\tab} {4.5cm} {\tat},clip]{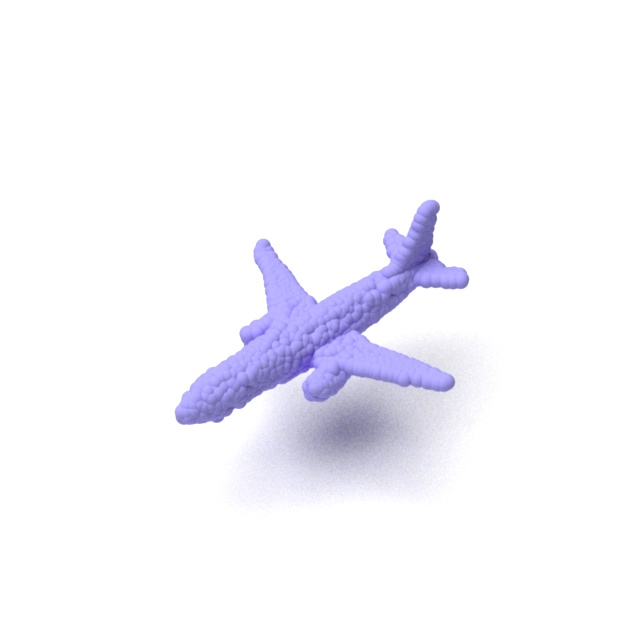} 
        \\
        \includegraphics[width=\sizea, trim={\tale} {\tab} {4.5cm} {\tat},clip]{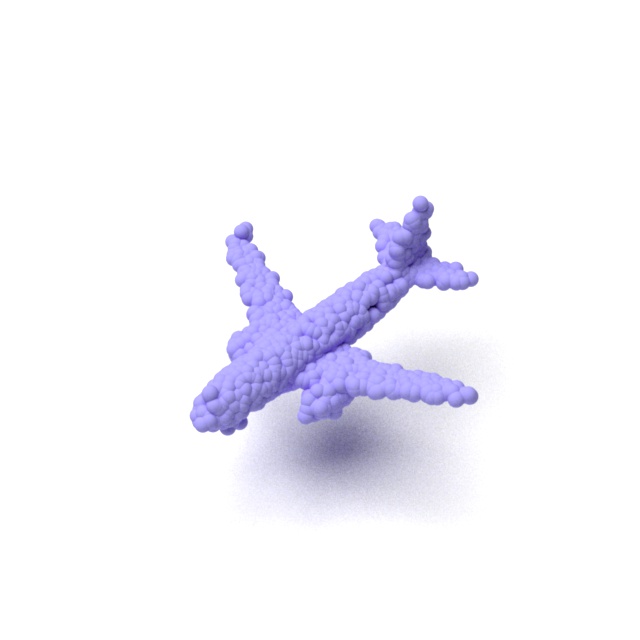} &
        \includegraphics[width=\sizea, trim={\tale} {\tab} {4.5cm} {\tat},clip]{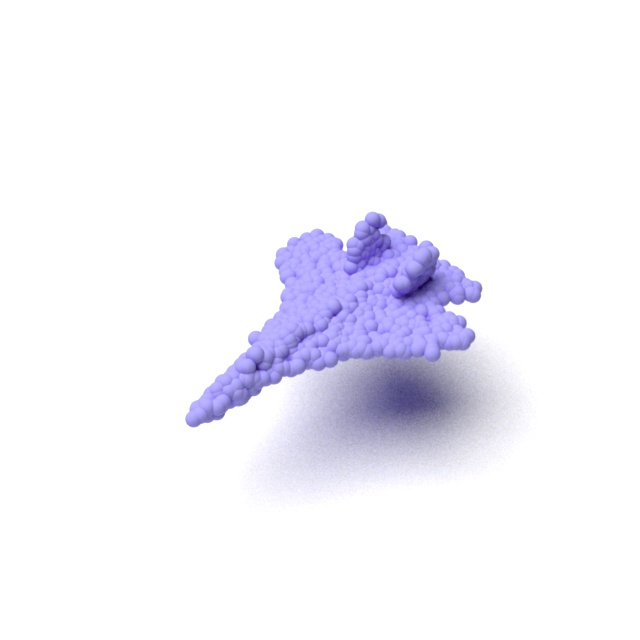} &
        \includegraphics[width=\sizea, trim={\tale} {\tab} {4.5cm} {\tat},clip]{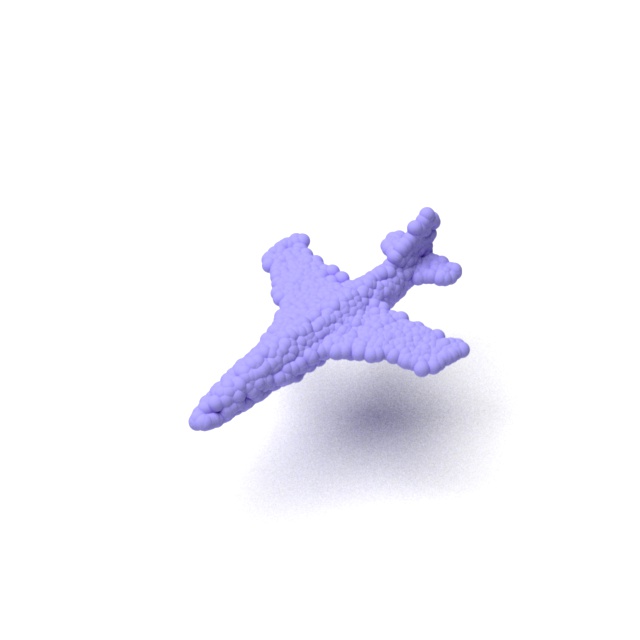} &
        \includegraphics[width=\sizea, trim={\tale} {\tab} {4.5cm} {\tat},clip]{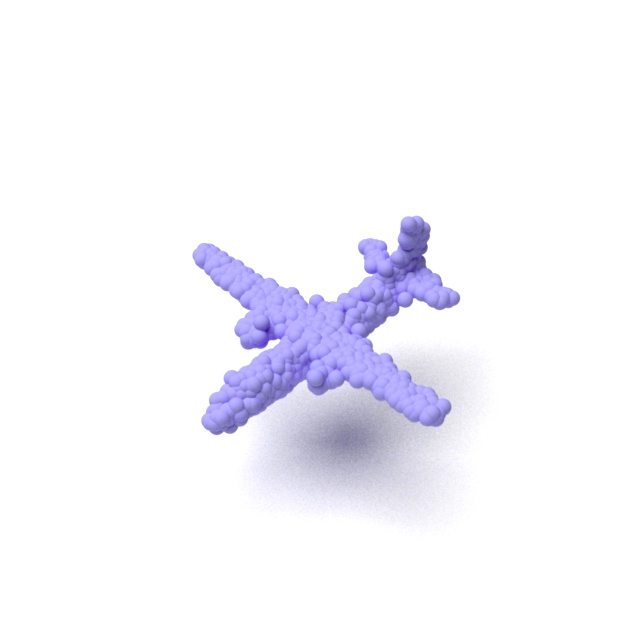} &
        \includegraphics[width=\sizea, trim={\tale} {\tab} {4.5cm} 
        {\tat},clip]{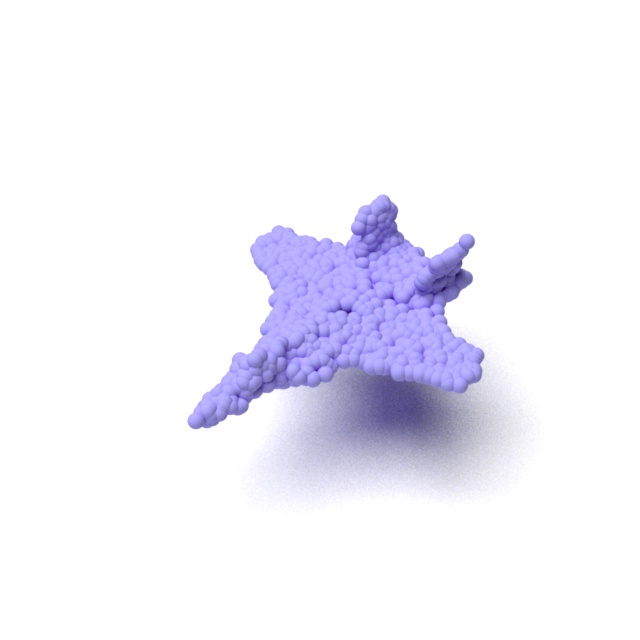} &
        \includegraphics[width=\sizea, trim={\tal} {\tab} {4.5cm} {\tat},clip]{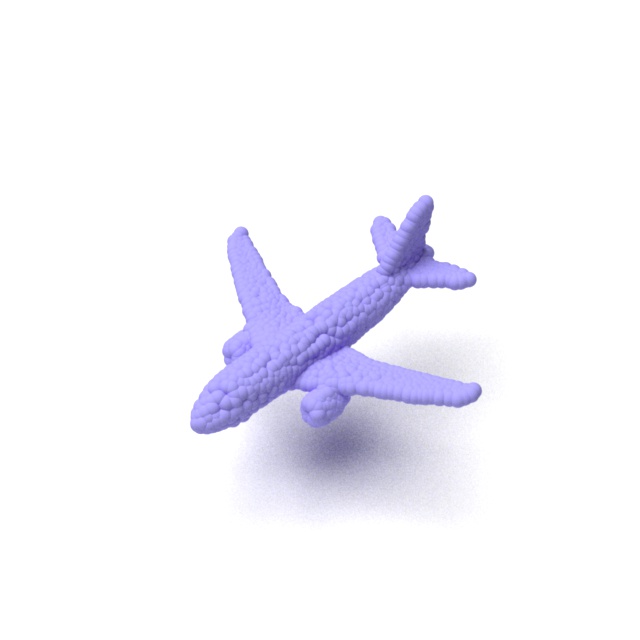} 
        \\
        \includegraphics[width=\sizea, trim={\tale} {\tab} {4.5cm} {\tat},clip]{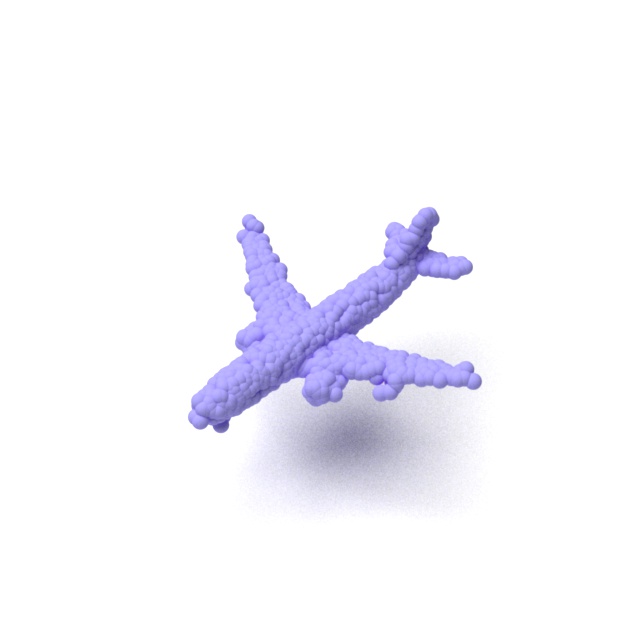} &
        \includegraphics[width=\sizea, trim={\tale} {\tab} {4.5cm} {\tat},clip]{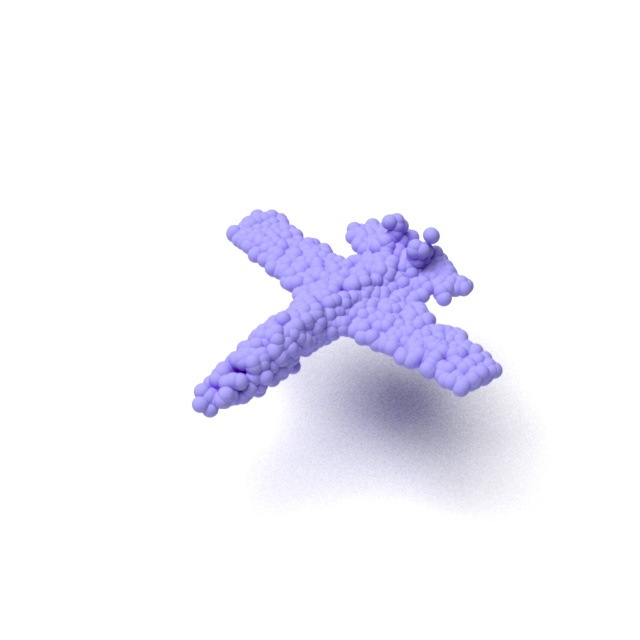} &
        \includegraphics[width=\sizea, trim={\tale} {\tab} {4.5cm} {\tat},clip]{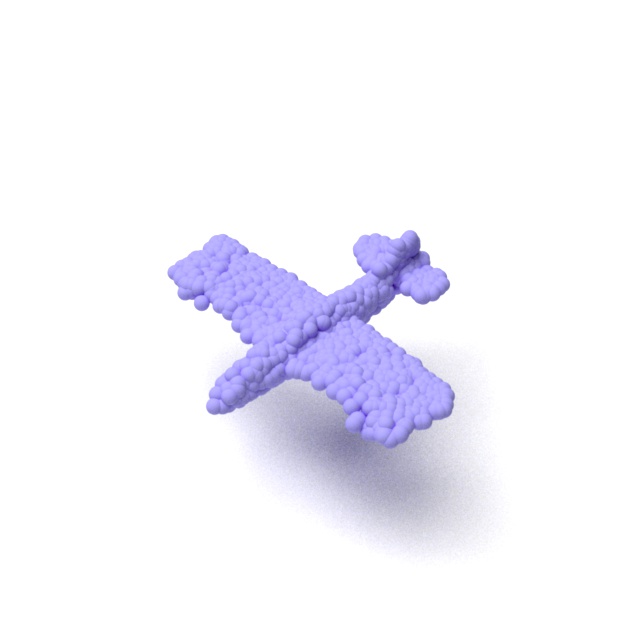} &
        \includegraphics[width=\sizea, trim={\tale} {\tab} {4.5cm} {\tat},clip]{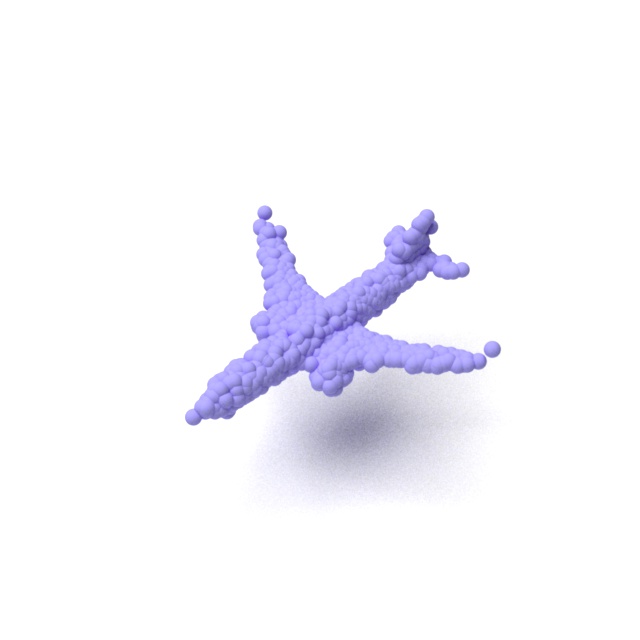} &
        \includegraphics[width=\sizea, trim={\tale} {\tab} {4.5cm} 
        {\tat},clip]{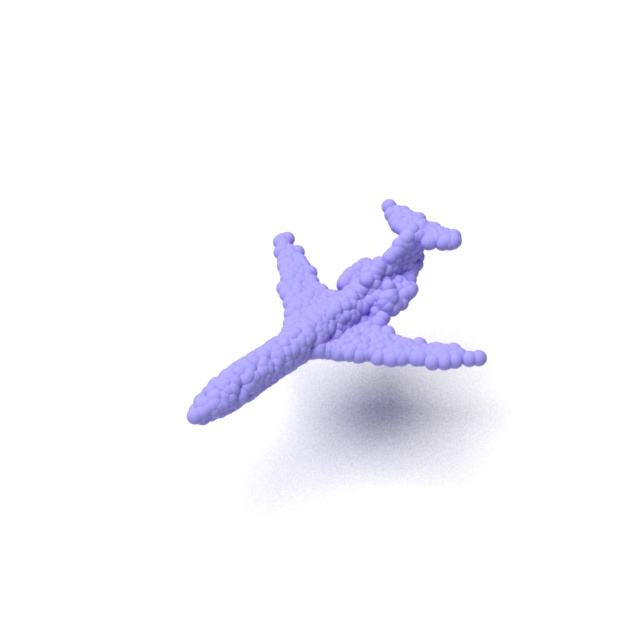} &
        \includegraphics[width=\sizea, trim={\tal} {\tab} {4.5cm} {\tat},clip]{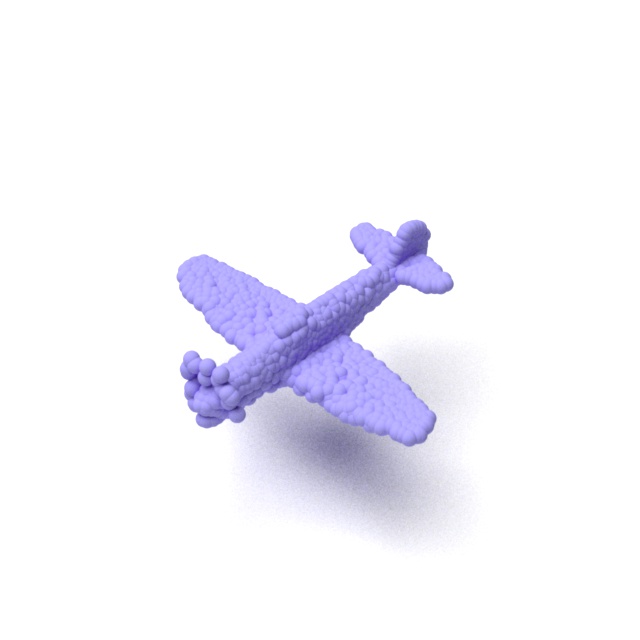} 
        \\
        PF & ShapeGF & SetVAE & DPM & PVD & Ours
    \end{tabular}

    \end{center}
    \caption{Shape generation results on ShapeNet Airplane. We shown results from PF (PointFlow)~\cite{yang2019pointflow}, ShapeGF~\cite{cai2020learning}, SetVAE~\cite{kim2021setvae}, DPM~\cite{luo2021diffusion}, and PVD~\cite{zhou20213d}.
    }
    \vspace{-8.5em}
    \label{fig:sup:gen:airplane}
\end{figure}

\begin{figure}[h]
    \begin{center}
    \newcommand{\sizea}{0.155\linewidth}
    \newcommand{\sizeb}{0.155\linewidth}
    \newcommand{\sizec}{0.155\linewidth}
    \newcommand{\tare}{5cm}
    \newcommand{\tale}{2.5cm}
    \newcommand{\tal}{3.5cm}
    \newcommand{\tab}{2.5cm}
    \newcommand{\tar}{3.5cm}
    \newcommand{\tat}{2.5cm}
    \newcommand{\tcl}{3.0cm}
    \newcommand{\tcb}{3cm}
    \newcommand{\tcr}{4cm}
    \newcommand{\tct}{4.2cm}
    \newcommand{\thl}{3.0cm}
    \newcommand{\thb}{0.0cm}
    \newcommand{\thr}{3cm}
    \newcommand{\tht}{2cm}
    \setlength{\tabcolsep}{0pt}
    \renewcommand{\arraystretch}{0}
    \begin{tabular}{@{}ccccc:c@{}}
        \includegraphics[width=\sizea, trim={\tale} {\tab} {2.5cm} {\tat},clip]{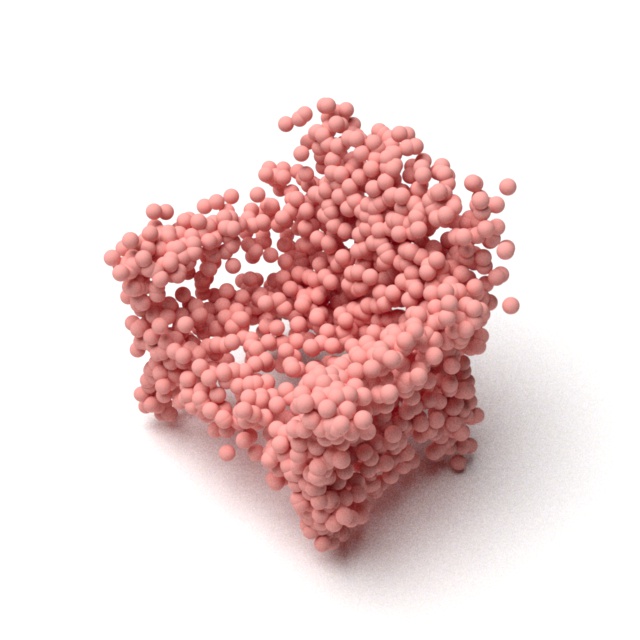} &
        \includegraphics[width=\sizea, trim={\tale} {\tab} {2.5cm} {\tat},clip]{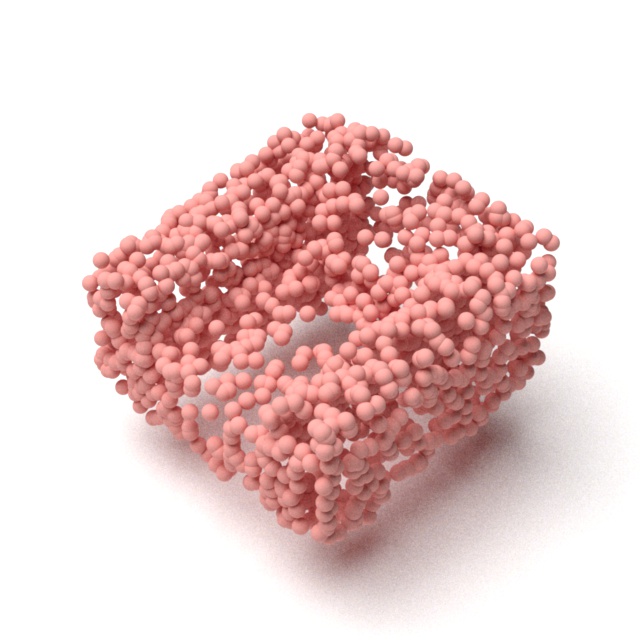} &
        \includegraphics[width=\sizea, trim={\tale} {\tab} {2.5cm} {\tat},clip]{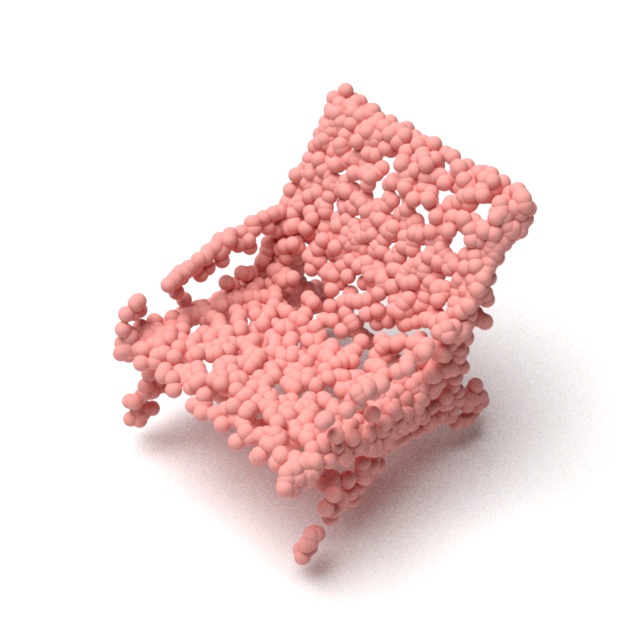} &
        \includegraphics[width=\sizea, trim={\tale} {\tab} {2.5cm} {\tat},clip]{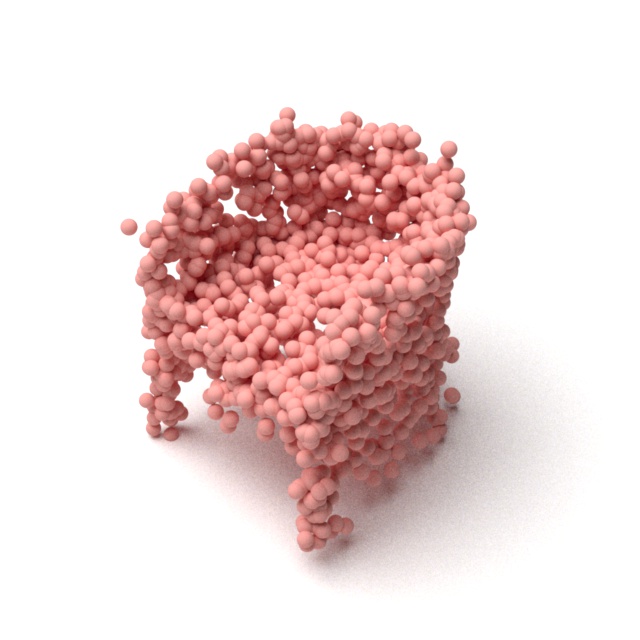} &
        \includegraphics[width=\sizea, trim={\tale} {\tab} {2.5cm} 
        {\tat},clip]{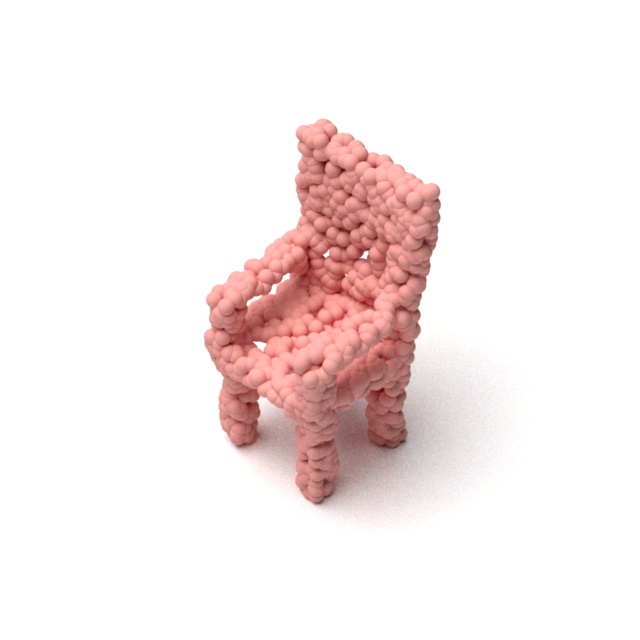} &
        \includegraphics[width=\sizea, trim={\tal} {\tab} {2.5cm} {\tat},clip]{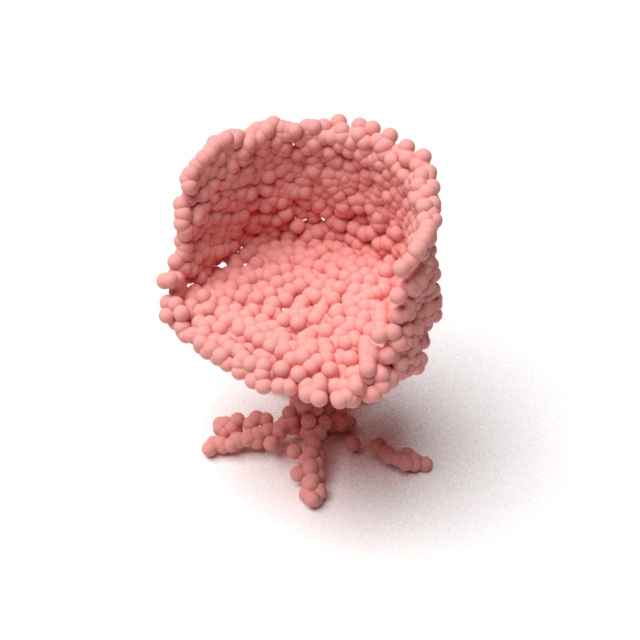} 
        \\
        \includegraphics[width=\sizea, trim={\tale} {\tab} {2.5cm} {\tat},clip]{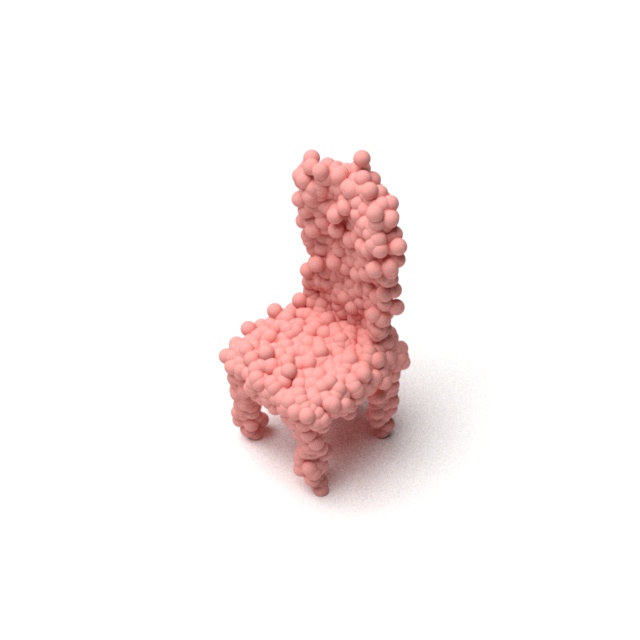} &
        \includegraphics[width=\sizea, trim={\tale} {\tab} {2.5cm} {\tat},clip]{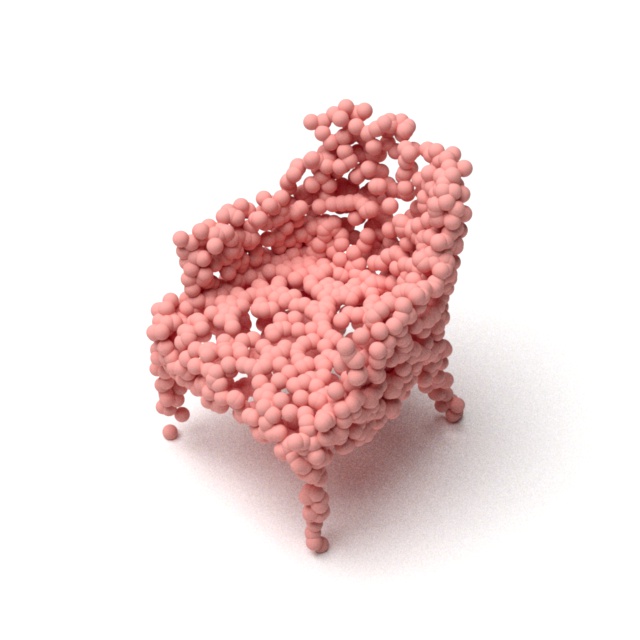} &
        \includegraphics[width=\sizea, trim={\tale} {\tab} {2.5cm} {\tat},clip]{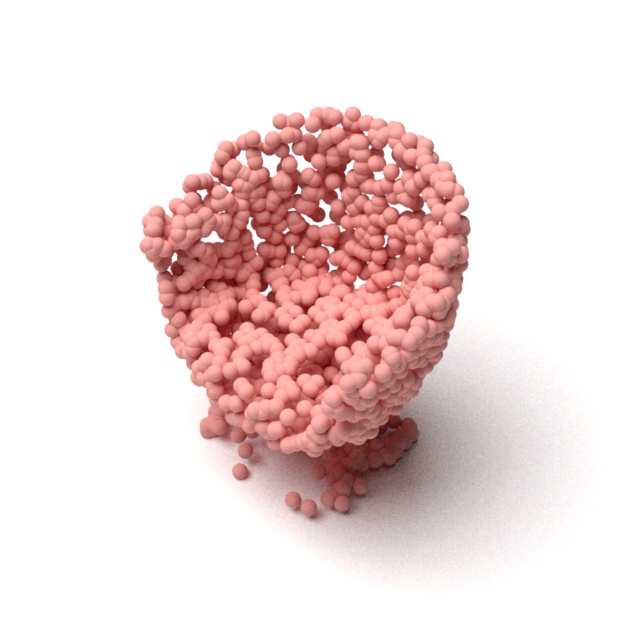} &
        \includegraphics[width=\sizea, trim={\tale} {\tab} {2.5cm} {\tat},clip]{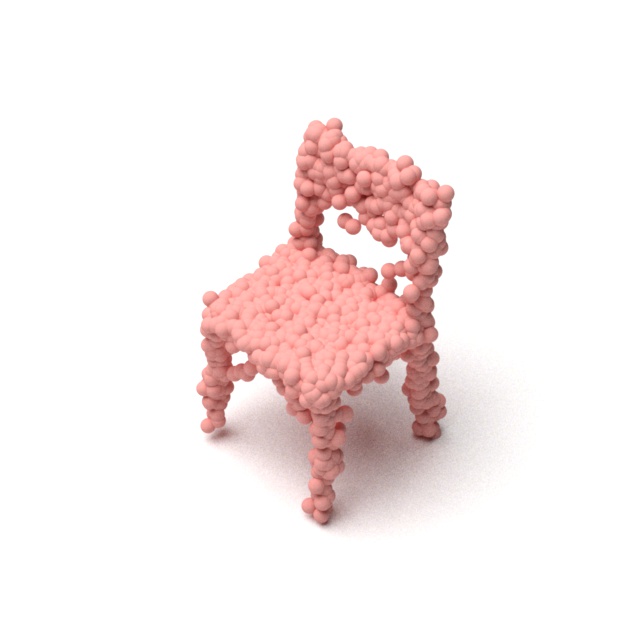} &
        \includegraphics[width=\sizea, trim={\tale} {\tab} {2.5cm} 
        {\tat},clip]{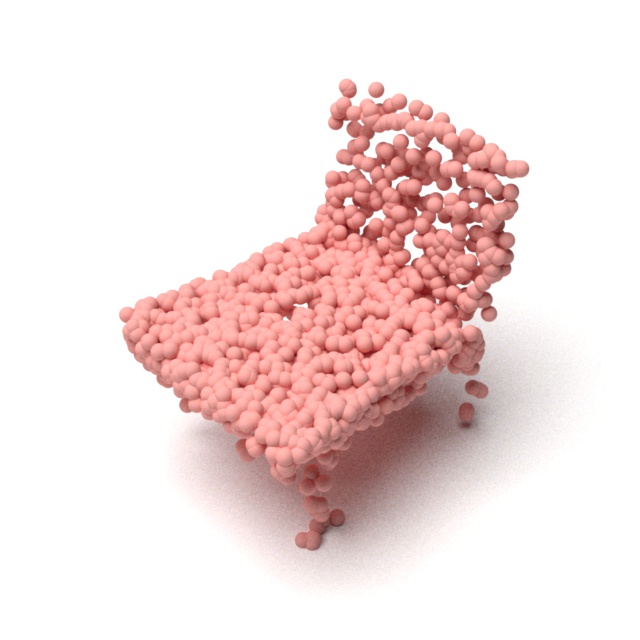} &
        \includegraphics[width=\sizea, trim={\tal} {\tab} {2.5cm} {\tat},clip]{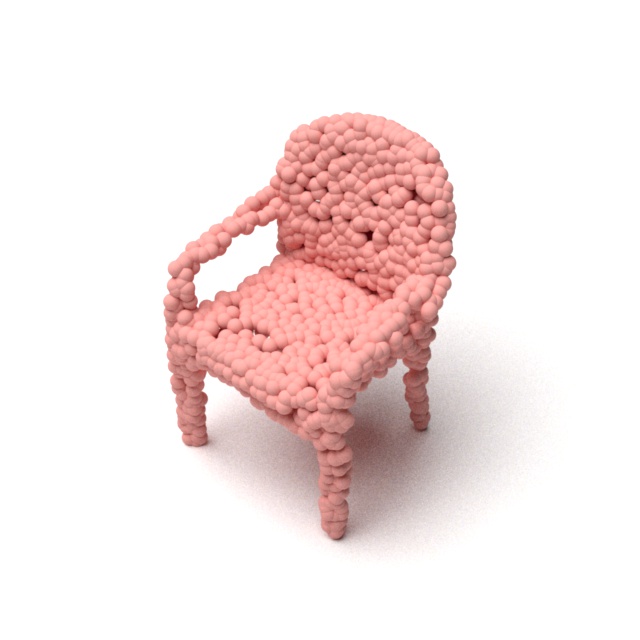} 
        \\
        \includegraphics[width=\sizea, trim={\tale} {\tab} {2.5cm} {\tat},clip]{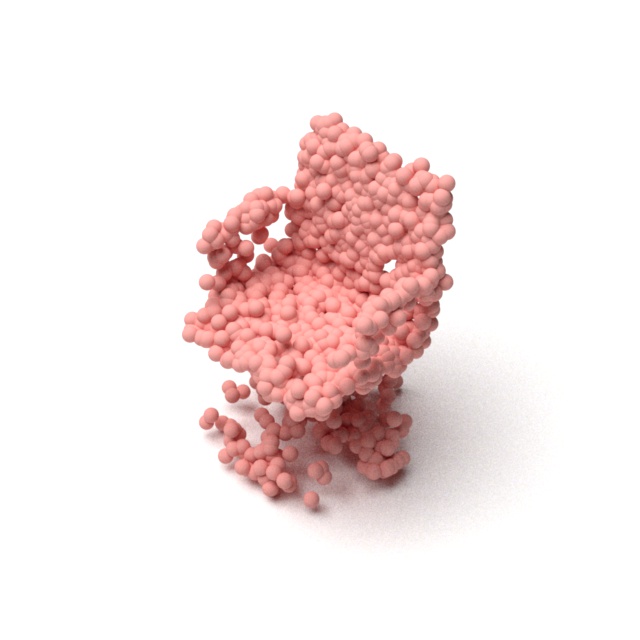} &
        \includegraphics[width=\sizea, trim={\tale} {\tab} {2.5cm} {\tat},clip]{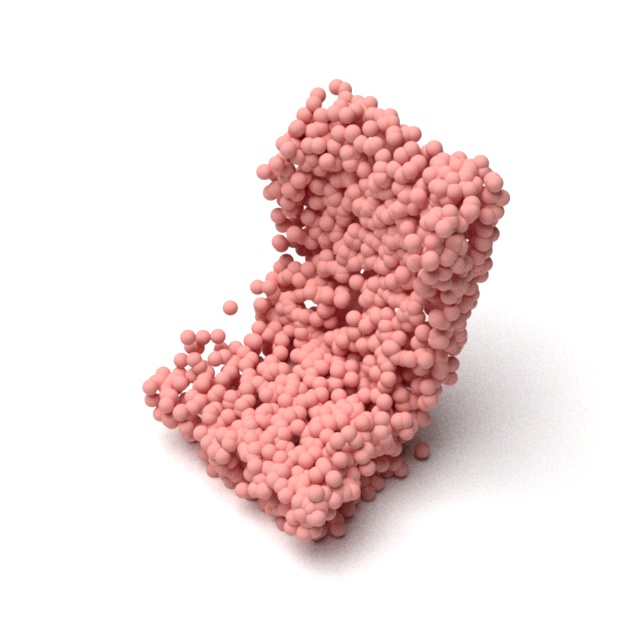} &
        \includegraphics[width=\sizea, trim={\tale} {\tab} {2.5cm} {\tat},clip]{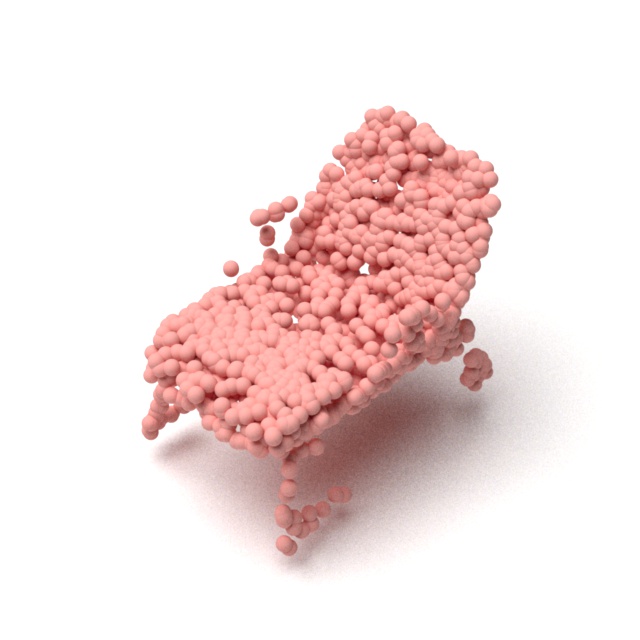} &
        \includegraphics[width=\sizea, trim={\tale} {\tab} {2.5cm} {\tat},clip]{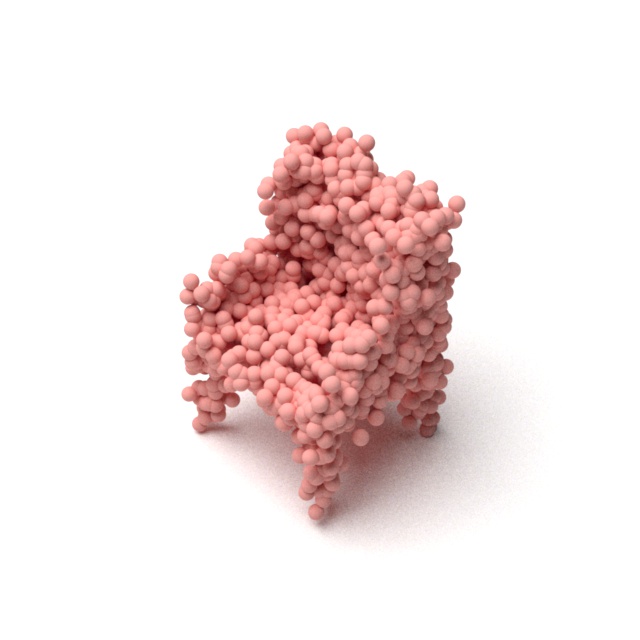} &
        \includegraphics[width=\sizea, trim={\tale} {\tab} {2.5cm} 
        {\tat},clip]{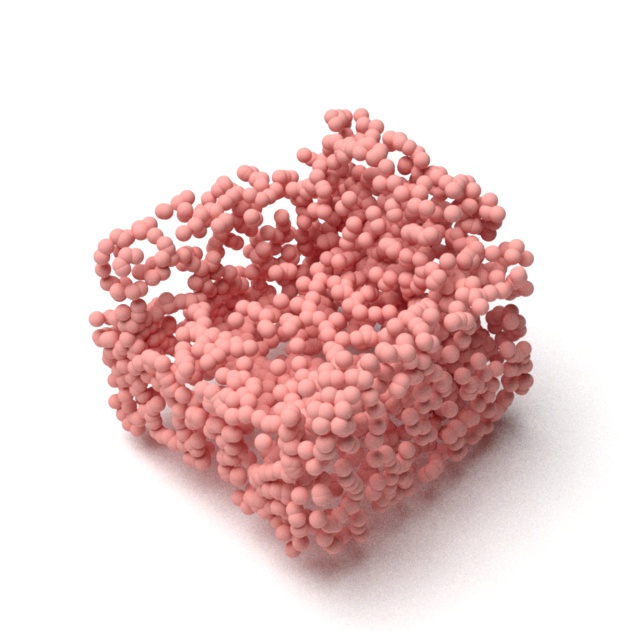} &
        \includegraphics[width=\sizea, trim={\tal} {\tab} {2.5cm} {\tat},clip]{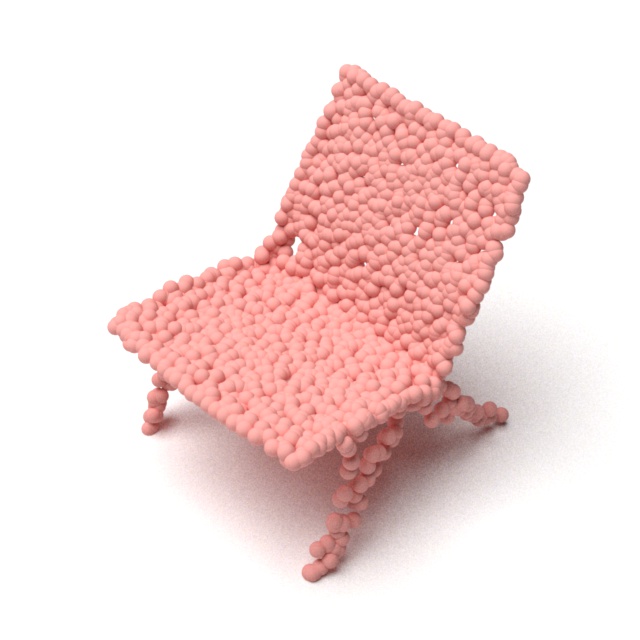} 
        \\
        \includegraphics[width=\sizea, trim={\tale} {\tab} {2.5cm} {\tat},clip]{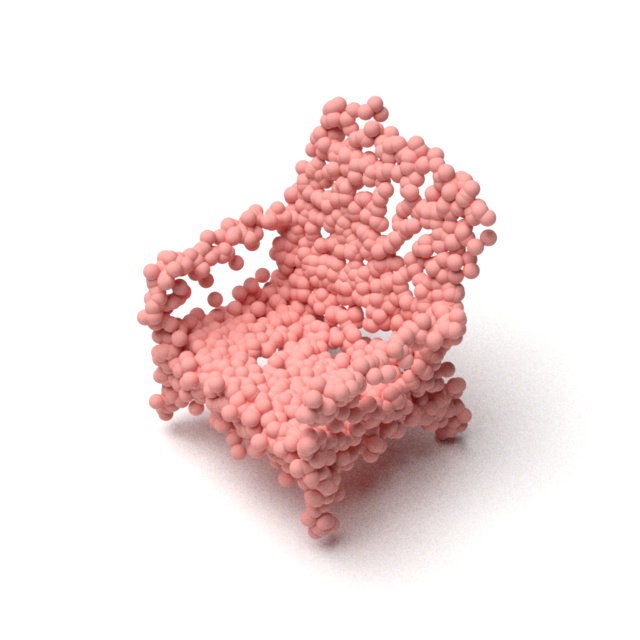} &
        \includegraphics[width=\sizea, trim={\tale} {\tab} {2.5cm} {\tat},clip]{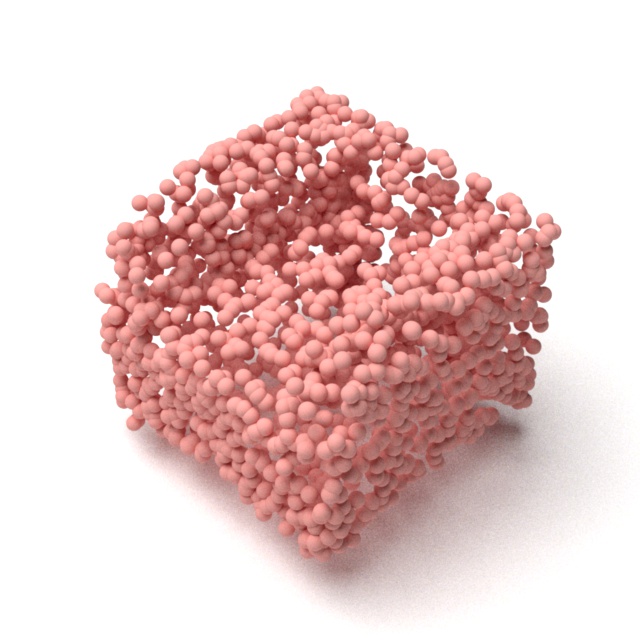} &
        \includegraphics[width=\sizea, trim={\tale} {\tab} {2.5cm} {\tat},clip]{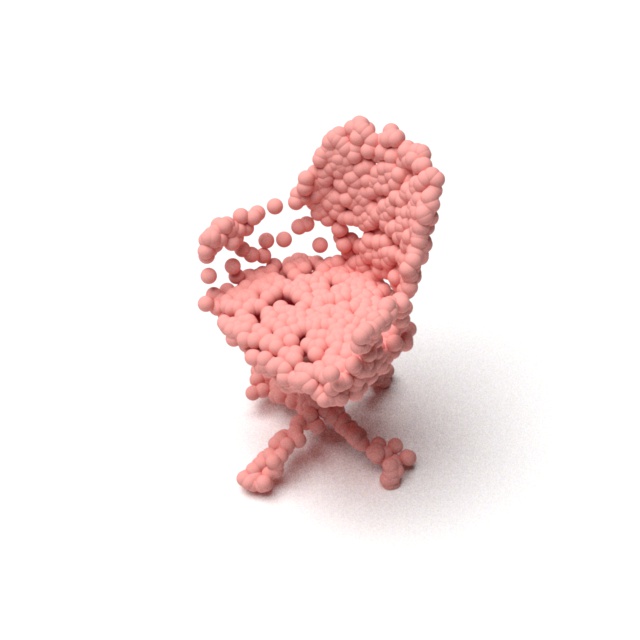} &
        \includegraphics[width=\sizea, trim={\tale} {\tab} {2.5cm} {\tat},clip]{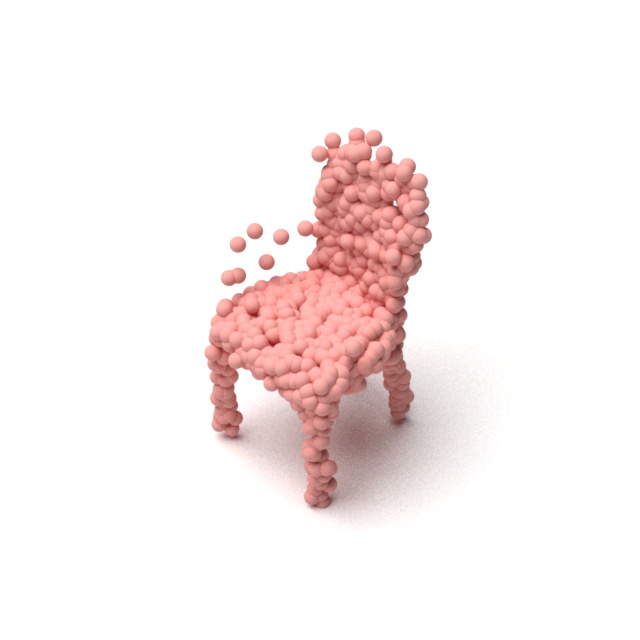} &
        \includegraphics[width=\sizea, trim={\tale} {\tab} {2.5cm} 
        {\tat},clip]{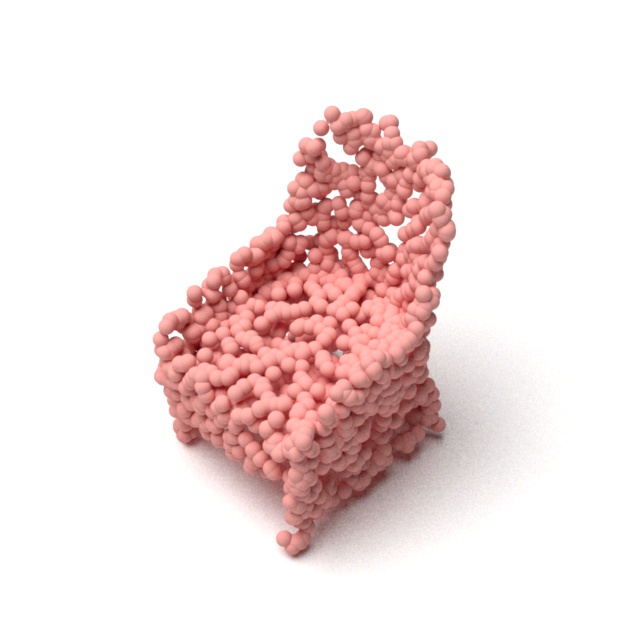} &
        \includegraphics[width=\sizea, trim={\tal} {\tab} {2.5cm} {\tat},clip]{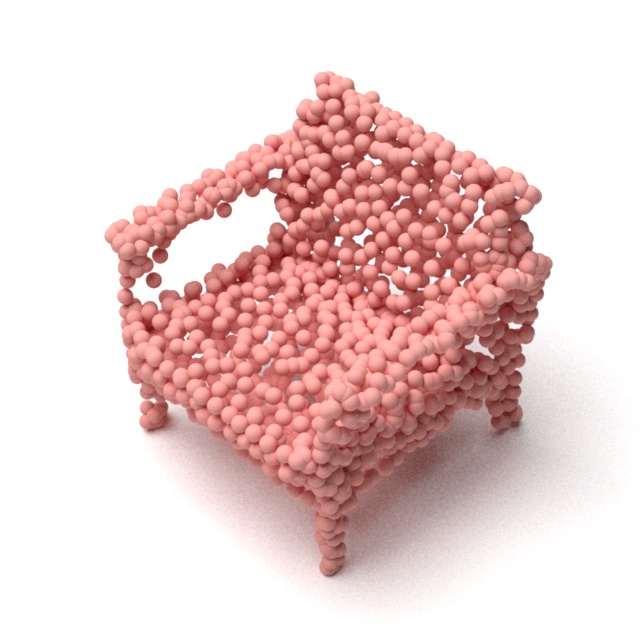} 
        \\
        \includegraphics[width=\sizea, trim={\tale} {\tab} {2.5cm} {\tat},clip]{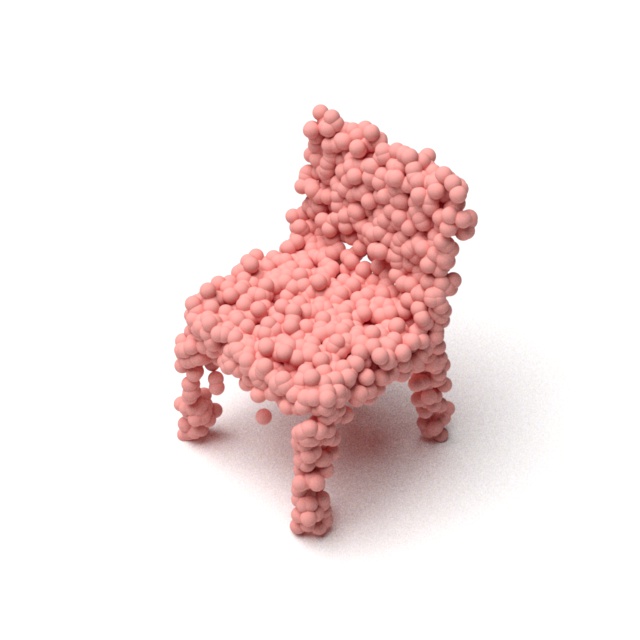} &
        \includegraphics[width=\sizea, trim={\tale} {\tab} {2.5cm} {\tat},clip]{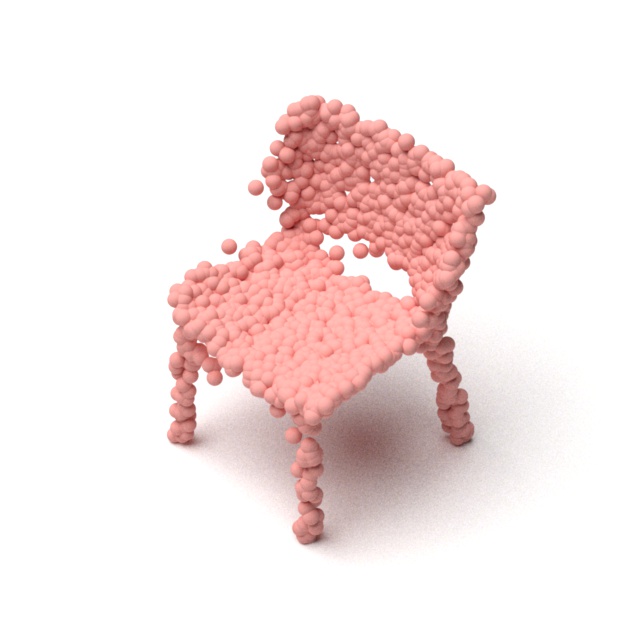} &
        \includegraphics[width=\sizea, trim={\tale} {\tab} {2.5cm} {\tat},clip]{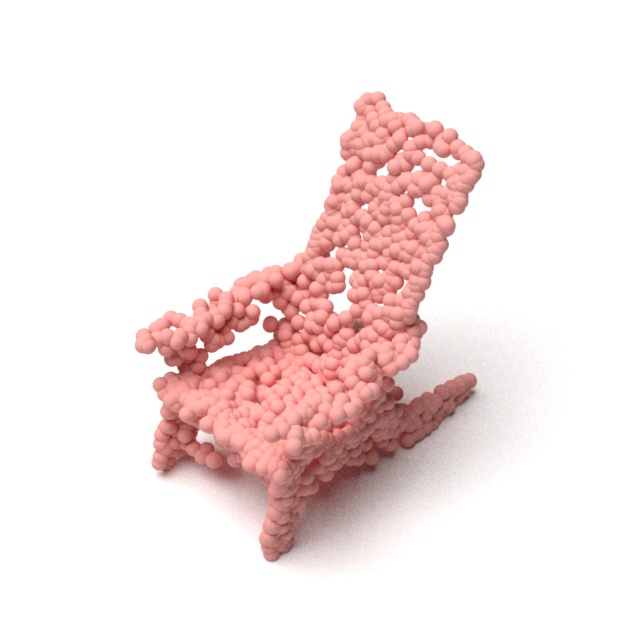} &
        \includegraphics[width=\sizea, trim={\tale} {\tab} {2.5cm} {\tat},clip]{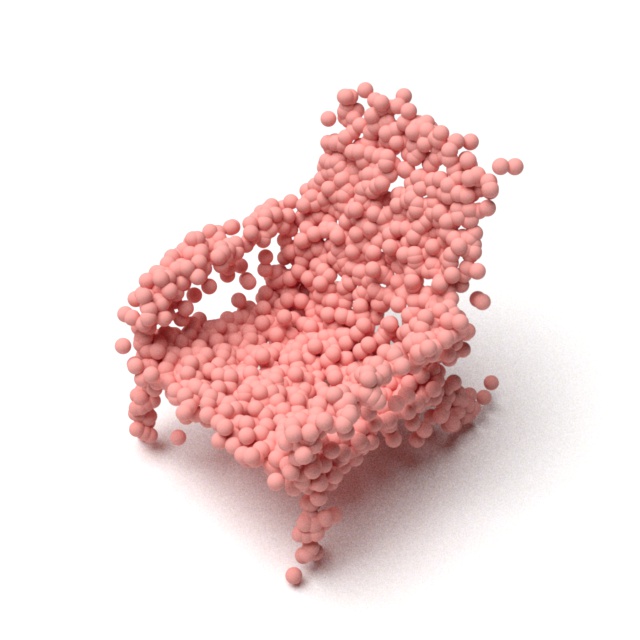} &
        \includegraphics[width=\sizea, trim={\tale} {\tab} {2.5cm} 
        {\tat},clip]{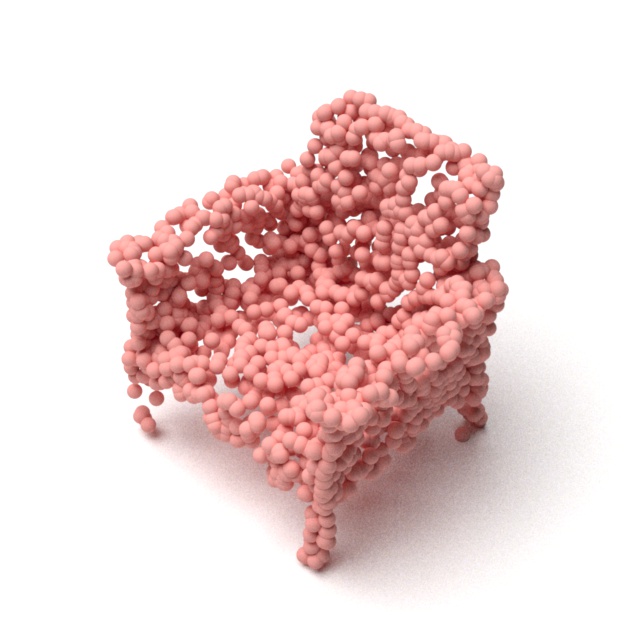} &
        \includegraphics[width=\sizea, trim={\tal} {\tab} {2.5cm} {\tat},clip]{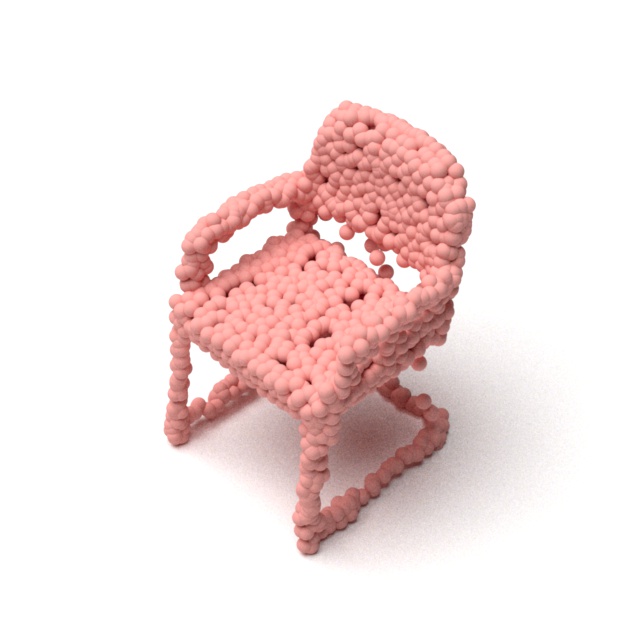} 
        \\
        \includegraphics[width=\sizea, trim={\tale} {\tab} {2.5cm} {\tat},clip]{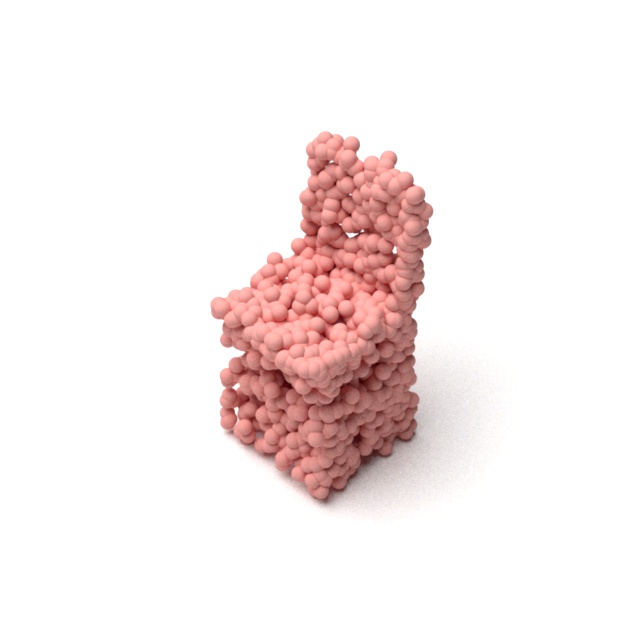} &
        \includegraphics[width=\sizea, trim={\tale} {\tab} {2.5cm} {\tat},clip]{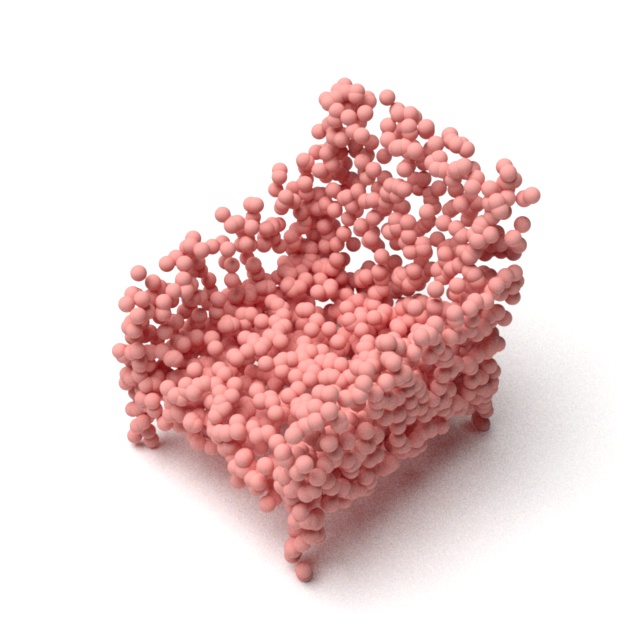} &
        \includegraphics[width=\sizea, trim={\tale} {\tab} {2.5cm} {\tat},clip]{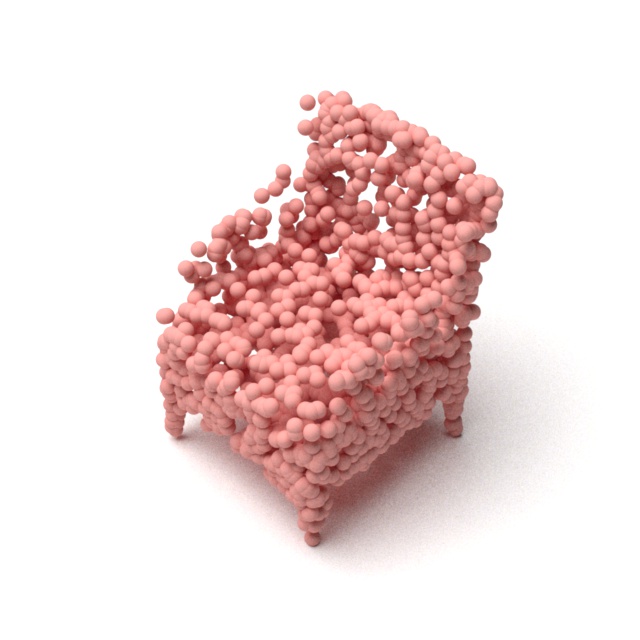} &
        \includegraphics[width=\sizea, trim={\tale} {\tab} {2.5cm} {\tat},clip]{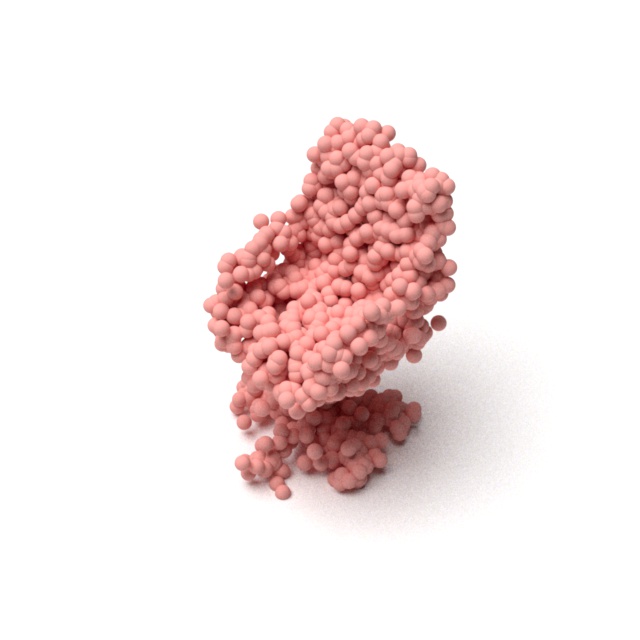} &
        \includegraphics[width=\sizea, trim={\tale} {\tab} {2.5cm} 
        {\tat},clip]{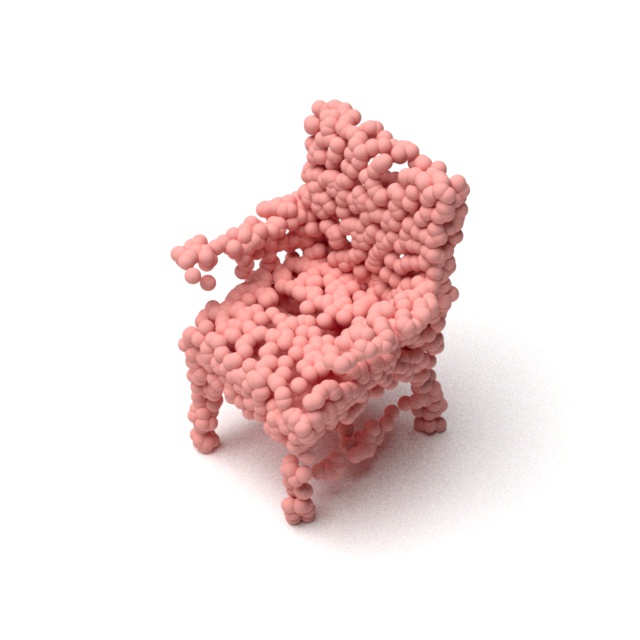} &
        \includegraphics[width=\sizea, trim={\tal} {\tab} {2.5cm} {\tat},clip]{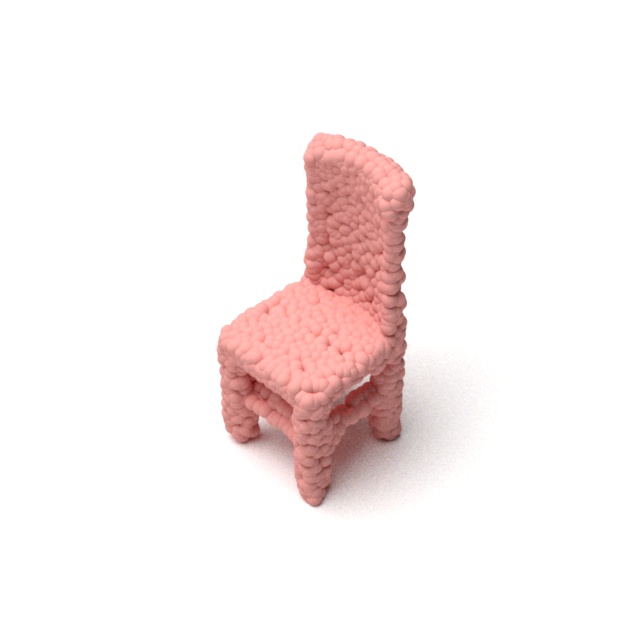} 
        \\
        \includegraphics[width=\sizea, trim={\tale} {\tab} {2.5cm} {\tat},clip]{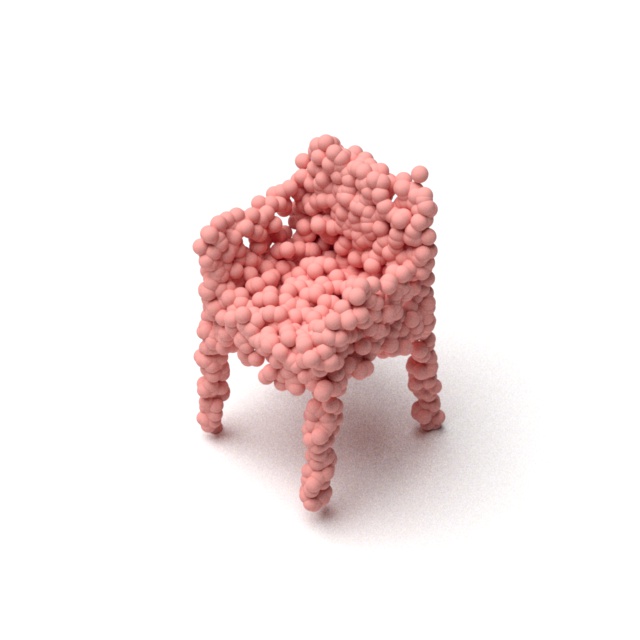} &
        \includegraphics[width=\sizea, trim={\tale} {\tab} {2.5cm} {\tat},clip]{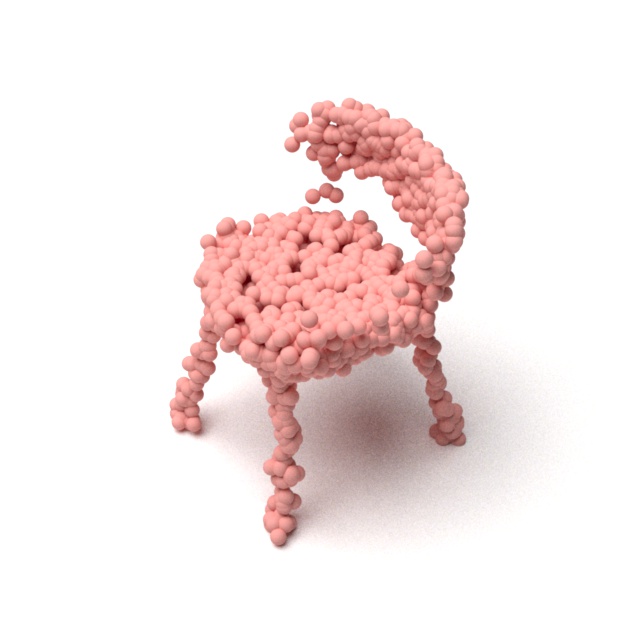} &
        \includegraphics[width=\sizea, trim={\tale} {\tab} {2.5cm} {\tat},clip]{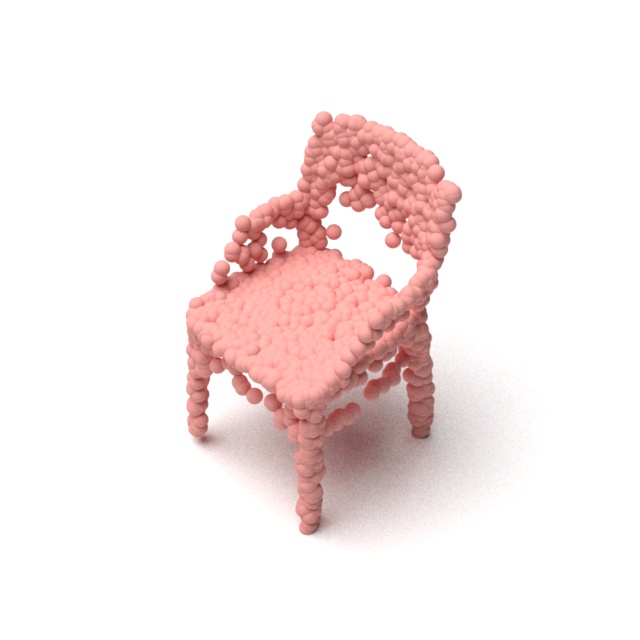} &
        \includegraphics[width=\sizea, trim={\tale} {\tab} {2.5cm} {\tat},clip]{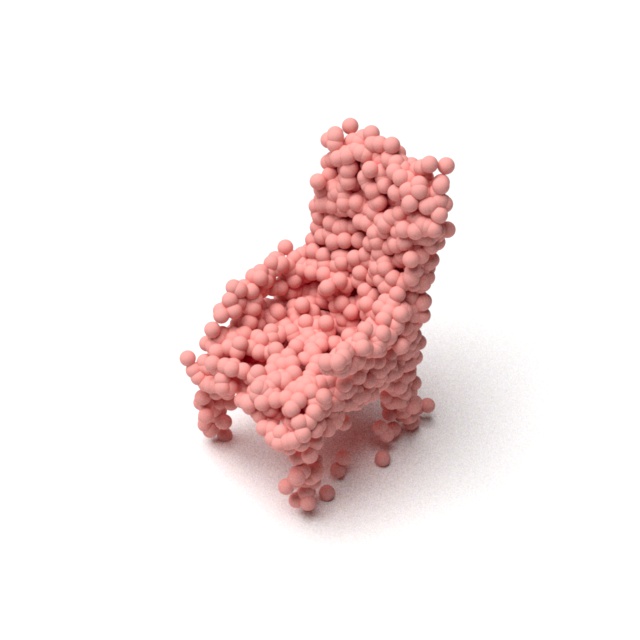} &
        \includegraphics[width=\sizea, trim={\tale} {\tab} {2.5cm} 
        {\tat},clip]{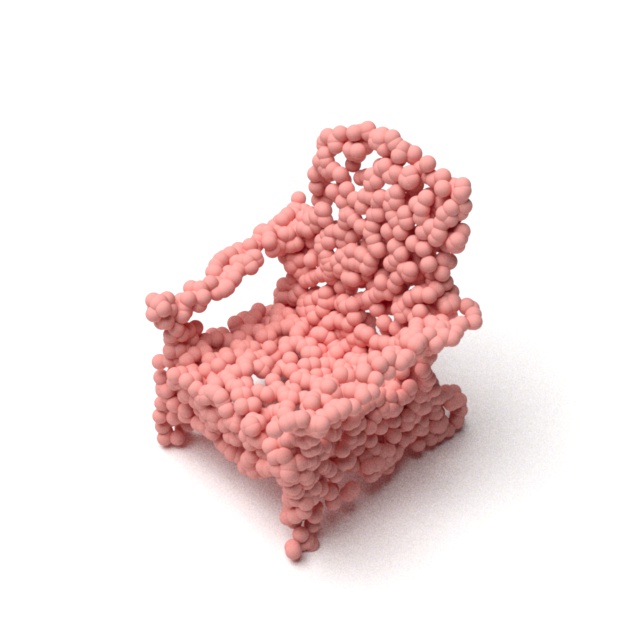} &
        \includegraphics[width=\sizea, trim={\tal} {\tab} {2.5cm} {\tat},clip]{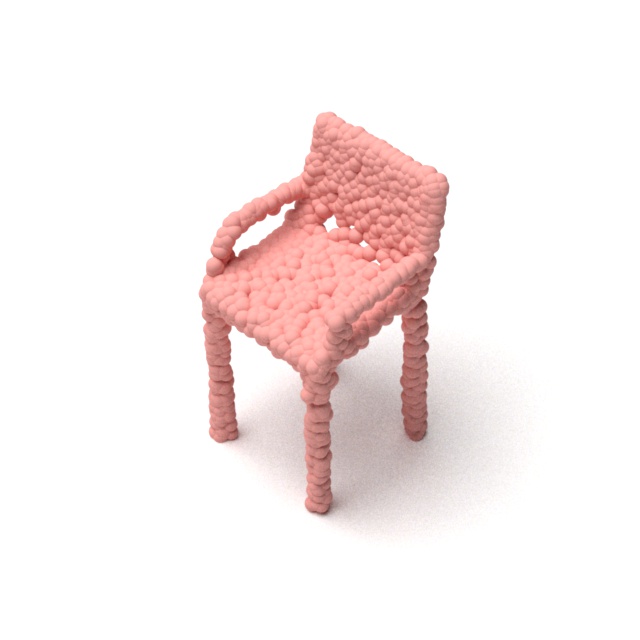} 
        \\
        \includegraphics[width=\sizea, trim={\tale} {\tab} {2.5cm} {\tat},clip]{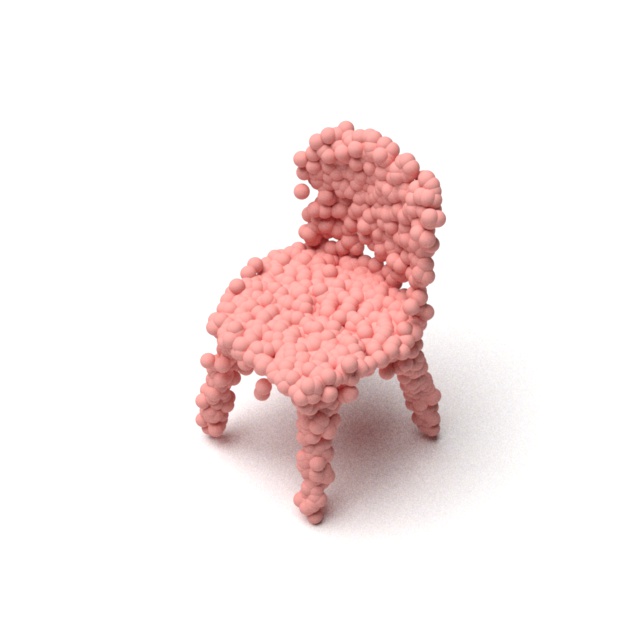} &
        \includegraphics[width=\sizea, trim={\tale} {\tab} {2.5cm} {\tat},clip]{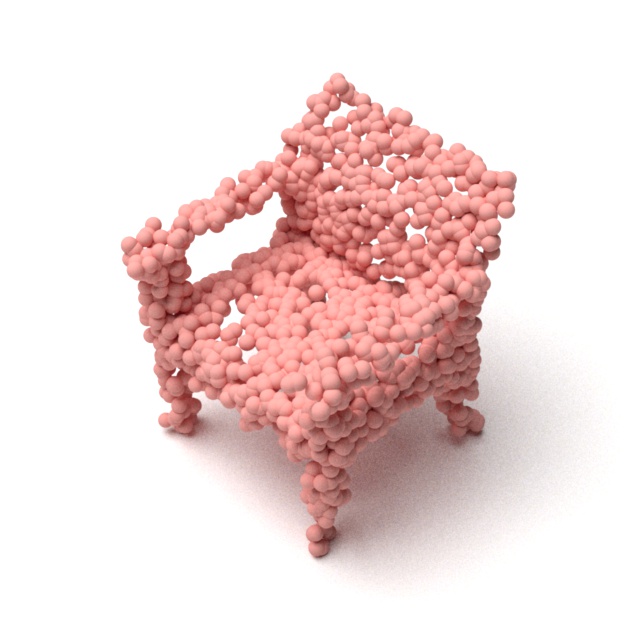} &
        \includegraphics[width=\sizea, trim={\tale} {\tab} {2.5cm} {\tat},clip]{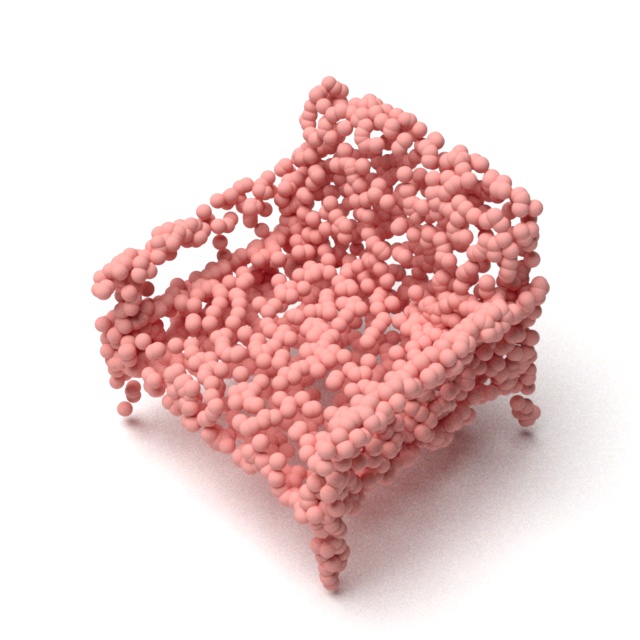} &
        \includegraphics[width=\sizea, trim={\tale} {\tab} {2.5cm} {\tat},clip]{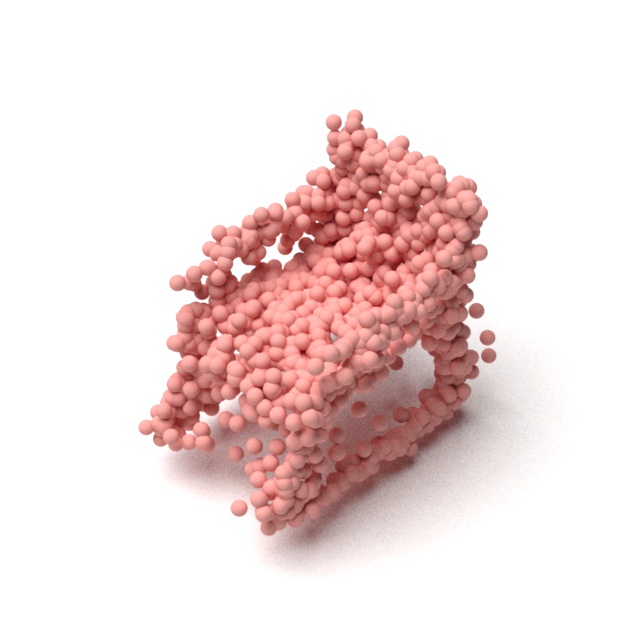} &
        \includegraphics[width=\sizea, trim={\tale} {\tab} {2.5cm} 
        {\tat},clip]{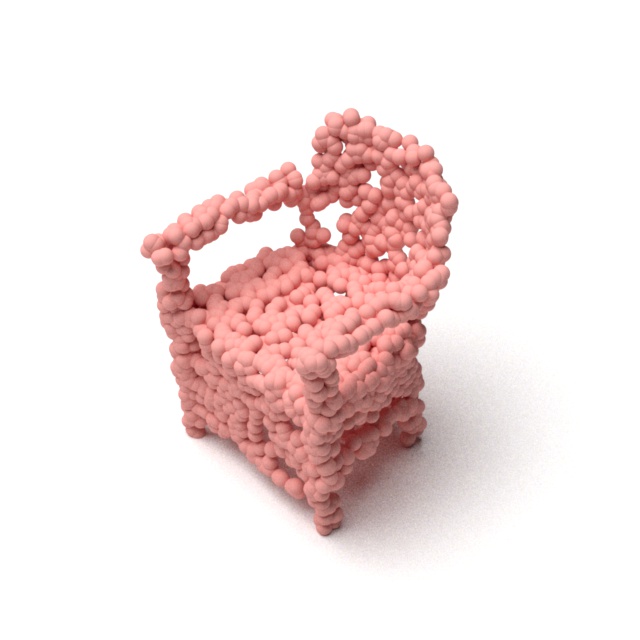} &
        \includegraphics[width=\sizea, trim={\tal} {\tab} {2.5cm} {\tat},clip]{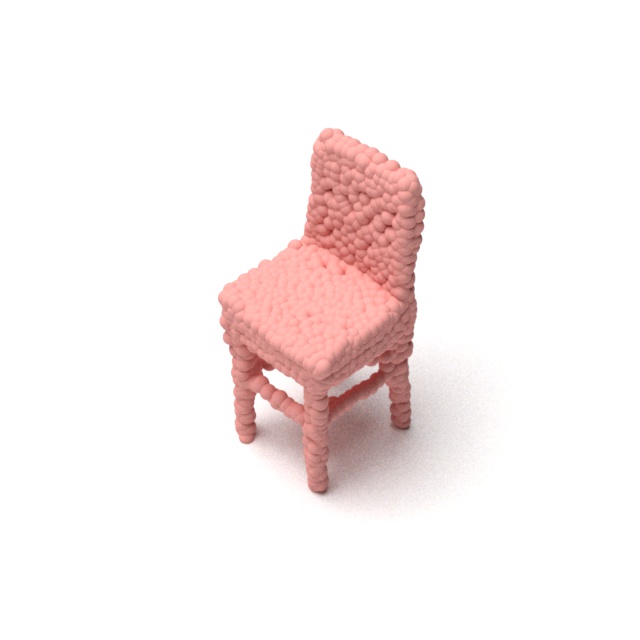} 
        \\
        \includegraphics[width=\sizea, trim={\tale} {\tab} {2.5cm} {\tat},clip]{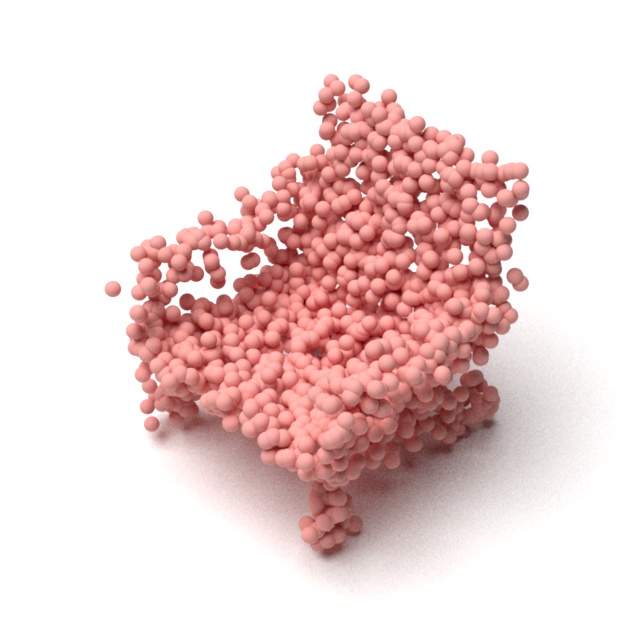} &
        \includegraphics[width=\sizea, trim={\tale} {\tab} {2.5cm} {\tat},clip]{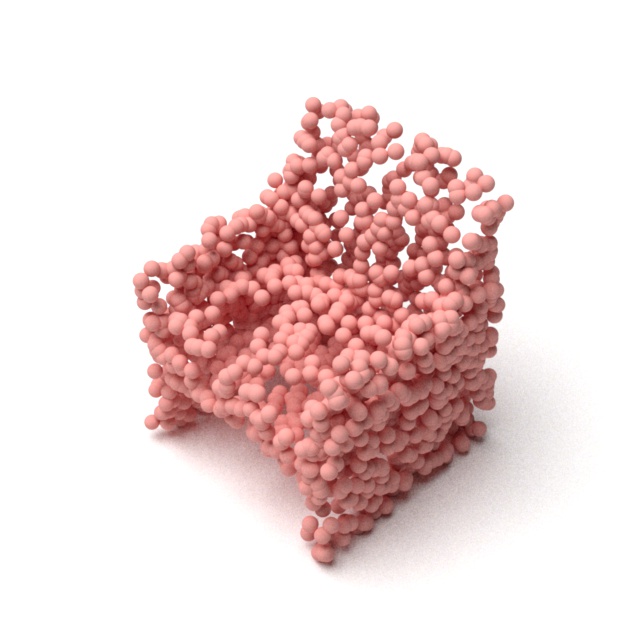} &
        \includegraphics[width=\sizea, trim={\tale} {\tab} {2.5cm} {\tat},clip]{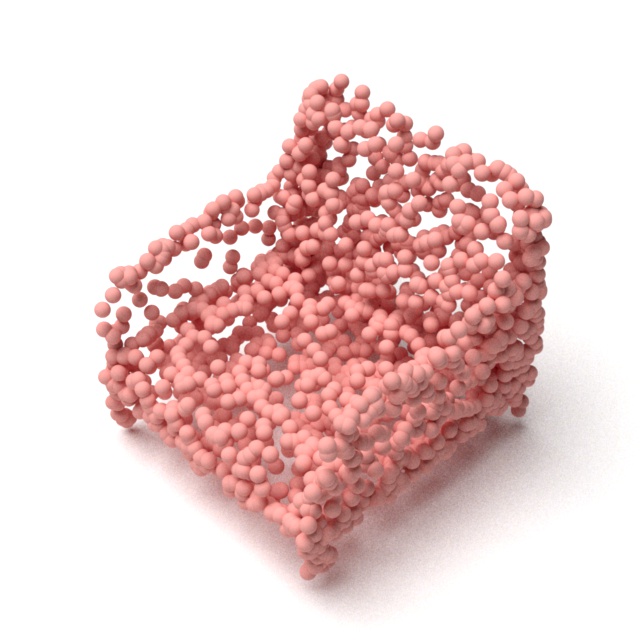} &
        \includegraphics[width=\sizea, trim={\tale} {\tab} {2.5cm} {\tat},clip]{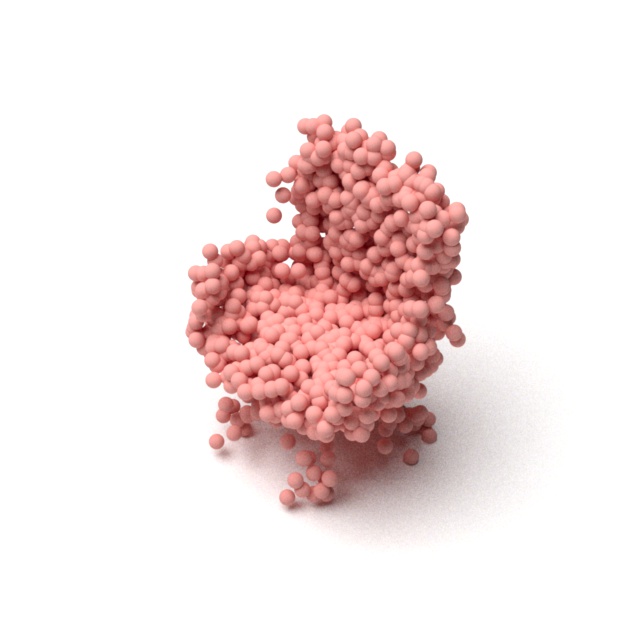} &
        \includegraphics[width=\sizea, trim={\tale} {\tab} {2.5cm} 
        {\tat},clip]{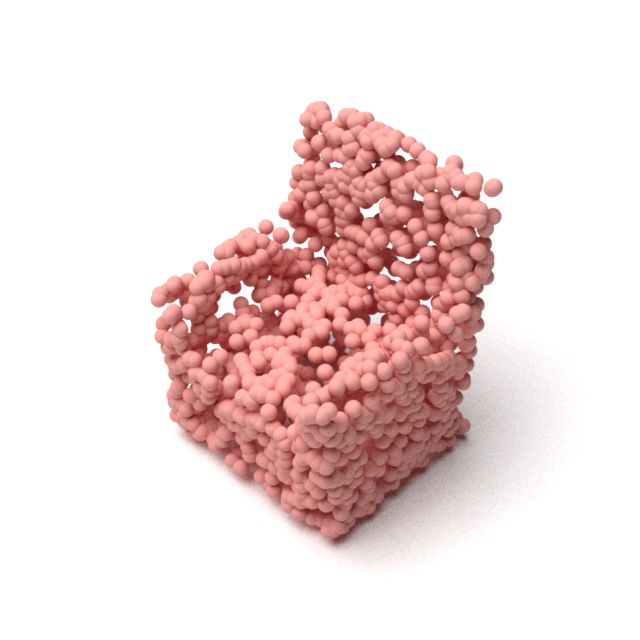} &
        \includegraphics[width=\sizea, trim={\tal} {\tab} {2.5cm} {\tat},clip]{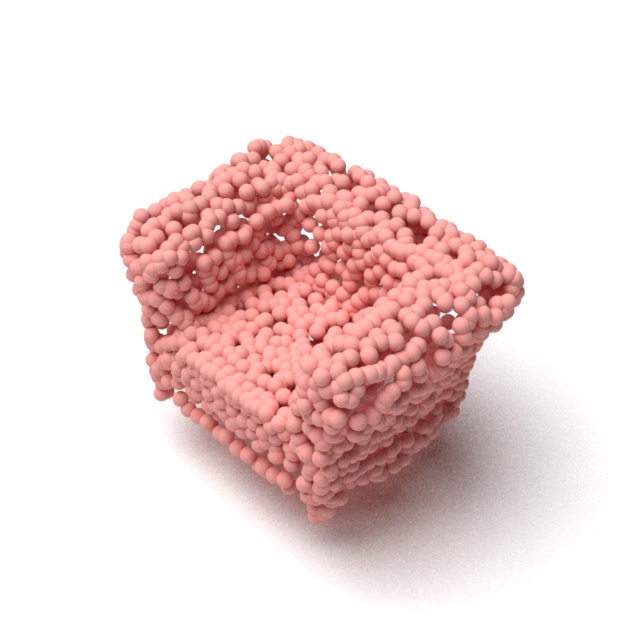} 
        \\
        PF & ShapeGF & SetVAE & DPM & PVD & Ours
    \end{tabular}

    \end{center}
    \caption{Shape generation results on ShapeNet Chair. We shown results from PF (PointFlow)~\cite{yang2019pointflow}, ShapeGF~\cite{cai2020learning}, SetVAE~\cite{kim2021setvae}, DPM~\cite{luo2021diffusion}, and PVD~\cite{zhou20213d}.
    }

    \label{fig:sup:gen:chair}
\end{figure}

\begin{figure}[h]
    \begin{center}
    \newcommand{\sizea}{0.165\linewidth}
    \newcommand{\sizeb}{0.165\linewidth}
    \newcommand{\sizec}{0.165\linewidth}
    \newcommand{\tare}{5cm}
    \newcommand{\tale}{3.5cm}
    \newcommand{\tal}{3.5cm}
    \newcommand{\tab}{3.5cm}
    \newcommand{\tar}{3.5cm}
    \newcommand{\tat}{5.5cm}
    \newcommand{\tcl}{3.0cm}
    \newcommand{\tcb}{3cm}
    \newcommand{\tcr}{4cm}
    \newcommand{\tct}{4.2cm}
    \newcommand{\thl}{3.0cm}
    \newcommand{\thb}{0.0cm}
    \newcommand{\thr}{3cm}
    \newcommand{\tht}{2cm}
    \setlength{\tabcolsep}{0pt}
    \renewcommand{\arraystretch}{0}
    \begin{tabular}{@{}ccccc:c@{}}
        \includegraphics[width=\sizea, trim={\tale} {\tab} {3.5cm} {\tat},clip]{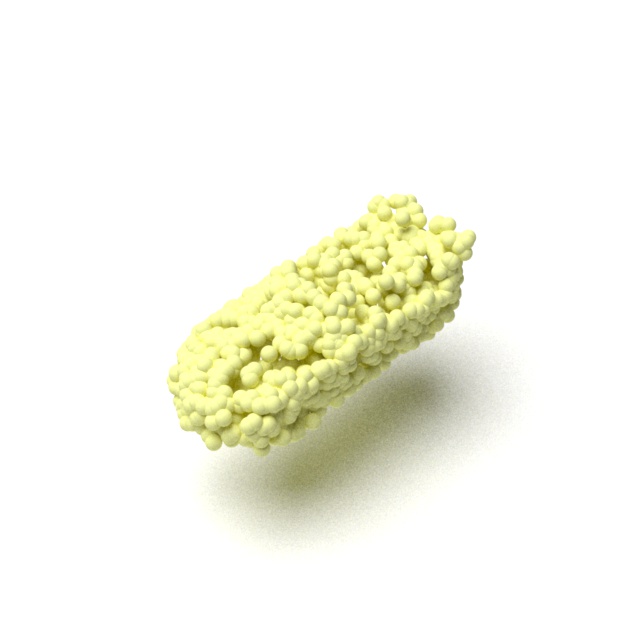} &
        \includegraphics[width=\sizea, trim={\tale} {\tab} {3.5cm} {\tat},clip]{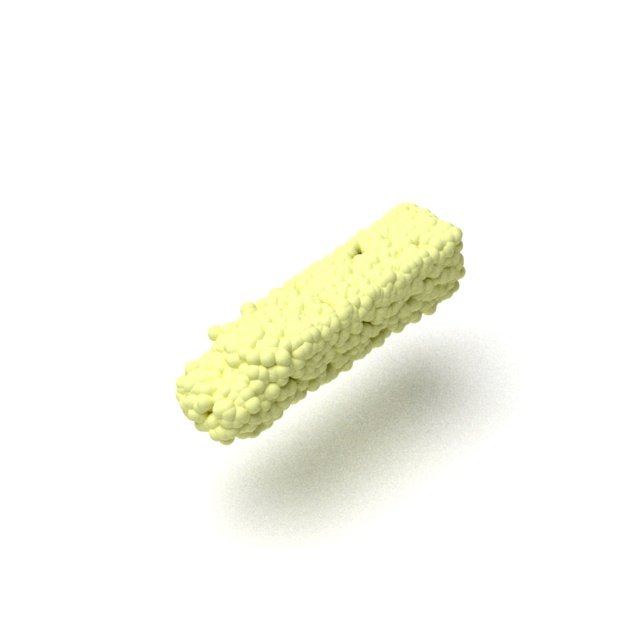} &
        \includegraphics[width=\sizea, trim={\tale} {\tab} {3.5cm} {\tat},clip]{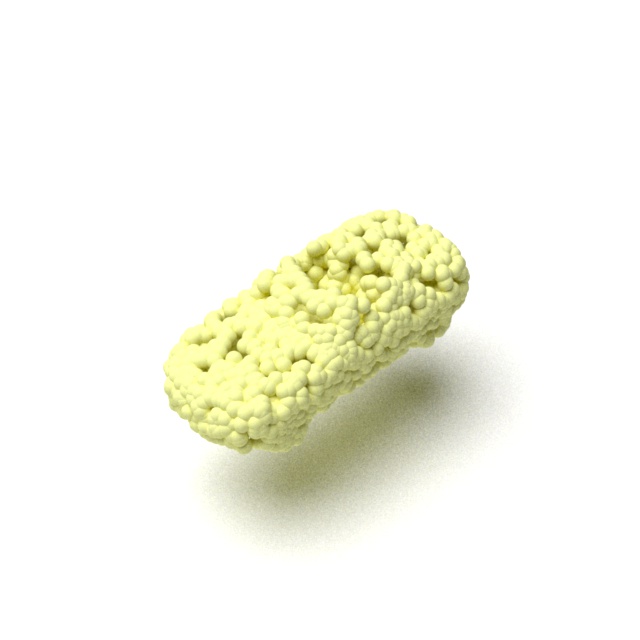} &
        \includegraphics[width=\sizea, trim={\tale} {\tab} {3.5cm} {\tat},clip]{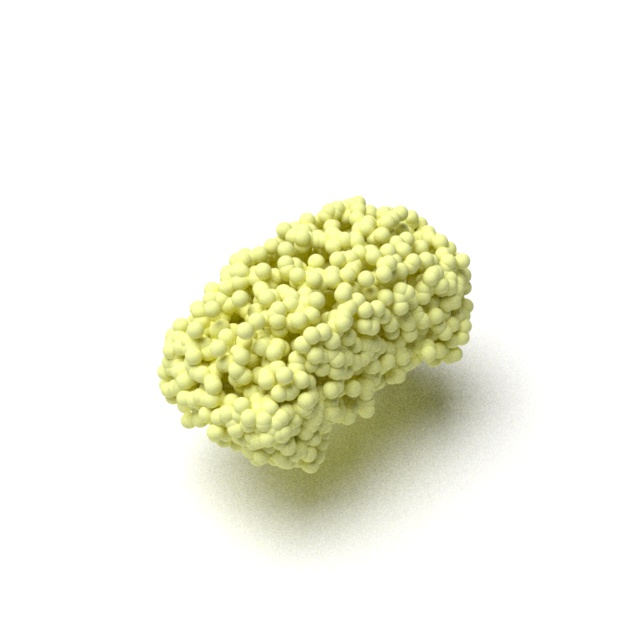} &
        \includegraphics[width=\sizea, trim={\tale} {\tab} {3.5cm} 
        {\tat},clip]{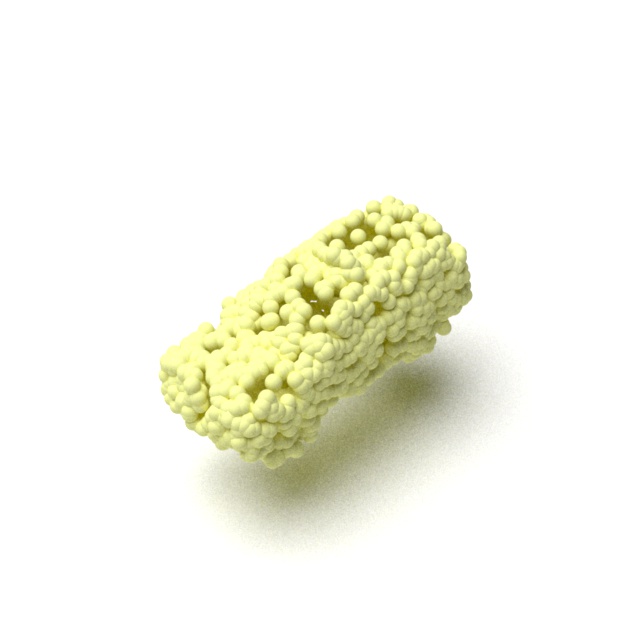} &
        \includegraphics[width=\sizea, trim={\tal} {\tab} {3.5cm} {\tat},clip]{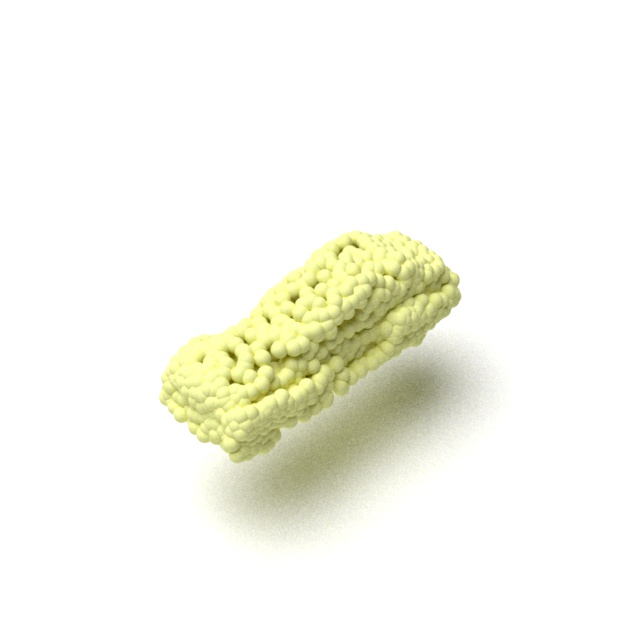} 
        \\
        \includegraphics[width=\sizea, trim={\tale} {\tab} {3.5cm} {\tat},clip]{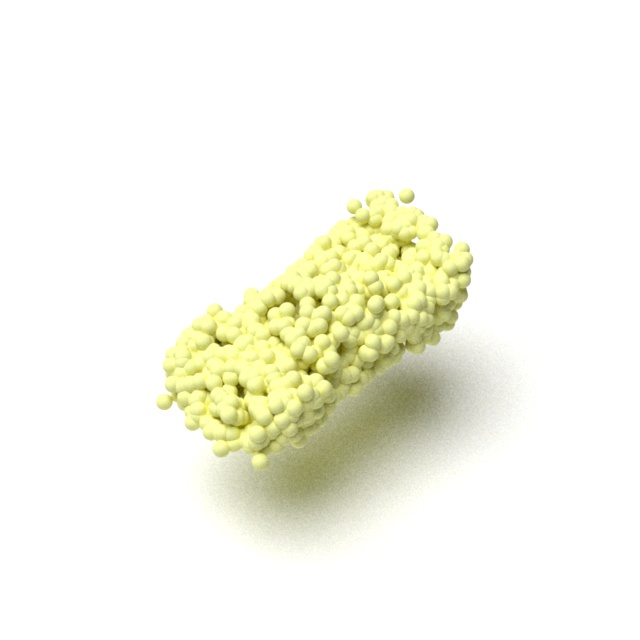} &
        \includegraphics[width=\sizea, trim={\tale} {\tab} {3.5cm} {\tat},clip]{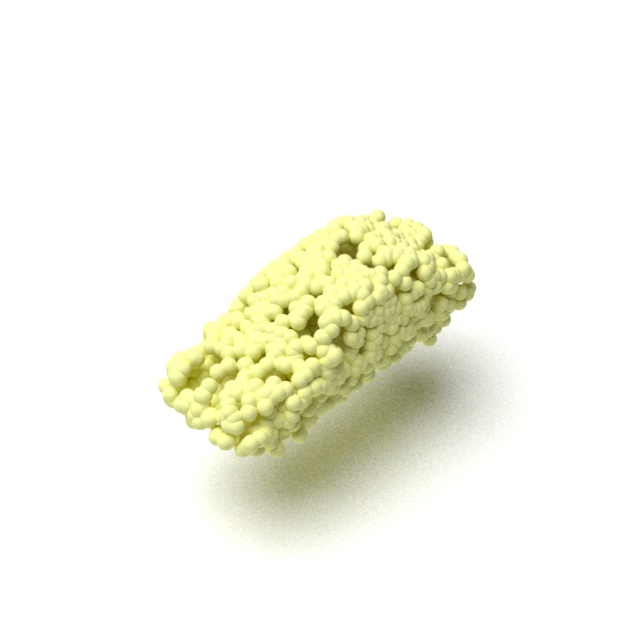} &
        \includegraphics[width=\sizea, trim={\tale} {\tab} {3.5cm} {\tat},clip]{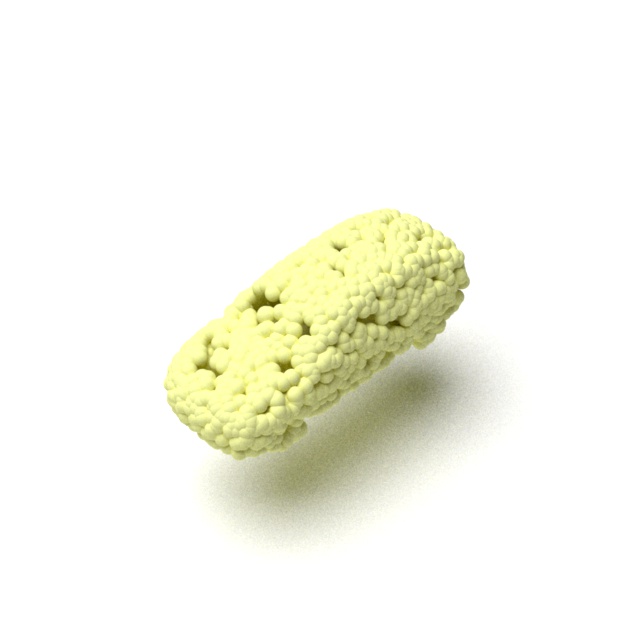} &
        \includegraphics[width=\sizea, trim={\tale} {\tab} {3.5cm} {\tat},clip]{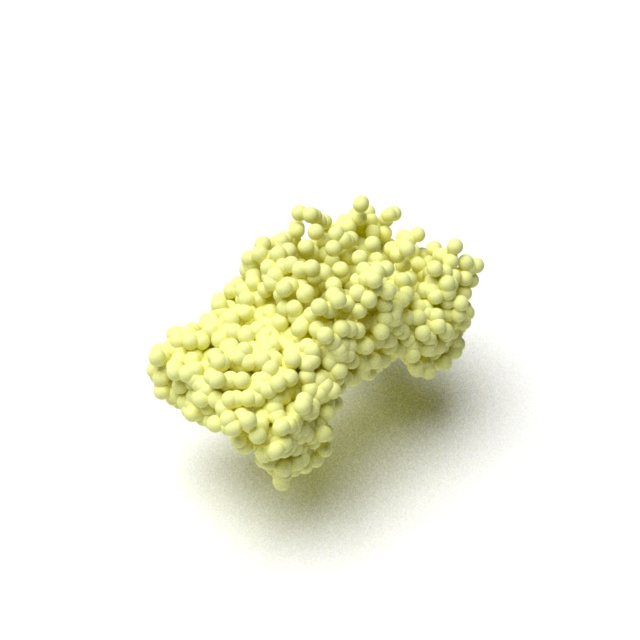} &
        \includegraphics[width=\sizea, trim={\tale} {\tab} {3.5cm} 
        {\tat},clip]{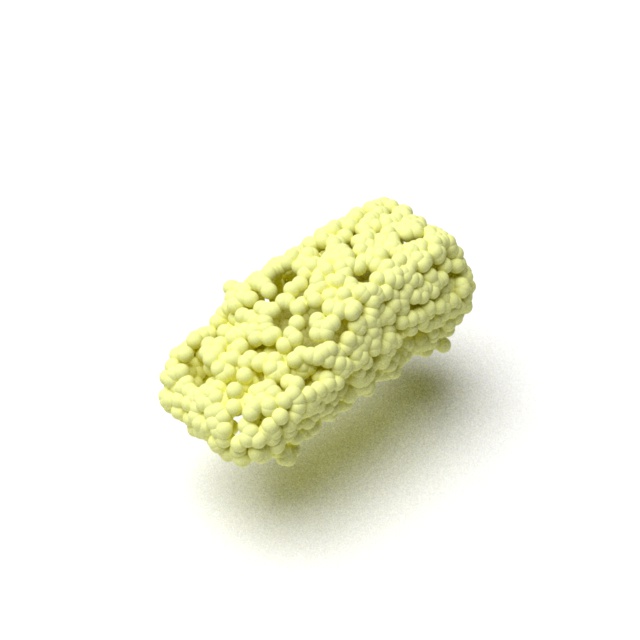} &
        \includegraphics[width=\sizea, trim={\tal} {\tab} {3.5cm} {\tat},clip]{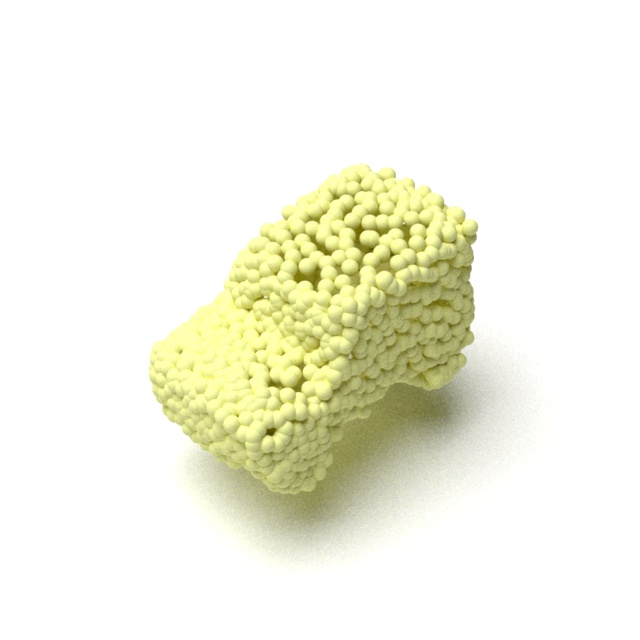} 
        \\
        \includegraphics[width=\sizea, trim={\tale} {\tab} {3.5cm} {\tat},clip]{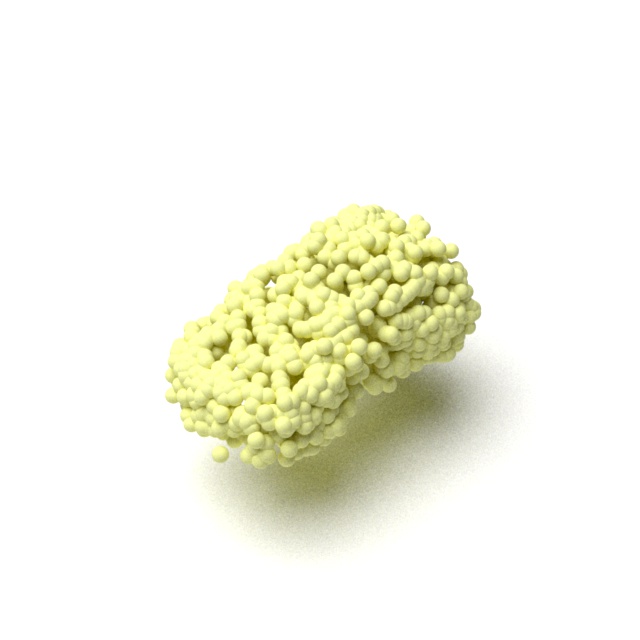} &
        \includegraphics[width=\sizea, trim={\tale} {\tab} {3.5cm} {\tat},clip]{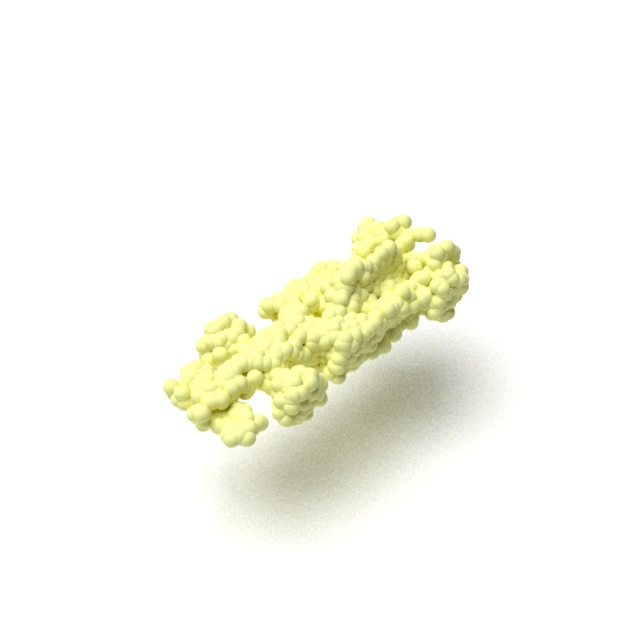} &
        \includegraphics[width=\sizea, trim={\tale} {\tab} {3.5cm} {\tat},clip]{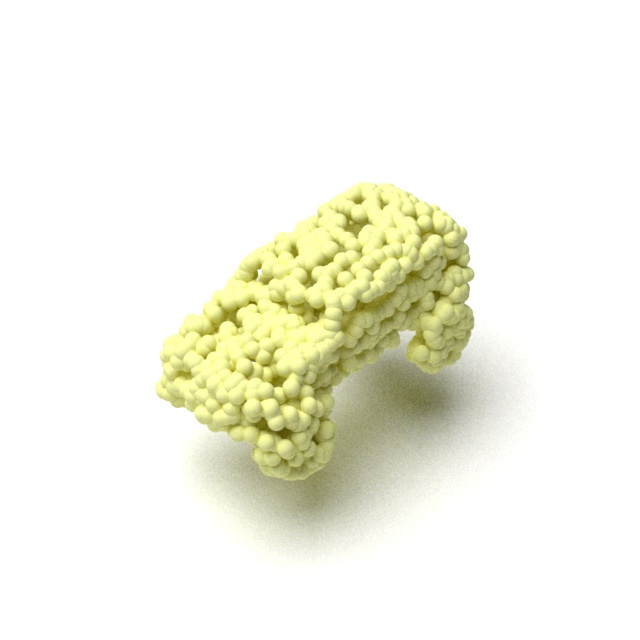} &
        \includegraphics[width=\sizea, trim={\tale} {\tab} {3.5cm} {\tat},clip]{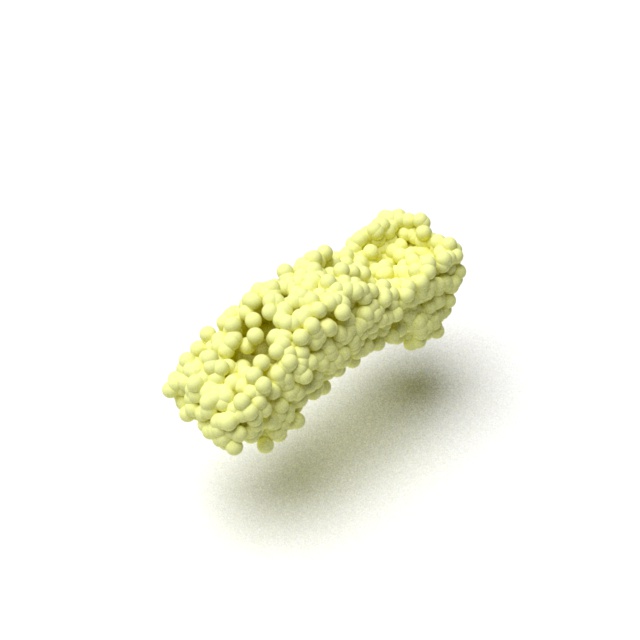} &
        \includegraphics[width=\sizea, trim={\tale} {\tab} {3.5cm} 
        {\tat},clip]{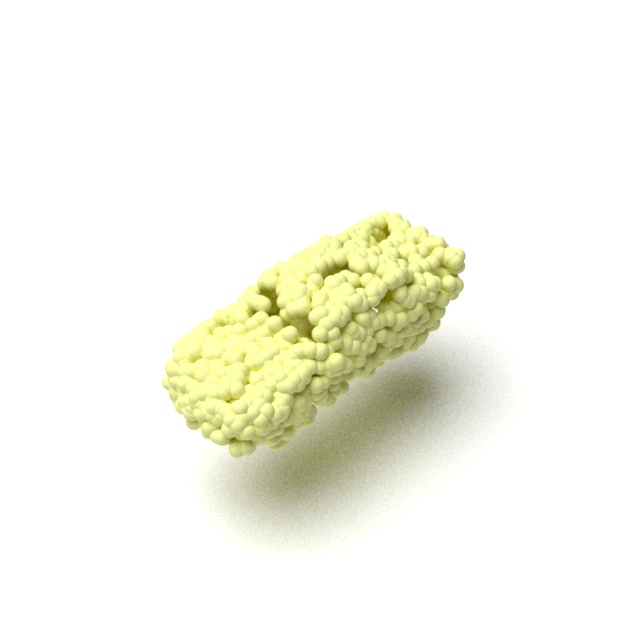} &
        \includegraphics[width=\sizea, trim={\tal} {\tab} {3.5cm} {\tat},clip]{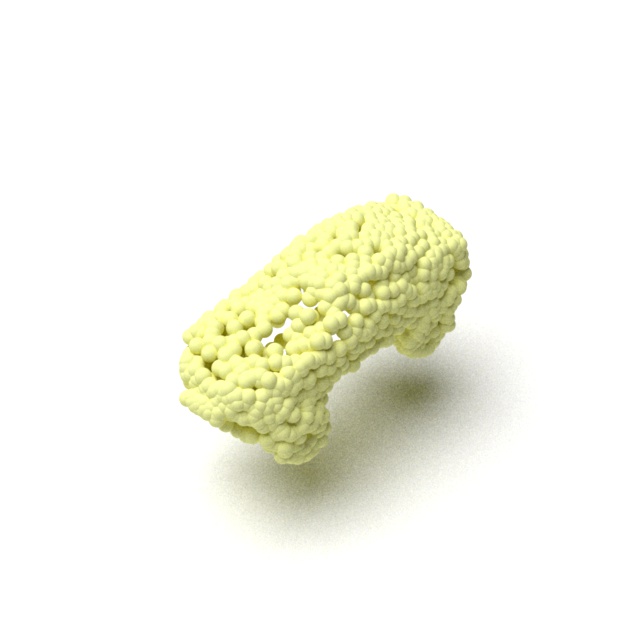} 
        \\
        \includegraphics[width=\sizea, trim={\tale} {\tab} {3.5cm} {\tat},clip]{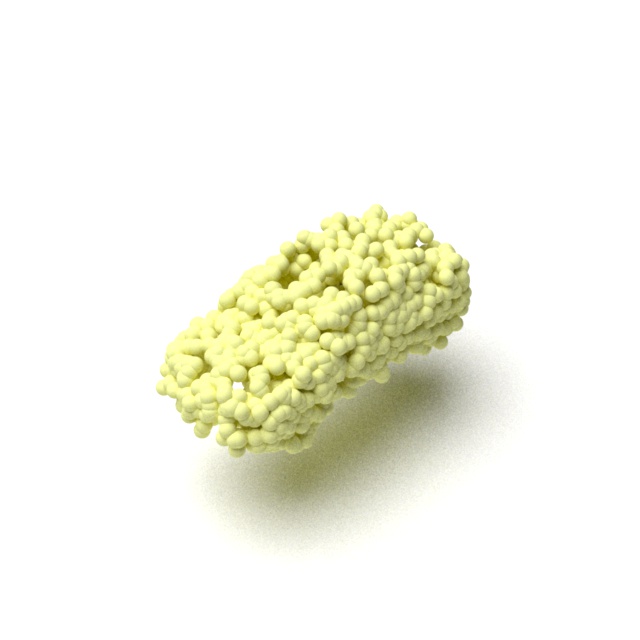} &
        \includegraphics[width=\sizea, trim={\tale} {\tab} {3.5cm} {\tat},clip]{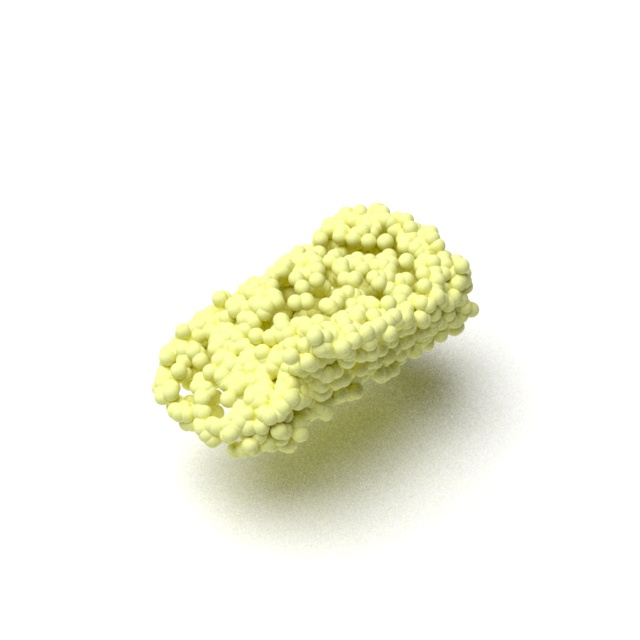} &
        \includegraphics[width=\sizea, trim={\tale} {\tab} {3.5cm} {\tat},clip]{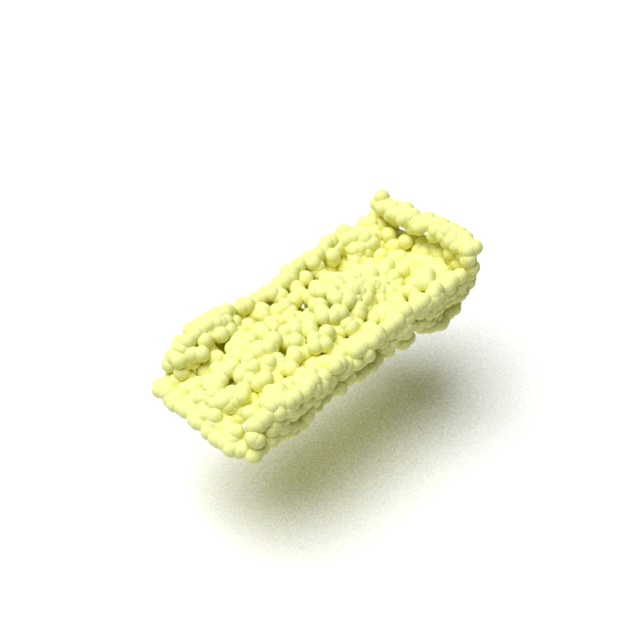} &
        \includegraphics[width=\sizea, trim={\tale} {\tab} {3.5cm} {\tat},clip]{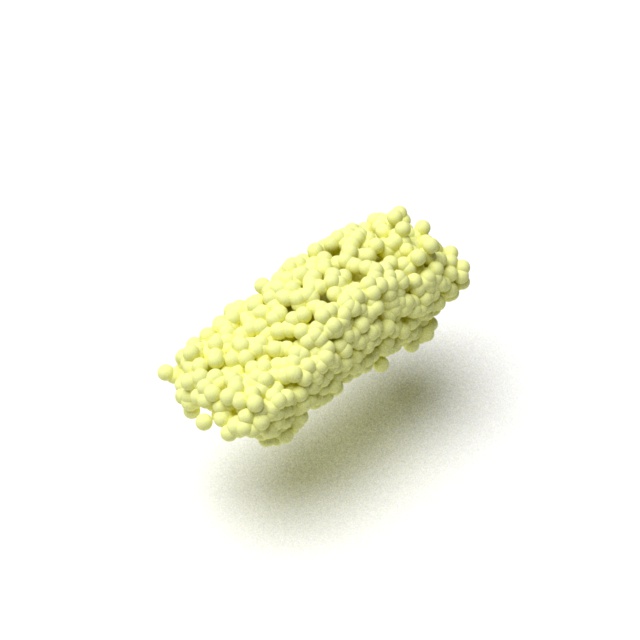} &
        \includegraphics[width=\sizea, trim={\tale} {\tab} {3.5cm} 
        {\tat},clip]{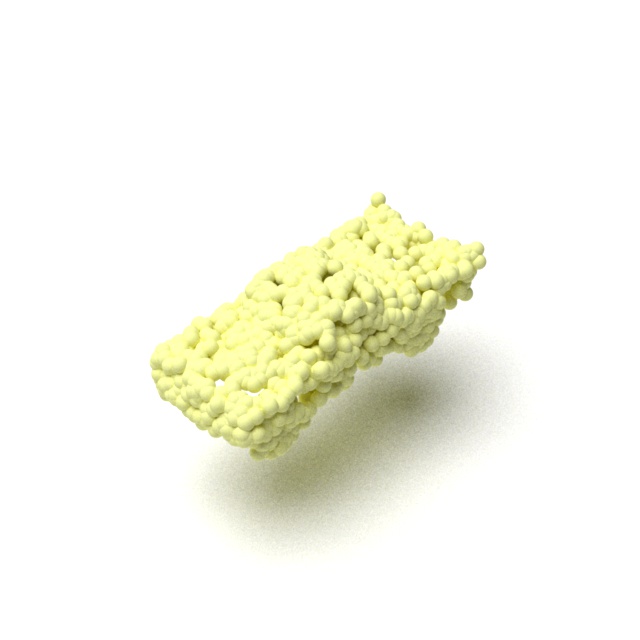} &
        \includegraphics[width=\sizea, trim={\tal} {\tab} {3.5cm} {\tat},clip]{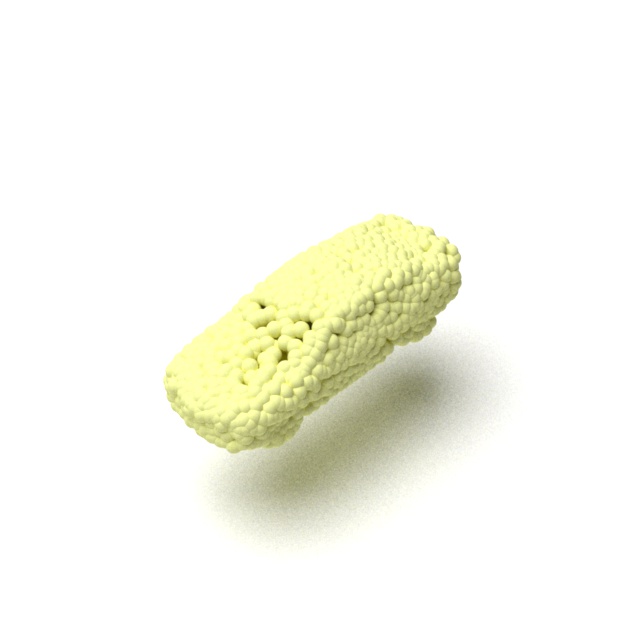} 
        \\
        \includegraphics[width=\sizea, trim={\tale} {\tab} {3.5cm} {\tat},clip]{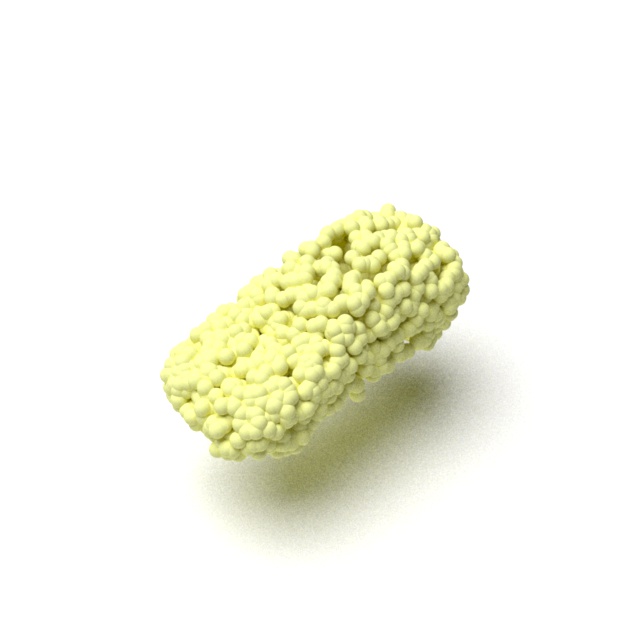} &
        \includegraphics[width=\sizea, trim={\tale} {\tab} {3.5cm} {\tat},clip]{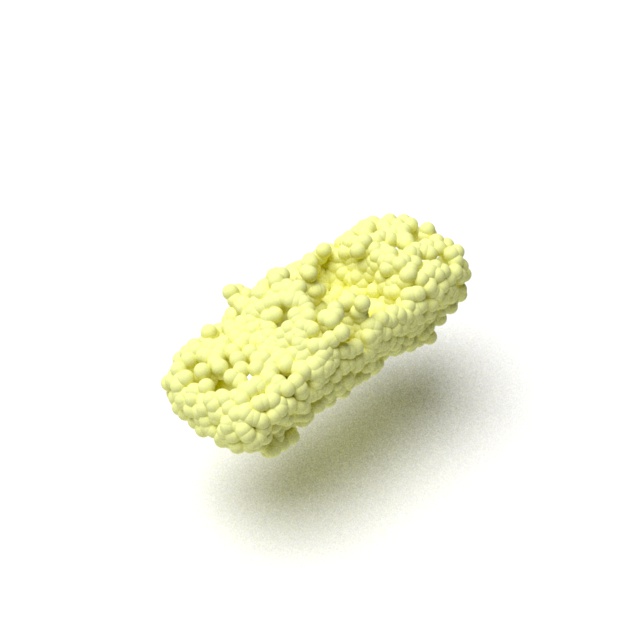} &
        \includegraphics[width=\sizea, trim={\tale} {\tab} {3.5cm} {\tat},clip]{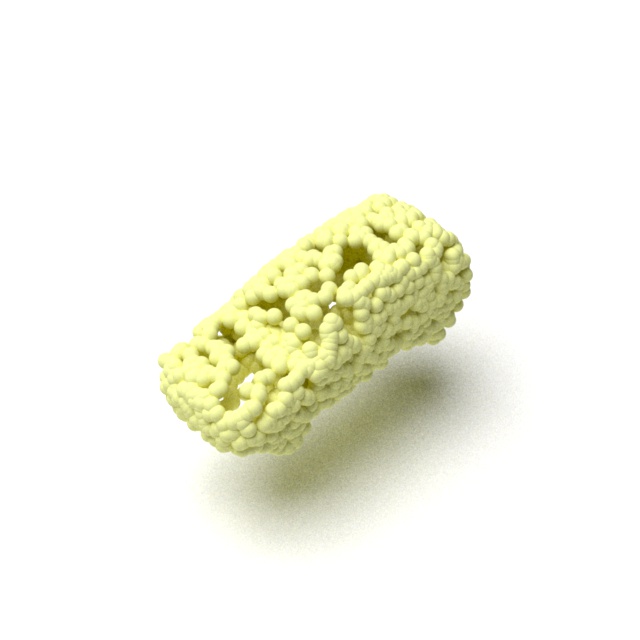} &
        \includegraphics[width=\sizea, trim={\tale} {\tab} {3.5cm} {\tat},clip]{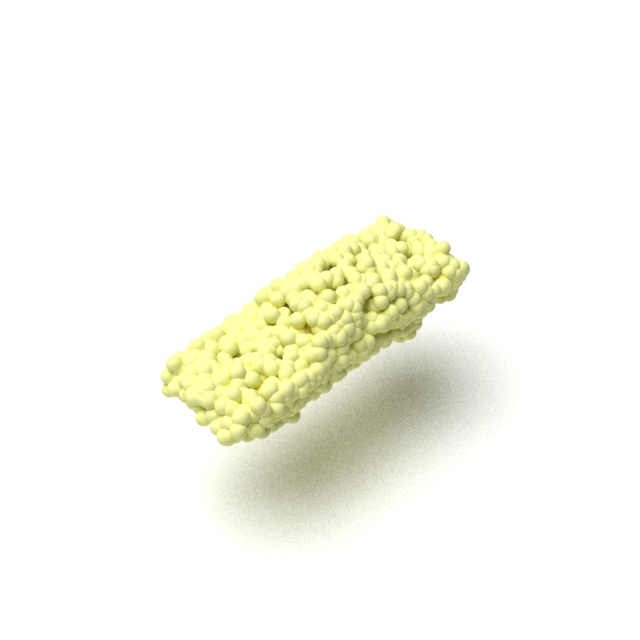} &
        \includegraphics[width=\sizea, trim={\tale} {\tab} {3.5cm} 
        {\tat},clip]{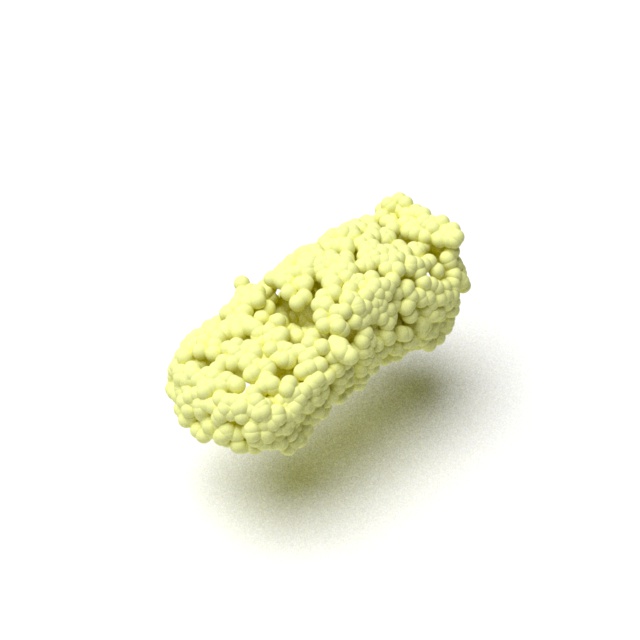} &
        \includegraphics[width=\sizea, trim={\tal} {\tab} {3.5cm} {\tat},clip]{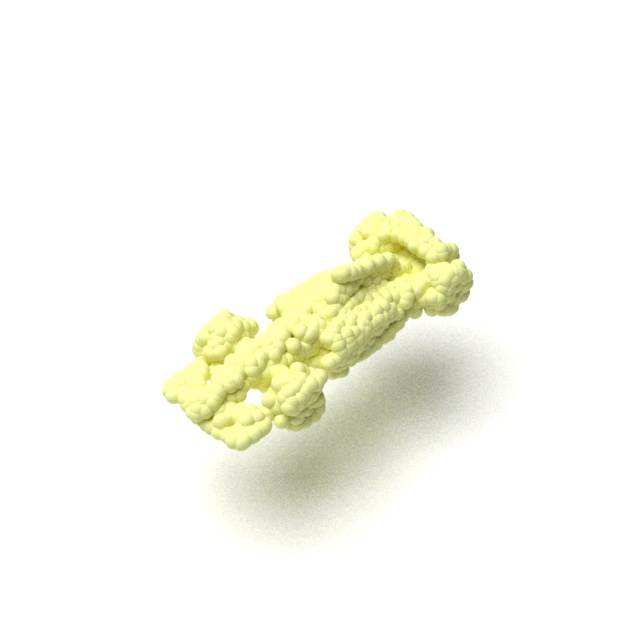} 
        \\
        \includegraphics[width=\sizea, trim={\tale} {\tab} {3.5cm} {\tat},clip]{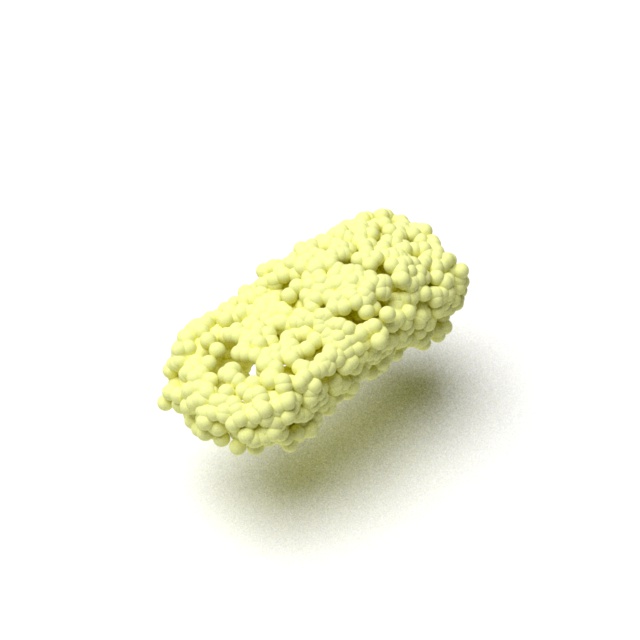} &
        \includegraphics[width=\sizea, trim={\tale} {\tab} {3.5cm} {\tat},clip]{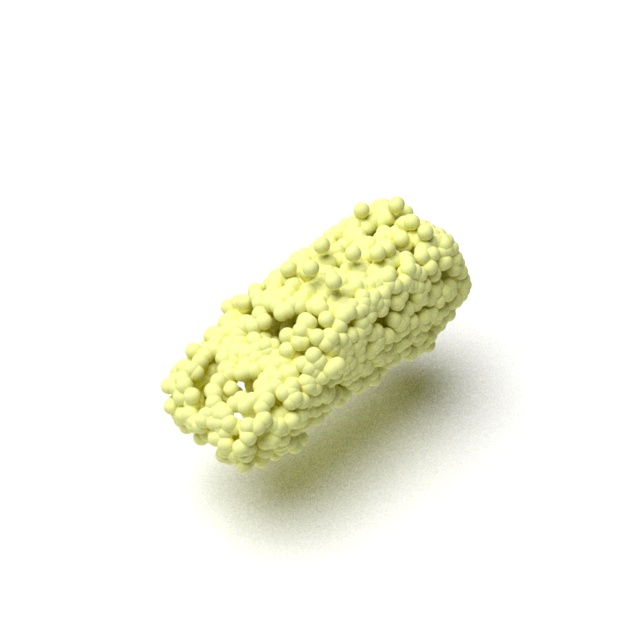} &
        \includegraphics[width=\sizea, trim={\tale} {\tab} {3.5cm} {\tat},clip]{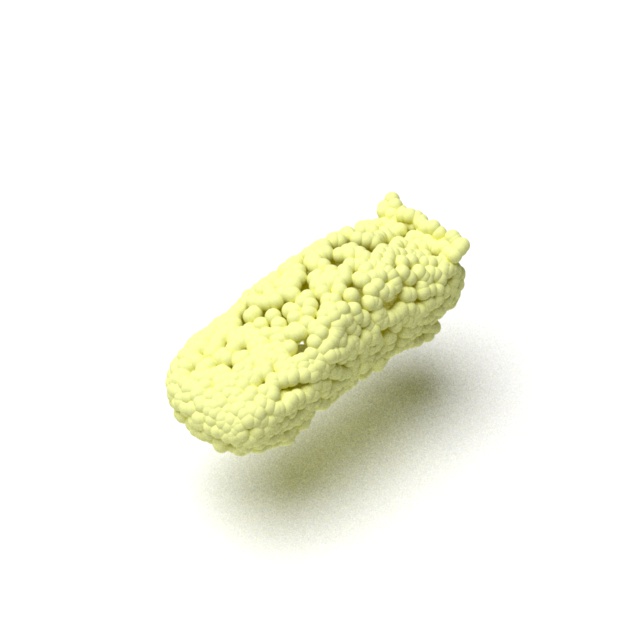} &
        \includegraphics[width=\sizea, trim={\tale} {\tab} {3.5cm} {\tat},clip]{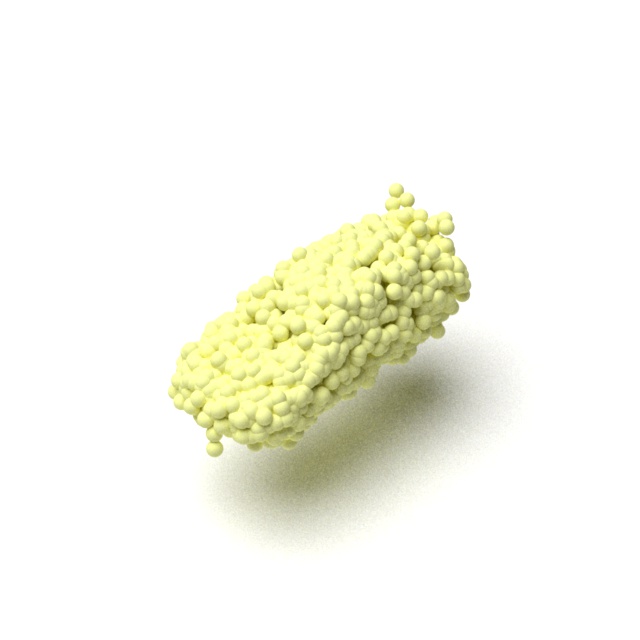} &
        \includegraphics[width=\sizea, trim={\tale} {\tab} {3.5cm} 
        {\tat},clip]{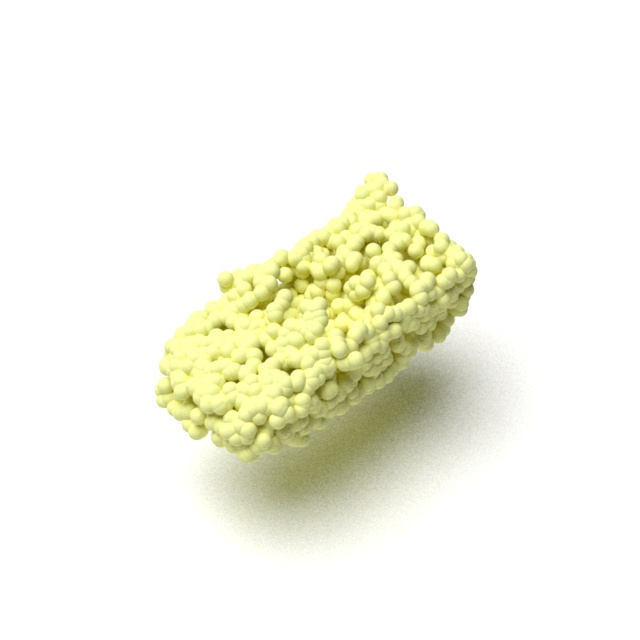} &
        \includegraphics[width=\sizea, trim={\tal} {\tab} {3.5cm} {\tat},clip]{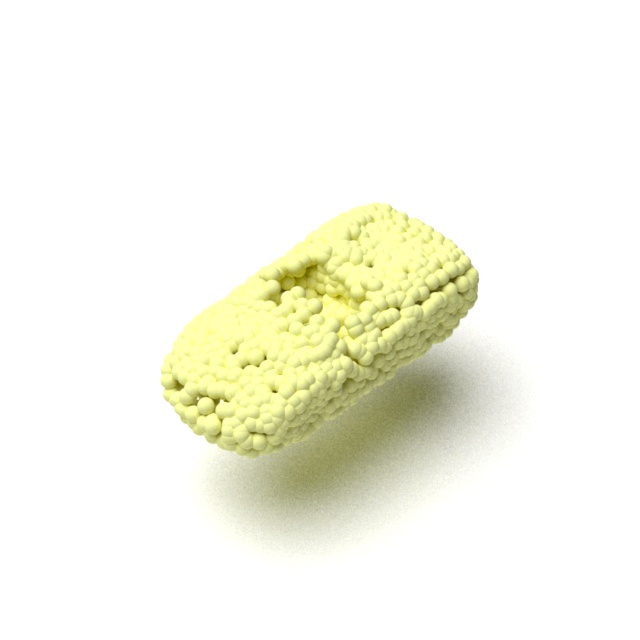} 
        \\
        \includegraphics[width=\sizea, trim={\tale} {\tab} {3.5cm} {\tat},clip]{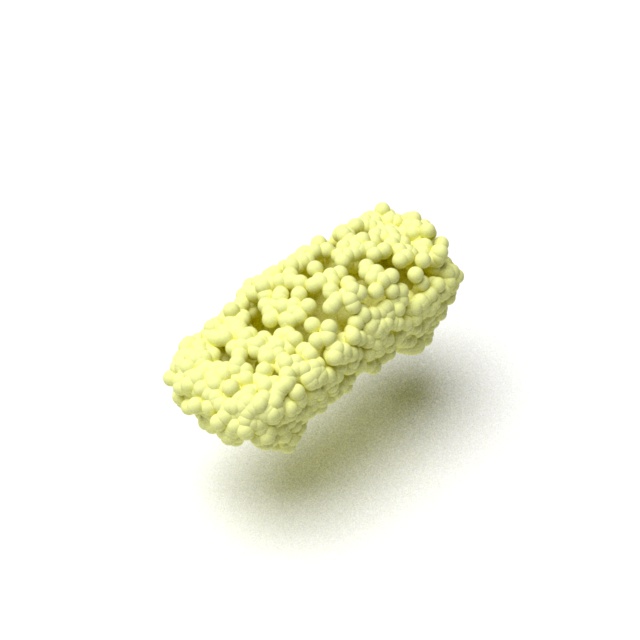} &
        \includegraphics[width=\sizea, trim={\tale} {\tab} {3.5cm} {\tat},clip]{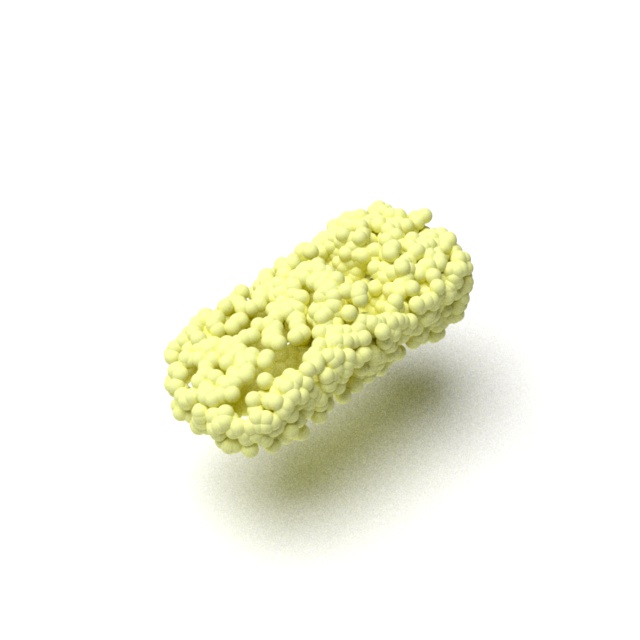} &
        \includegraphics[width=\sizea, trim={\tale} {\tab} {3.5cm} {\tat},clip]{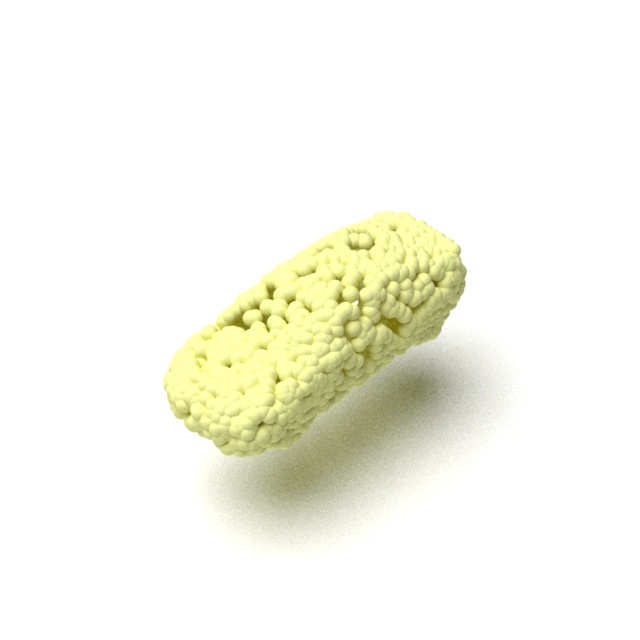} &
        \includegraphics[width=\sizea, trim={\tale} {\tab} {3.5cm} {\tat},clip]{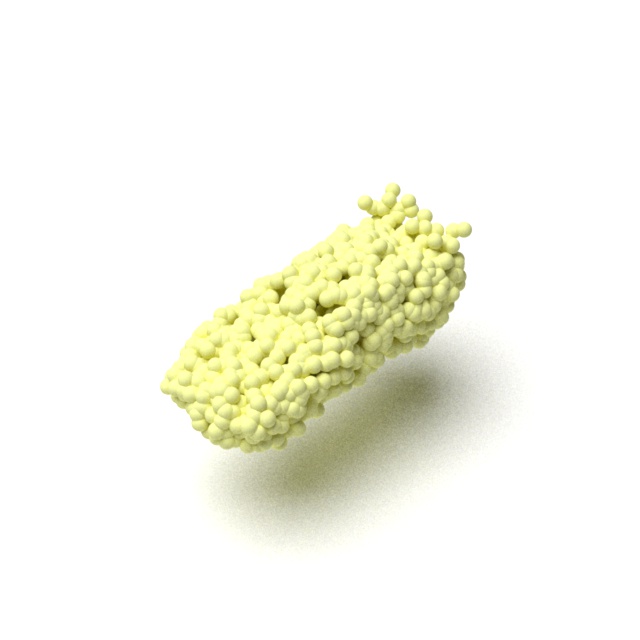} &
        \includegraphics[width=\sizea, trim={\tale} {\tab} {3.5cm} 
        {\tat},clip]{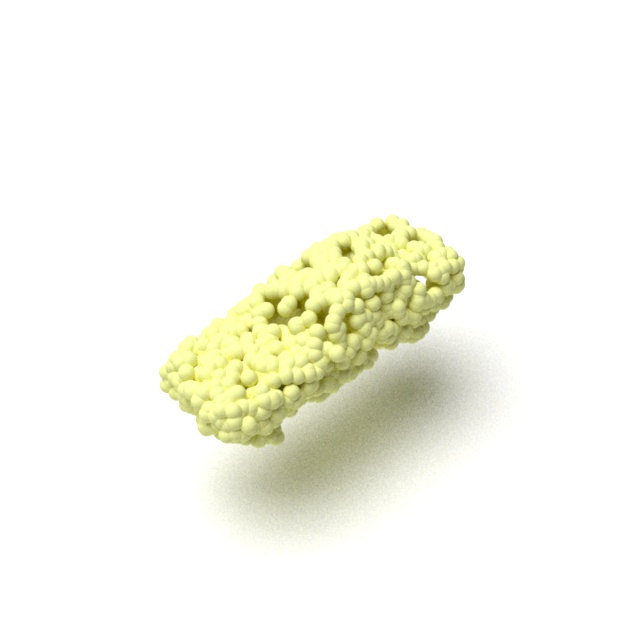} &
        \includegraphics[width=\sizea, trim={\tal} {\tab} {3.5cm} {\tat},clip]{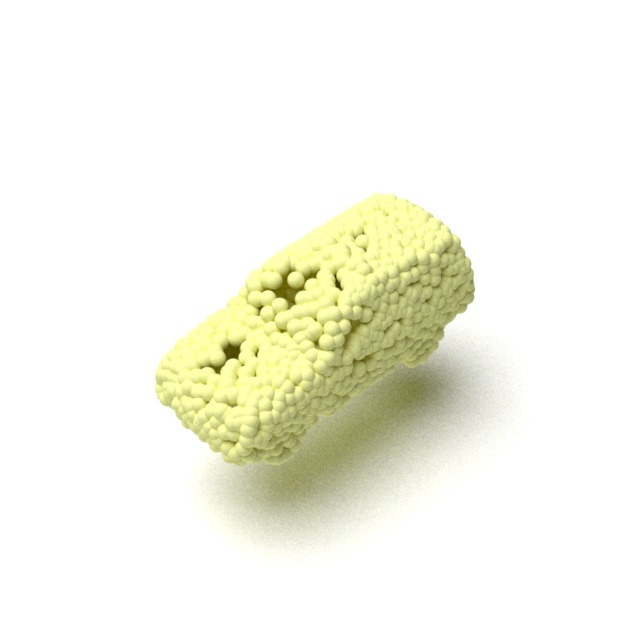} 
        \\
        \includegraphics[width=\sizea, trim={\tale} {\tab} {3.5cm} {\tat},clip]{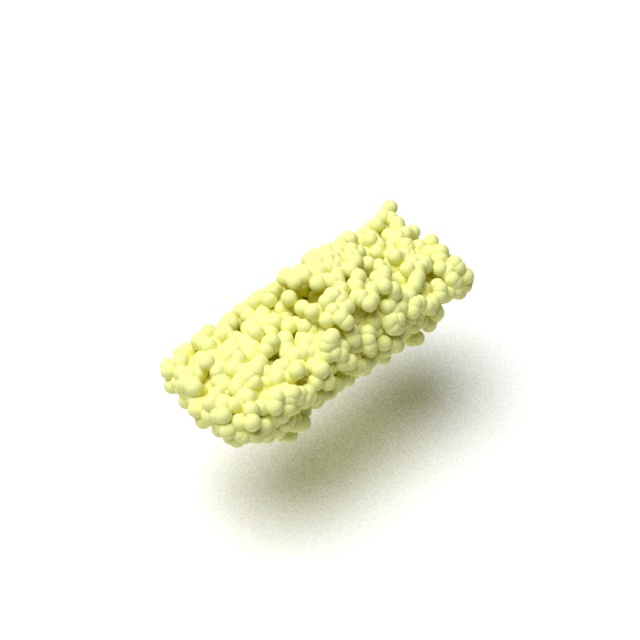} &
        \includegraphics[width=\sizea, trim={\tale} {\tab} {3.5cm} {\tat},clip]{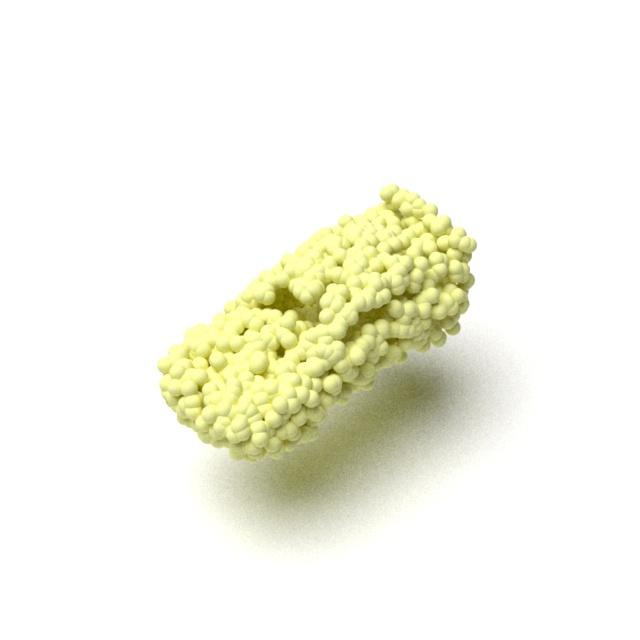} &
        \includegraphics[width=\sizea, trim={\tale} {\tab} {3.5cm} {\tat},clip]{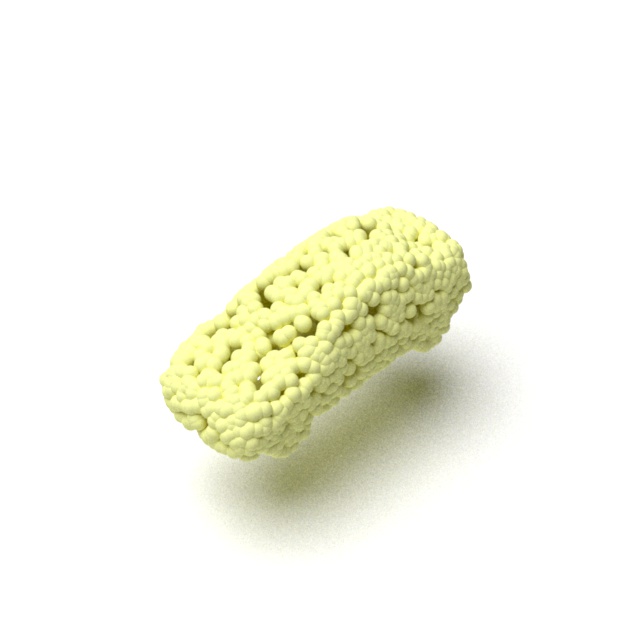} &
        \includegraphics[width=\sizea, trim={\tale} {\tab} {3.5cm} {\tat},clip]{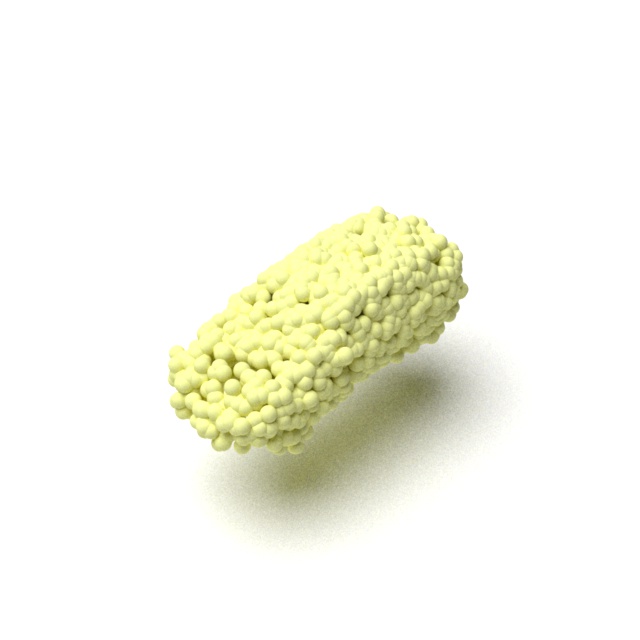} &
        \includegraphics[width=\sizea, trim={\tale} {\tab} {3.5cm} 
        {\tat},clip]{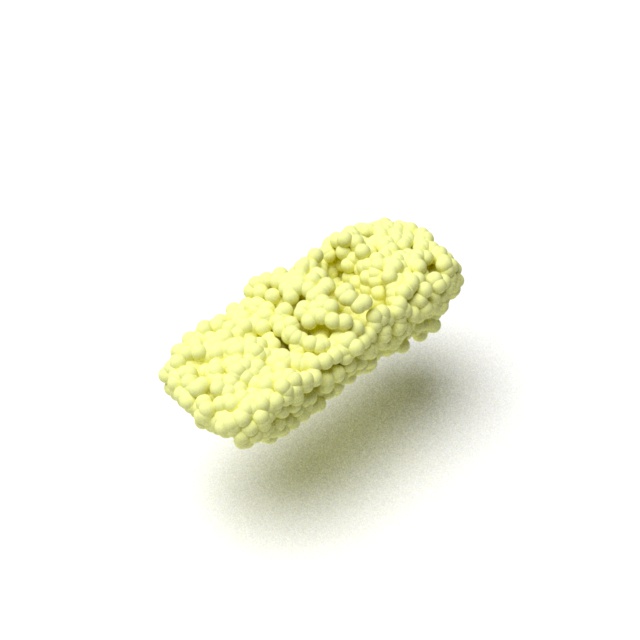} &
        \includegraphics[width=\sizea, trim={\tal} {\tab} {3.5cm} {\tat},clip]{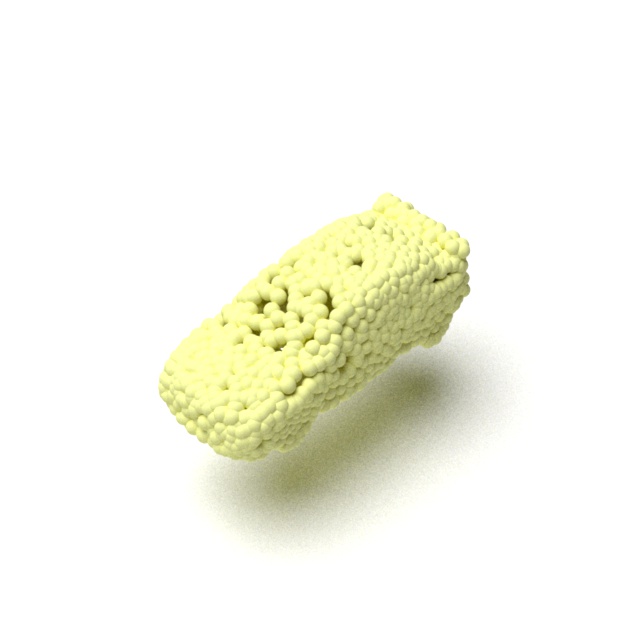} 
        \\
        \includegraphics[width=\sizea, trim={\tale} {\tab} {3.5cm} {\tat},clip]{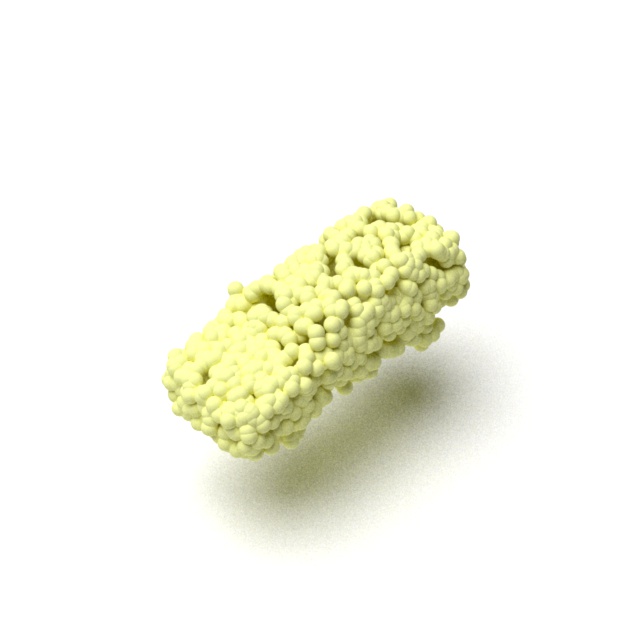} &
        \includegraphics[width=\sizea, trim={\tale} {\tab} {3.5cm} {\tat},clip]{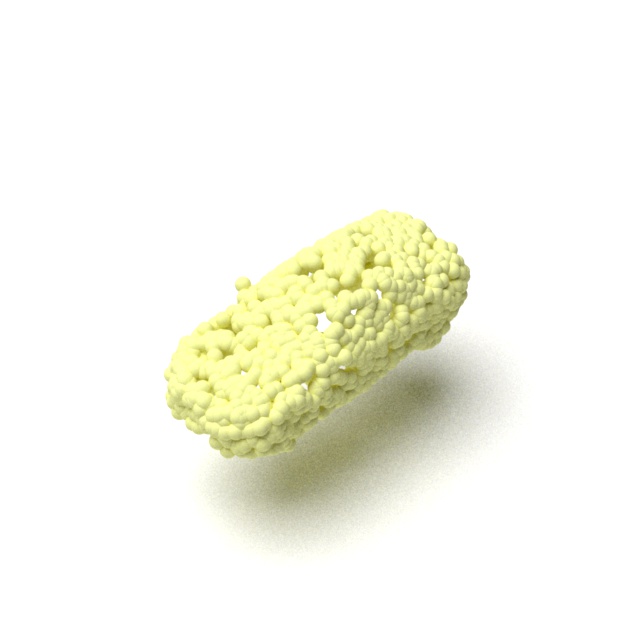} &
        \includegraphics[width=\sizea, trim={\tale} {\tab} {3.5cm} {\tat},clip]{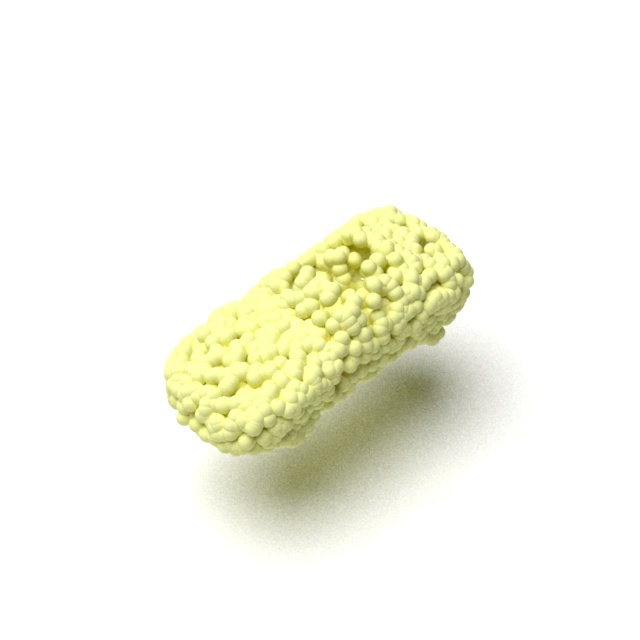} &
        \includegraphics[width=\sizea, trim={\tale} {\tab} {3.5cm} {\tat},clip]{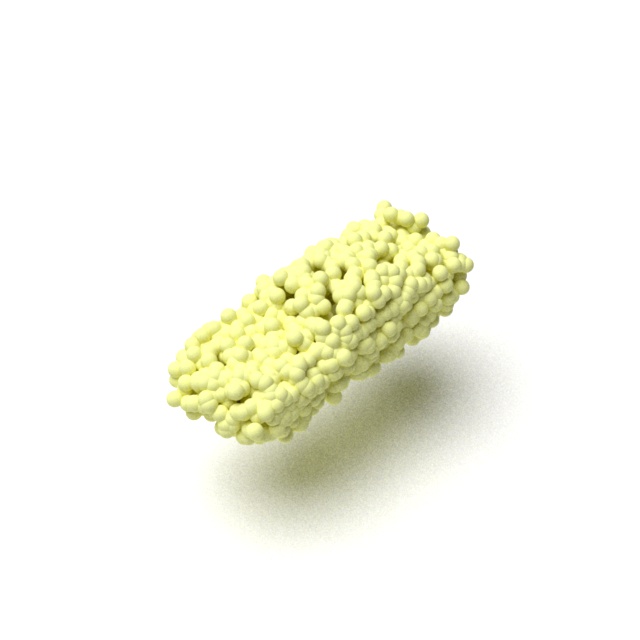} &
        \includegraphics[width=\sizea, trim={\tale} {\tab} {3.5cm} 
        {\tat},clip]{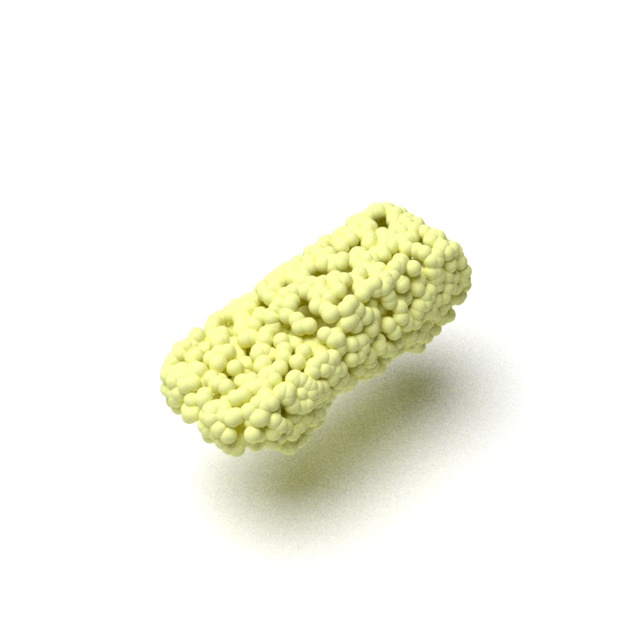} &
        \includegraphics[width=\sizea, trim={\tal} {\tab} {3.5cm} {\tat},clip]{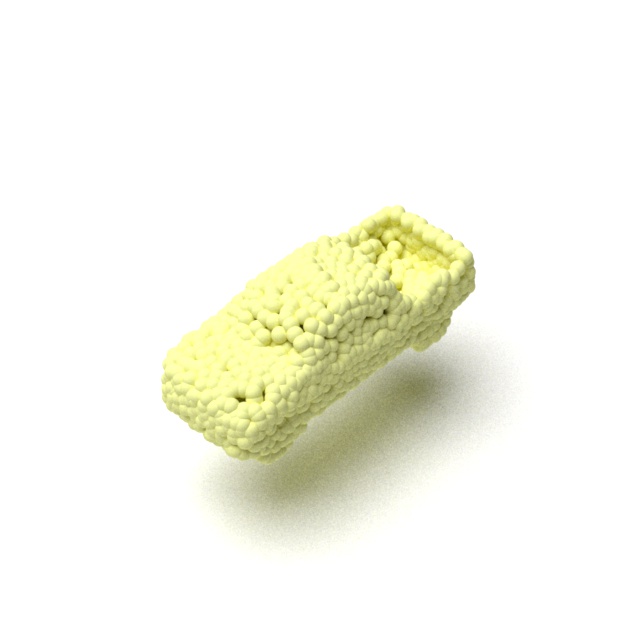} 
        \\
        PF & ShapeGF & SetVAE & DPM & PVD & Ours
    \end{tabular}

    \end{center}
    \caption{Shape generation results on ShapeNet Car. We shown results from PF (PointFlow)~\cite{yang2019pointflow}, ShapeGF~\cite{cai2020learning}, SetVAE~\cite{kim2021setvae}, DPM~\cite{luo2021diffusion}, and PVD~\cite{zhou20213d}.
    }

    \label{fig:sup:gen:car}
\end{figure}

In Figure~\ref{fig:sup:gen:airplane}, Figure~\ref{fig:sup:gen:chair}, and Figure~\ref{fig:sup:gen:car}, we show more unconditional generation results. The results suggest that our model can generate diverse shape in high fidelity.

\clearpage
\section{Qualitative Results of Conditional Generation}
\label{sec:con_gen}
\begin{figure}[h]
    \begin{center}
    \newcommand{\sizea}{0.124\linewidth}
    \newcommand{\sizeb}{0.124\linewidth}
    \newcommand{\sizec}{0.124\linewidth}
    \newcommand{\tare}{5cm}
    \newcommand{\tale}{3.5cm}
    \newcommand{\tal}{3.5cm}
    \newcommand{\tab}{2.5cm}
    \newcommand{\tar}{3.5cm}
    \newcommand{\tat}{3cm}
    \newcommand{\tcl}{3.0cm}
    \newcommand{\tcb}{3cm}
    \newcommand{\tcr}{4cm}
    \newcommand{\tct}{4.2cm}
    \newcommand{\thl}{3.0cm}
    \newcommand{\thb}{0.0cm}
    \newcommand{\thr}{3cm}
    \newcommand{\tht}{2cm}
    \setlength{\tabcolsep}{0pt}
    \renewcommand{\arraystretch}{0}
    \begin{tabular}{@{}cccc:cccc@{}}
        \includegraphics[width=\sizea, trim={\tale} {\tab} {2cm} {\tat},clip]{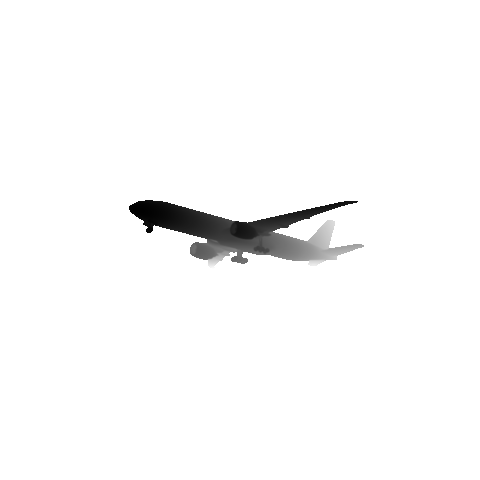} &
        \includegraphics[width=\sizec, trim={\tale} {\tab} {2cm} {\tat},clip]{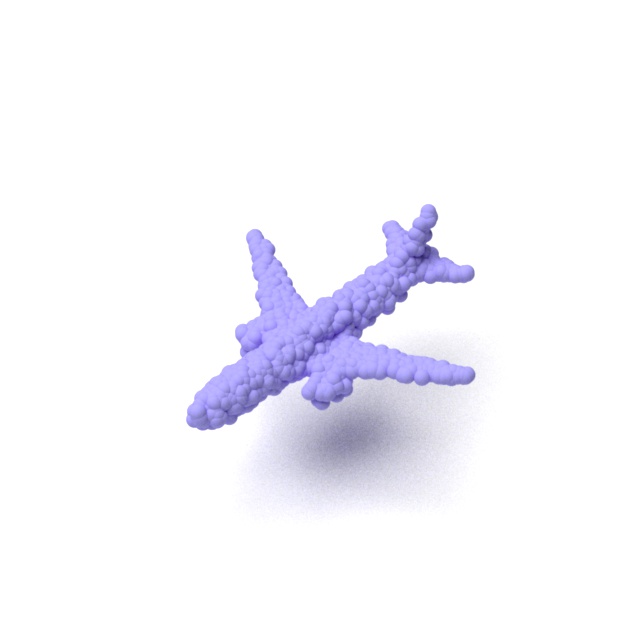} &
        \includegraphics[width=\sizeb, trim={\tale} {\tab} {2cm} {\tat},clip]{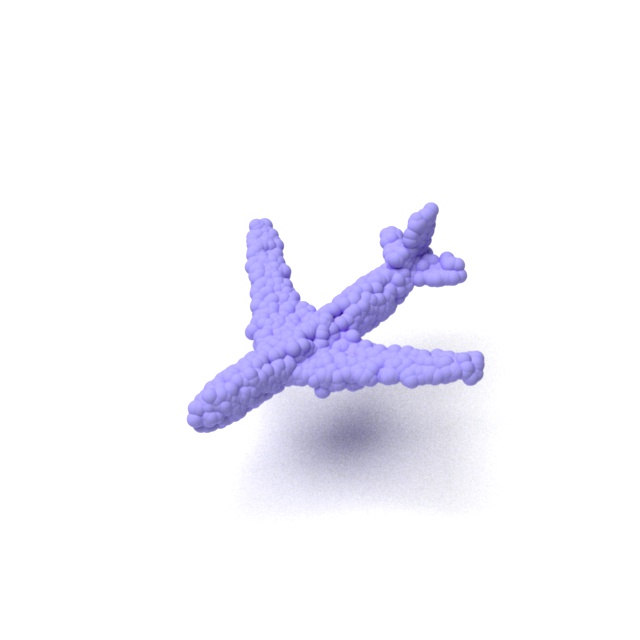} &
        \includegraphics[width=\sizea, trim={\tale} {\tab} {2cm} {\tat},clip]{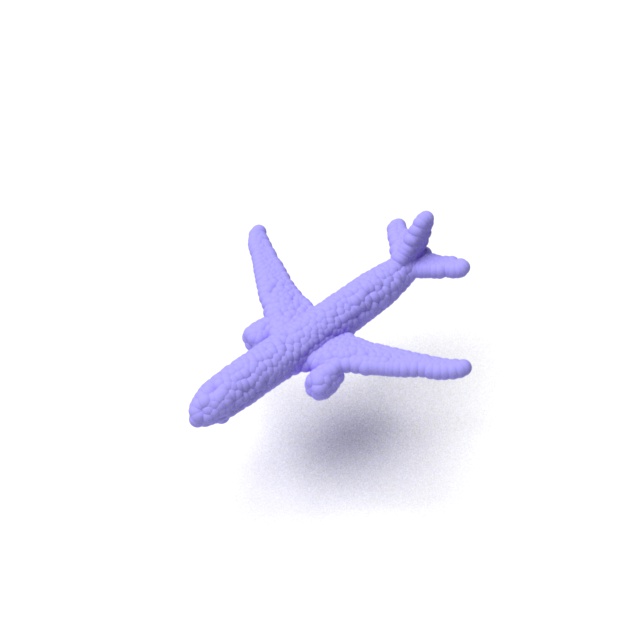} &
        \includegraphics[width=\sizea, trim={\tale} {\tab} {2cm} {\tat},clip]{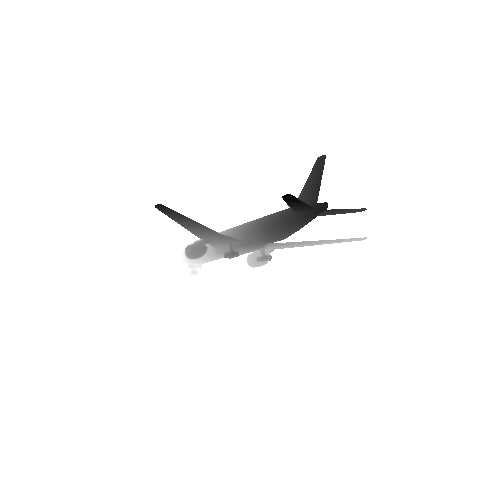} &
        \includegraphics[width=\sizec, trim={\tale} {\tab} {2cm} {\tat},clip]{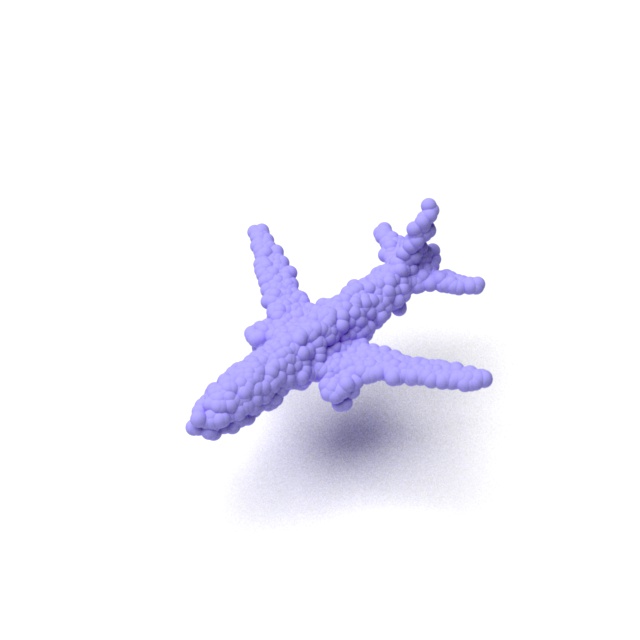} &
        \includegraphics[width=\sizeb, trim={\tale} {\tab} {2cm} {\tat},clip]{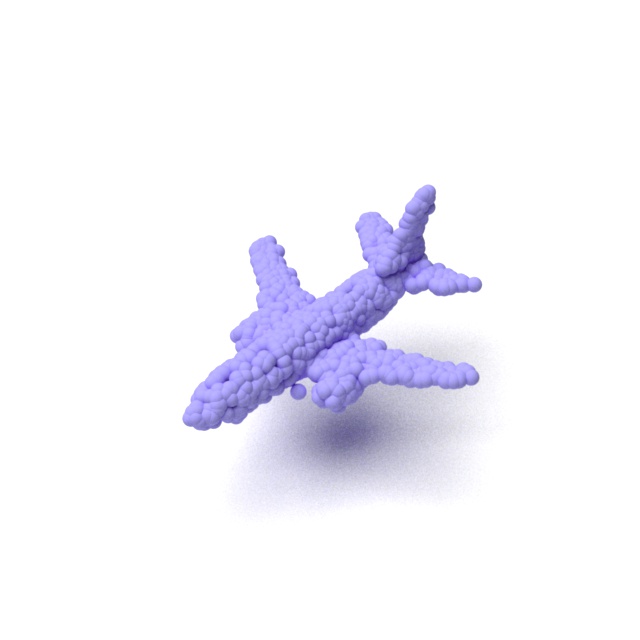} &
        \includegraphics[width=\sizea, trim={\tale} {\tab} {2cm} {\tat},clip]{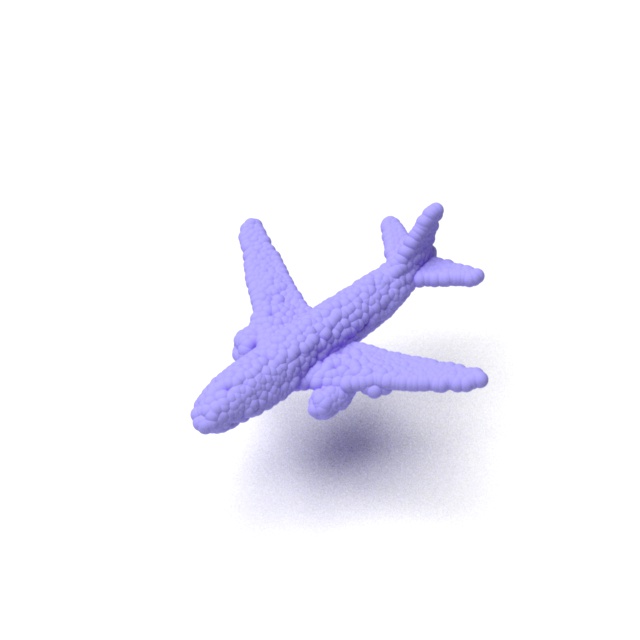} 
        \\
        \includegraphics[width=\sizea, trim={\tale} {\tab} {2cm} {\tat},clip]{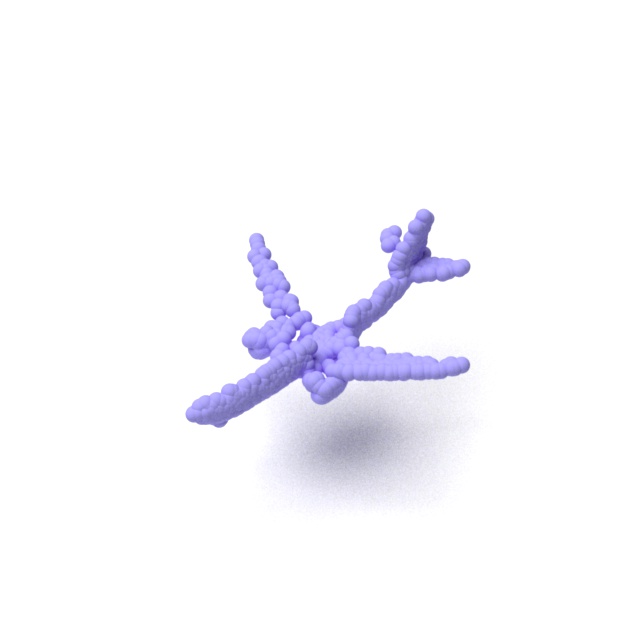} &
        \includegraphics[width=\sizec, trim={\tale} {\tab} {2cm} {\tat},clip]{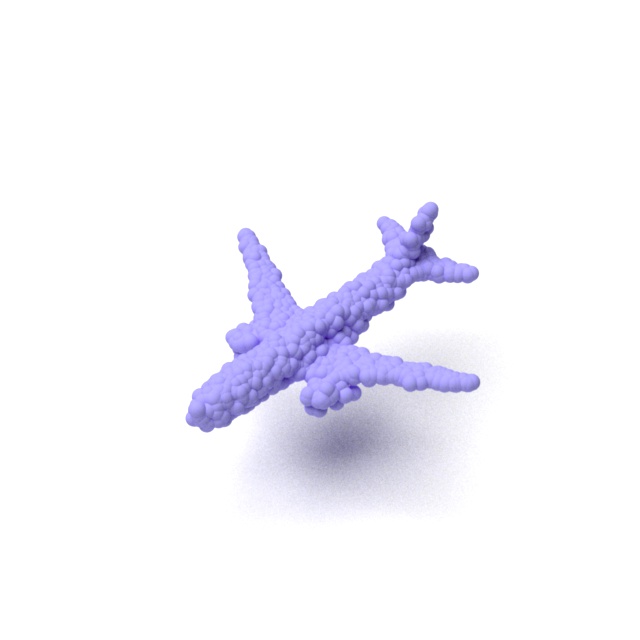} &
        \includegraphics[width=\sizeb, trim={\tale} {\tab} {2cm} {\tat},clip]{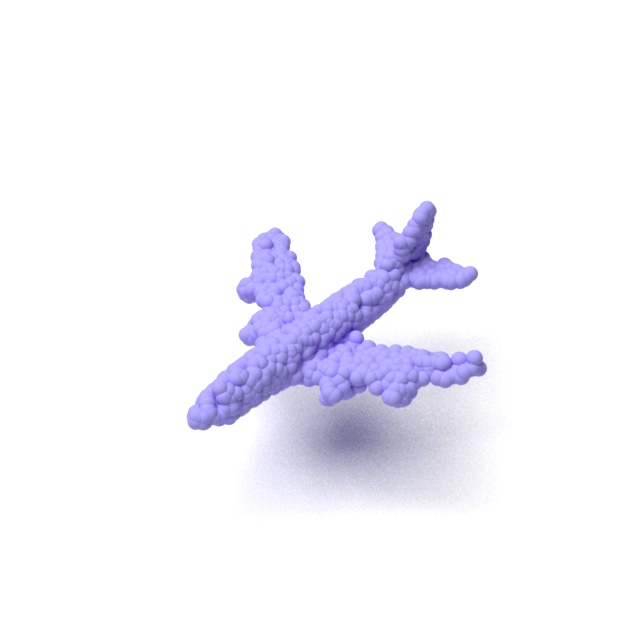} &
        \includegraphics[width=\sizea, trim={\tale} {\tab} {2cm} {\tat},clip]{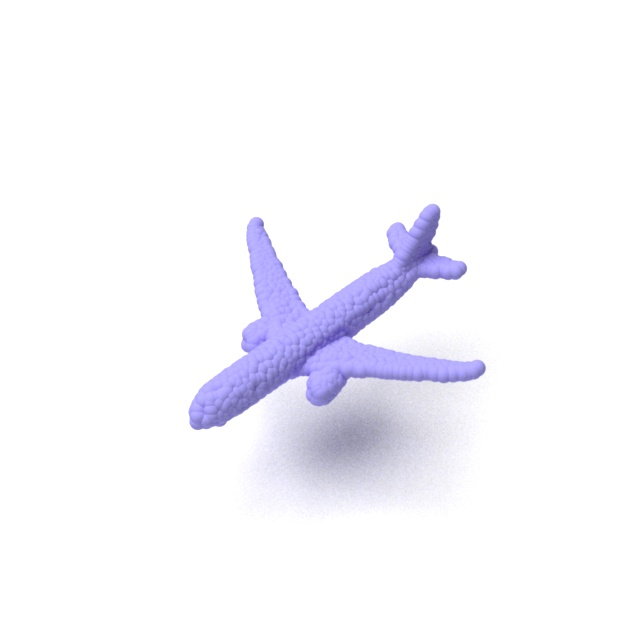} &
        \includegraphics[width=\sizea, trim={\tale} {\tab} {2cm} {\tat},clip]{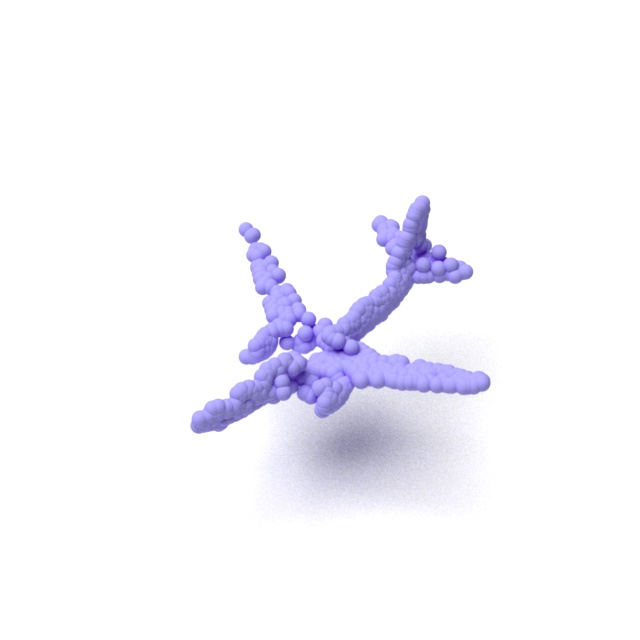} &
        \includegraphics[width=\sizec, trim={\tale} {\tab} {2cm} {\tat},clip]{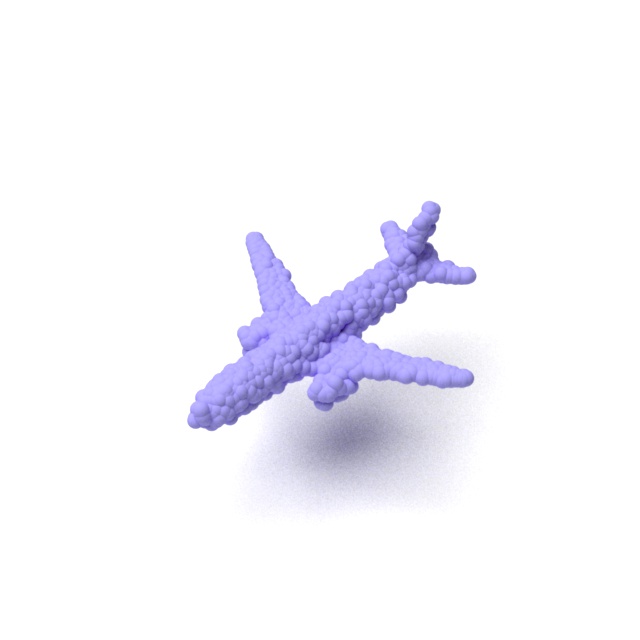} &
        \includegraphics[width=\sizeb, trim={\tale} {\tab} {2cm} {\tat},clip]{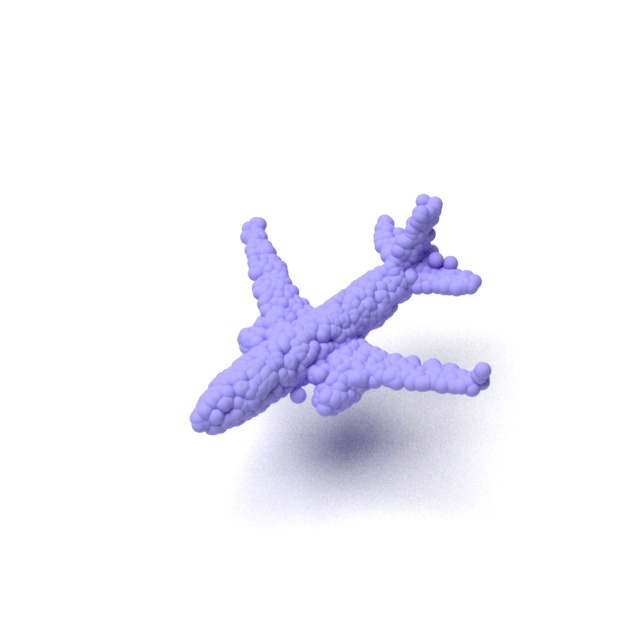} &
        \includegraphics[width=\sizea, trim={\tale} {\tab} {2cm} {\tat},clip]{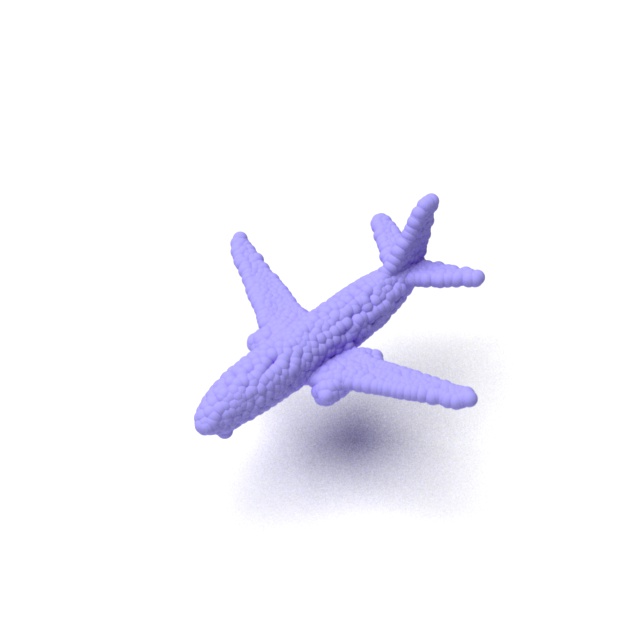} 
        \\
        \includegraphics[width=\sizea, trim={\tale} {\tab} {2cm} {\tat},clip]{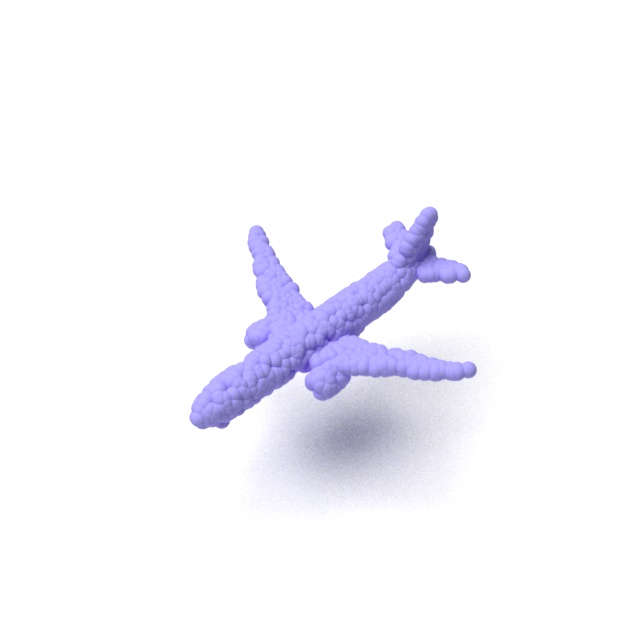} &
        \includegraphics[width=\sizec, trim={\tale} {\tab} {2cm} {\tat},clip]{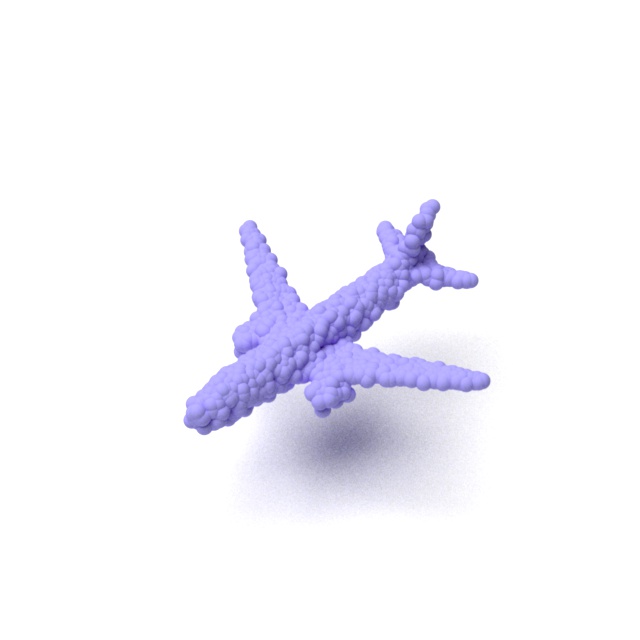} &
        \includegraphics[width=\sizeb, trim={\tale} {\tab} {2cm} {\tat},clip]{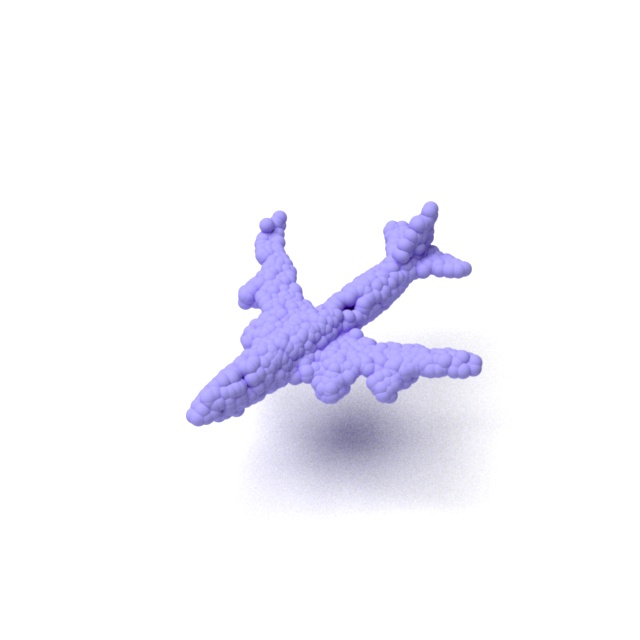} &
        \includegraphics[width=\sizea, trim={\tale} {\tab} {2cm} {\tat},clip]{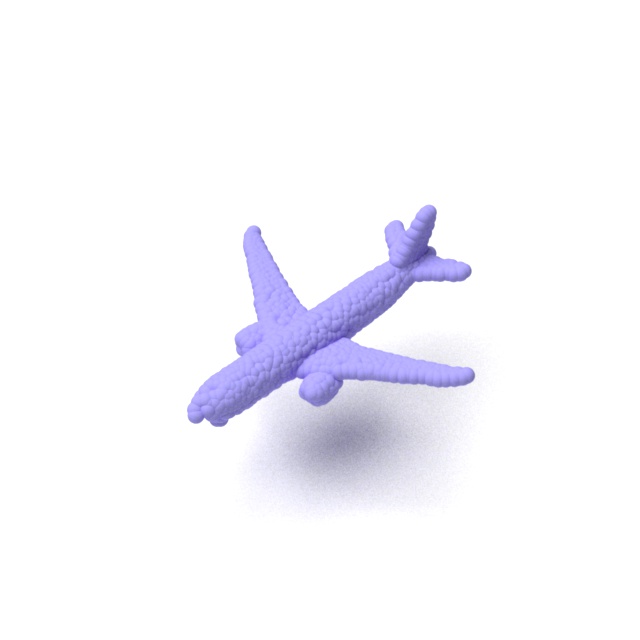} &
        \includegraphics[width=\sizea, trim={\tale} {\tab} {2cm} {\tat},clip]{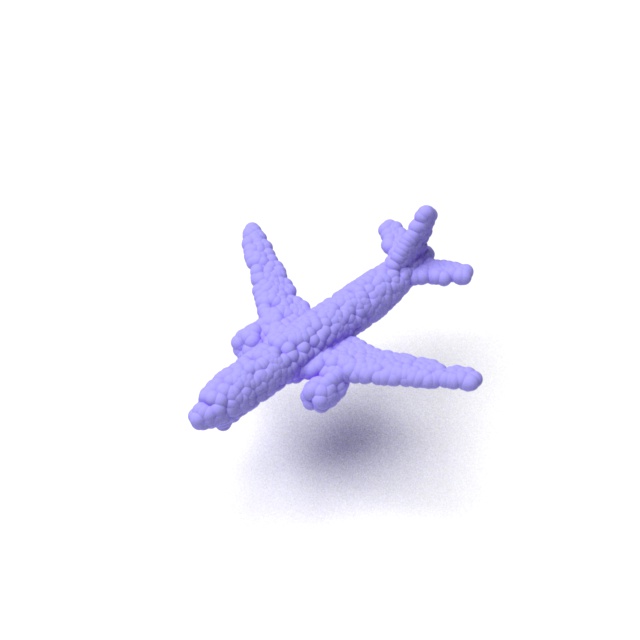} &
        \includegraphics[width=\sizec, trim={\tale} {\tab} {2cm} {\tat},clip]{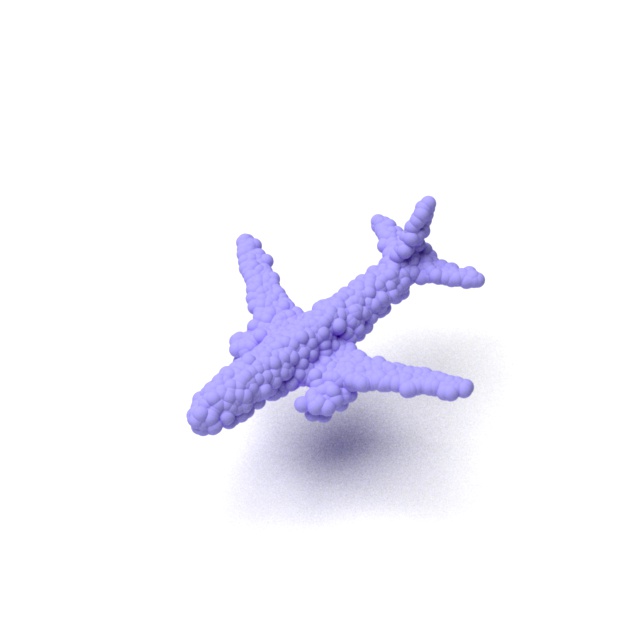} &
        \includegraphics[width=\sizeb, trim={\tale} {\tab} {2cm} {\tat},clip]{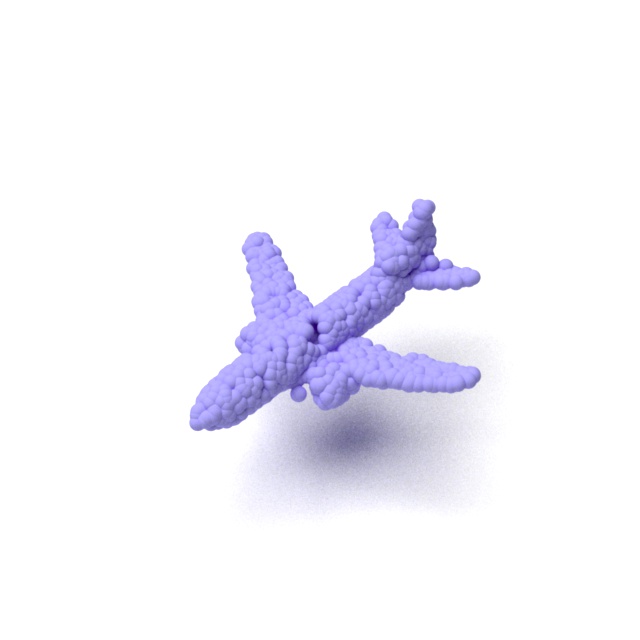} &
        \includegraphics[width=\sizea, trim={\tale} {\tab} {2cm} {\tat},clip]{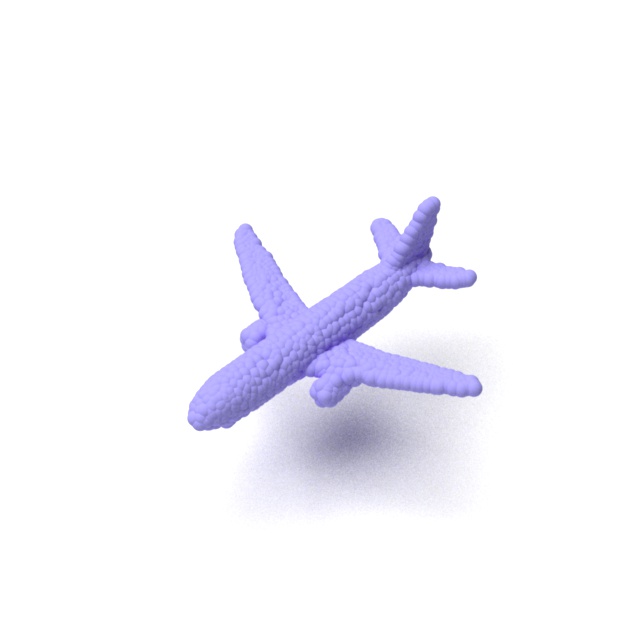}
        \\ \midrule
        \includegraphics[width=\sizea, trim={\tale} {\tab} {2cm} {\tat},clip]{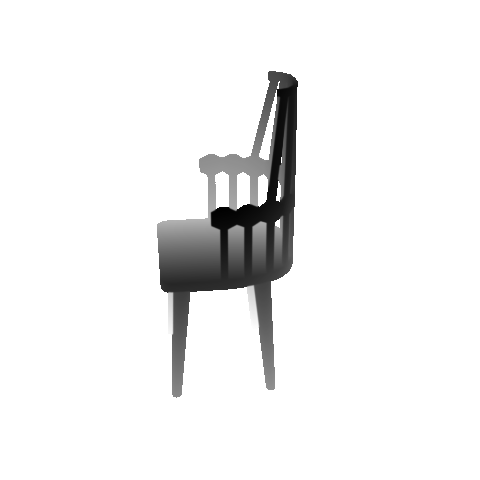} &
        \includegraphics[width=\sizec, trim={\tale} {\tab} {2cm} {\tat},clip]{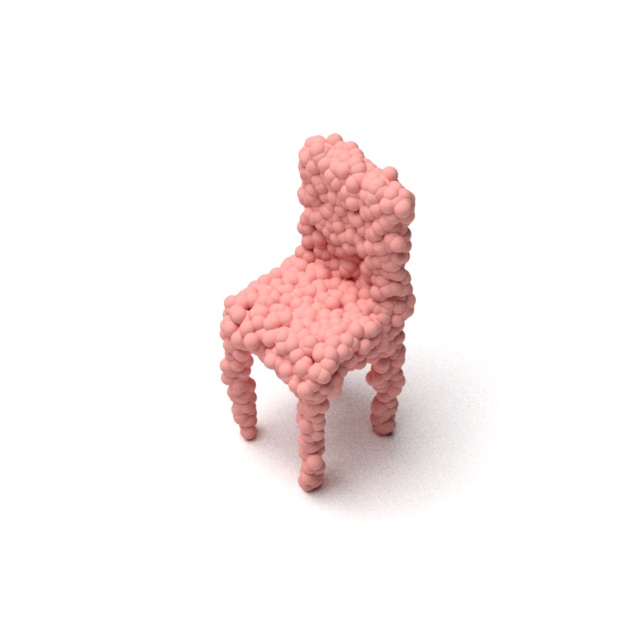} &
        \includegraphics[width=\sizeb, trim={\tale} {\tab} {2cm} {\tat},clip]{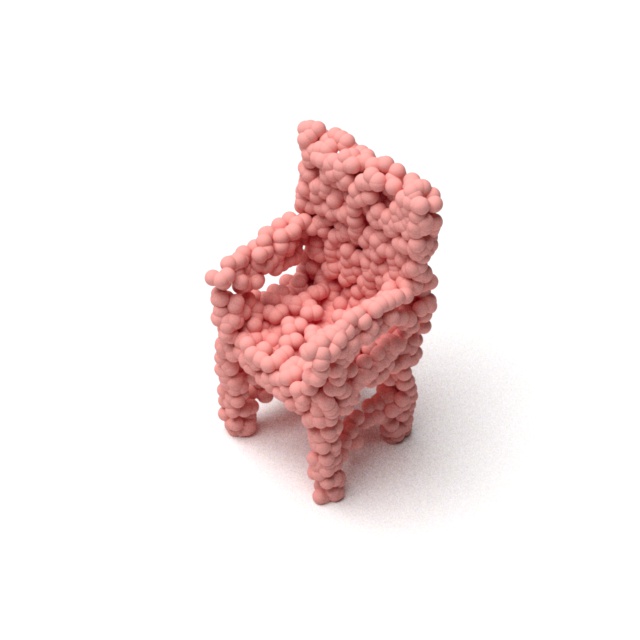} &
        \includegraphics[width=\sizeb, trim={\tale} {\tab} {2cm} {\tat},clip]{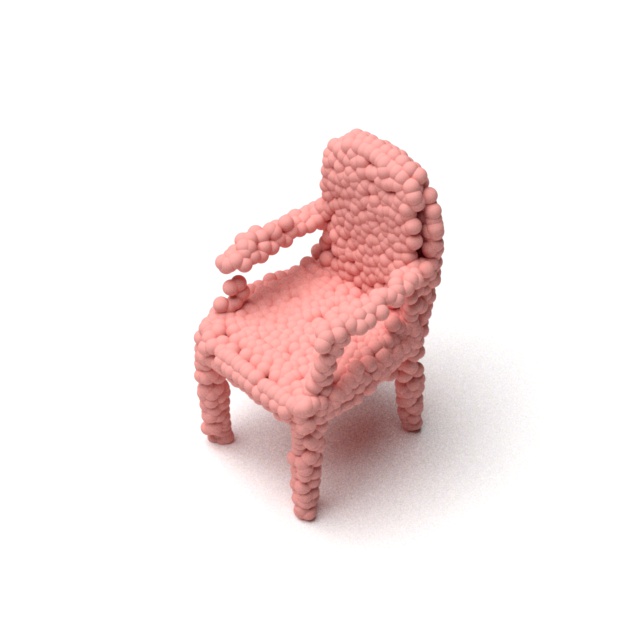} &
        \includegraphics[width=\sizea, trim={\tale} {\tab} {2cm} {\tat},clip]{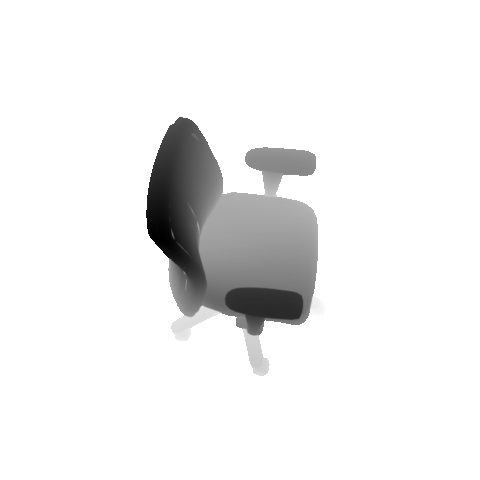} &
        \includegraphics[width=\sizec, trim={\tale} {\tab} {2cm} {\tat},clip]{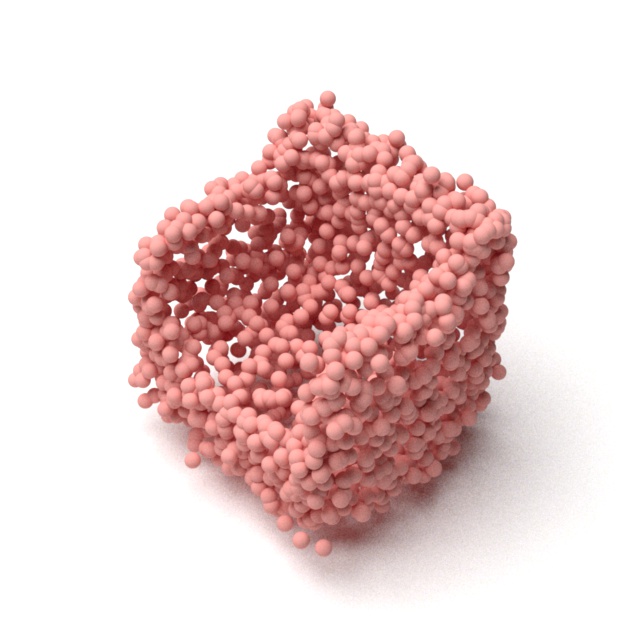} &
        \includegraphics[width=\sizeb, trim={\tale} {\tab} {2cm} {\tat},clip]{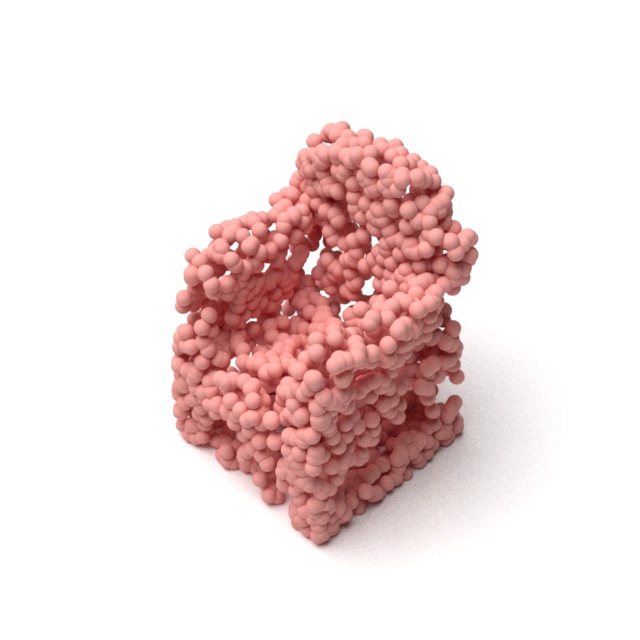} &
        \includegraphics[width=\sizea, trim={\tale} {\tab} {2cm} {\tat},clip]{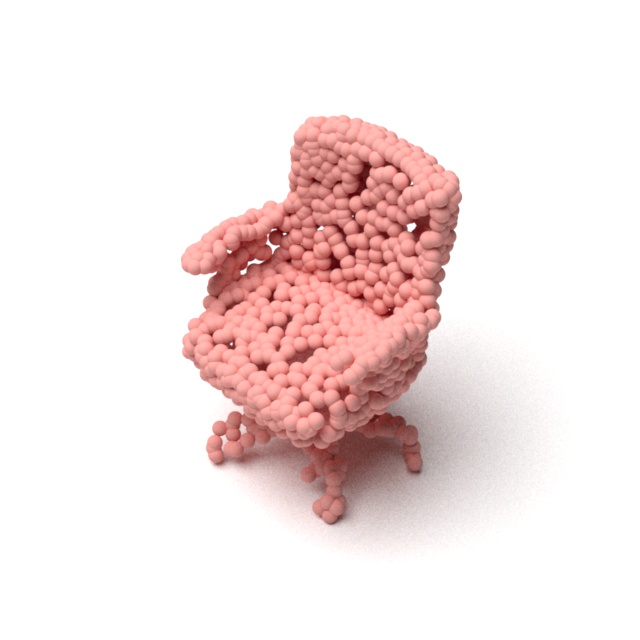} 
        \\
        \includegraphics[width=\sizea, trim={\tale} {\tab} {2cm} {\tat},clip]{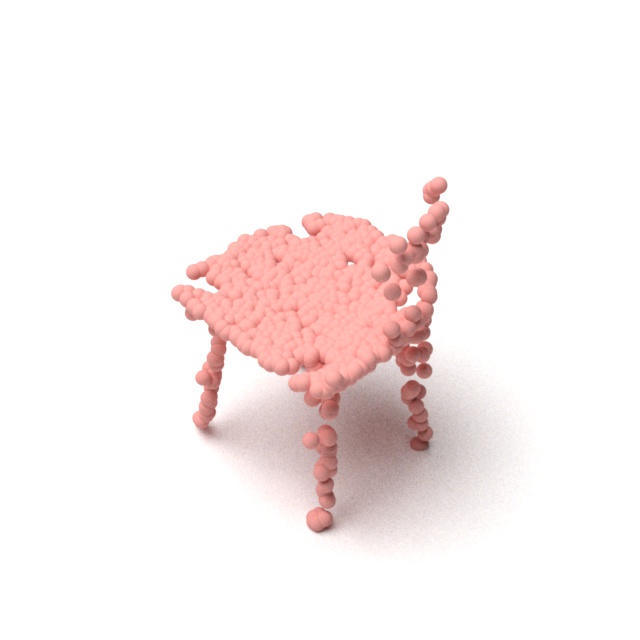} &
        \includegraphics[width=\sizec, trim={\tale} {\tab} {2cm} {\tat},clip]{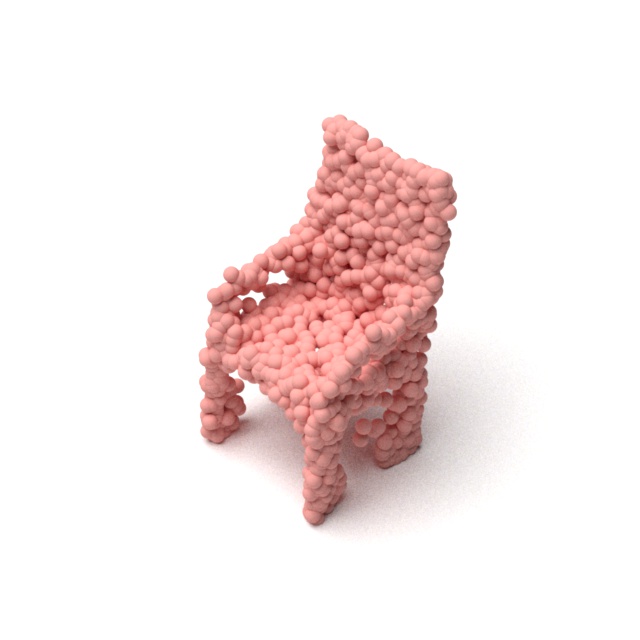} &
        \includegraphics[width=\sizeb, trim={\tale} {\tab} {2cm} {\tat},clip]{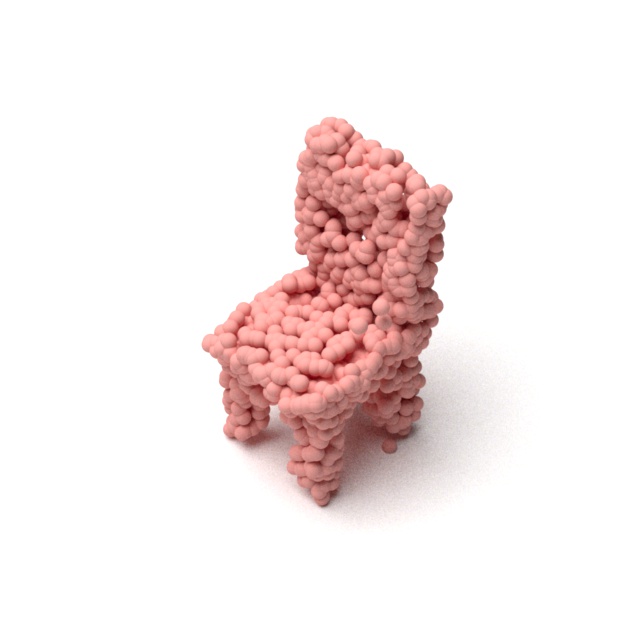} &
        \includegraphics[width=\sizeb, trim={\tale} {\tab} {2cm} {\tat},clip]{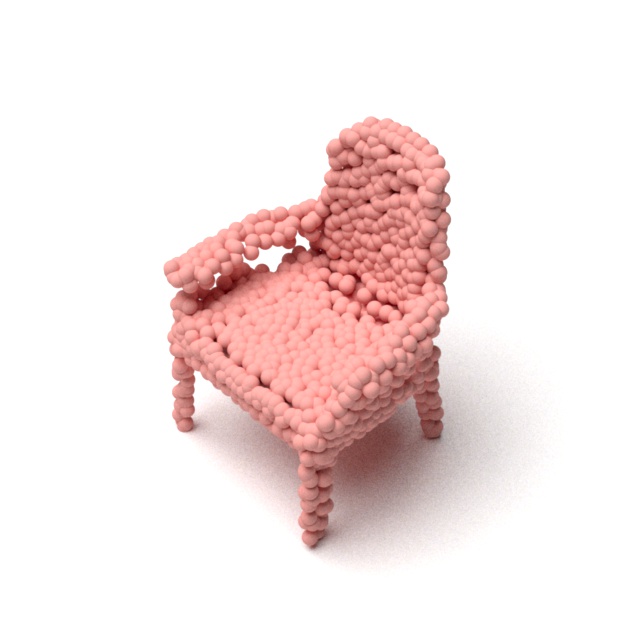} &
        \includegraphics[width=\sizeb, trim={\tale} {\tab} {2cm} {\tat},clip]{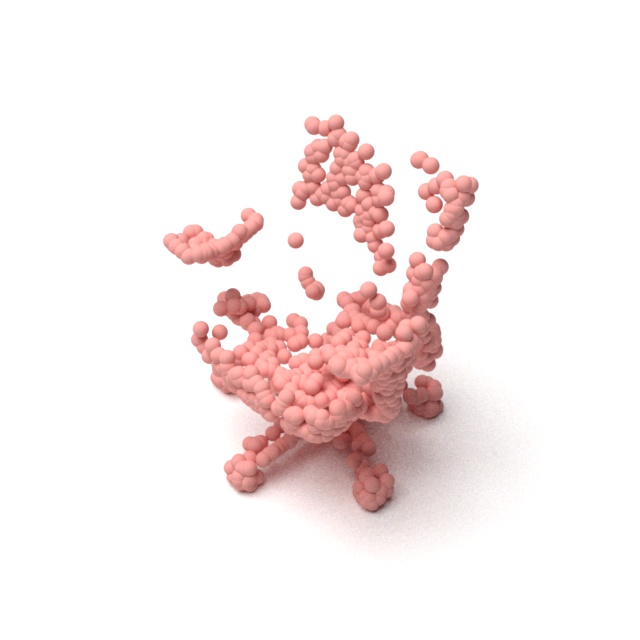} &
        \includegraphics[width=\sizec, trim={\tale} {\tab} {2cm} {\tat},clip]{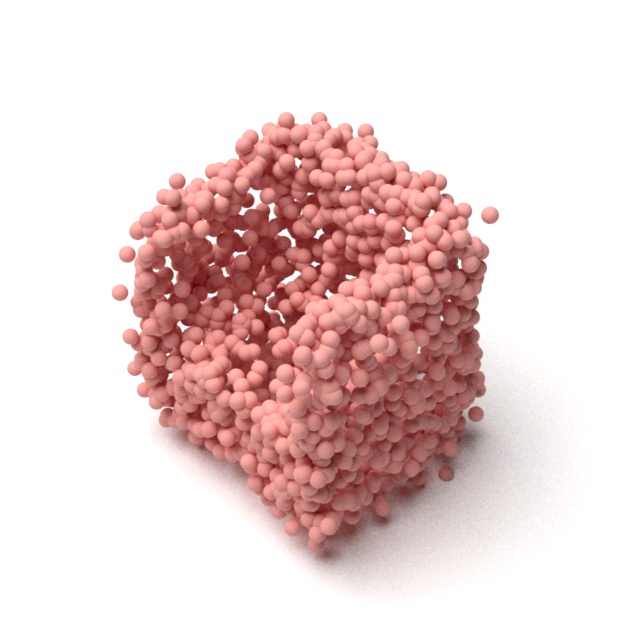} &
        \includegraphics[width=\sizeb, trim={\tale} {\tab} {2cm} {\tat},clip]{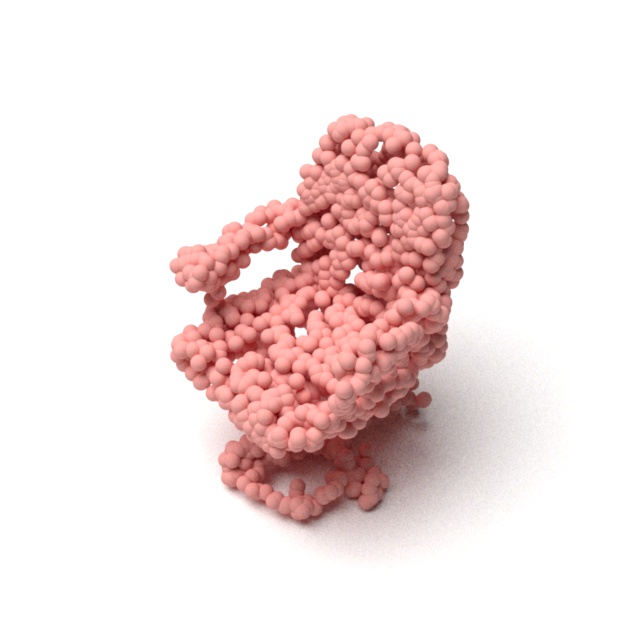} &
        \includegraphics[width=\sizea, trim={5cm} {\tab} {2cm} {\tat},clip]{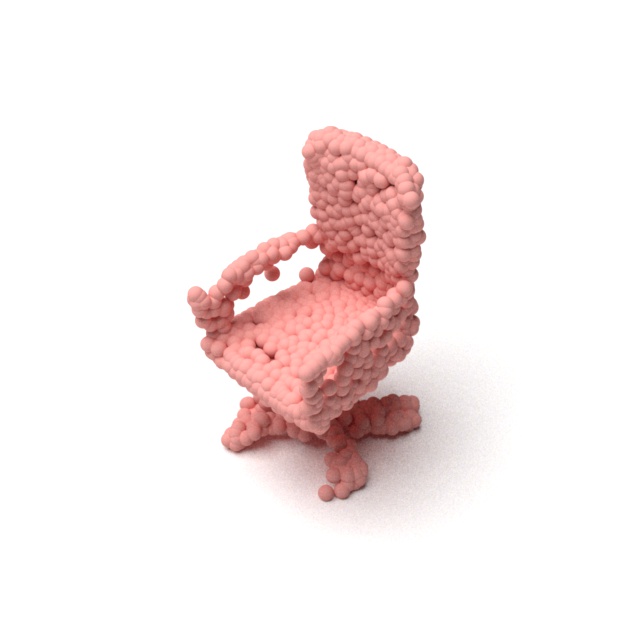}
        \\
        \includegraphics[width=\sizea, trim={\tale} {\tab} {2cm} {\tat},clip]{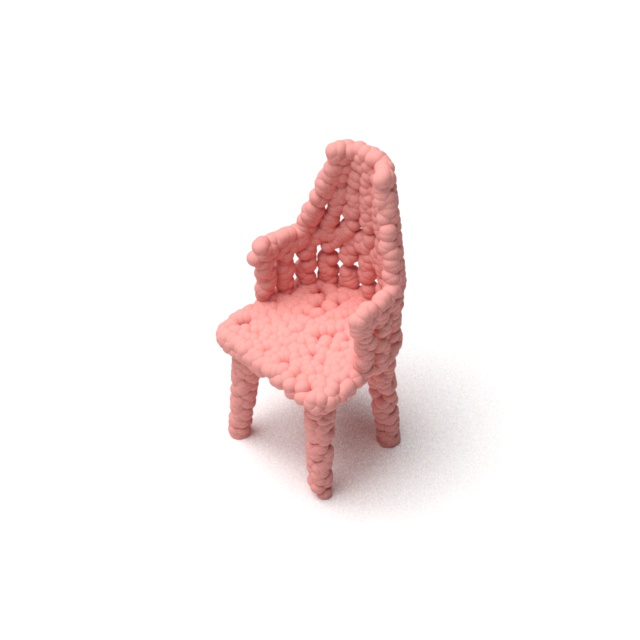} &
        \includegraphics[width=\sizec, trim={\tale} {\tab} {2cm} {\tat},clip]{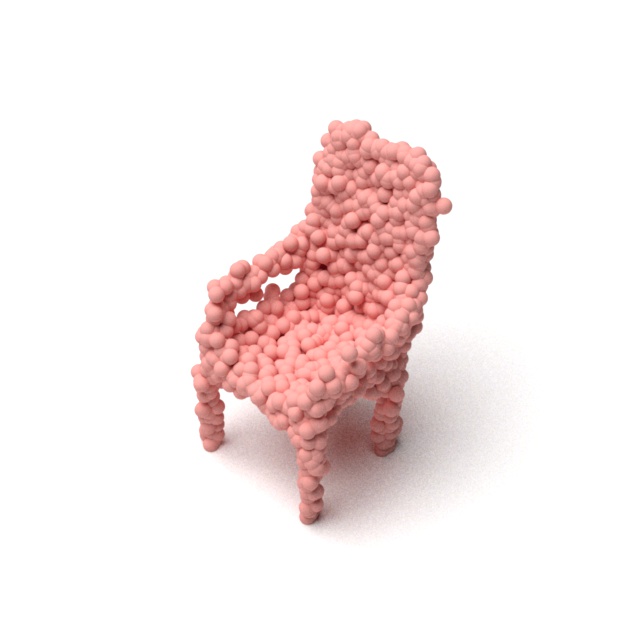} &
        \includegraphics[width=\sizeb, trim={\tale} {\tab} {2cm} {\tat},clip]{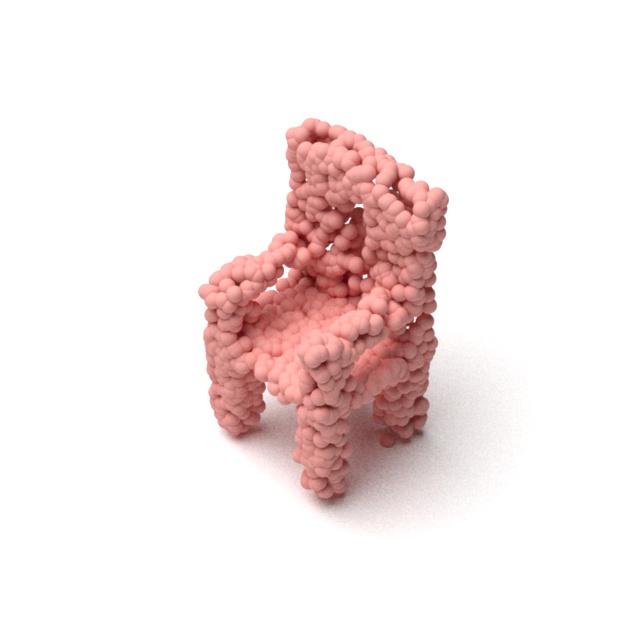} &
        \includegraphics[width=\sizeb, trim={\tale} {\tab} {2cm} {\tat},clip]{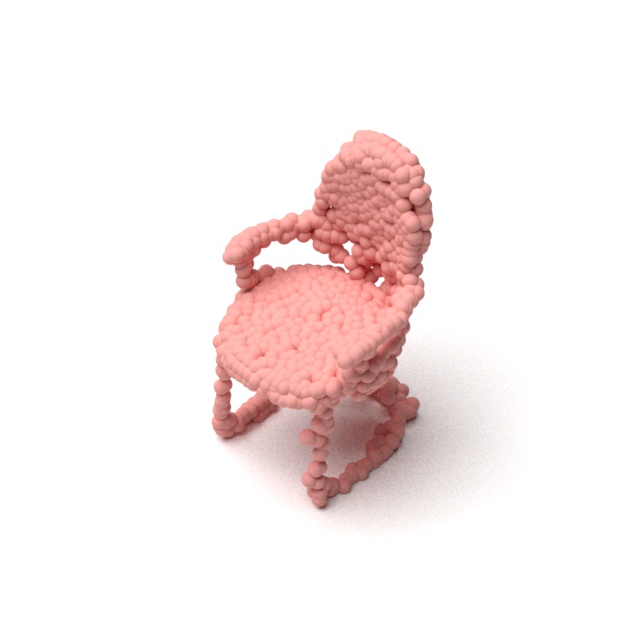} &
        \includegraphics[width=\sizea, trim={\tale} {\tab} {2cm} {\tat},clip]{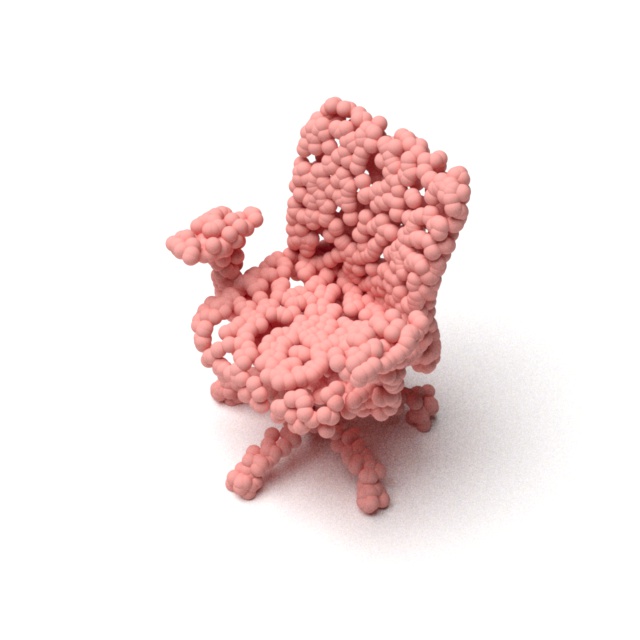} &
        \includegraphics[width=\sizec, trim={\tale} {\tab} {2cm} {\tat},clip]{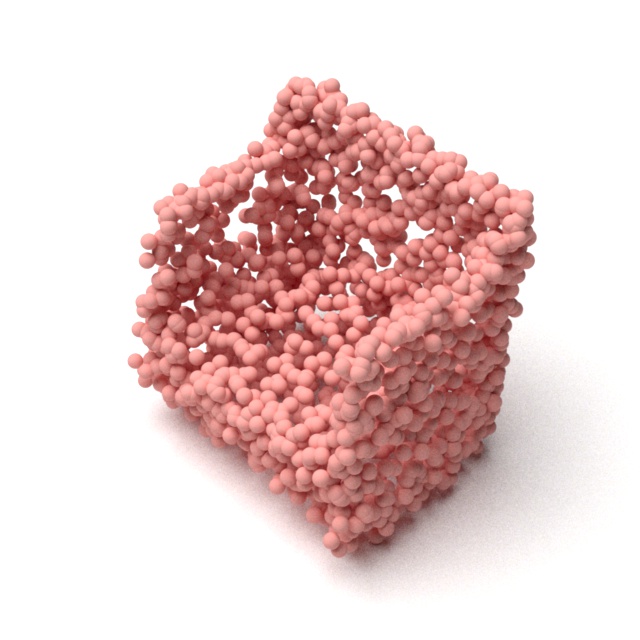} &
        \includegraphics[width=\sizeb, trim={\tale} {\tab} {2cm} {\tat},clip]{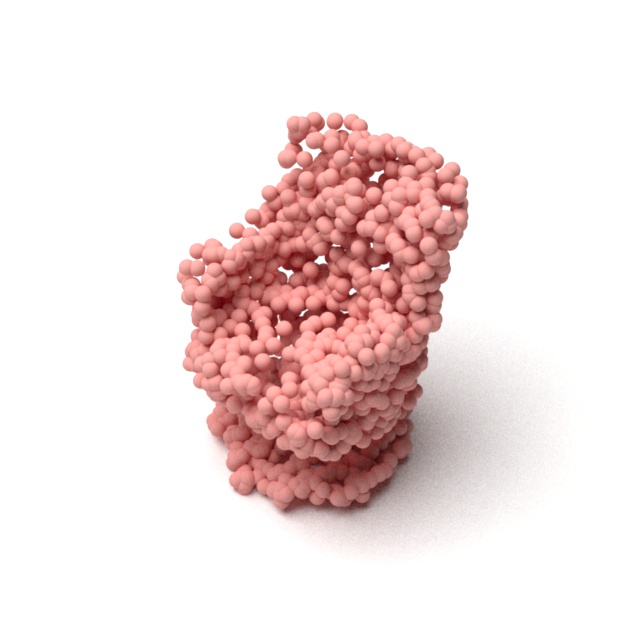} &
        \includegraphics[width=\sizea, trim={5cm} {\tab} {2cm} {\tat},clip]{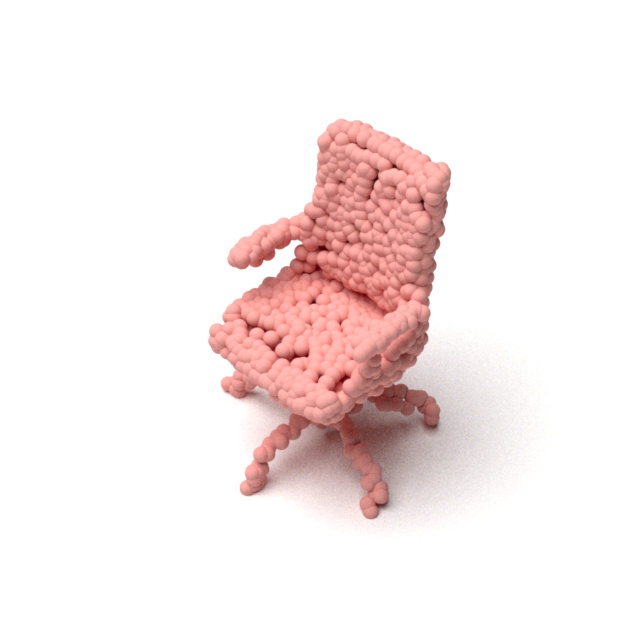}
        \\
        Input$/$GT & MSC & PVD & Ours & Input$/$GT & MSC & PVD & Ours
    \end{tabular}
    \end{center}
    \caption{Multi-modal shape completion results. We shown 4 samples comparing to MSC~\cite{wu2020multimodal} and PVD~\cite{zhou20213d}. The input depth-map, partial point cloud, and reference ground-truth shape for each sample is shown in the first column, respectively (from top to bottom). 
    }

    \label{fig:sup:completion}
\end{figure}

In Figure~\ref{fig:sup:completion}, we show more conditional generation results. Our shape completion results tend to show more variation and have better visual quality comparing to MSC~\cite{wu2020multimodal} and PVD~\cite{zhou20213d}.

\section{Limitations}
\label{sec:limitation}

\begin{figure}[h]

    \begin{center}
    \newcommand{\sizea}{0.15\linewidth}
    \newcommand{\sizeb}{0.15\linewidth}
    \newcommand{\sizec}{0.15\linewidth}
    \newcommand{\tare}{5cm}
    \newcommand{\tale}{2.5cm}
    \newcommand{\tal}{3.5cm}
    \newcommand{\tab}{2.5cm}
    \newcommand{\tar}{3.5cm}
    \newcommand{\tat}{2.5cm}
    \newcommand{\tcl}{3.0cm}
    \newcommand{\tcb}{3cm}
    \newcommand{\tcr}{4cm}
    \newcommand{\tct}{4.2cm}
    \newcommand{\thl}{3.0cm}
    \newcommand{\thb}{0.0cm}
    \newcommand{\thr}{3cm}
    \newcommand{\tht}{2cm}
    \setlength{\tabcolsep}{0pt}
    \renewcommand{\arraystretch}{0}
    \begin{tabular}{@{}cc:cc:cc@{}}
        \includegraphics[width=\sizea, trim={\tale} {\tab} {2.5cm} 
        {\tat},clip]{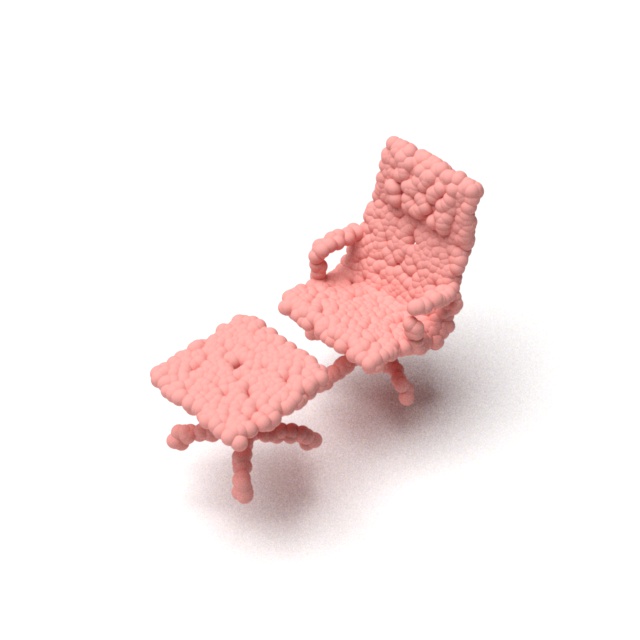} &
        \includegraphics[width=\sizea, trim={\tal} {\tab} {2.5cm} {\tat},clip]{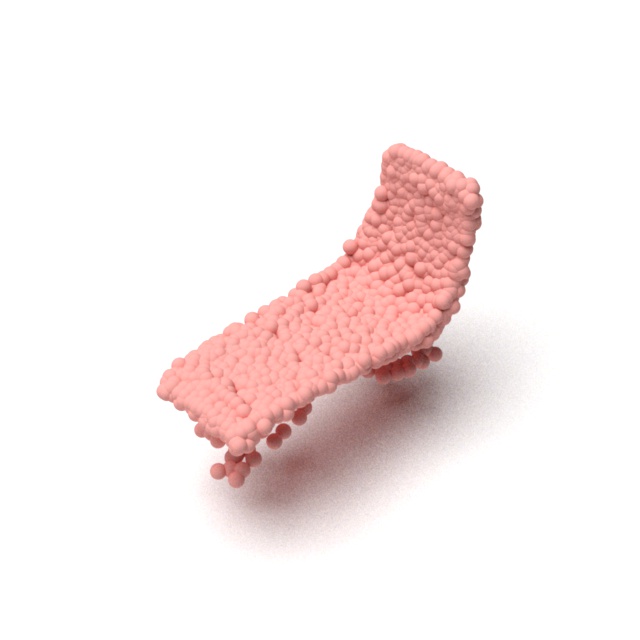} &
        \includegraphics[width=\sizea, trim={\tale} {\tab} {2.5cm} 
        {\tat},clip]{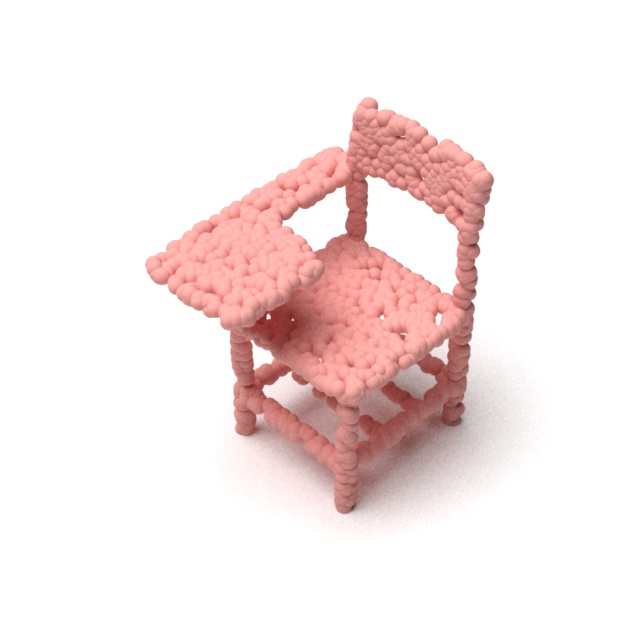} &
        \includegraphics[width=\sizea, trim={\tal} {\tab} {2.5cm} {\tat},clip]{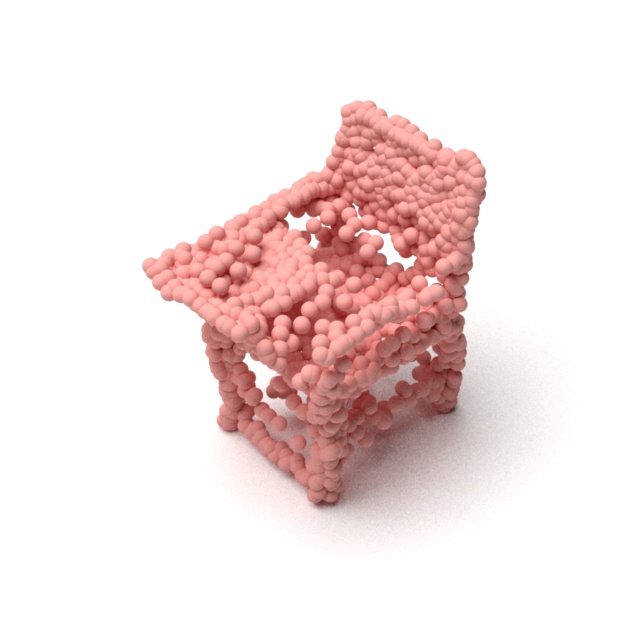} &
        \includegraphics[width=\sizea, trim={\tale} {\tab} {2.5cm} 
        {\tat},clip]{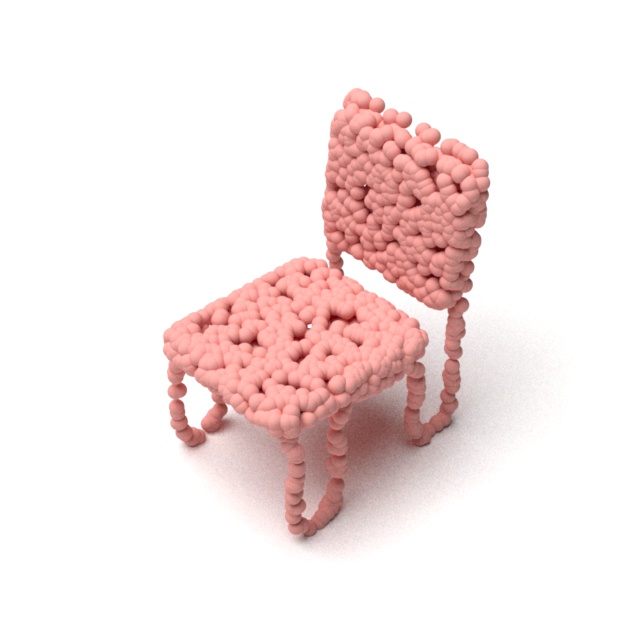} &
        \includegraphics[width=\sizea, trim={\tal} {\tab} {2.5cm} {\tat},clip]{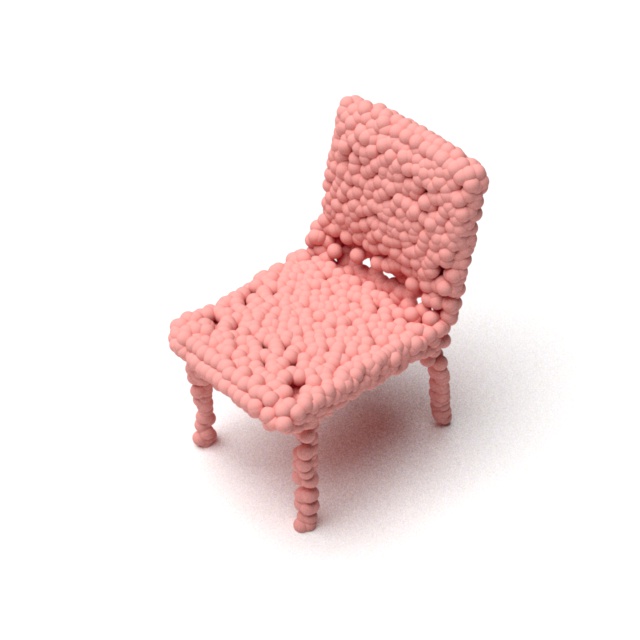} 
        \\
        Input & Ours & Input & Ours & Input & Ours
    \end{tabular}
    \end{center}
    \caption{Failure cases.
    }
    \label{fig:limitation}
\end{figure}

Our model relies on the learned correspondence from the canonical mapping function, therefore, inherits similar limitations from Cheng et al.~\cite{cheng2021learning}. Our model fails to reconstruct certain samples with holes or with complex topology. We show some failure cases of our model in Figure~\ref{fig:limitation}.



\clearpage

\bibliographystyle{unsrt}
\bibliography{main}
\end{document}